%% file: main.tex
\PassOptionsToPackage{table}{xcolor}
\documentclass[a4paper,fleqn]{cas-sc}

\usepackage[authoryear]{natbib}
\usepackage{booktabs}
\usepackage{amsmath,amssymb}
\usepackage{graphicx}
\usepackage{xcolor}
\usepackage{url}
\usepackage{placeins}
\usepackage{paper4a_tables}

\AtBeginEnvironment{table}{\let\sffamily\rmfamily}
\AtBeginEnvironment{figure}{\let\sffamily\rmfamily}

\newtheorem{proposition}{Proposition}
\newproof{pf}{Proof}

\input{numbers.tex}

\begin{document}

\let\WriteBookmarks\relax
\def\floatpagepagefraction{1}
\def\textpagefraction{.001}

\shorttitle{Benchmarking System One decision models for decision gates}
\shortauthors{A. Rafe and S. Das}

\title [mode = title]{Benchmarking System One decision models against trained classifiers and language models for automated decision gates}

\author[1]{Amir Rafe}[orcid=0000-0002-4089-2088]
\cormark[1]
\ead{amir.rafe@txstate.edu}

\author[1]{Subasish Das}[orcid=0000-0002-1671-2753]
\ead{subasish@txstate.edu}

\affiliation[1]{organization={Civil Engineering, Texas State University},
    addressline={601 University Drive},
    city={San Marcos},
    postcode={78666},
    state={Texas},
    country={USA}}

\cortext[1]{Corresponding author}

\input{sections/abstract}


\maketitle

\input{sections/introduction}
\input{sections/related}
\input{sections/methods}
\input{sections/models_datasets}
\input{sections/results}
\input{sections/engineering}
\input{sections/discussion}
\input{sections/conclusion}

\appendix
\counterwithin{figure}{section}
\counterwithin{table}{section}
\section*{Appendices}
\input{sections/appendix_proofs}
\input{sections/appendix}
\FloatBarrier

\input{sections/declarations}

\bibliographystyle{cas-model2-names}
\bibliography{refs}

\end{document}

%% file: numbers.tex
\makeatletter
\newcommand{\num}[1]{%
  \expandafter\ifx\csname cdnum@#1\endcsname\relax
    \GenericError{}{MISSING NUMBER: #1}{}{}\textbf{??}%
  \else\csname cdnum@#1\endcsname\fi}
\makeatother
\expandafter\def\csname cdnum@bb.bbnine.toxicity.kswap.flips\endcsname{7.2}
\expandafter\def\csname cdnum@bb.bbninebase.doneneutral.acc\endcsname{0.562}
\expandafter\def\csname cdnum@bb.bbninebase.dtwo.acc\endcsname{0.728}
\expandafter\def\csname cdnum@bb.bbtwobase.toxicity.kswap.flips\endcsname{100.0}
\expandafter\def\csname cdnum@bb.decider.bbtwobase.donek01.flips.diff\endcsname{0.7}
\expandafter\def\csname cdnum@bb.decider.bbtwobase.donek01.flips.hi\endcsname{2.3}
\expandafter\def\csname cdnum@bb.decider.bbtwobase.donek01.flips.lo\endcsname{\ensuremath{-}1.0}
\expandafter\def\csname cdnum@bb.decider.bbtwobase.donekrand.flips.diff\endcsname{0.0}
\expandafter\def\csname cdnum@bb.decider.bbtwobase.donekrand.flips.hi\endcsname{2.3}
\expandafter\def\csname cdnum@bb.decider.bbtwobase.donekrand.flips.lo\endcsname{\ensuremath{-}2.3}
\expandafter\def\csname cdnum@bb.decider.bbtwobase.donekswap.flips.diff\endcsname{0.0}
\expandafter\def\csname cdnum@bb.decider.bbtwobase.donekswap.flips.hi\endcsname{2.4}
\expandafter\def\csname cdnum@bb.decider.bbtwobase.donekswap.flips.lo\endcsname{\ensuremath{-}2.3}
\expandafter\def\csname cdnum@bb.decider.bbtwobase.doneneutral.acc.diff\endcsname{17.3}
\expandafter\def\csname cdnum@bb.decider.bbtwobase.doneneutral.acc.hi\endcsname{20.2}
\expandafter\def\csname cdnum@bb.decider.bbtwobase.doneneutral.acc.lo\endcsname{14.4}
\expandafter\def\csname cdnum@bb.decider.bbtwobase.doneneutral.ece.diff\endcsname{\ensuremath{-}15.8}
\expandafter\def\csname cdnum@bb.decider.bbtwobase.doneneutral.ece.hi\endcsname{\ensuremath{-}13.3}
\expandafter\def\csname cdnum@bb.decider.bbtwobase.doneneutral.ece.lo\endcsname{\ensuremath{-}18.1}
\expandafter\def\csname cdnum@bb.decider.bbtwobase.dtwo.acc.diff\endcsname{34.8}
\expandafter\def\csname cdnum@bb.decider.bbtwobase.dtwo.acc.hi\endcsname{39.3}
\expandafter\def\csname cdnum@bb.decider.bbtwobase.dtwo.acc.lo\endcsname{30.5}
\expandafter\def\csname cdnum@bb.decider.bbtwobase.dtwo.ece.diff\endcsname{\ensuremath{-}20.7}
\expandafter\def\csname cdnum@bb.decider.bbtwobase.dtwo.ece.hi\endcsname{\ensuremath{-}16.2}
\expandafter\def\csname cdnum@bb.decider.bbtwobase.dtwo.ece.lo\endcsname{\ensuremath{-}23.9}
\expandafter\def\csname cdnum@bb.decider.bbtwobase.emotion.acc.diff\endcsname{36.9}
\expandafter\def\csname cdnum@bb.decider.bbtwobase.emotion.acc.hi\endcsname{41.6}
\expandafter\def\csname cdnum@bb.decider.bbtwobase.emotion.acc.lo\endcsname{32.1}
\expandafter\def\csname cdnum@bb.decider.bbtwobase.emotion.ece.diff\endcsname{\ensuremath{-}5.5}
\expandafter\def\csname cdnum@bb.decider.bbtwobase.emotion.ece.hi\endcsname{\ensuremath{-}1.2}
\expandafter\def\csname cdnum@bb.decider.bbtwobase.emotion.ece.lo\endcsname{\ensuremath{-}9.5}
\expandafter\def\csname cdnum@bb.decider.bbtwobase.politeness.acc.diff\endcsname{4.8}
\expandafter\def\csname cdnum@bb.decider.bbtwobase.politeness.acc.hi\endcsname{11.4}
\expandafter\def\csname cdnum@bb.decider.bbtwobase.politeness.acc.lo\endcsname{\ensuremath{-}1.6}
\expandafter\def\csname cdnum@bb.decider.bbtwobase.politeness.ece.diff\endcsname{9.6}
\expandafter\def\csname cdnum@bb.decider.bbtwobase.politeness.ece.hi\endcsname{15.1}
\expandafter\def\csname cdnum@bb.decider.bbtwobase.politeness.ece.lo\endcsname{3.7}
\expandafter\def\csname cdnum@bb.decider.bbtwobase.power.acc.diff\endcsname{1.4}
\expandafter\def\csname cdnum@bb.decider.bbtwobase.power.acc.hi\endcsname{8.2}
\expandafter\def\csname cdnum@bb.decider.bbtwobase.power.acc.lo\endcsname{\ensuremath{-}5.2}
\expandafter\def\csname cdnum@bb.decider.bbtwobase.power.ece.diff\endcsname{12.0}
\expandafter\def\csname cdnum@bb.decider.bbtwobase.power.ece.hi\endcsname{16.8}
\expandafter\def\csname cdnum@bb.decider.bbtwobase.power.ece.lo\endcsname{4.8}
\expandafter\def\csname cdnum@bb.decider.bbtwobase.powerkswap.flips.diff\endcsname{\ensuremath{-}93.2}
\expandafter\def\csname cdnum@bb.decider.bbtwobase.powerkswap.flips.hi\endcsname{\ensuremath{-}90.8}
\expandafter\def\csname cdnum@bb.decider.bbtwobase.powerkswap.flips.lo\endcsname{\ensuremath{-}95.4}
\expandafter\def\csname cdnum@bb.decider.bbtwobase.toxicity.acc.diff\endcsname{1.8}
\expandafter\def\csname cdnum@bb.decider.bbtwobase.toxicity.acc.hi\endcsname{4.2}
\expandafter\def\csname cdnum@bb.decider.bbtwobase.toxicity.acc.lo\endcsname{\ensuremath{-}0.6}
\expandafter\def\csname cdnum@bb.decider.bbtwobase.toxicity.ece.diff\endcsname{24.2}
\expandafter\def\csname cdnum@bb.decider.bbtwobase.toxicity.ece.hi\endcsname{26.7}
\expandafter\def\csname cdnum@bb.decider.bbtwobase.toxicity.ece.lo\endcsname{21.7}
\expandafter\def\csname cdnum@bb.decider.bbtwobase.toxicitykrand.flips.diff\endcsname{\ensuremath{-}35.0}
\expandafter\def\csname cdnum@bb.decider.bbtwobase.toxicitykrand.flips.hi\endcsname{\ensuremath{-}30.8}
\expandafter\def\csname cdnum@bb.decider.bbtwobase.toxicitykrand.flips.lo\endcsname{\ensuremath{-}39.2}
\expandafter\def\csname cdnum@bb.decider.bbtwobase.toxicitykswap.flips.diff\endcsname{\ensuremath{-}99.0}
\expandafter\def\csname cdnum@bb.decider.bbtwobase.toxicitykswap.flips.hi\endcsname{\ensuremath{-}98.0}
\expandafter\def\csname cdnum@bb.decider.bbtwobase.toxicitykswap.flips.lo\endcsname{\ensuremath{-}99.8}
\expandafter\def\csname cdnum@bb.kevlarge.bbninebase.donek01.flips.diff\endcsname{\ensuremath{-}0.7}
\expandafter\def\csname cdnum@bb.kevlarge.bbninebase.donek01.flips.hi\endcsname{1.8}
\expandafter\def\csname cdnum@bb.kevlarge.bbninebase.donek01.flips.lo\endcsname{\ensuremath{-}3.1}
\expandafter\def\csname cdnum@bb.kevlarge.bbninebase.donekrand.flips.diff\endcsname{\ensuremath{-}6.5}
\expandafter\def\csname cdnum@bb.kevlarge.bbninebase.donekrand.flips.hi\endcsname{\ensuremath{-}3.3}
\expandafter\def\csname cdnum@bb.kevlarge.bbninebase.donekrand.flips.lo\endcsname{\ensuremath{-}9.7}
\expandafter\def\csname cdnum@bb.kevlarge.bbninebase.donekswap.flips.diff\endcsname{\ensuremath{-}34.0}
\expandafter\def\csname cdnum@bb.kevlarge.bbninebase.donekswap.flips.hi\endcsname{\ensuremath{-}29.7}
\expandafter\def\csname cdnum@bb.kevlarge.bbninebase.donekswap.flips.lo\endcsname{\ensuremath{-}38.2}
\expandafter\def\csname cdnum@bb.kevlarge.bbninebase.doneneutral.acc.diff\endcsname{14.9}
\expandafter\def\csname cdnum@bb.kevlarge.bbninebase.doneneutral.acc.hi\endcsname{17.7}
\expandafter\def\csname cdnum@bb.kevlarge.bbninebase.doneneutral.acc.lo\endcsname{12.2}
\expandafter\def\csname cdnum@bb.kevlarge.bbninebase.doneneutral.ece.diff\endcsname{\ensuremath{-}4.0}
\expandafter\def\csname cdnum@bb.kevlarge.bbninebase.doneneutral.ece.hi\endcsname{\ensuremath{-}0.9}
\expandafter\def\csname cdnum@bb.kevlarge.bbninebase.doneneutral.ece.lo\endcsname{\ensuremath{-}7.3}
\expandafter\def\csname cdnum@bb.kevlarge.bbninebase.dtwo.acc.diff\endcsname{12.2}
\expandafter\def\csname cdnum@bb.kevlarge.bbninebase.dtwo.acc.hi\endcsname{15.5}
\expandafter\def\csname cdnum@bb.kevlarge.bbninebase.dtwo.acc.lo\endcsname{8.8}
\expandafter\def\csname cdnum@bb.kevlarge.bbninebase.dtwo.ece.diff\endcsname{35.0}
\expandafter\def\csname cdnum@bb.kevlarge.bbninebase.dtwo.ece.hi\endcsname{36.6}
\expandafter\def\csname cdnum@bb.kevlarge.bbninebase.dtwo.ece.lo\endcsname{29.7}
\expandafter\def\csname cdnum@bb.kevlarge.bbninebase.emotion.acc.diff\endcsname{4.2}
\expandafter\def\csname cdnum@bb.kevlarge.bbninebase.emotion.acc.hi\endcsname{7.4}
\expandafter\def\csname cdnum@bb.kevlarge.bbninebase.emotion.acc.lo\endcsname{1.0}
\expandafter\def\csname cdnum@bb.kevlarge.bbninebase.emotion.ece.diff\endcsname{\ensuremath{-}13.0}
\expandafter\def\csname cdnum@bb.kevlarge.bbninebase.emotion.ece.hi\endcsname{\ensuremath{-}6.4}
\expandafter\def\csname cdnum@bb.kevlarge.bbninebase.emotion.ece.lo\endcsname{\ensuremath{-}16.7}
\expandafter\def\csname cdnum@bb.kevlarge.bbninebase.politeness.acc.diff\endcsname{\ensuremath{-}5.4}
\expandafter\def\csname cdnum@bb.kevlarge.bbninebase.politeness.acc.hi\endcsname{1.0}
\expandafter\def\csname cdnum@bb.kevlarge.bbninebase.politeness.acc.lo\endcsname{\ensuremath{-}11.6}
\expandafter\def\csname cdnum@bb.kevlarge.bbninebase.politeness.ece.diff\endcsname{2.0}
\expandafter\def\csname cdnum@bb.kevlarge.bbninebase.politeness.ece.hi\endcsname{8.4}
\expandafter\def\csname cdnum@bb.kevlarge.bbninebase.politeness.ece.lo\endcsname{\ensuremath{-}3.6}
\expandafter\def\csname cdnum@bb.kevlarge.bbninebase.power.acc.diff\endcsname{\ensuremath{-}1.4}
\expandafter\def\csname cdnum@bb.kevlarge.bbninebase.power.acc.hi\endcsname{2.8}
\expandafter\def\csname cdnum@bb.kevlarge.bbninebase.power.acc.lo\endcsname{\ensuremath{-}5.6}
\expandafter\def\csname cdnum@bb.kevlarge.bbninebase.power.ece.diff\endcsname{6.3}
\expandafter\def\csname cdnum@bb.kevlarge.bbninebase.power.ece.hi\endcsname{9.2}
\expandafter\def\csname cdnum@bb.kevlarge.bbninebase.power.ece.lo\endcsname{1.2}
\expandafter\def\csname cdnum@bb.kevlarge.bbninebase.powerkswap.flips.diff\endcsname{\ensuremath{-}60.0}
\expandafter\def\csname cdnum@bb.kevlarge.bbninebase.powerkswap.flips.hi\endcsname{\ensuremath{-}55.0}
\expandafter\def\csname cdnum@bb.kevlarge.bbninebase.powerkswap.flips.lo\endcsname{\ensuremath{-}64.8}
\expandafter\def\csname cdnum@bb.kevlarge.bbninebase.toxicity.acc.diff\endcsname{\ensuremath{-}4.0}
\expandafter\def\csname cdnum@bb.kevlarge.bbninebase.toxicity.acc.hi\endcsname{\ensuremath{-}0.4}
\expandafter\def\csname cdnum@bb.kevlarge.bbninebase.toxicity.acc.lo\endcsname{\ensuremath{-}7.6}
\expandafter\def\csname cdnum@bb.kevlarge.bbninebase.toxicity.ece.diff\endcsname{12.5}
\expandafter\def\csname cdnum@bb.kevlarge.bbninebase.toxicity.ece.hi\endcsname{15.7}
\expandafter\def\csname cdnum@bb.kevlarge.bbninebase.toxicity.ece.lo\endcsname{9.2}
\expandafter\def\csname cdnum@bb.kevlarge.bbninebase.toxicitykrand.flips.diff\endcsname{\ensuremath{-}7.0}
\expandafter\def\csname cdnum@bb.kevlarge.bbninebase.toxicitykrand.flips.hi\endcsname{\ensuremath{-}4.6}
\expandafter\def\csname cdnum@bb.kevlarge.bbninebase.toxicitykrand.flips.lo\endcsname{\ensuremath{-}9.4}
\expandafter\def\csname cdnum@bb.kevlarge.bbninebase.toxicitykswap.flips.diff\endcsname{\ensuremath{-}19.2}
\expandafter\def\csname cdnum@bb.kevlarge.bbninebase.toxicitykswap.flips.hi\endcsname{\ensuremath{-}15.6}
\expandafter\def\csname cdnum@bb.kevlarge.bbninebase.toxicitykswap.flips.lo\endcsname{\ensuremath{-}22.6}
\expandafter\def\csname cdnum@bb.kevsmall.bbsmallbase.donek01.flips.diff\endcsname{6.2}
\expandafter\def\csname cdnum@bb.kevsmall.bbsmallbase.donek01.flips.hi\endcsname{8.2}
\expandafter\def\csname cdnum@bb.kevsmall.bbsmallbase.donek01.flips.lo\endcsname{4.3}
\expandafter\def\csname cdnum@bb.kevsmall.bbsmallbase.donekrand.flips.diff\endcsname{16.3}
\expandafter\def\csname cdnum@bb.kevsmall.bbsmallbase.donekrand.flips.hi\endcsname{20.0}
\expandafter\def\csname cdnum@bb.kevsmall.bbsmallbase.donekrand.flips.lo\endcsname{12.9}
\expandafter\def\csname cdnum@bb.kevsmall.bbsmallbase.donekswap.flips.diff\endcsname{3.0}
\expandafter\def\csname cdnum@bb.kevsmall.bbsmallbase.donekswap.flips.hi\endcsname{6.8}
\expandafter\def\csname cdnum@bb.kevsmall.bbsmallbase.donekswap.flips.lo\endcsname{\ensuremath{-}0.7}
\expandafter\def\csname cdnum@bb.kevsmall.bbsmallbase.doneneutral.acc.diff\endcsname{0.1}
\expandafter\def\csname cdnum@bb.kevsmall.bbsmallbase.doneneutral.acc.hi\endcsname{3.1}
\expandafter\def\csname cdnum@bb.kevsmall.bbsmallbase.doneneutral.acc.lo\endcsname{\ensuremath{-}2.8}
\expandafter\def\csname cdnum@bb.kevsmall.bbsmallbase.doneneutral.ece.diff\endcsname{\ensuremath{-}17.0}
\expandafter\def\csname cdnum@bb.kevsmall.bbsmallbase.doneneutral.ece.hi\endcsname{\ensuremath{-}13.9}
\expandafter\def\csname cdnum@bb.kevsmall.bbsmallbase.doneneutral.ece.lo\endcsname{\ensuremath{-}19.5}
\expandafter\def\csname cdnum@bb.kevsmall.bbsmallbase.dtwo.acc.diff\endcsname{56.5}
\expandafter\def\csname cdnum@bb.kevsmall.bbsmallbase.dtwo.acc.hi\endcsname{60.7}
\expandafter\def\csname cdnum@bb.kevsmall.bbsmallbase.dtwo.acc.lo\endcsname{52.2}
\expandafter\def\csname cdnum@bb.kevsmall.bbsmallbase.dtwo.ece.diff\endcsname{20.6}
\expandafter\def\csname cdnum@bb.kevsmall.bbsmallbase.dtwo.ece.hi\endcsname{23.4}
\expandafter\def\csname cdnum@bb.kevsmall.bbsmallbase.dtwo.ece.lo\endcsname{16.4}
\expandafter\def\csname cdnum@bb.kevsmall.bbsmallbase.emotion.acc.diff\endcsname{8.2}
\expandafter\def\csname cdnum@bb.kevsmall.bbsmallbase.emotion.acc.hi\endcsname{12.0}
\expandafter\def\csname cdnum@bb.kevsmall.bbsmallbase.emotion.acc.lo\endcsname{4.4}
\expandafter\def\csname cdnum@bb.kevsmall.bbsmallbase.emotion.ece.diff\endcsname{\ensuremath{-}8.6}
\expandafter\def\csname cdnum@bb.kevsmall.bbsmallbase.emotion.ece.hi\endcsname{\ensuremath{-}2.5}
\expandafter\def\csname cdnum@bb.kevsmall.bbsmallbase.emotion.ece.lo\endcsname{\ensuremath{-}13.3}
\expandafter\def\csname cdnum@bb.kevsmall.bbsmallbase.politeness.acc.diff\endcsname{14.3}
\expandafter\def\csname cdnum@bb.kevsmall.bbsmallbase.politeness.acc.hi\endcsname{20.5}
\expandafter\def\csname cdnum@bb.kevsmall.bbsmallbase.politeness.acc.lo\endcsname{8.0}
\expandafter\def\csname cdnum@bb.kevsmall.bbsmallbase.politeness.ece.diff\endcsname{\ensuremath{-}38.6}
\expandafter\def\csname cdnum@bb.kevsmall.bbsmallbase.politeness.ece.hi\endcsname{\ensuremath{-}32.1}
\expandafter\def\csname cdnum@bb.kevsmall.bbsmallbase.politeness.ece.lo\endcsname{\ensuremath{-}43.0}
\expandafter\def\csname cdnum@bb.kevsmall.bbsmallbase.power.acc.diff\endcsname{2.0}
\expandafter\def\csname cdnum@bb.kevsmall.bbsmallbase.power.acc.hi\endcsname{10.2}
\expandafter\def\csname cdnum@bb.kevsmall.bbsmallbase.power.acc.lo\endcsname{\ensuremath{-}6.2}
\expandafter\def\csname cdnum@bb.kevsmall.bbsmallbase.power.ece.diff\endcsname{\ensuremath{-}6.5}
\expandafter\def\csname cdnum@bb.kevsmall.bbsmallbase.power.ece.hi\endcsname{0.9}
\expandafter\def\csname cdnum@bb.kevsmall.bbsmallbase.power.ece.lo\endcsname{\ensuremath{-}13.7}
\expandafter\def\csname cdnum@bb.kevsmall.bbsmallbase.powerkswap.flips.diff\endcsname{\ensuremath{-}44.6}
\expandafter\def\csname cdnum@bb.kevsmall.bbsmallbase.powerkswap.flips.hi\endcsname{\ensuremath{-}39.6}
\expandafter\def\csname cdnum@bb.kevsmall.bbsmallbase.powerkswap.flips.lo\endcsname{\ensuremath{-}49.6}
\expandafter\def\csname cdnum@bb.kevsmall.bbsmallbase.toxicity.acc.diff\endcsname{\ensuremath{-}4.2}
\expandafter\def\csname cdnum@bb.kevsmall.bbsmallbase.toxicity.acc.hi\endcsname{3.2}
\expandafter\def\csname cdnum@bb.kevsmall.bbsmallbase.toxicity.acc.lo\endcsname{\ensuremath{-}11.6}
\expandafter\def\csname cdnum@bb.kevsmall.bbsmallbase.toxicity.ece.diff\endcsname{12.0}
\expandafter\def\csname cdnum@bb.kevsmall.bbsmallbase.toxicity.ece.hi\endcsname{16.0}
\expandafter\def\csname cdnum@bb.kevsmall.bbsmallbase.toxicity.ece.lo\endcsname{6.1}
\expandafter\def\csname cdnum@bb.kevsmall.bbsmallbase.toxicitykrand.flips.diff\endcsname{\ensuremath{-}19.2}
\expandafter\def\csname cdnum@bb.kevsmall.bbsmallbase.toxicitykrand.flips.hi\endcsname{\ensuremath{-}15.6}
\expandafter\def\csname cdnum@bb.kevsmall.bbsmallbase.toxicitykrand.flips.lo\endcsname{\ensuremath{-}22.8}
\expandafter\def\csname cdnum@bb.kevsmall.bbsmallbase.toxicitykswap.flips.diff\endcsname{\ensuremath{-}96.2}
\expandafter\def\csname cdnum@bb.kevsmall.bbsmallbase.toxicitykswap.flips.hi\endcsname{\ensuremath{-}94.4}
\expandafter\def\csname cdnum@bb.kevsmall.bbsmallbase.toxicitykswap.flips.lo\endcsname{\ensuremath{-}97.8}
\expandafter\def\csname cdnum@bb.nimble.bbnine.donek01.flips.diff\endcsname{\ensuremath{-}9.3}
\expandafter\def\csname cdnum@bb.nimble.bbnine.donek01.flips.hi\endcsname{\ensuremath{-}6.8}
\expandafter\def\csname cdnum@bb.nimble.bbnine.donek01.flips.lo\endcsname{\ensuremath{-}11.9}
\expandafter\def\csname cdnum@bb.nimble.bbnine.donekrand.flips.diff\endcsname{\ensuremath{-}10.3}
\expandafter\def\csname cdnum@bb.nimble.bbnine.donekrand.flips.hi\endcsname{\ensuremath{-}7.2}
\expandafter\def\csname cdnum@bb.nimble.bbnine.donekrand.flips.lo\endcsname{\ensuremath{-}13.5}
\expandafter\def\csname cdnum@bb.nimble.bbnine.donekswap.flips.diff\endcsname{\ensuremath{-}30.2}
\expandafter\def\csname cdnum@bb.nimble.bbnine.donekswap.flips.hi\endcsname{\ensuremath{-}26.4}
\expandafter\def\csname cdnum@bb.nimble.bbnine.donekswap.flips.lo\endcsname{\ensuremath{-}33.7}
\expandafter\def\csname cdnum@bb.nimble.bbnine.doneneutral.acc.diff\endcsname{6.9}
\expandafter\def\csname cdnum@bb.nimble.bbnine.doneneutral.acc.hi\endcsname{9.4}
\expandafter\def\csname cdnum@bb.nimble.bbnine.doneneutral.acc.lo\endcsname{4.2}
\expandafter\def\csname cdnum@bb.nimble.bbnine.doneneutral.ece.diff\endcsname{\ensuremath{-}4.2}
\expandafter\def\csname cdnum@bb.nimble.bbnine.doneneutral.ece.hi\endcsname{\ensuremath{-}1.8}
\expandafter\def\csname cdnum@bb.nimble.bbnine.doneneutral.ece.lo\endcsname{\ensuremath{-}6.7}
\expandafter\def\csname cdnum@bb.nimble.bbnine.dtwo.acc.diff\endcsname{7.8}
\expandafter\def\csname cdnum@bb.nimble.bbnine.dtwo.acc.hi\endcsname{10.8}
\expandafter\def\csname cdnum@bb.nimble.bbnine.dtwo.acc.lo\endcsname{4.8}
\expandafter\def\csname cdnum@bb.nimble.bbnine.dtwo.ece.diff\endcsname{2.2}
\expandafter\def\csname cdnum@bb.nimble.bbnine.dtwo.ece.hi\endcsname{3.9}
\expandafter\def\csname cdnum@bb.nimble.bbnine.dtwo.ece.lo\endcsname{\ensuremath{-}1.3}
\expandafter\def\csname cdnum@bb.nimble.bbnine.emotion.acc.diff\endcsname{1.8}
\expandafter\def\csname cdnum@bb.nimble.bbnine.emotion.acc.hi\endcsname{4.6}
\expandafter\def\csname cdnum@bb.nimble.bbnine.emotion.acc.lo\endcsname{\ensuremath{-}1.0}
\expandafter\def\csname cdnum@bb.nimble.bbnine.emotion.ece.diff\endcsname{17.8}
\expandafter\def\csname cdnum@bb.nimble.bbnine.emotion.ece.hi\endcsname{20.1}
\expandafter\def\csname cdnum@bb.nimble.bbnine.emotion.ece.lo\endcsname{13.3}
\expandafter\def\csname cdnum@bb.nimble.bbnine.politeness.acc.diff\endcsname{\ensuremath{-}2.8}
\expandafter\def\csname cdnum@bb.nimble.bbnine.politeness.acc.hi\endcsname{0.8}
\expandafter\def\csname cdnum@bb.nimble.bbnine.politeness.acc.lo\endcsname{\ensuremath{-}6.4}
\expandafter\def\csname cdnum@bb.nimble.bbnine.politeness.ece.diff\endcsname{9.8}
\expandafter\def\csname cdnum@bb.nimble.bbnine.politeness.ece.hi\endcsname{13.1}
\expandafter\def\csname cdnum@bb.nimble.bbnine.politeness.ece.lo\endcsname{3.3}
\expandafter\def\csname cdnum@bb.nimble.bbnine.power.acc.diff\endcsname{\ensuremath{-}4.4}
\expandafter\def\csname cdnum@bb.nimble.bbnine.power.acc.hi\endcsname{\ensuremath{-}0.2}
\expandafter\def\csname cdnum@bb.nimble.bbnine.power.acc.lo\endcsname{\ensuremath{-}8.8}
\expandafter\def\csname cdnum@bb.nimble.bbnine.power.ece.diff\endcsname{21.7}
\expandafter\def\csname cdnum@bb.nimble.bbnine.power.ece.hi\endcsname{25.1}
\expandafter\def\csname cdnum@bb.nimble.bbnine.power.ece.lo\endcsname{17.5}
\expandafter\def\csname cdnum@bb.nimble.bbnine.powerkswap.flips.diff\endcsname{\ensuremath{-}43.2}
\expandafter\def\csname cdnum@bb.nimble.bbnine.powerkswap.flips.hi\endcsname{\ensuremath{-}38.6}
\expandafter\def\csname cdnum@bb.nimble.bbnine.powerkswap.flips.lo\endcsname{\ensuremath{-}47.8}
\expandafter\def\csname cdnum@bb.nimble.bbnine.toxicity.acc.diff\endcsname{\ensuremath{-}0.4}
\expandafter\def\csname cdnum@bb.nimble.bbnine.toxicity.acc.hi\endcsname{1.8}
\expandafter\def\csname cdnum@bb.nimble.bbnine.toxicity.acc.lo\endcsname{\ensuremath{-}2.8}
\expandafter\def\csname cdnum@bb.nimble.bbnine.toxicity.ece.diff\endcsname{17.5}
\expandafter\def\csname cdnum@bb.nimble.bbnine.toxicity.ece.hi\endcsname{19.5}
\expandafter\def\csname cdnum@bb.nimble.bbnine.toxicity.ece.lo\endcsname{14.3}
\expandafter\def\csname cdnum@bb.nimble.bbnine.toxicitykrand.flips.diff\endcsname{\ensuremath{-}9.8}
\expandafter\def\csname cdnum@bb.nimble.bbnine.toxicitykrand.flips.hi\endcsname{\ensuremath{-}6.8}
\expandafter\def\csname cdnum@bb.nimble.bbnine.toxicitykrand.flips.lo\endcsname{\ensuremath{-}13.0}
\expandafter\def\csname cdnum@bb.nimble.bbnine.toxicitykswap.flips.diff\endcsname{\ensuremath{-}5.4}
\expandafter\def\csname cdnum@bb.nimble.bbnine.toxicitykswap.flips.hi\endcsname{\ensuremath{-}3.0}
\expandafter\def\csname cdnum@bb.nimble.bbnine.toxicitykswap.flips.lo\endcsname{\ensuremath{-}8.0}
\expandafter\def\csname cdnum@bb.thisthat.bbtwobase.donek01.flips.diff\endcsname{0.7}
\expandafter\def\csname cdnum@bb.thisthat.bbtwobase.donek01.flips.hi\endcsname{2.4}
\expandafter\def\csname cdnum@bb.thisthat.bbtwobase.donek01.flips.lo\endcsname{\ensuremath{-}1.0}
\expandafter\def\csname cdnum@bb.thisthat.bbtwobase.donekrand.flips.diff\endcsname{0.8}
\expandafter\def\csname cdnum@bb.thisthat.bbtwobase.donekrand.flips.hi\endcsname{3.2}
\expandafter\def\csname cdnum@bb.thisthat.bbtwobase.donekrand.flips.lo\endcsname{\ensuremath{-}1.5}
\expandafter\def\csname cdnum@bb.thisthat.bbtwobase.donekswap.flips.diff\endcsname{\ensuremath{-}0.5}
\expandafter\def\csname cdnum@bb.thisthat.bbtwobase.donekswap.flips.hi\endcsname{1.9}
\expandafter\def\csname cdnum@bb.thisthat.bbtwobase.donekswap.flips.lo\endcsname{\ensuremath{-}3.0}
\expandafter\def\csname cdnum@bb.thisthat.bbtwobase.doneneutral.acc.diff\endcsname{18.3}
\expandafter\def\csname cdnum@bb.thisthat.bbtwobase.doneneutral.acc.hi\endcsname{21.4}
\expandafter\def\csname cdnum@bb.thisthat.bbtwobase.doneneutral.acc.lo\endcsname{15.2}
\expandafter\def\csname cdnum@bb.thisthat.bbtwobase.doneneutral.ece.diff\endcsname{\ensuremath{-}5.6}
\expandafter\def\csname cdnum@bb.thisthat.bbtwobase.doneneutral.ece.hi\endcsname{\ensuremath{-}3.0}
\expandafter\def\csname cdnum@bb.thisthat.bbtwobase.doneneutral.ece.lo\endcsname{\ensuremath{-}8.7}
\expandafter\def\csname cdnum@bb.thisthat.bbtwobase.dtwo.acc.diff\endcsname{35.2}
\expandafter\def\csname cdnum@bb.thisthat.bbtwobase.dtwo.acc.hi\endcsname{39.7}
\expandafter\def\csname cdnum@bb.thisthat.bbtwobase.dtwo.acc.lo\endcsname{30.8}
\expandafter\def\csname cdnum@bb.thisthat.bbtwobase.dtwo.ece.diff\endcsname{\ensuremath{-}18.7}
\expandafter\def\csname cdnum@bb.thisthat.bbtwobase.dtwo.ece.hi\endcsname{\ensuremath{-}14.3}
\expandafter\def\csname cdnum@bb.thisthat.bbtwobase.dtwo.ece.lo\endcsname{\ensuremath{-}22.8}
\expandafter\def\csname cdnum@bb.thisthat.bbtwobase.emotion.acc.diff\endcsname{37.1}
\expandafter\def\csname cdnum@bb.thisthat.bbtwobase.emotion.acc.hi\endcsname{42.0}
\expandafter\def\csname cdnum@bb.thisthat.bbtwobase.emotion.acc.lo\endcsname{32.3}
\expandafter\def\csname cdnum@bb.thisthat.bbtwobase.emotion.ece.diff\endcsname{1.3}
\expandafter\def\csname cdnum@bb.thisthat.bbtwobase.emotion.ece.hi\endcsname{4.5}
\expandafter\def\csname cdnum@bb.thisthat.bbtwobase.emotion.ece.lo\endcsname{\ensuremath{-}4.3}
\expandafter\def\csname cdnum@bb.thisthat.bbtwobase.politeness.acc.diff\endcsname{\ensuremath{-}0.2}
\expandafter\def\csname cdnum@bb.thisthat.bbtwobase.politeness.acc.hi\endcsname{6.8}
\expandafter\def\csname cdnum@bb.thisthat.bbtwobase.politeness.acc.lo\endcsname{\ensuremath{-}7.2}
\expandafter\def\csname cdnum@bb.thisthat.bbtwobase.politeness.ece.diff\endcsname{31.1}
\expandafter\def\csname cdnum@bb.thisthat.bbtwobase.politeness.ece.hi\endcsname{37.0}
\expandafter\def\csname cdnum@bb.thisthat.bbtwobase.politeness.ece.lo\endcsname{25.2}
\expandafter\def\csname cdnum@bb.thisthat.bbtwobase.power.acc.diff\endcsname{1.0}
\expandafter\def\csname cdnum@bb.thisthat.bbtwobase.power.acc.hi\endcsname{7.6}
\expandafter\def\csname cdnum@bb.thisthat.bbtwobase.power.acc.lo\endcsname{\ensuremath{-}5.4}
\expandafter\def\csname cdnum@bb.thisthat.bbtwobase.power.ece.diff\endcsname{18.8}
\expandafter\def\csname cdnum@bb.thisthat.bbtwobase.power.ece.hi\endcsname{23.5}
\expandafter\def\csname cdnum@bb.thisthat.bbtwobase.power.ece.lo\endcsname{11.9}
\expandafter\def\csname cdnum@bb.thisthat.bbtwobase.powerkswap.flips.diff\endcsname{\ensuremath{-}88.6}
\expandafter\def\csname cdnum@bb.thisthat.bbtwobase.powerkswap.flips.hi\endcsname{\ensuremath{-}85.6}
\expandafter\def\csname cdnum@bb.thisthat.bbtwobase.powerkswap.flips.lo\endcsname{\ensuremath{-}91.4}
\expandafter\def\csname cdnum@bb.thisthat.bbtwobase.toxicity.acc.diff\endcsname{2.0}
\expandafter\def\csname cdnum@bb.thisthat.bbtwobase.toxicity.acc.hi\endcsname{4.4}
\expandafter\def\csname cdnum@bb.thisthat.bbtwobase.toxicity.acc.lo\endcsname{\ensuremath{-}0.4}
\expandafter\def\csname cdnum@bb.thisthat.bbtwobase.toxicity.ece.diff\endcsname{26.3}
\expandafter\def\csname cdnum@bb.thisthat.bbtwobase.toxicity.ece.hi\endcsname{28.8}
\expandafter\def\csname cdnum@bb.thisthat.bbtwobase.toxicity.ece.lo\endcsname{23.9}
\expandafter\def\csname cdnum@bb.thisthat.bbtwobase.toxicitykrand.flips.diff\endcsname{\ensuremath{-}35.0}
\expandafter\def\csname cdnum@bb.thisthat.bbtwobase.toxicitykrand.flips.hi\endcsname{\ensuremath{-}30.8}
\expandafter\def\csname cdnum@bb.thisthat.bbtwobase.toxicitykrand.flips.lo\endcsname{\ensuremath{-}39.2}
\expandafter\def\csname cdnum@bb.thisthat.bbtwobase.toxicitykswap.flips.diff\endcsname{\ensuremath{-}99.8}
\expandafter\def\csname cdnum@bb.thisthat.bbtwobase.toxicitykswap.flips.hi\endcsname{\ensuremath{-}99.4}
\expandafter\def\csname cdnum@bb.thisthat.bbtwobase.toxicitykswap.flips.lo\endcsname{\ensuremath{-}100.0}
\expandafter\def\csname cdnum@bd.bbnine.dtwo.accdesc\endcsname{0.105}
\expandafter\def\csname cdnum@bd.bbnine.dtwo.acckey\endcsname{0.782}
\expandafter\def\csname cdnum@bl.bertbaseclinc.d2k150.acc\endcsname{0.940}
\expandafter\def\csname cdnum@bl.bertbaseclinc.d2k150.aurocoos\endcsname{0.975}
\expandafter\def\csname cdnum@bl.bertbaseclinc.d2k150.cov5\endcsname{0.98}
\expandafter\def\csname cdnum@bl.bertbaseclinc.d2k150.ecescaled\endcsname{0.029}
\expandafter\def\csname cdnum@bl.bertbaseclinc.d2k150.eceshipped\endcsname{0.028}
\expandafter\def\csname cdnum@bl.bertbaseclinc.d2k150.latencymsp50gpu\endcsname{6}
\expandafter\def\csname cdnum@bl.bgesmalllrclinc.d2k150.acc\endcsname{0.948}
\expandafter\def\csname cdnum@bl.bgesmalllrclinc.d2k150.aurocoos\endcsname{0.955}
\expandafter\def\csname cdnum@bl.bgesmalllrclinc.d2k150.cov5\endcsname{0.99}
\expandafter\def\csname cdnum@bl.bgesmalllrclinc.d2k150.ecescaled\endcsname{0.011}
\expandafter\def\csname cdnum@bl.bgesmalllrclinc.d2k150.eceshipped\endcsname{0.017}
\expandafter\def\csname cdnum@bl.bgesmalllrclinc.d2k150.latencymsp50gpu\endcsname{7}
\expandafter\def\csname cdnum@bl.bgesmalllrd1.d1neutral.acc\endcsname{0.702}
\expandafter\def\csname cdnum@bl.bgesmalllrd1.d1neutral.latencymsp50gpu\endcsname{9}
\expandafter\def\csname cdnum@bl.bgesmalllrd1.d1neutral.softacc\endcsname{0.563}
\expandafter\def\csname cdnum@bl.bgesmalllrd3.d3convgoawry.acc\endcsname{0.528}
\expandafter\def\csname cdnum@bl.bgesmalllrd3.d3emotion.acc\endcsname{0.675}
\expandafter\def\csname cdnum@bl.bgesmalllrd3.d3wikicorpus.acc\endcsname{0.592}
\expandafter\def\csname cdnum@bl.bgesmalllrd3.d3wikipoliteness.acc\endcsname{0.580}
\expandafter\def\csname cdnum@bl.bgesmalllrd3.dthree.macrofone\endcsname{59.4}
\expandafter\def\csname cdnum@bl.debertalargeclinc.d2k150.acc\endcsname{0.953}
\expandafter\def\csname cdnum@bl.debertalargeclinc.d2k150.aurocoos\endcsname{0.989}
\expandafter\def\csname cdnum@bl.debertalargeclinc.d2k150.cov5\endcsname{1.00}
\expandafter\def\csname cdnum@bl.debertalargeclinc.d2k150.ecescaled\endcsname{0.035}
\expandafter\def\csname cdnum@bl.debertalargeclinc.d2k150.eceshipped\endcsname{0.036}
\expandafter\def\csname cdnum@bl.debertalargeclinc.d2k150.latencymsp50l4\endcsname{58.0}
\expandafter\def\csname cdnum@bl.debertalarged1.d1neutral.acc\endcsname{0.793}
\expandafter\def\csname cdnum@bl.debertalarged1.d1neutral.latencymsp50l4\endcsname{126.5}
\expandafter\def\csname cdnum@bl.debertalarged1.d1neutral.softacc\endcsname{0.616}
\expandafter\def\csname cdnum@bl.debertalarged3.d3convgoawry.acc\endcsname{0.592}
\expandafter\def\csname cdnum@bl.debertalarged3.d3emotion.acc\endcsname{0.938}
\expandafter\def\csname cdnum@bl.debertalarged3.d3wikicorpus.acc\endcsname{0.586}
\expandafter\def\csname cdnum@bl.debertalarged3.d3wikipoliteness.acc\endcsname{0.729}
\expandafter\def\csname cdnum@bl.debertalarged3.dthree.macrofone\endcsname{70.9}
\expandafter\def\csname cdnum@bl.nlidebertav3base.d1neutral.acc\endcsname{0.411}
\expandafter\def\csname cdnum@bl.nlidebertav3base.d1neutral.softacc\endcsname{0.383}
\expandafter\def\csname cdnum@bl.nlidebertav3base.d2k150.acc\endcsname{0.507}
\expandafter\def\csname cdnum@bl.nlidebertav3base.d2k150.aurocoos\endcsname{0.619}
\expandafter\def\csname cdnum@bl.nlidebertav3base.d2k150.cov5\endcsname{0.21}
\expandafter\def\csname cdnum@bl.nlidebertav3base.d2k150.ecescaled\endcsname{0.133}
\expandafter\def\csname cdnum@bl.nlidebertav3base.d2k150.eceshipped\endcsname{0.094}
\expandafter\def\csname cdnum@bl.nlidebertav3base.d2k150.latencymsp50gpu\endcsname{3140}
\expandafter\def\csname cdnum@bl.nlidebertav3base.d3convgoawry.acc\endcsname{0.514}
\expandafter\def\csname cdnum@bl.nlidebertav3base.d3emotion.acc\endcsname{0.651}
\expandafter\def\csname cdnum@bl.nlidebertav3base.d3wikicorpus.acc\endcsname{0.514}
\expandafter\def\csname cdnum@bl.nlidebertav3base.d3wikipoliteness.acc\endcsname{0.400}
\expandafter\def\csname cdnum@bl.nlidebertav3base.dthree.macrofone\endcsname{48.0}
\expandafter\def\csname cdnum@bp.doneneutral.trained.compgem.diff\endcsname{3.9}
\expandafter\def\csname cdnum@bp.dtwo.trainedlarge.compgem.diff\endcsname{\ensuremath{-}1.2}
\expandafter\def\csname cdnum@bp.dtwo.trainedlarge.compgem.holm\endcsname{0.5}
\expandafter\def\csname cdnum@d.banking.k\endcsname{77}
\expandafter\def\csname cdnum@d.banking.license\endcsname{CC BY 4.0}
\expandafter\def\csname cdnum@d.d1.license\endcsname{Apache-2.0}
\expandafter\def\csname cdnum@d.d2.license\endcsname{CC BY 3.0}
\expandafter\def\csname cdnum@d.d3convgoawry.k\endcsname{2}
\expandafter\def\csname cdnum@d.d3convgoawry.license\endcsname{unstated}
\expandafter\def\csname cdnum@d.d3emotion.k\endcsname{6}
\expandafter\def\csname cdnum@d.d3emotion.license\endcsname{research use only}
\expandafter\def\csname cdnum@d.d3wikicorpus.k\endcsname{2}
\expandafter\def\csname cdnum@d.d3wikicorpus.license\endcsname{CC BY-SA 4.0}
\expandafter\def\csname cdnum@d.d3wikipoliteness.k\endcsname{3}
\expandafter\def\csname cdnum@d.d3wikipoliteness.license\endcsname{CC BY 4.0}
\expandafter\def\csname cdnum@d.goemo.k\endcsname{6}
\expandafter\def\csname cdnum@d.goemo.license\endcsname{Apache-2.0}
\expandafter\def\csname cdnum@e1.compgem.doneneutral.acc\endcsname{0.741}
\expandafter\def\csname cdnum@e1.compgem.doneneutral.brier\endcsname{0.411}
\expandafter\def\csname cdnum@e1.compgem.doneneutral.covfive\endcsname{0.12}
\expandafter\def\csname cdnum@e1.compgem.doneneutral.ece\endcsname{0.106}
\expandafter\def\csname cdnum@e1.compgem.doneneutral.ecescaled\endcsname{0.043}
\expandafter\def\csname cdnum@e1.compgem.doneneutral.nll\endcsname{0.761}
\expandafter\def\csname cdnum@e1.compgem.doneneutral.softacc\endcsname{0.592}
\expandafter\def\csname cdnum@e1.compgem.dthree.covfive\endcsname{0.00}
\expandafter\def\csname cdnum@e1.compgem.dthree.ece\endcsname{0.258}
\expandafter\def\csname cdnum@e1.compgem.dthree.ecescaled\endcsname{0.033}
\expandafter\def\csname cdnum@e1.compgem.dthree.macrofone\endcsname{55.8}
\expandafter\def\csname cdnum@e1.compgem.dtwo.acc\endcsname{0.942}
\expandafter\def\csname cdnum@e1.compgem.dtwo.brier\endcsname{0.104}
\expandafter\def\csname cdnum@e1.compgem.dtwo.covfive\endcsname{0.97}
\expandafter\def\csname cdnum@e1.compgem.dtwo.ece\endcsname{0.022}
\expandafter\def\csname cdnum@e1.compgem.dtwo.ecescaled\endcsname{0.026}
\expandafter\def\csname cdnum@e1.compgem.dtwo.nll\endcsname{0.325}
\expandafter\def\csname cdnum@e1.compgem.emotion.acc\endcsname{0.474}
\expandafter\def\csname cdnum@e1.compgem.emotion.brier\endcsname{0.785}
\expandafter\def\csname cdnum@e1.compgem.emotion.nll\endcsname{1.856}
\expandafter\def\csname cdnum@e1.compgem.oos.auroc\endcsname{0.925}
\expandafter\def\csname cdnum@e1.compgem.politeness.acc\endcsname{0.661}
\expandafter\def\csname cdnum@e1.compgem.politeness.brier\endcsname{0.506}
\expandafter\def\csname cdnum@e1.compgem.politeness.nll\endcsname{0.851}
\expandafter\def\csname cdnum@e1.compgem.power.acc\endcsname{0.614}
\expandafter\def\csname cdnum@e1.compgem.power.brier\endcsname{0.666}
\expandafter\def\csname cdnum@e1.compgem.power.nll\endcsname{1.129}
\expandafter\def\csname cdnum@e1.compgem.toxicity.acc\endcsname{0.588}
\expandafter\def\csname cdnum@e1.compgem.toxicity.brier\endcsname{0.666}
\expandafter\def\csname cdnum@e1.compgem.toxicity.nll\endcsname{1.085}
\expandafter\def\csname cdnum@e1.compgemll.doneneutral.acc\endcsname{0.742}
\expandafter\def\csname cdnum@e1.compgemll.doneneutral.brier\endcsname{0.495}
\expandafter\def\csname cdnum@e1.compgemll.doneneutral.covfive\endcsname{0.28}
\expandafter\def\csname cdnum@e1.compgemll.doneneutral.ece\endcsname{0.244}
\expandafter\def\csname cdnum@e1.compgemll.doneneutral.eceraw\endcsname{0.244}
\expandafter\def\csname cdnum@e1.compgemll.doneneutral.ecescaled\endcsname{0.064}
\expandafter\def\csname cdnum@e1.compgemll.doneneutral.nll\endcsname{2.220}
\expandafter\def\csname cdnum@e1.compgemll.doneneutral.softacc\endcsname{0.592}
\expandafter\def\csname cdnum@e1.compgemll.dthree.covfive\endcsname{0.00}
\expandafter\def\csname cdnum@e1.compgemll.dthree.ece\endcsname{0.409}
\expandafter\def\csname cdnum@e1.compgemll.dthree.eceraw\endcsname{0.409}
\expandafter\def\csname cdnum@e1.compgemll.dthree.ecescaled\endcsname{0.071}
\expandafter\def\csname cdnum@e1.compgemll.dthree.macrofone\endcsname{55.7}
\expandafter\def\csname cdnum@e1.compgemll.dtwo.acc\endcsname{0.940}
\expandafter\def\csname cdnum@e1.compgemll.dtwo.brier\endcsname{0.115}
\expandafter\def\csname cdnum@e1.compgemll.dtwo.covfive\endcsname{0.98}
\expandafter\def\csname cdnum@e1.compgemll.dtwo.ece\endcsname{0.058}
\expandafter\def\csname cdnum@e1.compgemll.dtwo.eceraw\endcsname{0.058}
\expandafter\def\csname cdnum@e1.compgemll.dtwo.ecescaled\endcsname{0.034}
\expandafter\def\csname cdnum@e1.compgemll.dtwo.nll\endcsname{0.606}
\expandafter\def\csname cdnum@e1.compgemll.emotion.acc\endcsname{0.474}
\expandafter\def\csname cdnum@e1.compgemll.emotion.brier\endcsname{1.032}
\expandafter\def\csname cdnum@e1.compgemll.emotion.nll\endcsname{6.278}
\expandafter\def\csname cdnum@e1.compgemll.oos.auroc\endcsname{0.902}
\expandafter\def\csname cdnum@e1.compgemll.politeness.acc\endcsname{0.661}
\expandafter\def\csname cdnum@e1.compgemll.politeness.brier\endcsname{0.666}
\expandafter\def\csname cdnum@e1.compgemll.politeness.covfive\endcsname{0.17}
\expandafter\def\csname cdnum@e1.compgemll.politeness.nll\endcsname{3.308}
\expandafter\def\csname cdnum@e1.compgemll.power.acc\endcsname{0.612}
\expandafter\def\csname cdnum@e1.compgemll.power.brier\endcsname{0.768}
\expandafter\def\csname cdnum@e1.compgemll.power.nll\endcsname{4.935}
\expandafter\def\csname cdnum@e1.compgemll.toxicity.acc\endcsname{0.588}
\expandafter\def\csname cdnum@e1.compgemll.toxicity.brier\endcsname{0.817}
\expandafter\def\csname cdnum@e1.compgemll.toxicity.nll\endcsname{5.203}
\expandafter\def\csname cdnum@e1.compmis.doneneutral.acc\endcsname{0.606}
\expandafter\def\csname cdnum@e1.compmis.doneneutral.brier\endcsname{0.552}
\expandafter\def\csname cdnum@e1.compmis.doneneutral.covfive\endcsname{0.00}
\expandafter\def\csname cdnum@e1.compmis.doneneutral.ece\endcsname{0.151}
\expandafter\def\csname cdnum@e1.compmis.doneneutral.ecescaled\endcsname{0.065}
\expandafter\def\csname cdnum@e1.compmis.doneneutral.nll\endcsname{1.010}
\expandafter\def\csname cdnum@e1.compmis.doneneutral.softacc\endcsname{0.510}
\expandafter\def\csname cdnum@e1.compmis.dthree.covfive\endcsname{0.00}
\expandafter\def\csname cdnum@e1.compmis.dthree.ece\endcsname{0.203}
\expandafter\def\csname cdnum@e1.compmis.dthree.ecescaled\endcsname{0.032}
\expandafter\def\csname cdnum@e1.compmis.dthree.macrofone\endcsname{47.0}
\expandafter\def\csname cdnum@e1.compmis.dtwo.acc\endcsname{0.757}
\expandafter\def\csname cdnum@e1.compmis.dtwo.brier\endcsname{0.419}
\expandafter\def\csname cdnum@e1.compmis.dtwo.covfive\endcsname{0.00}
\expandafter\def\csname cdnum@e1.compmis.dtwo.ece\endcsname{0.135}
\expandafter\def\csname cdnum@e1.compmis.dtwo.ecescaled\endcsname{0.186}
\expandafter\def\csname cdnum@e1.compmis.dtwo.nll\endcsname{2.030}
\expandafter\def\csname cdnum@e1.compmis.emotion.acc\endcsname{0.243}
\expandafter\def\csname cdnum@e1.compmis.emotion.brier\endcsname{0.802}
\expandafter\def\csname cdnum@e1.compmis.emotion.nll\endcsname{1.718}
\expandafter\def\csname cdnum@e1.compmis.oos.auroc\endcsname{0.724}
\expandafter\def\csname cdnum@e1.compmis.politeness.acc\endcsname{0.651}
\expandafter\def\csname cdnum@e1.compmis.politeness.brier\endcsname{0.566}
\expandafter\def\csname cdnum@e1.compmis.politeness.nll\endcsname{0.996}
\expandafter\def\csname cdnum@e1.compmis.power.acc\endcsname{0.558}
\expandafter\def\csname cdnum@e1.compmis.power.brier\endcsname{0.709}
\expandafter\def\csname cdnum@e1.compmis.power.nll\endcsname{1.045}
\expandafter\def\csname cdnum@e1.compmis.toxicity.acc\endcsname{0.586}
\expandafter\def\csname cdnum@e1.compmis.toxicity.brier\endcsname{0.668}
\expandafter\def\csname cdnum@e1.compmis.toxicity.nll\endcsname{0.982}
\expandafter\def\csname cdnum@e1.compmisll.doneneutral.acc\endcsname{0.655}
\expandafter\def\csname cdnum@e1.compmisll.doneneutral.brier\endcsname{0.473}
\expandafter\def\csname cdnum@e1.compmisll.doneneutral.covfive\endcsname{0.08}
\expandafter\def\csname cdnum@e1.compmisll.doneneutral.ece\endcsname{0.112}
\expandafter\def\csname cdnum@e1.compmisll.doneneutral.eceraw\endcsname{0.112}
\expandafter\def\csname cdnum@e1.compmisll.doneneutral.ecescaled\endcsname{0.045}
\expandafter\def\csname cdnum@e1.compmisll.doneneutral.nll\endcsname{0.832}
\expandafter\def\csname cdnum@e1.compmisll.doneneutral.softacc\endcsname{0.547}
\expandafter\def\csname cdnum@e1.compmisll.dthree.covfive\endcsname{0.01}
\expandafter\def\csname cdnum@e1.compmisll.dthree.ece\endcsname{0.292}
\expandafter\def\csname cdnum@e1.compmisll.dthree.eceraw\endcsname{0.292}
\expandafter\def\csname cdnum@e1.compmisll.dthree.ecescaled\endcsname{0.026}
\expandafter\def\csname cdnum@e1.compmisll.dthree.macrofone\endcsname{54.3}
\expandafter\def\csname cdnum@e1.compmisll.dtwo.acc\endcsname{0.782}
\expandafter\def\csname cdnum@e1.compmisll.dtwo.brier\endcsname{0.320}
\expandafter\def\csname cdnum@e1.compmisll.dtwo.covfive\endcsname{0.41}
\expandafter\def\csname cdnum@e1.compmisll.dtwo.ece\endcsname{0.055}
\expandafter\def\csname cdnum@e1.compmisll.dtwo.eceraw\endcsname{0.055}
\expandafter\def\csname cdnum@e1.compmisll.dtwo.ecescaled\endcsname{0.053}
\expandafter\def\csname cdnum@e1.compmisll.dtwo.nll\endcsname{0.850}
\expandafter\def\csname cdnum@e1.compmisll.emotion.acc\endcsname{0.454}
\expandafter\def\csname cdnum@e1.compmisll.emotion.brier\endcsname{0.876}
\expandafter\def\csname cdnum@e1.compmisll.emotion.nll\endcsname{2.224}
\expandafter\def\csname cdnum@e1.compmisll.oos.auroc\endcsname{0.876}
\expandafter\def\csname cdnum@e1.compmisll.politeness.acc\endcsname{0.689}
\expandafter\def\csname cdnum@e1.compmisll.politeness.brier\endcsname{0.446}
\expandafter\def\csname cdnum@e1.compmisll.politeness.nll\endcsname{0.772}
\expandafter\def\csname cdnum@e1.compmisll.power.acc\endcsname{0.580}
\expandafter\def\csname cdnum@e1.compmisll.power.brier\endcsname{0.762}
\expandafter\def\csname cdnum@e1.compmisll.power.nll\endcsname{1.680}
\expandafter\def\csname cdnum@e1.compmisll.toxicity.acc\endcsname{0.574}
\expandafter\def\csname cdnum@e1.compmisll.toxicity.brier\endcsname{0.746}
\expandafter\def\csname cdnum@e1.compmisll.toxicity.nll\endcsname{1.677}
\expandafter\def\csname cdnum@e1.compopen.doneneutral.acc\endcsname{0.680}
\expandafter\def\csname cdnum@e1.compopen.doneneutral.brier\endcsname{0.493}
\expandafter\def\csname cdnum@e1.compopen.doneneutral.covfive\endcsname{0.00}
\expandafter\def\csname cdnum@e1.compopen.doneneutral.ece\endcsname{0.117}
\expandafter\def\csname cdnum@e1.compopen.doneneutral.ecescaled\endcsname{0.091}
\expandafter\def\csname cdnum@e1.compopen.doneneutral.nll\endcsname{1.021}
\expandafter\def\csname cdnum@e1.compopen.doneneutral.softacc\endcsname{0.555}
\expandafter\def\csname cdnum@e1.compopen.dthree.covfive\endcsname{0.00}
\expandafter\def\csname cdnum@e1.compopen.dthree.ece\endcsname{0.197}
\expandafter\def\csname cdnum@e1.compopen.dthree.ecescaled\endcsname{0.032}
\expandafter\def\csname cdnum@e1.compopen.dthree.macrofone\endcsname{52.7}
\expandafter\def\csname cdnum@e1.compopen.dtwo.acc\endcsname{0.728}
\expandafter\def\csname cdnum@e1.compopen.dtwo.brier\endcsname{0.494}
\expandafter\def\csname cdnum@e1.compopen.dtwo.covfive\endcsname{0.01}
\expandafter\def\csname cdnum@e1.compopen.dtwo.ece\endcsname{0.204}
\expandafter\def\csname cdnum@e1.compopen.dtwo.ecescaled\endcsname{0.096}
\expandafter\def\csname cdnum@e1.compopen.dtwo.nll\endcsname{2.234}
\expandafter\def\csname cdnum@e1.compopen.emotion.acc\endcsname{0.464}
\expandafter\def\csname cdnum@e1.compopen.emotion.brier\endcsname{0.738}
\expandafter\def\csname cdnum@e1.compopen.emotion.nll\endcsname{1.576}
\expandafter\def\csname cdnum@e1.compopen.oos.auroc\endcsname{0.600}
\expandafter\def\csname cdnum@e1.compopen.politeness.acc\endcsname{0.653}
\expandafter\def\csname cdnum@e1.compopen.politeness.brier\endcsname{0.524}
\expandafter\def\csname cdnum@e1.compopen.politeness.nll\endcsname{0.938}
\expandafter\def\csname cdnum@e1.compopen.power.acc\endcsname{0.576}
\expandafter\def\csname cdnum@e1.compopen.power.brier\endcsname{0.638}
\expandafter\def\csname cdnum@e1.compopen.power.nll\endcsname{0.914}
\expandafter\def\csname cdnum@e1.compopen.toxicity.acc\endcsname{0.554}
\expandafter\def\csname cdnum@e1.compopen.toxicity.brier\endcsname{0.667}
\expandafter\def\csname cdnum@e1.compopen.toxicity.nll\endcsname{0.945}
\expandafter\def\csname cdnum@e1.comptwo.doneneutral.acc\endcsname{0.725}
\expandafter\def\csname cdnum@e1.comptwo.doneneutral.brier\endcsname{0.445}
\expandafter\def\csname cdnum@e1.comptwo.doneneutral.covfive\endcsname{0.04}
\expandafter\def\csname cdnum@e1.comptwo.doneneutral.ece\endcsname{0.126}
\expandafter\def\csname cdnum@e1.comptwo.doneneutral.ecescaled\endcsname{0.063}
\expandafter\def\csname cdnum@e1.comptwo.doneneutral.nll\endcsname{0.841}
\expandafter\def\csname cdnum@e1.comptwo.doneneutral.softacc\endcsname{0.584}
\expandafter\def\csname cdnum@e1.comptwo.dthree.covfive\endcsname{0.00}
\expandafter\def\csname cdnum@e1.comptwo.dthree.ece\endcsname{0.293}
\expandafter\def\csname cdnum@e1.comptwo.dthree.ecescaled\endcsname{0.029}
\expandafter\def\csname cdnum@e1.comptwo.dthree.macrofone\endcsname{50.2}
\expandafter\def\csname cdnum@e1.comptwo.dtwo.acc\endcsname{0.683}
\expandafter\def\csname cdnum@e1.comptwo.dtwo.brier\endcsname{0.356}
\expandafter\def\csname cdnum@e1.comptwo.dtwo.covfive\endcsname{0.69}
\expandafter\def\csname cdnum@e1.comptwo.dtwo.ece\endcsname{0.023}
\expandafter\def\csname cdnum@e1.comptwo.dtwo.ecescaled\endcsname{0.027}
\expandafter\def\csname cdnum@e1.comptwo.dtwo.nll\endcsname{1.633}
\expandafter\def\csname cdnum@e1.comptwo.emotion.acc\endcsname{0.398}
\expandafter\def\csname cdnum@e1.comptwo.emotion.brier\endcsname{0.828}
\expandafter\def\csname cdnum@e1.comptwo.emotion.nll\endcsname{1.792}
\expandafter\def\csname cdnum@e1.comptwo.oos.auroc\endcsname{0.649}
\expandafter\def\csname cdnum@e1.comptwo.politeness.acc\endcsname{0.655}
\expandafter\def\csname cdnum@e1.comptwo.politeness.brier\endcsname{0.559}
\expandafter\def\csname cdnum@e1.comptwo.politeness.nll\endcsname{0.947}
\expandafter\def\csname cdnum@e1.comptwo.power.acc\endcsname{0.616}
\expandafter\def\csname cdnum@e1.comptwo.power.brier\endcsname{0.639}
\expandafter\def\csname cdnum@e1.comptwo.power.nll\endcsname{0.966}
\expandafter\def\csname cdnum@e1.comptwo.toxicity.acc\endcsname{0.534}
\expandafter\def\csname cdnum@e1.comptwo.toxicity.brier\endcsname{0.793}
\expandafter\def\csname cdnum@e1.comptwo.toxicity.nll\endcsname{1.255}
\expandafter\def\csname cdnum@e1.comptwoll.doneneutral.acc\endcsname{0.733}
\expandafter\def\csname cdnum@e1.comptwoll.doneneutral.brier\endcsname{0.355}
\expandafter\def\csname cdnum@e1.comptwoll.doneneutral.covfive\endcsname{0.42}
\expandafter\def\csname cdnum@e1.comptwoll.doneneutral.ece\endcsname{0.061}
\expandafter\def\csname cdnum@e1.comptwoll.doneneutral.eceraw\endcsname{0.061}
\expandafter\def\csname cdnum@e1.comptwoll.doneneutral.ecescaled\endcsname{0.029}
\expandafter\def\csname cdnum@e1.comptwoll.doneneutral.nll\endcsname{0.606}
\expandafter\def\csname cdnum@e1.comptwoll.doneneutral.softacc\endcsname{0.587}
\expandafter\def\csname cdnum@e1.comptwoll.dthree.covfive\endcsname{0.00}
\expandafter\def\csname cdnum@e1.comptwoll.dthree.ece\endcsname{0.312}
\expandafter\def\csname cdnum@e1.comptwoll.dthree.eceraw\endcsname{0.312}
\expandafter\def\csname cdnum@e1.comptwoll.dthree.ecescaled\endcsname{0.035}
\expandafter\def\csname cdnum@e1.comptwoll.dthree.macrofone\endcsname{52.3}
\expandafter\def\csname cdnum@e1.comptwoll.dtwo.acc\endcsname{0.912}
\expandafter\def\csname cdnum@e1.comptwoll.dtwo.brier\endcsname{0.155}
\expandafter\def\csname cdnum@e1.comptwoll.dtwo.covfive\endcsname{0.86}
\expandafter\def\csname cdnum@e1.comptwoll.dtwo.ece\endcsname{0.044}
\expandafter\def\csname cdnum@e1.comptwoll.dtwo.eceraw\endcsname{0.044}
\expandafter\def\csname cdnum@e1.comptwoll.dtwo.ecescaled\endcsname{0.035}
\expandafter\def\csname cdnum@e1.comptwoll.dtwo.nll\endcsname{0.403}
\expandafter\def\csname cdnum@e1.comptwoll.emotion.acc\endcsname{0.464}
\expandafter\def\csname cdnum@e1.comptwoll.emotion.brier\endcsname{0.870}
\expandafter\def\csname cdnum@e1.comptwoll.emotion.nll\endcsname{2.405}
\expandafter\def\csname cdnum@e1.comptwoll.oos.auroc\endcsname{0.932}
\expandafter\def\csname cdnum@e1.comptwoll.politeness.acc\endcsname{0.659}
\expandafter\def\csname cdnum@e1.comptwoll.politeness.brier\endcsname{0.500}
\expandafter\def\csname cdnum@e1.comptwoll.politeness.nll\endcsname{0.935}
\expandafter\def\csname cdnum@e1.comptwoll.power.acc\endcsname{0.618}
\expandafter\def\csname cdnum@e1.comptwoll.power.brier\endcsname{0.657}
\expandafter\def\csname cdnum@e1.comptwoll.power.nll\endcsname{1.376}
\expandafter\def\csname cdnum@e1.comptwoll.toxicity.acc\endcsname{0.534}
\expandafter\def\csname cdnum@e1.comptwoll.toxicity.brier\endcsname{0.859}
\expandafter\def\csname cdnum@e1.comptwoll.toxicity.nll\endcsname{2.498}
\expandafter\def\csname cdnum@e1.comptwothink.doneneutral.acc\endcsname{0.721}
\expandafter\def\csname cdnum@e1.comptwothink.doneneutral.brier\endcsname{0.433}
\expandafter\def\csname cdnum@e1.comptwothink.doneneutral.covfive\endcsname{0.10}
\expandafter\def\csname cdnum@e1.comptwothink.doneneutral.ece\endcsname{0.129}
\expandafter\def\csname cdnum@e1.comptwothink.doneneutral.ecescaled\endcsname{0.045}
\expandafter\def\csname cdnum@e1.comptwothink.doneneutral.nll\endcsname{0.774}
\expandafter\def\csname cdnum@e1.comptwothink.doneneutral.softacc\endcsname{0.583}
\expandafter\def\csname cdnum@e1.comptwothink.dthree.covfive\endcsname{0.00}
\expandafter\def\csname cdnum@e1.comptwothink.dthree.ece\endcsname{0.271}
\expandafter\def\csname cdnum@e1.comptwothink.dthree.ecescaled\endcsname{0.026}
\expandafter\def\csname cdnum@e1.comptwothink.dthree.macrofone\endcsname{55.5}
\expandafter\def\csname cdnum@e1.comptwothink.dtwo.acc\endcsname{0.928}
\expandafter\def\csname cdnum@e1.comptwothink.dtwo.brier\endcsname{0.127}
\expandafter\def\csname cdnum@e1.comptwothink.dtwo.covfive\endcsname{0.85}
\expandafter\def\csname cdnum@e1.comptwothink.dtwo.ece\endcsname{0.038}
\expandafter\def\csname cdnum@e1.comptwothink.dtwo.ecescaled\endcsname{0.032}
\expandafter\def\csname cdnum@e1.comptwothink.dtwo.nll\endcsname{0.389}
\expandafter\def\csname cdnum@e1.comptwothink.emotion.acc\endcsname{0.518}
\expandafter\def\csname cdnum@e1.comptwothink.emotion.brier\endcsname{0.731}
\expandafter\def\csname cdnum@e1.comptwothink.emotion.nll\endcsname{1.528}
\expandafter\def\csname cdnum@e1.comptwothink.oos.auroc\endcsname{0.879}
\expandafter\def\csname cdnum@e1.comptwothink.politeness.acc\endcsname{0.645}
\expandafter\def\csname cdnum@e1.comptwothink.politeness.brier\endcsname{0.544}
\expandafter\def\csname cdnum@e1.comptwothink.politeness.nll\endcsname{0.942}
\expandafter\def\csname cdnum@e1.comptwothink.power.acc\endcsname{0.634}
\expandafter\def\csname cdnum@e1.comptwothink.power.brier\endcsname{0.626}
\expandafter\def\csname cdnum@e1.comptwothink.power.nll\endcsname{1.002}
\expandafter\def\csname cdnum@e1.comptwothink.toxicity.acc\endcsname{0.542}
\expandafter\def\csname cdnum@e1.comptwothink.toxicity.brier\endcsname{0.725}
\expandafter\def\csname cdnum@e1.comptwothink.toxicity.nll\endcsname{1.116}
\expandafter\def\csname cdnum@e1.decider.doneneutral.acc\endcsname{0.588}
\expandafter\def\csname cdnum@e1.decider.doneneutral.brier\endcsname{0.555}
\expandafter\def\csname cdnum@e1.decider.doneneutral.covfive\endcsname{0.06}
\expandafter\def\csname cdnum@e1.decider.doneneutral.ece\endcsname{0.094}
\expandafter\def\csname cdnum@e1.decider.doneneutral.eceraw\endcsname{0.139}
\expandafter\def\csname cdnum@e1.decider.doneneutral.ecescaled\endcsname{0.036}
\expandafter\def\csname cdnum@e1.decider.doneneutral.nll\endcsname{0.967}
\expandafter\def\csname cdnum@e1.decider.doneneutral.softacc\endcsname{0.504}
\expandafter\def\csname cdnum@e1.decider.dthree.covfive\endcsname{0.00}
\expandafter\def\csname cdnum@e1.decider.dthree.ece\endcsname{0.232}
\expandafter\def\csname cdnum@e1.decider.dthree.eceraw\endcsname{0.272}
\expandafter\def\csname cdnum@e1.decider.dthree.ecescaled\endcsname{0.032}
\expandafter\def\csname cdnum@e1.decider.dthree.macrofone\endcsname{51.5}
\expandafter\def\csname cdnum@e1.decider.dtwo.acc\endcsname{0.860}
\expandafter\def\csname cdnum@e1.decider.dtwo.brier\endcsname{0.209}
\expandafter\def\csname cdnum@e1.decider.dtwo.covfive\endcsname{0.75}
\expandafter\def\csname cdnum@e1.decider.dtwo.ece\endcsname{0.035}
\expandafter\def\csname cdnum@e1.decider.dtwo.eceraw\endcsname{0.035}
\expandafter\def\csname cdnum@e1.decider.dtwo.ecescaled\endcsname{0.025}
\expandafter\def\csname cdnum@e1.decider.dtwo.nll\endcsname{0.571}
\expandafter\def\csname cdnum@e1.decider.emotion.acc\endcsname{0.803}
\expandafter\def\csname cdnum@e1.decider.emotion.brier\endcsname{0.286}
\expandafter\def\csname cdnum@e1.decider.emotion.nll\endcsname{0.543}
\expandafter\def\csname cdnum@e1.decider.oos.auroc\endcsname{0.885}
\expandafter\def\csname cdnum@e1.decider.politeness.acc\endcsname{0.464}
\expandafter\def\csname cdnum@e1.decider.politeness.brier\endcsname{0.707}
\expandafter\def\csname cdnum@e1.decider.politeness.nll\endcsname{1.217}
\expandafter\def\csname cdnum@e1.decider.power.acc\endcsname{0.534}
\expandafter\def\csname cdnum@e1.decider.power.brier\endcsname{0.606}
\expandafter\def\csname cdnum@e1.decider.power.nll\endcsname{0.856}
\expandafter\def\csname cdnum@e1.decider.toxicity.acc\endcsname{0.514}
\expandafter\def\csname cdnum@e1.decider.toxicity.brier\endcsname{0.894}
\expandafter\def\csname cdnum@e1.decider.toxicity.nll\endcsname{1.851}
\expandafter\def\csname cdnum@e1.jev.donenative.acc\endcsname{0.736}
\expandafter\def\csname cdnum@e1.jev.donenative.softacc\endcsname{0.582}
\expandafter\def\csname cdnum@e1.jev.doneneutral.acc\endcsname{0.732}
\expandafter\def\csname cdnum@e1.jev.doneneutral.brier\endcsname{0.371}
\expandafter\def\csname cdnum@e1.jev.doneneutral.covfive\endcsname{0.13}
\expandafter\def\csname cdnum@e1.jev.doneneutral.ece\endcsname{0.033}
\expandafter\def\csname cdnum@e1.jev.doneneutral.ecescaled\endcsname{0.032}
\expandafter\def\csname cdnum@e1.jev.doneneutral.nll\endcsname{0.689}
\expandafter\def\csname cdnum@e1.jev.doneneutral.softacc\endcsname{0.582}
\expandafter\def\csname cdnum@e1.jev.dthree.covfive\endcsname{0.00}
\expandafter\def\csname cdnum@e1.jev.dthree.ece\endcsname{0.251}
\expandafter\def\csname cdnum@e1.jev.dthree.ecescaled\endcsname{0.038}
\expandafter\def\csname cdnum@e1.jev.dthree.macrofone\endcsname{55.0}
\expandafter\def\csname cdnum@e1.jev.dtwo.acc\endcsname{0.913}
\expandafter\def\csname cdnum@e1.jev.dtwo.acchi\endcsname{0.935}
\expandafter\def\csname cdnum@e1.jev.dtwo.acclo\endcsname{0.890}
\expandafter\def\csname cdnum@e1.jev.dtwo.brier\endcsname{0.131}
\expandafter\def\csname cdnum@e1.jev.dtwo.covfive\endcsname{0.93}
\expandafter\def\csname cdnum@e1.jev.dtwo.ece\endcsname{0.032}
\expandafter\def\csname cdnum@e1.jev.dtwo.ecescaled\endcsname{0.032}
\expandafter\def\csname cdnum@e1.jev.dtwo.nll\endcsname{0.390}
\expandafter\def\csname cdnum@e1.jev.emotion.acc\endcsname{0.504}
\expandafter\def\csname cdnum@e1.jev.emotion.brier\endcsname{0.749}
\expandafter\def\csname cdnum@e1.jev.emotion.nll\endcsname{2.945}
\expandafter\def\csname cdnum@e1.jev.oos.auroc\endcsname{0.921}
\expandafter\def\csname cdnum@e1.jev.politeness.acc\endcsname{0.657}
\expandafter\def\csname cdnum@e1.jev.politeness.brier\endcsname{0.470}
\expandafter\def\csname cdnum@e1.jev.politeness.nll\endcsname{0.810}
\expandafter\def\csname cdnum@e1.jev.power.acc\endcsname{0.608}
\expandafter\def\csname cdnum@e1.jev.power.brier\endcsname{0.613}
\expandafter\def\csname cdnum@e1.jev.power.nll\endcsname{1.166}
\expandafter\def\csname cdnum@e1.jev.toxicity.acc\endcsname{0.558}
\expandafter\def\csname cdnum@e1.jev.toxicity.brier\endcsname{0.738}
\expandafter\def\csname cdnum@e1.jev.toxicity.nll\endcsname{1.810}
\expandafter\def\csname cdnum@e1.kevlarge.doneneutral.acc\endcsname{0.712}
\expandafter\def\csname cdnum@e1.kevlarge.doneneutral.brier\endcsname{0.415}
\expandafter\def\csname cdnum@e1.kevlarge.doneneutral.covfive\endcsname{0.12}
\expandafter\def\csname cdnum@e1.kevlarge.doneneutral.ece\endcsname{0.087}
\expandafter\def\csname cdnum@e1.kevlarge.doneneutral.eceraw\endcsname{0.092}
\expandafter\def\csname cdnum@e1.kevlarge.doneneutral.ecescaled\endcsname{0.047}
\expandafter\def\csname cdnum@e1.kevlarge.doneneutral.nll\endcsname{0.741}
\expandafter\def\csname cdnum@e1.kevlarge.doneneutral.softacc\endcsname{0.574}
\expandafter\def\csname cdnum@e1.kevlarge.dthree.covfive\endcsname{0.00}
\expandafter\def\csname cdnum@e1.kevlarge.dthree.ece\endcsname{0.138}
\expandafter\def\csname cdnum@e1.kevlarge.dthree.eceraw\endcsname{0.295}
\expandafter\def\csname cdnum@e1.kevlarge.dthree.ecescaled\endcsname{0.029}
\expandafter\def\csname cdnum@e1.kevlarge.dthree.macrofone\endcsname{48.3}
\expandafter\def\csname cdnum@e1.kevlarge.dtwo.acc\endcsname{0.850}
\expandafter\def\csname cdnum@e1.kevlarge.dtwo.brier\endcsname{0.402}
\expandafter\def\csname cdnum@e1.kevlarge.dtwo.covfive\endcsname{0.61}
\expandafter\def\csname cdnum@e1.kevlarge.dtwo.ece\endcsname{0.377}
\expandafter\def\csname cdnum@e1.kevlarge.dtwo.eceraw\endcsname{0.051}
\expandafter\def\csname cdnum@e1.kevlarge.dtwo.ecescaled\endcsname{0.046}
\expandafter\def\csname cdnum@e1.kevlarge.dtwo.nll\endcsname{1.162}
\expandafter\def\csname cdnum@e1.kevlarge.emotion.acc\endcsname{0.528}
\expandafter\def\csname cdnum@e1.kevlarge.emotion.brier\endcsname{0.619}
\expandafter\def\csname cdnum@e1.kevlarge.emotion.nll\endcsname{1.286}
\expandafter\def\csname cdnum@e1.kevlarge.oos.auroc\endcsname{0.933}
\expandafter\def\csname cdnum@e1.kevlarge.politeness.acc\endcsname{0.476}
\expandafter\def\csname cdnum@e1.kevlarge.politeness.brier\endcsname{0.635}
\expandafter\def\csname cdnum@e1.kevlarge.politeness.nll\endcsname{1.067}
\expandafter\def\csname cdnum@e1.kevlarge.power.acc\endcsname{0.594}
\expandafter\def\csname cdnum@e1.kevlarge.power.brier\endcsname{0.509}
\expandafter\def\csname cdnum@e1.kevlarge.power.nll\endcsname{0.711}
\expandafter\def\csname cdnum@e1.kevlarge.toxicity.acc\endcsname{0.530}
\expandafter\def\csname cdnum@e1.kevlarge.toxicity.brier\endcsname{0.631}
\expandafter\def\csname cdnum@e1.kevlarge.toxicity.nll\endcsname{0.889}
\expandafter\def\csname cdnum@e1.kevsmall.doneneutral.acc\endcsname{0.451}
\expandafter\def\csname cdnum@e1.kevsmall.doneneutral.brier\endcsname{0.608}
\expandafter\def\csname cdnum@e1.kevsmall.doneneutral.covfive\endcsname{0.02}
\expandafter\def\csname cdnum@e1.kevsmall.doneneutral.ece\endcsname{0.060}
\expandafter\def\csname cdnum@e1.kevsmall.doneneutral.eceraw\endcsname{0.161}
\expandafter\def\csname cdnum@e1.kevsmall.doneneutral.ecescaled\endcsname{0.070}
\expandafter\def\csname cdnum@e1.kevsmall.doneneutral.nll\endcsname{1.065}
\expandafter\def\csname cdnum@e1.kevsmall.doneneutral.softacc\endcsname{0.415}
\expandafter\def\csname cdnum@e1.kevsmall.dthree.covfive\endcsname{0.00}
\expandafter\def\csname cdnum@e1.kevsmall.dthree.ece\endcsname{0.125}
\expandafter\def\csname cdnum@e1.kevsmall.dthree.eceraw\endcsname{0.290}
\expandafter\def\csname cdnum@e1.kevsmall.dthree.ecescaled\endcsname{0.026}
\expandafter\def\csname cdnum@e1.kevsmall.dthree.macrofone\endcsname{44.7}
\expandafter\def\csname cdnum@e1.kevsmall.dtwo.acc\endcsname{0.685}
\expandafter\def\csname cdnum@e1.kevsmall.dtwo.brier\endcsname{0.522}
\expandafter\def\csname cdnum@e1.kevsmall.dtwo.covfive\endcsname{0.22}
\expandafter\def\csname cdnum@e1.kevsmall.dtwo.ece\endcsname{0.254}
\expandafter\def\csname cdnum@e1.kevsmall.dtwo.eceraw\endcsname{0.090}
\expandafter\def\csname cdnum@e1.kevsmall.dtwo.ecescaled\endcsname{0.087}
\expandafter\def\csname cdnum@e1.kevsmall.dtwo.nll\endcsname{1.495}
\expandafter\def\csname cdnum@e1.kevsmall.emotion.acc\endcsname{0.516}
\expandafter\def\csname cdnum@e1.kevsmall.emotion.brier\endcsname{0.644}
\expandafter\def\csname cdnum@e1.kevsmall.emotion.nll\endcsname{1.327}
\expandafter\def\csname cdnum@e1.kevsmall.oos.auroc\endcsname{0.761}
\expandafter\def\csname cdnum@e1.kevsmall.politeness.acc\endcsname{0.476}
\expandafter\def\csname cdnum@e1.kevsmall.politeness.brier\endcsname{0.639}
\expandafter\def\csname cdnum@e1.kevsmall.politeness.nll\endcsname{1.037}
\expandafter\def\csname cdnum@e1.kevsmall.power.acc\endcsname{0.520}
\expandafter\def\csname cdnum@e1.kevsmall.power.brier\endcsname{0.555}
\expandafter\def\csname cdnum@e1.kevsmall.power.nll\endcsname{0.757}
\expandafter\def\csname cdnum@e1.kevsmall.toxicity.acc\endcsname{0.530}
\expandafter\def\csname cdnum@e1.kevsmall.toxicity.brier\endcsname{0.559}
\expandafter\def\csname cdnum@e1.kevsmall.toxicity.nll\endcsname{0.764}
\expandafter\def\csname cdnum@e1.layaen.donenative.acc\endcsname{0.358}
\expandafter\def\csname cdnum@e1.layaen.doneneutral.acc\endcsname{0.366}
\expandafter\def\csname cdnum@e1.layaen.doneneutral.brier\endcsname{0.741}
\expandafter\def\csname cdnum@e1.layaen.doneneutral.covfive\endcsname{0.00}
\expandafter\def\csname cdnum@e1.layaen.doneneutral.ece\endcsname{0.156}
\expandafter\def\csname cdnum@e1.layaen.doneneutral.eceraw\endcsname{0.232}
\expandafter\def\csname cdnum@e1.layaen.doneneutral.ecescaled\endcsname{0.068}
\expandafter\def\csname cdnum@e1.layaen.doneneutral.nll\endcsname{1.294}
\expandafter\def\csname cdnum@e1.layaen.doneneutral.softacc\endcsname{0.358}
\expandafter\def\csname cdnum@e1.layaen.dthree.covfive\endcsname{0.01}
\expandafter\def\csname cdnum@e1.layaen.dthree.ece\endcsname{0.158}
\expandafter\def\csname cdnum@e1.layaen.dthree.eceraw\endcsname{0.225}
\expandafter\def\csname cdnum@e1.layaen.dthree.ecescaled\endcsname{0.029}
\expandafter\def\csname cdnum@e1.layaen.dthree.macrofone\endcsname{45.8}
\expandafter\def\csname cdnum@e1.layaen.emotion.acc\endcsname{0.504}
\expandafter\def\csname cdnum@e1.layaen.emotion.brier\endcsname{0.792}
\expandafter\def\csname cdnum@e1.layaen.emotion.nll\endcsname{2.027}
\expandafter\def\csname cdnum@e1.layaen.politeness.acc\endcsname{0.373}
\expandafter\def\csname cdnum@e1.layaen.politeness.brier\endcsname{0.725}
\expandafter\def\csname cdnum@e1.layaen.politeness.nll\endcsname{1.197}
\expandafter\def\csname cdnum@e1.layaen.power.acc\endcsname{0.508}
\expandafter\def\csname cdnum@e1.layaen.power.brier\endcsname{0.513}
\expandafter\def\csname cdnum@e1.layaen.power.nll\endcsname{0.708}
\expandafter\def\csname cdnum@e1.layaen.toxicity.acc\endcsname{0.568}
\expandafter\def\csname cdnum@e1.layaen.toxicity.brier\endcsname{0.498}
\expandafter\def\csname cdnum@e1.layaen.toxicity.nll\endcsname{0.693}
\expandafter\def\csname cdnum@e1.layaml.doneneutral.acc\endcsname{0.349}
\expandafter\def\csname cdnum@e1.layaml.doneneutral.brier\endcsname{0.853}
\expandafter\def\csname cdnum@e1.layaml.doneneutral.covfive\endcsname{0.01}
\expandafter\def\csname cdnum@e1.layaml.doneneutral.ece\endcsname{0.279}
\expandafter\def\csname cdnum@e1.layaml.doneneutral.eceraw\endcsname{0.279}
\expandafter\def\csname cdnum@e1.layaml.doneneutral.ecescaled\endcsname{0.044}
\expandafter\def\csname cdnum@e1.layaml.doneneutral.nll\endcsname{1.618}
\expandafter\def\csname cdnum@e1.layaml.doneneutral.softacc\endcsname{0.339}
\expandafter\def\csname cdnum@e1.layaml.dthree.covfive\endcsname{0.00}
\expandafter\def\csname cdnum@e1.layaml.dthree.ece\endcsname{0.204}
\expandafter\def\csname cdnum@e1.layaml.dthree.eceraw\endcsname{0.204}
\expandafter\def\csname cdnum@e1.layaml.dthree.ecescaled\endcsname{0.029}
\expandafter\def\csname cdnum@e1.layaml.dthree.macrofone\endcsname{41.7}
\expandafter\def\csname cdnum@e1.layaml.dtwo.acc\endcsname{0.005}
\expandafter\def\csname cdnum@e1.layaml.dtwo.brier\endcsname{0.999}
\expandafter\def\csname cdnum@e1.layaml.dtwo.covfive\endcsname{0.00}
\expandafter\def\csname cdnum@e1.layaml.dtwo.ece\endcsname{0.031}
\expandafter\def\csname cdnum@e1.layaml.dtwo.eceraw\endcsname{0.031}
\expandafter\def\csname cdnum@e1.layaml.dtwo.ecescaled\endcsname{0.010}
\expandafter\def\csname cdnum@e1.layaml.dtwo.nll\endcsname{5.160}
\expandafter\def\csname cdnum@e1.layaml.emotion.acc\endcsname{0.396}
\expandafter\def\csname cdnum@e1.layaml.emotion.brier\endcsname{0.840}
\expandafter\def\csname cdnum@e1.layaml.emotion.nll\endcsname{1.940}
\expandafter\def\csname cdnum@e1.layaml.oos.auroc\endcsname{0.548}
\expandafter\def\csname cdnum@e1.layaml.politeness.acc\endcsname{0.365}
\expandafter\def\csname cdnum@e1.layaml.politeness.brier\endcsname{0.715}
\expandafter\def\csname cdnum@e1.layaml.politeness.nll\endcsname{1.170}
\expandafter\def\csname cdnum@e1.layaml.power.acc\endcsname{0.482}
\expandafter\def\csname cdnum@e1.layaml.power.brier\endcsname{0.609}
\expandafter\def\csname cdnum@e1.layaml.power.nll\endcsname{0.843}
\expandafter\def\csname cdnum@e1.layaml.toxicity.acc\endcsname{0.552}
\expandafter\def\csname cdnum@e1.layaml.toxicity.brier\endcsname{0.564}
\expandafter\def\csname cdnum@e1.layaml.toxicity.nll\endcsname{0.805}
\expandafter\def\csname cdnum@e1.nimble.doneneutral.acc\endcsname{0.685}
\expandafter\def\csname cdnum@e1.nimble.doneneutral.brier\endcsname{0.430}
\expandafter\def\csname cdnum@e1.nimble.doneneutral.covfive\endcsname{0.20}
\expandafter\def\csname cdnum@e1.nimble.doneneutral.ece\endcsname{0.057}
\expandafter\def\csname cdnum@e1.nimble.doneneutral.eceraw\endcsname{0.057}
\expandafter\def\csname cdnum@e1.nimble.doneneutral.ecescaled\endcsname{0.026}
\expandafter\def\csname cdnum@e1.nimble.doneneutral.nll\endcsname{0.797}
\expandafter\def\csname cdnum@e1.nimble.doneneutral.softacc\endcsname{0.559}
\expandafter\def\csname cdnum@e1.nimble.dthree.covfive\endcsname{0.00}
\expandafter\def\csname cdnum@e1.nimble.dthree.ece\endcsname{0.285}
\expandafter\def\csname cdnum@e1.nimble.dthree.eceraw\endcsname{0.285}
\expandafter\def\csname cdnum@e1.nimble.dthree.ecescaled\endcsname{0.031}
\expandafter\def\csname cdnum@e1.nimble.dthree.macrofone\endcsname{48.3}
\expandafter\def\csname cdnum@e1.nimble.dtwo.acc\endcsname{0.860}
\expandafter\def\csname cdnum@e1.nimble.dtwo.brier\endcsname{0.220}
\expandafter\def\csname cdnum@e1.nimble.dtwo.covfive\endcsname{0.66}
\expandafter\def\csname cdnum@e1.nimble.dtwo.ece\endcsname{0.069}
\expandafter\def\csname cdnum@e1.nimble.dtwo.eceraw\endcsname{0.069}
\expandafter\def\csname cdnum@e1.nimble.dtwo.ecescaled\endcsname{0.037}
\expandafter\def\csname cdnum@e1.nimble.dtwo.nll\endcsname{0.644}
\expandafter\def\csname cdnum@e1.nimble.emotion.acc\endcsname{0.514}
\expandafter\def\csname cdnum@e1.nimble.emotion.brier\endcsname{0.747}
\expandafter\def\csname cdnum@e1.nimble.emotion.nll\endcsname{1.794}
\expandafter\def\csname cdnum@e1.nimble.oos.auroc\endcsname{0.923}
\expandafter\def\csname cdnum@e1.nimble.politeness.acc\endcsname{0.516}
\expandafter\def\csname cdnum@e1.nimble.politeness.brier\endcsname{0.618}
\expandafter\def\csname cdnum@e1.nimble.politeness.nll\endcsname{1.096}
\expandafter\def\csname cdnum@e1.nimble.power.acc\endcsname{0.554}
\expandafter\def\csname cdnum@e1.nimble.power.brier\endcsname{0.743}
\expandafter\def\csname cdnum@e1.nimble.power.nll\endcsname{1.272}
\expandafter\def\csname cdnum@e1.nimble.toxicity.acc\endcsname{0.556}
\expandafter\def\csname cdnum@e1.nimble.toxicity.brier\endcsname{0.737}
\expandafter\def\csname cdnum@e1.nimble.toxicity.nll\endcsname{1.357}
\expandafter\def\csname cdnum@e1.thisthat.doneneutral.acc\endcsname{0.597}
\expandafter\def\csname cdnum@e1.thisthat.doneneutral.brier\endcsname{0.564}
\expandafter\def\csname cdnum@e1.thisthat.doneneutral.covfive\endcsname{0.04}
\expandafter\def\csname cdnum@e1.thisthat.doneneutral.ece\endcsname{0.196}
\expandafter\def\csname cdnum@e1.thisthat.doneneutral.eceraw\endcsname{0.196}
\expandafter\def\csname cdnum@e1.thisthat.doneneutral.ecescaled\endcsname{0.040}
\expandafter\def\csname cdnum@e1.thisthat.doneneutral.nll\endcsname{1.090}
\expandafter\def\csname cdnum@e1.thisthat.doneneutral.softacc\endcsname{0.516}
\expandafter\def\csname cdnum@e1.thisthat.dthree.covfive\endcsname{0.00}
\expandafter\def\csname cdnum@e1.thisthat.dthree.ece\endcsname{0.331}
\expandafter\def\csname cdnum@e1.thisthat.dthree.eceraw\endcsname{0.331}
\expandafter\def\csname cdnum@e1.thisthat.dthree.ecescaled\endcsname{0.038}
\expandafter\def\csname cdnum@e1.thisthat.dthree.macrofone\endcsname{50.1}
\expandafter\def\csname cdnum@e1.thisthat.dtwo.acc\endcsname{0.863}
\expandafter\def\csname cdnum@e1.thisthat.dtwo.brier\endcsname{0.208}
\expandafter\def\csname cdnum@e1.thisthat.dtwo.covfive\endcsname{0.78}
\expandafter\def\csname cdnum@e1.thisthat.dtwo.ece\endcsname{0.055}
\expandafter\def\csname cdnum@e1.thisthat.dtwo.eceraw\endcsname{0.055}
\expandafter\def\csname cdnum@e1.thisthat.dtwo.ecescaled\endcsname{0.023}
\expandafter\def\csname cdnum@e1.thisthat.dtwo.nll\endcsname{0.646}
\expandafter\def\csname cdnum@e1.thisthat.emotion.acc\endcsname{0.805}
\expandafter\def\csname cdnum@e1.thisthat.emotion.brier\endcsname{0.292}
\expandafter\def\csname cdnum@e1.thisthat.emotion.nll\endcsname{0.626}
\expandafter\def\csname cdnum@e1.thisthat.oos.auroc\endcsname{0.860}
\expandafter\def\csname cdnum@e1.thisthat.politeness.acc\endcsname{0.414}
\expandafter\def\csname cdnum@e1.thisthat.politeness.brier\endcsname{0.990}
\expandafter\def\csname cdnum@e1.thisthat.politeness.nll\endcsname{2.152}
\expandafter\def\csname cdnum@e1.thisthat.power.acc\endcsname{0.530}
\expandafter\def\csname cdnum@e1.thisthat.power.brier\endcsname{0.692}
\expandafter\def\csname cdnum@e1.thisthat.power.nll\endcsname{1.153}
\expandafter\def\csname cdnum@e1.thisthat.toxicity.acc\endcsname{0.516}
\expandafter\def\csname cdnum@e1.thisthat.toxicity.brier\endcsname{0.936}
\expandafter\def\csname cdnum@e1.thisthat.toxicity.nll\endcsname{2.987}
\expandafter\def\csname cdnum@e2.compgem.done.k01.flips\endcsname{2.8}
\expandafter\def\csname cdnum@e2.compgem.done.kny.auc\endcsname{0.91}
\expandafter\def\csname cdnum@e2.compgem.done.krand.flips\endcsname{3.2}
\expandafter\def\csname cdnum@e2.compgem.done.kswap.flips\endcsname{17.2}
\expandafter\def\csname cdnum@e2.compgem.power.k01.flips\endcsname{1.8}
\expandafter\def\csname cdnum@e2.compgem.power.kny.auc\endcsname{0.67}
\expandafter\def\csname cdnum@e2.compgem.power.krand.flips\endcsname{2.2}
\expandafter\def\csname cdnum@e2.compgem.power.kswap.flips\endcsname{31.0}
\expandafter\def\csname cdnum@e2.compgem.toxicity.k01.flips\endcsname{2.4}
\expandafter\def\csname cdnum@e2.compgem.toxicity.kny.auc\endcsname{0.66}
\expandafter\def\csname cdnum@e2.compgem.toxicity.krand.flips\endcsname{3.6}
\expandafter\def\csname cdnum@e2.compgem.toxicity.kswap.flips\endcsname{15.6}
\expandafter\def\csname cdnum@e2.compgemll.done.k01.flips\endcsname{2.3}
\expandafter\def\csname cdnum@e2.compgemll.done.kny.auc\endcsname{0.94}
\expandafter\def\csname cdnum@e2.compgemll.done.krand.flips\endcsname{2.8}
\expandafter\def\csname cdnum@e2.compgemll.done.kswap.flips\endcsname{16.8}
\expandafter\def\csname cdnum@e2.compgemll.power.k01.flips\endcsname{2.0}
\expandafter\def\csname cdnum@e2.compgemll.power.kny.auc\endcsname{0.65}
\expandafter\def\csname cdnum@e2.compgemll.power.krand.flips\endcsname{1.8}
\expandafter\def\csname cdnum@e2.compgemll.power.kswap.flips\endcsname{31.6}
\expandafter\def\csname cdnum@e2.compgemll.toxicity.k01.flips\endcsname{2.4}
\expandafter\def\csname cdnum@e2.compgemll.toxicity.kny.auc\endcsname{0.65}
\expandafter\def\csname cdnum@e2.compgemll.toxicity.krand.flips\endcsname{3.6}
\expandafter\def\csname cdnum@e2.compgemll.toxicity.kswap.flips\endcsname{15.8}
\expandafter\def\csname cdnum@e2.compmis.done.k01.flips\endcsname{9.0}
\expandafter\def\csname cdnum@e2.compmis.done.kny.auc\endcsname{0.78}
\expandafter\def\csname cdnum@e2.compmis.done.krand.flips\endcsname{14.8}
\expandafter\def\csname cdnum@e2.compmis.done.kswap.flips\endcsname{40.3}
\expandafter\def\csname cdnum@e2.compmis.power.k01.flips\endcsname{6.4}
\expandafter\def\csname cdnum@e2.compmis.power.kny.auc\endcsname{0.61}
\expandafter\def\csname cdnum@e2.compmis.power.krand.flips\endcsname{5.8}
\expandafter\def\csname cdnum@e2.compmis.power.kswap.flips\endcsname{81.0}
\expandafter\def\csname cdnum@e2.compmis.toxicity.k01.flips\endcsname{12.8}
\expandafter\def\csname cdnum@e2.compmis.toxicity.kny.auc\endcsname{0.61}
\expandafter\def\csname cdnum@e2.compmis.toxicity.krand.flips\endcsname{5.6}
\expandafter\def\csname cdnum@e2.compmis.toxicity.kswap.flips\endcsname{84.2}
\expandafter\def\csname cdnum@e2.compmisll.done.k01.flips\endcsname{11.0}
\expandafter\def\csname cdnum@e2.compmisll.done.kny.auc\endcsname{0.83}
\expandafter\def\csname cdnum@e2.compmisll.done.krand.flips\endcsname{17.2}
\expandafter\def\csname cdnum@e2.compmisll.done.kswap.flips\endcsname{40.7}
\expandafter\def\csname cdnum@e2.compmisll.power.k01.flips\endcsname{5.0}
\expandafter\def\csname cdnum@e2.compmisll.power.kny.auc\endcsname{0.61}
\expandafter\def\csname cdnum@e2.compmisll.power.krand.flips\endcsname{5.6}
\expandafter\def\csname cdnum@e2.compmisll.power.kswap.flips\endcsname{83.0}
\expandafter\def\csname cdnum@e2.compmisll.toxicity.k01.flips\endcsname{4.4}
\expandafter\def\csname cdnum@e2.compmisll.toxicity.kny.auc\endcsname{0.66}
\expandafter\def\csname cdnum@e2.compmisll.toxicity.krand.flips\endcsname{6.0}
\expandafter\def\csname cdnum@e2.compmisll.toxicity.kswap.flips\endcsname{80.2}
\expandafter\def\csname cdnum@e2.comptwo.done.k01.flips\endcsname{9.7}
\expandafter\def\csname cdnum@e2.comptwo.done.kny.auc\endcsname{0.90}
\expandafter\def\csname cdnum@e2.comptwo.done.krand.flips\endcsname{11.0}
\expandafter\def\csname cdnum@e2.comptwo.done.kswap.flips\endcsname{47.8}
\expandafter\def\csname cdnum@e2.comptwo.power.k01.flips\endcsname{4.4}
\expandafter\def\csname cdnum@e2.comptwo.power.kny.auc\endcsname{0.63}
\expandafter\def\csname cdnum@e2.comptwo.power.krand.flips\endcsname{6.6}
\expandafter\def\csname cdnum@e2.comptwo.power.kswap.flips\endcsname{85.0}
\expandafter\def\csname cdnum@e2.comptwo.toxicity.k01.flips\endcsname{2.2}
\expandafter\def\csname cdnum@e2.comptwo.toxicity.kny.auc\endcsname{0.61}
\expandafter\def\csname cdnum@e2.comptwo.toxicity.krand.flips\endcsname{3.8}
\expandafter\def\csname cdnum@e2.comptwo.toxicity.kswap.flips\endcsname{36.2}
\expandafter\def\csname cdnum@e2.comptwoll.done.k01.flips\endcsname{7.0}
\expandafter\def\csname cdnum@e2.comptwoll.done.kny.auc\endcsname{0.93}
\expandafter\def\csname cdnum@e2.comptwoll.done.krand.flips\endcsname{8.8}
\expandafter\def\csname cdnum@e2.comptwoll.done.kswap.flips\endcsname{48.7}
\expandafter\def\csname cdnum@e2.comptwoll.power.k01.flips\endcsname{3.8}
\expandafter\def\csname cdnum@e2.comptwoll.power.kny.auc\endcsname{0.65}
\expandafter\def\csname cdnum@e2.comptwoll.power.krand.flips\endcsname{6.4}
\expandafter\def\csname cdnum@e2.comptwoll.power.kswap.flips\endcsname{83.4}
\expandafter\def\csname cdnum@e2.comptwoll.toxicity.k01.flips\endcsname{1.8}
\expandafter\def\csname cdnum@e2.comptwoll.toxicity.kny.auc\endcsname{0.64}
\expandafter\def\csname cdnum@e2.comptwoll.toxicity.krand.flips\endcsname{2.6}
\expandafter\def\csname cdnum@e2.comptwoll.toxicity.kswap.flips\endcsname{18.4}
\expandafter\def\csname cdnum@e2.decider.done.floor\endcsname{0.0}
\expandafter\def\csname cdnum@e2.decider.done.k01.flips\endcsname{2.7}
\expandafter\def\csname cdnum@e2.decider.done.kny.auc\endcsname{0.75}
\expandafter\def\csname cdnum@e2.decider.done.krand.flips\endcsname{5.3}
\expandafter\def\csname cdnum@e2.decider.done.kswap.absdauc\endcsname{0.06}
\expandafter\def\csname cdnum@e2.decider.done.kswap.flips\endcsname{4.8}
\expandafter\def\csname cdnum@e2.decider.power.floor\endcsname{0.0}
\expandafter\def\csname cdnum@e2.decider.power.k01.flips\endcsname{3.4}
\expandafter\def\csname cdnum@e2.decider.power.kny.auc\endcsname{0.60}
\expandafter\def\csname cdnum@e2.decider.power.krand.flips\endcsname{4.0}
\expandafter\def\csname cdnum@e2.decider.power.kswap.flips\endcsname{5.2}
\expandafter\def\csname cdnum@e2.decider.toxicity.floor\endcsname{0.0}
\expandafter\def\csname cdnum@e2.decider.toxicity.k01.flips\endcsname{0.0}
\expandafter\def\csname cdnum@e2.decider.toxicity.kny.auc\endcsname{0.64}
\expandafter\def\csname cdnum@e2.decider.toxicity.krand.flips\endcsname{0.4}
\expandafter\def\csname cdnum@e2.decider.toxicity.kswap.flips\endcsname{1.0}
\expandafter\def\csname cdnum@e2.decoderheads.kswap.maxflips\endcsname{9.8}
\expandafter\def\csname cdnum@e2.jev.done.floor\endcsname{1.3}
\expandafter\def\csname cdnum@e2.jev.done.k01.flips\endcsname{2.5}
\expandafter\def\csname cdnum@e2.jev.done.kny.auc\endcsname{0.89}
\expandafter\def\csname cdnum@e2.jev.done.krand.flips\endcsname{2.2}
\expandafter\def\csname cdnum@e2.jev.done.kswap.auc\endcsname{0.31}
\expandafter\def\csname cdnum@e2.jev.done.kswap.flips\endcsname{50.5}
\expandafter\def\csname cdnum@e2.jev.neutralnames.maxflips\endcsname{2.8}
\expandafter\def\csname cdnum@e2.jev.power.floor\endcsname{0.8}
\expandafter\def\csname cdnum@e2.jev.power.k01.flips\endcsname{1.8}
\expandafter\def\csname cdnum@e2.jev.power.kny.auc\endcsname{0.63}
\expandafter\def\csname cdnum@e2.jev.power.krand.flips\endcsname{2.8}
\expandafter\def\csname cdnum@e2.jev.power.kswap.flips\endcsname{84.0}
\expandafter\def\csname cdnum@e2.jev.toxicity.floor\endcsname{0.4}
\expandafter\def\csname cdnum@e2.jev.toxicity.k01.flips\endcsname{0.8}
\expandafter\def\csname cdnum@e2.jev.toxicity.kny.auc\endcsname{0.66}
\expandafter\def\csname cdnum@e2.jev.toxicity.krand.flips\endcsname{1.0}
\expandafter\def\csname cdnum@e2.jev.toxicity.kswap.flips\endcsname{97.2}
\expandafter\def\csname cdnum@e2.kevlarge.done.floor\endcsname{0.0}
\expandafter\def\csname cdnum@e2.kevlarge.done.k01.flips\endcsname{4.7}
\expandafter\def\csname cdnum@e2.kevlarge.done.kny.auc\endcsname{0.89}
\expandafter\def\csname cdnum@e2.kevlarge.done.krand.flips\endcsname{4.7}
\expandafter\def\csname cdnum@e2.kevlarge.done.kswap.flips\endcsname{6.5}
\expandafter\def\csname cdnum@e2.kevlarge.power.floor\endcsname{0.0}
\expandafter\def\csname cdnum@e2.kevlarge.power.k01.flips\endcsname{1.8}
\expandafter\def\csname cdnum@e2.kevlarge.power.kny.auc\endcsname{0.65}
\expandafter\def\csname cdnum@e2.kevlarge.power.krand.flips\endcsname{1.8}
\expandafter\def\csname cdnum@e2.kevlarge.power.kswap.flips\endcsname{6.0}
\expandafter\def\csname cdnum@e2.kevlarge.toxicity.floor\endcsname{0.0}
\expandafter\def\csname cdnum@e2.kevlarge.toxicity.k01.flips\endcsname{0.6}
\expandafter\def\csname cdnum@e2.kevlarge.toxicity.kny.auc\endcsname{0.65}
\expandafter\def\csname cdnum@e2.kevlarge.toxicity.krand.flips\endcsname{0.8}
\expandafter\def\csname cdnum@e2.kevlarge.toxicity.kswap.flips\endcsname{0.8}
\expandafter\def\csname cdnum@e2.kevsmall.done.floor\endcsname{0.0}
\expandafter\def\csname cdnum@e2.kevsmall.done.k01.flips\endcsname{6.2}
\expandafter\def\csname cdnum@e2.kevsmall.done.kny.auc\endcsname{0.64}
\expandafter\def\csname cdnum@e2.kevsmall.done.krand.flips\endcsname{20.2}
\expandafter\def\csname cdnum@e2.kevsmall.done.kswap.flips\endcsname{14.3}
\expandafter\def\csname cdnum@e2.kevsmall.power.floor\endcsname{0.0}
\expandafter\def\csname cdnum@e2.kevsmall.power.k01.flips\endcsname{1.4}
\expandafter\def\csname cdnum@e2.kevsmall.power.kny.auc\endcsname{0.56}
\expandafter\def\csname cdnum@e2.kevsmall.power.krand.flips\endcsname{1.2}
\expandafter\def\csname cdnum@e2.kevsmall.power.kswap.flips\endcsname{5.6}
\expandafter\def\csname cdnum@e2.kevsmall.toxicity.floor\endcsname{0.0}
\expandafter\def\csname cdnum@e2.kevsmall.toxicity.k01.flips\endcsname{0.4}
\expandafter\def\csname cdnum@e2.kevsmall.toxicity.kny.auc\endcsname{0.63}
\expandafter\def\csname cdnum@e2.kevsmall.toxicity.krand.flips\endcsname{0.8}
\expandafter\def\csname cdnum@e2.kevsmall.toxicity.kswap.flips\endcsname{1.4}
\expandafter\def\csname cdnum@e2.layaen.done.floor\endcsname{0.0}
\expandafter\def\csname cdnum@e2.layaen.done.k01.flips\endcsname{22.8}
\expandafter\def\csname cdnum@e2.layaen.done.kny.auc\endcsname{0.45}
\expandafter\def\csname cdnum@e2.layaen.done.krand.flips\endcsname{31.5}
\expandafter\def\csname cdnum@e2.layaen.done.kswap.flips\endcsname{91.3}
\expandafter\def\csname cdnum@e2.layaen.power.floor\endcsname{0.0}
\expandafter\def\csname cdnum@e2.layaen.power.k01.flips\endcsname{19.4}
\expandafter\def\csname cdnum@e2.layaen.power.kny.auc\endcsname{0.55}
\expandafter\def\csname cdnum@e2.layaen.power.krand.flips\endcsname{24.8}
\expandafter\def\csname cdnum@e2.layaen.power.kswap.flips\endcsname{90.6}
\expandafter\def\csname cdnum@e2.layaen.toxicity.floor\endcsname{0.0}
\expandafter\def\csname cdnum@e2.layaen.toxicity.k01.flips\endcsname{11.6}
\expandafter\def\csname cdnum@e2.layaen.toxicity.kny.auc\endcsname{0.62}
\expandafter\def\csname cdnum@e2.layaen.toxicity.krand.flips\endcsname{18.8}
\expandafter\def\csname cdnum@e2.layaen.toxicity.kswap.flips\endcsname{98.8}
\expandafter\def\csname cdnum@e2.layaml.done.floor\endcsname{0.0}
\expandafter\def\csname cdnum@e2.layaml.done.k01.flips\endcsname{26.2}
\expandafter\def\csname cdnum@e2.layaml.done.kny.auc\endcsname{0.42}
\expandafter\def\csname cdnum@e2.layaml.done.krand.flips\endcsname{29.3}
\expandafter\def\csname cdnum@e2.layaml.done.kswap.flips\endcsname{86.8}
\expandafter\def\csname cdnum@e2.layaml.power.floor\endcsname{0.0}
\expandafter\def\csname cdnum@e2.layaml.power.k01.flips\endcsname{32.0}
\expandafter\def\csname cdnum@e2.layaml.power.kny.auc\endcsname{0.49}
\expandafter\def\csname cdnum@e2.layaml.power.krand.flips\endcsname{43.4}
\expandafter\def\csname cdnum@e2.layaml.power.kswap.flips\endcsname{71.4}
\expandafter\def\csname cdnum@e2.layaml.toxicity.floor\endcsname{0.0}
\expandafter\def\csname cdnum@e2.layaml.toxicity.k01.flips\endcsname{42.0}
\expandafter\def\csname cdnum@e2.layaml.toxicity.kny.auc\endcsname{0.55}
\expandafter\def\csname cdnum@e2.layaml.toxicity.krand.flips\endcsname{50.6}
\expandafter\def\csname cdnum@e2.layaml.toxicity.kswap.flips\endcsname{83.0}
\expandafter\def\csname cdnum@e2.nimble.done.floor\endcsname{0.0}
\expandafter\def\csname cdnum@e2.nimble.done.k01.flips\endcsname{1.0}
\expandafter\def\csname cdnum@e2.nimble.done.kny.auc\endcsname{0.91}
\expandafter\def\csname cdnum@e2.nimble.done.krand.flips\endcsname{3.3}
\expandafter\def\csname cdnum@e2.nimble.done.kswap.flips\endcsname{4.2}
\expandafter\def\csname cdnum@e2.nimble.power.floor\endcsname{0.0}
\expandafter\def\csname cdnum@e2.nimble.power.k01.flips\endcsname{2.6}
\expandafter\def\csname cdnum@e2.nimble.power.kny.auc\endcsname{0.59}
\expandafter\def\csname cdnum@e2.nimble.power.krand.flips\endcsname{1.8}
\expandafter\def\csname cdnum@e2.nimble.power.kswap.flips\endcsname{1.8}
\expandafter\def\csname cdnum@e2.nimble.toxicity.floor\endcsname{0.0}
\expandafter\def\csname cdnum@e2.nimble.toxicity.k01.flips\endcsname{2.4}
\expandafter\def\csname cdnum@e2.nimble.toxicity.kny.auc\endcsname{0.65}
\expandafter\def\csname cdnum@e2.nimble.toxicity.krand.flips\endcsname{2.2}
\expandafter\def\csname cdnum@e2.nimble.toxicity.kswap.flips\endcsname{1.8}
\expandafter\def\csname cdnum@e2.thisthat.done.floor\endcsname{0.0}
\expandafter\def\csname cdnum@e2.thisthat.done.k01.flips\endcsname{2.7}
\expandafter\def\csname cdnum@e2.thisthat.done.kny.auc\endcsname{0.81}
\expandafter\def\csname cdnum@e2.thisthat.done.krand.flips\endcsname{6.2}
\expandafter\def\csname cdnum@e2.thisthat.done.kswap.flips\endcsname{4.3}
\expandafter\def\csname cdnum@e2.thisthat.power.floor\endcsname{0.0}
\expandafter\def\csname cdnum@e2.thisthat.power.k01.flips\endcsname{4.6}
\expandafter\def\csname cdnum@e2.thisthat.power.kny.auc\endcsname{0.59}
\expandafter\def\csname cdnum@e2.thisthat.power.krand.flips\endcsname{4.6}
\expandafter\def\csname cdnum@e2.thisthat.power.kswap.flips\endcsname{9.8}
\expandafter\def\csname cdnum@e2.thisthat.toxicity.floor\endcsname{0.0}
\expandafter\def\csname cdnum@e2.thisthat.toxicity.k01.flips\endcsname{0.2}
\expandafter\def\csname cdnum@e2.thisthat.toxicity.kny.auc\endcsname{0.64}
\expandafter\def\csname cdnum@e2.thisthat.toxicity.krand.flips\endcsname{0.4}
\expandafter\def\csname cdnum@e2.thisthat.toxicity.kswap.flips\endcsname{0.2}
\expandafter\def\csname cdnum@e3.compgem.all.acc\endcsname{0.942}
\expandafter\def\csname cdnum@e3.compgem.all.ece\endcsname{0.022}
\expandafter\def\csname cdnum@e3.compgem.drop\endcsname{0.053}
\expandafter\def\csname cdnum@e3.compgem.fifty.acc\endcsname{0.965}
\expandafter\def\csname cdnum@e3.compgem.five.acc\endcsname{0.995}
\expandafter\def\csname cdnum@e3.compgem.twenty.acc\endcsname{0.980}
\expandafter\def\csname cdnum@e3.compgemll.all.acc\endcsname{0.940}
\expandafter\def\csname cdnum@e3.compgemll.all.ece\endcsname{0.058}
\expandafter\def\csname cdnum@e3.compgemll.drop\endcsname{0.055}
\expandafter\def\csname cdnum@e3.compgemll.fifty.acc\endcsname{0.967}
\expandafter\def\csname cdnum@e3.compgemll.five.acc\endcsname{0.995}
\expandafter\def\csname cdnum@e3.compgemll.twenty.acc\endcsname{0.980}
\expandafter\def\csname cdnum@e3.compmis.all.acc\endcsname{0.757}
\expandafter\def\csname cdnum@e3.compmis.all.ece\endcsname{0.135}
\expandafter\def\csname cdnum@e3.compmis.drop\endcsname{0.117}
\expandafter\def\csname cdnum@e3.compmis.fifty.acc\endcsname{0.860}
\expandafter\def\csname cdnum@e3.compmis.five.acc\endcsname{0.873}
\expandafter\def\csname cdnum@e3.compmis.twenty.acc\endcsname{0.970}
\expandafter\def\csname cdnum@e3.compmisll.all.acc\endcsname{0.782}
\expandafter\def\csname cdnum@e3.compmisll.all.ece\endcsname{0.055}
\expandafter\def\csname cdnum@e3.compmisll.drop\endcsname{0.207}
\expandafter\def\csname cdnum@e3.compmisll.fifty.acc\endcsname{0.862}
\expandafter\def\csname cdnum@e3.compmisll.five.acc\endcsname{0.988}
\expandafter\def\csname cdnum@e3.compmisll.twenty.acc\endcsname{0.962}
\expandafter\def\csname cdnum@e3.compopen.all.acc\endcsname{0.728}
\expandafter\def\csname cdnum@e3.compopen.all.ece\endcsname{0.204}
\expandafter\def\csname cdnum@e3.compopen.drop\endcsname{0.250}
\expandafter\def\csname cdnum@e3.compopen.fifty.acc\endcsname{0.875}
\expandafter\def\csname cdnum@e3.compopen.five.acc\endcsname{0.978}
\expandafter\def\csname cdnum@e3.compopen.twenty.acc\endcsname{0.955}
\expandafter\def\csname cdnum@e3.comptwo.all.acc\endcsname{0.683}
\expandafter\def\csname cdnum@e3.comptwo.all.ece\endcsname{0.023}
\expandafter\def\csname cdnum@e3.comptwo.drop\endcsname{0.298}
\expandafter\def\csname cdnum@e3.comptwo.fifty.acc\endcsname{0.415}
\expandafter\def\csname cdnum@e3.comptwo.five.acc\endcsname{0.982}
\expandafter\def\csname cdnum@e3.comptwo.twenty.acc\endcsname{0.863}
\expandafter\def\csname cdnum@e3.comptwoll.all.acc\endcsname{0.912}
\expandafter\def\csname cdnum@e3.comptwoll.all.ece\endcsname{0.044}
\expandafter\def\csname cdnum@e3.comptwoll.drop\endcsname{0.083}
\expandafter\def\csname cdnum@e3.comptwoll.fifty.acc\endcsname{0.952}
\expandafter\def\csname cdnum@e3.comptwoll.five.acc\endcsname{0.995}
\expandafter\def\csname cdnum@e3.comptwoll.twenty.acc\endcsname{0.982}
\expandafter\def\csname cdnum@e3.decider.all.acc\endcsname{0.860}
\expandafter\def\csname cdnum@e3.decider.all.ece\endcsname{0.035}
\expandafter\def\csname cdnum@e3.decider.drop\endcsname{0.135}
\expandafter\def\csname cdnum@e3.decider.fifty.acc\endcsname{0.947}
\expandafter\def\csname cdnum@e3.decider.five.acc\endcsname{0.995}
\expandafter\def\csname cdnum@e3.decider.hier.acc\endcsname{0.518}
\expandafter\def\csname cdnum@e3.decider.twenty.acc\endcsname{0.978}
\expandafter\def\csname cdnum@e3.jev.all.acc\endcsname{0.913}
\expandafter\def\csname cdnum@e3.jev.all.ece\endcsname{0.032}
\expandafter\def\csname cdnum@e3.jev.drop\endcsname{0.077}
\expandafter\def\csname cdnum@e3.jev.fifty.acc\endcsname{0.957}
\expandafter\def\csname cdnum@e3.jev.five.acc\endcsname{0.990}
\expandafter\def\csname cdnum@e3.jev.hier.acc\endcsname{0.673}
\expandafter\def\csname cdnum@e3.jev.hier.domacc\endcsname{0.703}
\expandafter\def\csname cdnum@e3.jev.twenty.acc\endcsname{0.975}
\expandafter\def\csname cdnum@e3.kevlarge.all.acc\endcsname{0.850}
\expandafter\def\csname cdnum@e3.kevlarge.all.ece\endcsname{0.377}
\expandafter\def\csname cdnum@e3.kevlarge.drop\endcsname{0.130}
\expandafter\def\csname cdnum@e3.kevlarge.fifty.acc\endcsname{0.920}
\expandafter\def\csname cdnum@e3.kevlarge.five.acc\endcsname{0.980}
\expandafter\def\csname cdnum@e3.kevlarge.hier.acc\endcsname{0.602}
\expandafter\def\csname cdnum@e3.kevlarge.twenty.acc\endcsname{0.952}
\expandafter\def\csname cdnum@e3.kevsmall.all.acc\endcsname{0.685}
\expandafter\def\csname cdnum@e3.kevsmall.all.ece\endcsname{0.254}
\expandafter\def\csname cdnum@e3.kevsmall.drop\endcsname{0.253}
\expandafter\def\csname cdnum@e3.kevsmall.fifty.acc\endcsname{0.750}
\expandafter\def\csname cdnum@e3.kevsmall.five.acc\endcsname{0.938}
\expandafter\def\csname cdnum@e3.kevsmall.hier.acc\endcsname{0.382}
\expandafter\def\csname cdnum@e3.kevsmall.twenty.acc\endcsname{0.838}
\expandafter\def\csname cdnum@e3.layaen.fifty.acc\endcsname{0.020}
\expandafter\def\csname cdnum@e3.layaen.five.acc\endcsname{0.932}
\expandafter\def\csname cdnum@e3.layaen.hier.acc\endcsname{0.365}
\expandafter\def\csname cdnum@e3.layaen.twenty.acc\endcsname{0.852}
\expandafter\def\csname cdnum@e3.layaml.all.acc\endcsname{0.005}
\expandafter\def\csname cdnum@e3.layaml.all.ece\endcsname{0.031}
\expandafter\def\csname cdnum@e3.layaml.drop\endcsname{0.938}
\expandafter\def\csname cdnum@e3.layaml.fifty.acc\endcsname{0.065}
\expandafter\def\csname cdnum@e3.layaml.five.acc\endcsname{0.943}
\expandafter\def\csname cdnum@e3.layaml.hier.acc\endcsname{0.302}
\expandafter\def\csname cdnum@e3.layaml.twenty.acc\endcsname{0.830}
\expandafter\def\csname cdnum@e3.nimble.all.acc\endcsname{0.860}
\expandafter\def\csname cdnum@e3.nimble.all.ece\endcsname{0.069}
\expandafter\def\csname cdnum@e3.nimble.drop\endcsname{0.127}
\expandafter\def\csname cdnum@e3.nimble.fifty.acc\endcsname{0.930}
\expandafter\def\csname cdnum@e3.nimble.five.acc\endcsname{0.987}
\expandafter\def\csname cdnum@e3.nimble.hier.acc\endcsname{0.593}
\expandafter\def\csname cdnum@e3.nimble.twenty.acc\endcsname{0.975}
\expandafter\def\csname cdnum@e3.thisthat.all.acc\endcsname{0.863}
\expandafter\def\csname cdnum@e3.thisthat.all.ece\endcsname{0.055}
\expandafter\def\csname cdnum@e3.thisthat.drop\endcsname{0.132}
\expandafter\def\csname cdnum@e3.thisthat.fifty.acc\endcsname{0.938}
\expandafter\def\csname cdnum@e3.thisthat.five.acc\endcsname{0.995}
\expandafter\def\csname cdnum@e3.thisthat.hier.acc\endcsname{0.523}
\expandafter\def\csname cdnum@e3.thisthat.twenty.acc\endcsname{0.982}
\expandafter\def\csname cdnum@e3p.jev.fifty.accrange\endcsname{0.013}
\expandafter\def\csname cdnum@e3p.kevsmall.twenty.accrange\endcsname{0.035}
\expandafter\def\csname cdnum@e3p.layaen.twenty.accrange\endcsname{0.058}
\expandafter\def\csname cdnum@e5.compgem.d1neutral.usd\endcsname{0.0623}
\expandafter\def\csname cdnum@e5.compgem.d2k150.pfifty\endcsname{885}
\expandafter\def\csname cdnum@e5.compgem.d2k150.pninetyfive\endcsname{885}
\expandafter\def\csname cdnum@e5.compgem.d2k150.usd\endcsname{0.2893}
\expandafter\def\csname cdnum@e5.compgembone.d1neutral.pfifty\endcsname{28464}
\expandafter\def\csname cdnum@e5.compgembone.d1neutral.pninetyfive\endcsname{29260}
\expandafter\def\csname cdnum@e5.compgemll.d1neutral.usd\endcsname{0.0220}
\expandafter\def\csname cdnum@e5.compgemllbone.d1neutral.pfifty\endcsname{668}
\expandafter\def\csname cdnum@e5.compgemllbone.d1neutral.pninetyfive\endcsname{677}
\expandafter\def\csname cdnum@e5.compmis.d1neutral.usd\endcsname{0.0427}
\expandafter\def\csname cdnum@e5.compmisbone.d1neutral.pfifty\endcsname{12670}
\expandafter\def\csname cdnum@e5.compmisbone.d1neutral.pninetyfive\endcsname{12874}
\expandafter\def\csname cdnum@e5.compmisll.d1neutral.usd\endcsname{0.0136}
\expandafter\def\csname cdnum@e5.compmisll.d2k150.usd\endcsname{0.1037}
\expandafter\def\csname cdnum@e5.compmisllbone.d1neutral.pfifty\endcsname{358}
\expandafter\def\csname cdnum@e5.compmisllbone.d1neutral.pninetyfive\endcsname{369}
\expandafter\def\csname cdnum@e5.compopen.d1neutral.usd\endcsname{0.0265}
\expandafter\def\csname cdnum@e5.compopenbone.d1neutral.pfifty\endcsname{11330}
\expandafter\def\csname cdnum@e5.compopenbone.d1neutral.pninetyfive\endcsname{11514}
\expandafter\def\csname cdnum@e5.comptwo.d1neutral.usd\endcsname{0.0707}
\expandafter\def\csname cdnum@e5.comptwobone.d1neutral.pfifty\endcsname{8387}
\expandafter\def\csname cdnum@e5.comptwobone.d1neutral.pninetyfive\endcsname{8692}
\expandafter\def\csname cdnum@e5.comptwoll.d1neutral.usd\endcsname{0.0581}
\expandafter\def\csname cdnum@e5.comptwoll.d2k150.pfifty\endcsname{3609}
\expandafter\def\csname cdnum@e5.comptwoll.d2k150.pninetyfive\endcsname{3609}
\expandafter\def\csname cdnum@e5.comptwoll.d2k150.usd\endcsname{1.1814}
\expandafter\def\csname cdnum@e5.comptwollbone.d1neutral.pfifty\endcsname{997}
\expandafter\def\csname cdnum@e5.comptwollbone.d1neutral.pninetyfive\endcsname{1019}
\expandafter\def\csname cdnum@e5.comptwothink.d1neutral.usd\endcsname{0.6215}
\expandafter\def\csname cdnum@e5.decider.d1neutral.pfifty\endcsname{426}
\expandafter\def\csname cdnum@e5.decider.d1neutral.pninetyfive\endcsname{601}
\expandafter\def\csname cdnum@e5.decider.d1neutral.usd\endcsname{0.0109}
\expandafter\def\csname cdnum@e5.decider.d2k150.pfifty\endcsname{149}
\expandafter\def\csname cdnum@e5.decider.d2k150.pninetyfive\endcsname{154}
\expandafter\def\csname cdnum@e5.decider.d2k150.usd\endcsname{0.0202}
\expandafter\def\csname cdnum@e5.jev.cloud.n\endcsname{40}
\expandafter\def\csname cdnum@e5.jev.cloud.pfifty\endcsname{248}
\expandafter\def\csname cdnum@e5.jev.cloudlocal.pfifty\endcsname{139}
\expandafter\def\csname cdnum@e5.jev.d1neutral.pfifty\endcsname{138}
\expandafter\def\csname cdnum@e5.jev.d1neutral.pninetyfive\endcsname{183}
\expandafter\def\csname cdnum@e5.jev.d1neutral.usd\endcsname{0.0079}
\expandafter\def\csname cdnum@e5.jev.d2k150.pfifty\endcsname{148}
\expandafter\def\csname cdnum@e5.jev.d2k150.pninetyfive\endcsname{212}
\expandafter\def\csname cdnum@e5.jev.d2k150.usd\endcsname{0.1095}
\expandafter\def\csname cdnum@e5.kevlarge.d1neutral.pfifty\endcsname{590}
\expandafter\def\csname cdnum@e5.kevlarge.d1neutral.pninetyfive\endcsname{798}
\expandafter\def\csname cdnum@e5.kevlarge.d1neutral.usd\endcsname{0.0147}
\expandafter\def\csname cdnum@e5.kevlarge.d2k150.pfifty\endcsname{428}
\expandafter\def\csname cdnum@e5.kevlarge.d2k150.pninetyfive\endcsname{435}
\expandafter\def\csname cdnum@e5.kevlarge.d2k150.usd\endcsname{0.0573}
\expandafter\def\csname cdnum@e5.kevsmall.d1neutral.pfifty\endcsname{84}
\expandafter\def\csname cdnum@e5.kevsmall.d1neutral.pninetyfive\endcsname{125}
\expandafter\def\csname cdnum@e5.kevsmall.d1neutral.usd\endcsname{0.0025}
\expandafter\def\csname cdnum@e5.layaen.d1neutral.pfifty\endcsname{60}
\expandafter\def\csname cdnum@e5.layaen.d1neutral.pninetyfive\endcsname{80}
\expandafter\def\csname cdnum@e5.layaen.d1neutral.usd\endcsname{0.0016}
\expandafter\def\csname cdnum@e5.layaml.d1neutral.pfifty\endcsname{32}
\expandafter\def\csname cdnum@e5.layaml.d1neutral.pninetyfive\endcsname{38}
\expandafter\def\csname cdnum@e5.layaml.d1neutral.usd\endcsname{0.0009}
\expandafter\def\csname cdnum@e5.nimble.d1neutral.pfifty\endcsname{2056}
\expandafter\def\csname cdnum@e5.nimble.d1neutral.pninetyfive\endcsname{2466}
\expandafter\def\csname cdnum@e5.nimble.d1neutral.usd\endcsname{0.0532}
\expandafter\def\csname cdnum@e5.thisthat.d1neutral.pfifty\endcsname{71}
\expandafter\def\csname cdnum@e5.thisthat.d1neutral.pninetyfive\endcsname{74}
\expandafter\def\csname cdnum@e5.thisthat.d1neutral.usd\endcsname{0.0020}
\expandafter\def\csname cdnum@e6.decider.compgem.doneneutral.accfirst\endcsname{0.588}
\expandafter\def\csname cdnum@e6.decider.compgem.doneneutral.accsecond\endcsname{0.741}
\expandafter\def\csname cdnum@e6.decider.compgem.dtwo.accfirst\endcsname{0.860}
\expandafter\def\csname cdnum@e6.decider.compgem.dtwo.accsecond\endcsname{0.942}
\expandafter\def\csname cdnum@e6.decider.compopen.doneneutral.accfirst\endcsname{0.588}
\expandafter\def\csname cdnum@e6.decider.compopen.doneneutral.accsecond\endcsname{0.680}
\expandafter\def\csname cdnum@e6.decider.compopen.dtwo.accfirst\endcsname{0.860}
\expandafter\def\csname cdnum@e6.decider.compopen.dtwo.accsecond\endcsname{0.728}
\expandafter\def\csname cdnum@e6.decider.comptwo.doneneutral.accfirst\endcsname{0.588}
\expandafter\def\csname cdnum@e6.decider.comptwo.doneneutral.accsecond\endcsname{0.725}
\expandafter\def\csname cdnum@e6.decider.comptwo.dtwo.accfirst\endcsname{0.860}
\expandafter\def\csname cdnum@e6.decider.comptwo.dtwo.accsecond\endcsname{0.683}
\expandafter\def\csname cdnum@e6.decider.comptwoll.doneneutral.accfirst\endcsname{0.588}
\expandafter\def\csname cdnum@e6.decider.comptwoll.doneneutral.accsecond\endcsname{0.733}
\expandafter\def\csname cdnum@e6.decider.comptwoll.dtwo.accfirst\endcsname{0.860}
\expandafter\def\csname cdnum@e6.decider.comptwoll.dtwo.accsecond\endcsname{0.912}
\expandafter\def\csname cdnum@e6.decider.jev.doneneutral.accfirst\endcsname{0.588}
\expandafter\def\csname cdnum@e6.decider.jev.doneneutral.accsecond\endcsname{0.732}
\expandafter\def\csname cdnum@e6.decider.jev.dtwo.accfirst\endcsname{0.860}
\expandafter\def\csname cdnum@e6.decider.jev.dtwo.accsecond\endcsname{0.913}
\expandafter\def\csname cdnum@e6.jev.compgem.doneneutral.accfirst\endcsname{0.732}
\expandafter\def\csname cdnum@e6.jev.compgem.doneneutral.accsecond\endcsname{0.741}
\expandafter\def\csname cdnum@e6.jev.compgem.dtwo.accfirst\endcsname{0.913}
\expandafter\def\csname cdnum@e6.jev.compgem.dtwo.accsecond\endcsname{0.942}
\expandafter\def\csname cdnum@e6.jev.compopen.doneneutral.accfirst\endcsname{0.732}
\expandafter\def\csname cdnum@e6.jev.compopen.doneneutral.accsecond\endcsname{0.680}
\expandafter\def\csname cdnum@e6.jev.compopen.dtwo.accfirst\endcsname{0.913}
\expandafter\def\csname cdnum@e6.jev.compopen.dtwo.accsecond\endcsname{0.728}
\expandafter\def\csname cdnum@e6.jev.comptwo.doneneutral.accfirst\endcsname{0.732}
\expandafter\def\csname cdnum@e6.jev.comptwo.doneneutral.accsecond\endcsname{0.725}
\expandafter\def\csname cdnum@e6.jev.comptwo.dtwo.accfirst\endcsname{0.913}
\expandafter\def\csname cdnum@e6.jev.comptwo.dtwo.accsecond\endcsname{0.683}
\expandafter\def\csname cdnum@e6.jev.comptwoll.doneneutral.accfirst\endcsname{0.732}
\expandafter\def\csname cdnum@e6.jev.comptwoll.doneneutral.accsecond\endcsname{0.733}
\expandafter\def\csname cdnum@e6.jev.comptwoll.dtwo.accfirst\endcsname{0.913}
\expandafter\def\csname cdnum@e6.jev.comptwoll.dtwo.accsecond\endcsname{0.912}
\expandafter\def\csname cdnum@e6.kevlarge.compgem.doneneutral.accfirst\endcsname{0.712}
\expandafter\def\csname cdnum@e6.kevlarge.compgem.doneneutral.accsecond\endcsname{0.741}
\expandafter\def\csname cdnum@e6.kevlarge.compgem.dtwo.accfirst\endcsname{0.850}
\expandafter\def\csname cdnum@e6.kevlarge.compgem.dtwo.accsecond\endcsname{0.942}
\expandafter\def\csname cdnum@e6.kevlarge.compopen.doneneutral.accfirst\endcsname{0.712}
\expandafter\def\csname cdnum@e6.kevlarge.compopen.doneneutral.accsecond\endcsname{0.680}
\expandafter\def\csname cdnum@e6.kevlarge.compopen.dtwo.accfirst\endcsname{0.850}
\expandafter\def\csname cdnum@e6.kevlarge.compopen.dtwo.accsecond\endcsname{0.728}
\expandafter\def\csname cdnum@e6.kevlarge.comptwo.doneneutral.accfirst\endcsname{0.712}
\expandafter\def\csname cdnum@e6.kevlarge.comptwo.doneneutral.accsecond\endcsname{0.725}
\expandafter\def\csname cdnum@e6.kevlarge.comptwo.dtwo.accfirst\endcsname{0.850}
\expandafter\def\csname cdnum@e6.kevlarge.comptwo.dtwo.accsecond\endcsname{0.683}
\expandafter\def\csname cdnum@e6.kevlarge.comptwoll.doneneutral.accfirst\endcsname{0.712}
\expandafter\def\csname cdnum@e6.kevlarge.comptwoll.doneneutral.accsecond\endcsname{0.733}
\expandafter\def\csname cdnum@e6.kevlarge.comptwoll.dtwo.accfirst\endcsname{0.850}
\expandafter\def\csname cdnum@e6.kevlarge.comptwoll.dtwo.accsecond\endcsname{0.912}
\expandafter\def\csname cdnum@e6.kevlarge.jev.doneneutral.accfirst\endcsname{0.712}
\expandafter\def\csname cdnum@e6.kevlarge.jev.doneneutral.accsecond\endcsname{0.732}
\expandafter\def\csname cdnum@e6.kevlarge.jev.dtwo.accfirst\endcsname{0.850}
\expandafter\def\csname cdnum@e6.kevlarge.jev.dtwo.accsecond\endcsname{0.913}
\expandafter\def\csname cdnum@e6.kevsmall.compgem.doneneutral.accfirst\endcsname{0.451}
\expandafter\def\csname cdnum@e6.kevsmall.compgem.doneneutral.accsecond\endcsname{0.741}
\expandafter\def\csname cdnum@e6.kevsmall.compgem.dtwo.accfirst\endcsname{0.685}
\expandafter\def\csname cdnum@e6.kevsmall.compgem.dtwo.accsecond\endcsname{0.942}
\expandafter\def\csname cdnum@e6.kevsmall.compopen.doneneutral.accfirst\endcsname{0.451}
\expandafter\def\csname cdnum@e6.kevsmall.compopen.doneneutral.accsecond\endcsname{0.680}
\expandafter\def\csname cdnum@e6.kevsmall.compopen.dtwo.accfirst\endcsname{0.685}
\expandafter\def\csname cdnum@e6.kevsmall.compopen.dtwo.accsecond\endcsname{0.728}
\expandafter\def\csname cdnum@e6.kevsmall.comptwo.doneneutral.accfirst\endcsname{0.451}
\expandafter\def\csname cdnum@e6.kevsmall.comptwo.doneneutral.accsecond\endcsname{0.725}
\expandafter\def\csname cdnum@e6.kevsmall.comptwo.dtwo.accfirst\endcsname{0.685}
\expandafter\def\csname cdnum@e6.kevsmall.comptwo.dtwo.accsecond\endcsname{0.683}
\expandafter\def\csname cdnum@e6.kevsmall.comptwoll.doneneutral.accfirst\endcsname{0.451}
\expandafter\def\csname cdnum@e6.kevsmall.comptwoll.doneneutral.accsecond\endcsname{0.733}
\expandafter\def\csname cdnum@e6.kevsmall.comptwoll.dtwo.accfirst\endcsname{0.685}
\expandafter\def\csname cdnum@e6.kevsmall.comptwoll.dtwo.accsecond\endcsname{0.912}
\expandafter\def\csname cdnum@e6.kevsmall.jev.doneneutral.accfirst\endcsname{0.451}
\expandafter\def\csname cdnum@e6.kevsmall.jev.doneneutral.accsecond\endcsname{0.732}
\expandafter\def\csname cdnum@e6.kevsmall.jev.dtwo.accfirst\endcsname{0.685}
\expandafter\def\csname cdnum@e6.kevsmall.jev.dtwo.accsecond\endcsname{0.913}
\expandafter\def\csname cdnum@e6.layaen.compgem.doneneutral.accfirst\endcsname{0.366}
\expandafter\def\csname cdnum@e6.layaen.compgem.doneneutral.accsecond\endcsname{0.741}
\expandafter\def\csname cdnum@e6.layaen.compopen.doneneutral.accfirst\endcsname{0.366}
\expandafter\def\csname cdnum@e6.layaen.compopen.doneneutral.accsecond\endcsname{0.680}
\expandafter\def\csname cdnum@e6.layaen.comptwo.doneneutral.accfirst\endcsname{0.366}
\expandafter\def\csname cdnum@e6.layaen.comptwo.doneneutral.accsecond\endcsname{0.725}
\expandafter\def\csname cdnum@e6.layaen.comptwoll.doneneutral.accfirst\endcsname{0.366}
\expandafter\def\csname cdnum@e6.layaen.comptwoll.doneneutral.accsecond\endcsname{0.733}
\expandafter\def\csname cdnum@e6.layaen.jev.doneneutral.accfirst\endcsname{0.366}
\expandafter\def\csname cdnum@e6.layaen.jev.doneneutral.accsecond\endcsname{0.732}
\expandafter\def\csname cdnum@e6.layaml.compgem.doneneutral.accfirst\endcsname{0.349}
\expandafter\def\csname cdnum@e6.layaml.compgem.doneneutral.accsecond\endcsname{0.741}
\expandafter\def\csname cdnum@e6.layaml.compgem.dtwo.accfirst\endcsname{0.005}
\expandafter\def\csname cdnum@e6.layaml.compgem.dtwo.accsecond\endcsname{0.942}
\expandafter\def\csname cdnum@e6.layaml.compopen.doneneutral.accfirst\endcsname{0.349}
\expandafter\def\csname cdnum@e6.layaml.compopen.doneneutral.accsecond\endcsname{0.680}
\expandafter\def\csname cdnum@e6.layaml.compopen.dtwo.accfirst\endcsname{0.005}
\expandafter\def\csname cdnum@e6.layaml.compopen.dtwo.accsecond\endcsname{0.728}
\expandafter\def\csname cdnum@e6.layaml.comptwo.doneneutral.accfirst\endcsname{0.349}
\expandafter\def\csname cdnum@e6.layaml.comptwo.doneneutral.accsecond\endcsname{0.725}
\expandafter\def\csname cdnum@e6.layaml.comptwo.dtwo.accfirst\endcsname{0.005}
\expandafter\def\csname cdnum@e6.layaml.comptwo.dtwo.accsecond\endcsname{0.683}
\expandafter\def\csname cdnum@e6.layaml.comptwoll.doneneutral.accfirst\endcsname{0.349}
\expandafter\def\csname cdnum@e6.layaml.comptwoll.doneneutral.accsecond\endcsname{0.733}
\expandafter\def\csname cdnum@e6.layaml.comptwoll.dtwo.accfirst\endcsname{0.005}
\expandafter\def\csname cdnum@e6.layaml.comptwoll.dtwo.accsecond\endcsname{0.912}
\expandafter\def\csname cdnum@e6.layaml.jev.doneneutral.accfirst\endcsname{0.349}
\expandafter\def\csname cdnum@e6.layaml.jev.doneneutral.accsecond\endcsname{0.732}
\expandafter\def\csname cdnum@e6.layaml.jev.dtwo.accfirst\endcsname{0.005}
\expandafter\def\csname cdnum@e6.layaml.jev.dtwo.accsecond\endcsname{0.913}
\expandafter\def\csname cdnum@e6.nimble.compgem.doneneutral.accfirst\endcsname{0.685}
\expandafter\def\csname cdnum@e6.nimble.compgem.doneneutral.accsecond\endcsname{0.741}
\expandafter\def\csname cdnum@e6.nimble.compgem.dtwo.accfirst\endcsname{0.860}
\expandafter\def\csname cdnum@e6.nimble.compgem.dtwo.accsecond\endcsname{0.942}
\expandafter\def\csname cdnum@e6.nimble.compopen.doneneutral.accfirst\endcsname{0.685}
\expandafter\def\csname cdnum@e6.nimble.compopen.doneneutral.accsecond\endcsname{0.680}
\expandafter\def\csname cdnum@e6.nimble.compopen.dtwo.accfirst\endcsname{0.860}
\expandafter\def\csname cdnum@e6.nimble.compopen.dtwo.accsecond\endcsname{0.728}
\expandafter\def\csname cdnum@e6.nimble.comptwo.doneneutral.accfirst\endcsname{0.685}
\expandafter\def\csname cdnum@e6.nimble.comptwo.doneneutral.accsecond\endcsname{0.725}
\expandafter\def\csname cdnum@e6.nimble.comptwo.dtwo.accfirst\endcsname{0.860}
\expandafter\def\csname cdnum@e6.nimble.comptwo.dtwo.accsecond\endcsname{0.683}
\expandafter\def\csname cdnum@e6.nimble.comptwoll.doneneutral.accfirst\endcsname{0.685}
\expandafter\def\csname cdnum@e6.nimble.comptwoll.doneneutral.accsecond\endcsname{0.733}
\expandafter\def\csname cdnum@e6.nimble.comptwoll.dtwo.accfirst\endcsname{0.860}
\expandafter\def\csname cdnum@e6.nimble.comptwoll.dtwo.accsecond\endcsname{0.912}
\expandafter\def\csname cdnum@e6.nimble.jev.doneneutral.accfirst\endcsname{0.685}
\expandafter\def\csname cdnum@e6.nimble.jev.doneneutral.accsecond\endcsname{0.732}
\expandafter\def\csname cdnum@e6.nimble.jev.dtwo.accfirst\endcsname{0.860}
\expandafter\def\csname cdnum@e6.nimble.jev.dtwo.accsecond\endcsname{0.913}
\expandafter\def\csname cdnum@e6.thisthat.compgem.doneneutral.accfirst\endcsname{0.597}
\expandafter\def\csname cdnum@e6.thisthat.compgem.doneneutral.accsecond\endcsname{0.741}
\expandafter\def\csname cdnum@e6.thisthat.compgem.dtwo.accfirst\endcsname{0.863}
\expandafter\def\csname cdnum@e6.thisthat.compgem.dtwo.accsecond\endcsname{0.942}
\expandafter\def\csname cdnum@e6.thisthat.compopen.doneneutral.accfirst\endcsname{0.597}
\expandafter\def\csname cdnum@e6.thisthat.compopen.doneneutral.accsecond\endcsname{0.680}
\expandafter\def\csname cdnum@e6.thisthat.compopen.dtwo.accfirst\endcsname{0.863}
\expandafter\def\csname cdnum@e6.thisthat.compopen.dtwo.accsecond\endcsname{0.728}
\expandafter\def\csname cdnum@e6.thisthat.comptwo.doneneutral.accfirst\endcsname{0.597}
\expandafter\def\csname cdnum@e6.thisthat.comptwo.doneneutral.accsecond\endcsname{0.725}
\expandafter\def\csname cdnum@e6.thisthat.comptwo.dtwo.accfirst\endcsname{0.863}
\expandafter\def\csname cdnum@e6.thisthat.comptwo.dtwo.accsecond\endcsname{0.683}
\expandafter\def\csname cdnum@e6.thisthat.comptwoll.doneneutral.accfirst\endcsname{0.597}
\expandafter\def\csname cdnum@e6.thisthat.comptwoll.doneneutral.accsecond\endcsname{0.733}
\expandafter\def\csname cdnum@e6.thisthat.comptwoll.dtwo.accfirst\endcsname{0.863}
\expandafter\def\csname cdnum@e6.thisthat.comptwoll.dtwo.accsecond\endcsname{0.912}
\expandafter\def\csname cdnum@e6.thisthat.jev.doneneutral.accfirst\endcsname{0.597}
\expandafter\def\csname cdnum@e6.thisthat.jev.doneneutral.accsecond\endcsname{0.732}
\expandafter\def\csname cdnum@e6.thisthat.jev.dtwo.accfirst\endcsname{0.863}
\expandafter\def\csname cdnum@e6.thisthat.jev.dtwo.accsecond\endcsname{0.913}
\expandafter\def\csname cdnum@lb.layaen.all.dropped\endcsname{24}
\expandafter\def\csname cdnum@lb.layaen.all.maxlen\endcsname{512}
\expandafter\def\csname cdnum@lb.layaen.fifty.head\endcsname{192}
\expandafter\def\csname cdnum@lb.layaen.fifty.opttokens\endcsname{302}
\expandafter\def\csname cdnum@lb.layaml.fifty.head\endcsname{256}
\expandafter\def\csname cdnum@lit.cheng2026thisthatmodel.thisThatAccuracy68q\endcsname{0.941}
\expandafter\def\csname cdnum@lit.cheng2026thisthatmodel.thisThatAccuracy68q.cmp\endcsname{0.765}
\expandafter\def\csname cdnum@lit.deusser2026evaluating.datasets\endcsname{37}
\expandafter\def\csname cdnum@lit.deusser2026evaluating.jevBankingAcc\endcsname{0.797}
\expandafter\def\csname cdnum@lit.deusser2026evaluating.jevClincInScope\endcsname{0.921}
\expandafter\def\csname cdnum@lit.deusser2026evaluating.jevOosRecall\endcsname{0.776}
\expandafter\def\csname cdnum@lit.deusser2026evaluating.jevPooledEce\endcsname{0.028}
\expandafter\def\csname cdnum@lit.greyKev.kev9bOodEce\endcsname{0.042}
\expandafter\def\csname cdnum@lit.greyLaya.layaEnglishZeroShotAccuracy\endcsname{0.362}
\expandafter\def\csname cdnum@lit.ibrahim2026evaluating.BespokeNimble9BMedianF1\endcsname{53.8}
\expandafter\def\csname cdnum@lit.ibrahim2026evaluating.decider2bMedianF1\endcsname{45.2}
\expandafter\def\csname cdnum@lit.ibrahim2026evaluating.jevGeminiCascadeCostFracAt0.8\endcsname{0.561}
\expandafter\def\csname cdnum@lit.ibrahim2026evaluating.jevGemmaCascadeCostFracAt0.8\endcsname{1.13}
\expandafter\def\csname cdnum@lit.ibrahim2026evaluating.jevMedianECE\endcsname{0.157}
\expandafter\def\csname cdnum@lit.ibrahim2026evaluating.jevMedianECETempScaled\endcsname{0.121}
\expandafter\def\csname cdnum@lit.li2026jevasajudge.jevAccuracyHighConfidenceQGe09\endcsname{95.8}
\expandafter\def\csname cdnum@lit.li2026jevasajudge.jevAccuracyLowConfidenceQLt06\endcsname{47.7}
\expandafter\def\csname cdnum@lit.li2026jevasajudge.jevFrozenCascadeAccuracyGpt6Tau09\endcsname{92.5}
\expandafter\def\csname cdnum@lit.li2026jevasajudge.jevFrozenCascadeFeePctGpt6Tau09\endcsname{56.8}
\expandafter\def\csname cdnum@lit.ling2026jev.projects\endcsname{2,170}
\expandafter\def\csname cdnum@lit.sun2026typesafe.hostedJevFlipRate\endcsname{32.5}
\expandafter\def\csname cdnum@lit.sun2026typesafe.markerNoYesFlipRate\endcsname{76.92}
\expandafter\def\csname cdnum@lit.webDecisionIndex.deciderBankingF1\endcsname{72.8}
\expandafter\def\csname cdnum@lit.webDecisionIndex.engines\endcsname{70}
\expandafter\def\csname cdnum@lit.webDecisionIndex.jevBankingF1\endcsname{79.7}
\expandafter\def\csname cdnum@lit.webDecisionIndex.kevlargeBankingF1\endcsname{84.8}
\expandafter\def\csname cdnum@lit.webDecisionIndex.kevsmallBankingF1\endcsname{76.2}
\expandafter\def\csname cdnum@lit.webDecisionIndex.panel\endcsname{38}
\expandafter\def\csname cdnum@lit.webDecisionIndex.requests\endcsname{120,340}
\expandafter\def\csname cdnum@m.comparator.quantbits\endcsname{4}
\expandafter\def\csname cdnum@m.comparator.temp\endcsname{0.0}
\expandafter\def\csname cdnum@m.decider.baseparams\endcsname{1.9B}
\expandafter\def\csname cdnum@m.decider.context\endcsname{32}
\expandafter\def\csname cdnum@m.decider.temp\endcsname{1.30}
\expandafter\def\csname cdnum@m.kev.lorarank\endcsname{16}
\expandafter\def\csname cdnum@m.kev08b.temp\endcsname{2.35}
\expandafter\def\csname cdnum@m.kev9b.temp\endcsname{2.30}
\expandafter\def\csname cdnum@m.laya.en.context\endcsname{512}
\expandafter\def\csname cdnum@m.laya.en.params\endcsname{421M}
\expandafter\def\csname cdnum@m.laya.ml.context\endcsname{1,024}
\expandafter\def\csname cdnum@m.laya.ml.params\endcsname{322M}
\expandafter\def\csname cdnum@m.laya.temp\endcsname{1.0}
\expandafter\def\csname cdnum@m.likelihood.temp\endcsname{1.0}
\expandafter\def\csname cdnum@m.nimble.context\endcsname{8,192}
\expandafter\def\csname cdnum@m.nimble.temp\endcsname{1.0}
\expandafter\def\csname cdnum@m.think.maxtokens\endcsname{8192}
\expandafter\def\csname cdnum@m.think.temp\endcsname{0.6}
\expandafter\def\csname cdnum@m.think.topp\endcsname{0.95}
\expandafter\def\csname cdnum@m.thisthat.params\endcsname{1.88B}
\expandafter\def\csname cdnum@m.thisthat.temp\endcsname{1.0}
\expandafter\def\csname cdnum@model.version\endcsname{jev-1.13.0}
\expandafter\def\csname cdnum@n.d1neutral.decisions\endcsname{2,000}
\expandafter\def\csname cdnum@n.d1neutral.requests\endcsname{400}
\expandafter\def\csname cdnum@n.d2banking77.decisions\endcsname{616}
\expandafter\def\csname cdnum@n.d2banking77.requests\endcsname{616}
\expandafter\def\csname cdnum@n.d2k150.decisions\endcsname{800}
\expandafter\def\csname cdnum@n.d2k150.requests\endcsname{800}
\expandafter\def\csname cdnum@n.d2k5.decisions\endcsname{600}
\expandafter\def\csname cdnum@n.d2k5.requests\endcsname{600}
\expandafter\def\csname cdnum@n.d3convgoawry.decisions\endcsname{500}
\expandafter\def\csname cdnum@n.d3emotion.decisions\endcsname{498}
\expandafter\def\csname cdnum@n.d3goemotions.decisions\endcsname{492}
\expandafter\def\csname cdnum@n.d3goemotions.requests\endcsname{492}
\expandafter\def\csname cdnum@n.d3wikicorpus.decisions\endcsname{500}
\expandafter\def\csname cdnum@n.d3wikipoliteness.decisions\endcsname{498}
\expandafter\def\csname cdnum@n.retest.states\endcsname{40}
\expandafter\def\csname cdnum@nt.bgebanking.banking.acc\endcsname{0.937}
\expandafter\def\csname cdnum@nt.bgebanking.banking.diff\endcsname{11.5}
\expandafter\def\csname cdnum@nt.bgebanking.banking.ece\endcsname{0.057}
\expandafter\def\csname cdnum@nt.bgebanking.banking.fone\endcsname{93.6}
\expandafter\def\csname cdnum@nt.bgegoemo.goemo.acc\endcsname{0.677}
\expandafter\def\csname cdnum@nt.bgegoemo.goemo.diff\endcsname{7.3}
\expandafter\def\csname cdnum@nt.bgegoemo.goemo.ece\endcsname{0.076}
\expandafter\def\csname cdnum@nt.bgegoemo.goemo.fone\endcsname{67.6}
\expandafter\def\csname cdnum@nt.compgem.banking.acc\endcsname{0.821}
\expandafter\def\csname cdnum@nt.compgem.banking.diff\endcsname{0.0}
\expandafter\def\csname cdnum@nt.compgem.banking.ece\endcsname{0.071}
\expandafter\def\csname cdnum@nt.compgem.banking.fone\endcsname{80.8}
\expandafter\def\csname cdnum@nt.compgem.goemo.acc\endcsname{0.573}
\expandafter\def\csname cdnum@nt.compgem.goemo.diff\endcsname{\ensuremath{-}3.0}
\expandafter\def\csname cdnum@nt.compgem.goemo.ece\endcsname{0.162}
\expandafter\def\csname cdnum@nt.compgem.goemo.fone\endcsname{57.8}
\expandafter\def\csname cdnum@nt.compgemll.banking.acc\endcsname{0.818}
\expandafter\def\csname cdnum@nt.compgemll.banking.diff\endcsname{\ensuremath{-}0.3}
\expandafter\def\csname cdnum@nt.compgemll.banking.ece\endcsname{0.176}
\expandafter\def\csname cdnum@nt.compgemll.banking.fone\endcsname{80.3}
\expandafter\def\csname cdnum@nt.compgemll.goemo.acc\endcsname{0.571}
\expandafter\def\csname cdnum@nt.compgemll.goemo.diff\endcsname{\ensuremath{-}3.3}
\expandafter\def\csname cdnum@nt.compgemll.goemo.ece\endcsname{0.416}
\expandafter\def\csname cdnum@nt.compgemll.goemo.fone\endcsname{57.5}
\expandafter\def\csname cdnum@nt.compmis.banking.acc\endcsname{0.679}
\expandafter\def\csname cdnum@nt.compmis.banking.diff\endcsname{\ensuremath{-}14.3}
\expandafter\def\csname cdnum@nt.compmis.banking.ece\endcsname{0.168}
\expandafter\def\csname cdnum@nt.compmis.banking.fone\endcsname{66.1}
\expandafter\def\csname cdnum@nt.compmis.goemo.acc\endcsname{0.226}
\expandafter\def\csname cdnum@nt.compmis.goemo.diff\endcsname{\ensuremath{-}37.8}
\expandafter\def\csname cdnum@nt.compmis.goemo.ece\endcsname{0.211}
\expandafter\def\csname cdnum@nt.compmis.goemo.fone\endcsname{18.9}
\expandafter\def\csname cdnum@nt.compmisll.banking.acc\endcsname{0.662}
\expandafter\def\csname cdnum@nt.compmisll.banking.diff\endcsname{\ensuremath{-}15.9}
\expandafter\def\csname cdnum@nt.compmisll.banking.ece\endcsname{0.057}
\expandafter\def\csname cdnum@nt.compmisll.banking.fone\endcsname{64.0}
\expandafter\def\csname cdnum@nt.compmisll.goemo.acc\endcsname{0.577}
\expandafter\def\csname cdnum@nt.compmisll.goemo.diff\endcsname{\ensuremath{-}2.6}
\expandafter\def\csname cdnum@nt.compmisll.goemo.ece\endcsname{0.210}
\expandafter\def\csname cdnum@nt.compmisll.goemo.fone\endcsname{57.8}
\expandafter\def\csname cdnum@nt.compopen.banking.acc\endcsname{0.659}
\expandafter\def\csname cdnum@nt.compopen.banking.diff\endcsname{\ensuremath{-}16.2}
\expandafter\def\csname cdnum@nt.compopen.banking.ece\endcsname{0.237}
\expandafter\def\csname cdnum@nt.compopen.banking.fone\endcsname{65.0}
\expandafter\def\csname cdnum@nt.compopen.goemo.acc\endcsname{0.585}
\expandafter\def\csname cdnum@nt.compopen.goemo.diff\endcsname{\ensuremath{-}1.8}
\expandafter\def\csname cdnum@nt.compopen.goemo.ece\endcsname{0.100}
\expandafter\def\csname cdnum@nt.compopen.goemo.fone\endcsname{58.8}
\expandafter\def\csname cdnum@nt.comptwo.banking.acc\endcsname{0.765}
\expandafter\def\csname cdnum@nt.comptwo.banking.diff\endcsname{\ensuremath{-}5.7}
\expandafter\def\csname cdnum@nt.comptwo.banking.ece\endcsname{0.106}
\expandafter\def\csname cdnum@nt.comptwo.banking.fone\endcsname{75.3}
\expandafter\def\csname cdnum@nt.comptwo.goemo.acc\endcsname{0.530}
\expandafter\def\csname cdnum@nt.comptwo.goemo.diff\endcsname{\ensuremath{-}7.3}
\expandafter\def\csname cdnum@nt.comptwo.goemo.ece\endcsname{0.211}
\expandafter\def\csname cdnum@nt.comptwo.goemo.fone\endcsname{53.0}
\expandafter\def\csname cdnum@nt.comptwoll.banking.acc\endcsname{0.800}
\expandafter\def\csname cdnum@nt.comptwoll.banking.diff\endcsname{\ensuremath{-}2.1}
\expandafter\def\csname cdnum@nt.comptwoll.banking.ece\endcsname{0.101}
\expandafter\def\csname cdnum@nt.comptwoll.banking.fone\endcsname{77.7}
\expandafter\def\csname cdnum@nt.comptwoll.goemo.acc\endcsname{0.585}
\expandafter\def\csname cdnum@nt.comptwoll.goemo.diff\endcsname{\ensuremath{-}1.8}
\expandafter\def\csname cdnum@nt.comptwoll.goemo.ece\endcsname{0.225}
\expandafter\def\csname cdnum@nt.comptwoll.goemo.fone\endcsname{58.7}
\expandafter\def\csname cdnum@nt.comptwothink.banking.acc\endcsname{0.825}
\expandafter\def\csname cdnum@nt.comptwothink.banking.diff\endcsname{0.3}
\expandafter\def\csname cdnum@nt.comptwothink.banking.ece\endcsname{0.113}
\expandafter\def\csname cdnum@nt.comptwothink.banking.fone\endcsname{80.9}
\expandafter\def\csname cdnum@nt.comptwothink.goemo.acc\endcsname{0.606}
\expandafter\def\csname cdnum@nt.comptwothink.goemo.diff\endcsname{0.2}
\expandafter\def\csname cdnum@nt.comptwothink.goemo.ece\endcsname{0.175}
\expandafter\def\csname cdnum@nt.comptwothink.goemo.fone\endcsname{60.5}
\expandafter\def\csname cdnum@nt.debbanking.banking.acc\endcsname{0.950}
\expandafter\def\csname cdnum@nt.debbanking.banking.diff\endcsname{12.8}
\expandafter\def\csname cdnum@nt.debbanking.banking.ece\endcsname{0.022}
\expandafter\def\csname cdnum@nt.debbanking.banking.fone\endcsname{94.8}
\expandafter\def\csname cdnum@nt.debgoemo.goemo.acc\endcsname{0.756}
\expandafter\def\csname cdnum@nt.debgoemo.goemo.diff\endcsname{15.2}
\expandafter\def\csname cdnum@nt.debgoemo.goemo.ece\endcsname{0.110}
\expandafter\def\csname cdnum@nt.debgoemo.goemo.fone\endcsname{75.7}
\expandafter\def\csname cdnum@nt.decider.banking.acc\endcsname{0.828}
\expandafter\def\csname cdnum@nt.decider.banking.diff\endcsname{0.6}
\expandafter\def\csname cdnum@nt.decider.banking.ece\endcsname{0.030}
\expandafter\def\csname cdnum@nt.decider.banking.fone\endcsname{81.0}
\expandafter\def\csname cdnum@nt.decider.goemo.acc\endcsname{0.618}
\expandafter\def\csname cdnum@nt.decider.goemo.diff\endcsname{1.4}
\expandafter\def\csname cdnum@nt.decider.goemo.ece\endcsname{0.072}
\expandafter\def\csname cdnum@nt.decider.goemo.fone\endcsname{60.6}
\expandafter\def\csname cdnum@nt.jev.banking.acc\endcsname{0.821}
\expandafter\def\csname cdnum@nt.jev.banking.ece\endcsname{0.073}
\expandafter\def\csname cdnum@nt.jev.banking.fone\endcsname{80.0}
\expandafter\def\csname cdnum@nt.jev.goemo.acc\endcsname{0.604}
\expandafter\def\csname cdnum@nt.jev.goemo.ece\endcsname{0.203}
\expandafter\def\csname cdnum@nt.jev.goemo.fone\endcsname{60.7}
\expandafter\def\csname cdnum@nt.kevlarge.banking.acc\endcsname{0.849}
\expandafter\def\csname cdnum@nt.kevlarge.banking.diff\endcsname{2.8}
\expandafter\def\csname cdnum@nt.kevlarge.banking.ece\endcsname{0.243}
\expandafter\def\csname cdnum@nt.kevlarge.banking.fone\endcsname{83.9}
\expandafter\def\csname cdnum@nt.kevlarge.banking.holm\endcsname{0.36}
\expandafter\def\csname cdnum@nt.kevlarge.goemo.acc\endcsname{0.608}
\expandafter\def\csname cdnum@nt.kevlarge.goemo.diff\endcsname{0.4}
\expandafter\def\csname cdnum@nt.kevlarge.goemo.ece\endcsname{0.092}
\expandafter\def\csname cdnum@nt.kevlarge.goemo.fone\endcsname{61.2}
\expandafter\def\csname cdnum@nt.kevsmall.banking.acc\endcsname{0.830}
\expandafter\def\csname cdnum@nt.kevsmall.banking.diff\endcsname{0.8}
\expandafter\def\csname cdnum@nt.kevsmall.banking.ece\endcsname{0.149}
\expandafter\def\csname cdnum@nt.kevsmall.banking.fone\endcsname{82.0}
\expandafter\def\csname cdnum@nt.kevsmall.goemo.acc\endcsname{0.555}
\expandafter\def\csname cdnum@nt.kevsmall.goemo.diff\endcsname{\ensuremath{-}4.9}
\expandafter\def\csname cdnum@nt.kevsmall.goemo.ece\endcsname{0.072}
\expandafter\def\csname cdnum@nt.kevsmall.goemo.fone\endcsname{55.2}
\expandafter\def\csname cdnum@nt.layaen.banking.acc\endcsname{0.010}
\expandafter\def\csname cdnum@nt.layaen.banking.diff\endcsname{\ensuremath{-}81.2}
\expandafter\def\csname cdnum@nt.layaen.banking.ece\endcsname{0.072}
\expandafter\def\csname cdnum@nt.layaen.banking.fone\endcsname{0.2}
\expandafter\def\csname cdnum@nt.layaen.goemo.acc\endcsname{0.524}
\expandafter\def\csname cdnum@nt.layaen.goemo.diff\endcsname{\ensuremath{-}7.9}
\expandafter\def\csname cdnum@nt.layaen.goemo.ece\endcsname{0.263}
\expandafter\def\csname cdnum@nt.layaen.goemo.fone\endcsname{53.0}
\expandafter\def\csname cdnum@nt.layaml.banking.acc\endcsname{0.013}
\expandafter\def\csname cdnum@nt.layaml.banking.diff\endcsname{\ensuremath{-}80.8}
\expandafter\def\csname cdnum@nt.layaml.banking.ece\endcsname{0.048}
\expandafter\def\csname cdnum@nt.layaml.banking.fone\endcsname{0.4}
\expandafter\def\csname cdnum@nt.layaml.goemo.acc\endcsname{0.526}
\expandafter\def\csname cdnum@nt.layaml.goemo.diff\endcsname{\ensuremath{-}7.7}
\expandafter\def\csname cdnum@nt.layaml.goemo.ece\endcsname{0.069}
\expandafter\def\csname cdnum@nt.layaml.goemo.fone\endcsname{51.5}
\expandafter\def\csname cdnum@nt.nimble.banking.acc\endcsname{0.843}
\expandafter\def\csname cdnum@nt.nimble.banking.diff\endcsname{2.1}
\expandafter\def\csname cdnum@nt.nimble.banking.ece\endcsname{0.067}
\expandafter\def\csname cdnum@nt.nimble.banking.fone\endcsname{82.9}
\expandafter\def\csname cdnum@nt.nimble.goemo.acc\endcsname{0.596}
\expandafter\def\csname cdnum@nt.nimble.goemo.diff\endcsname{\ensuremath{-}0.8}
\expandafter\def\csname cdnum@nt.nimble.goemo.ece\endcsname{0.173}
\expandafter\def\csname cdnum@nt.nimble.goemo.fone\endcsname{60.0}
\expandafter\def\csname cdnum@nt.nlizs.banking.acc\endcsname{0.683}
\expandafter\def\csname cdnum@nt.nlizs.banking.diff\endcsname{\ensuremath{-}13.8}
\expandafter\def\csname cdnum@nt.nlizs.banking.ece\endcsname{0.147}
\expandafter\def\csname cdnum@nt.nlizs.banking.fone\endcsname{66.2}
\expandafter\def\csname cdnum@nt.nlizs.goemo.acc\endcsname{0.490}
\expandafter\def\csname cdnum@nt.nlizs.goemo.diff\endcsname{\ensuremath{-}11.4}
\expandafter\def\csname cdnum@nt.nlizs.goemo.ece\endcsname{0.329}
\expandafter\def\csname cdnum@nt.nlizs.goemo.fone\endcsname{48.2}
\expandafter\def\csname cdnum@nt.thisthat.banking.acc\endcsname{0.828}
\expandafter\def\csname cdnum@nt.thisthat.banking.diff\endcsname{0.6}
\expandafter\def\csname cdnum@nt.thisthat.banking.ece\endcsname{0.070}
\expandafter\def\csname cdnum@nt.thisthat.banking.fone\endcsname{81.7}
\expandafter\def\csname cdnum@nt.thisthat.goemo.acc\endcsname{0.612}
\expandafter\def\csname cdnum@nt.thisthat.goemo.diff\endcsname{0.8}
\expandafter\def\csname cdnum@nt.thisthat.goemo.ece\endcsname{0.185}
\expandafter\def\csname cdnum@nt.thisthat.goemo.fone\endcsname{60.1}
\expandafter\def\csname cdnum@oo.compgem.accin\endcsname{0.938}
\expandafter\def\csname cdnum@oo.compgem.falsereject\endcsname{0.003}
\expandafter\def\csname cdnum@oo.compgem.gcov\endcsname{0.945}
\expandafter\def\csname cdnum@oo.compgem.gcovhi\endcsname{0.998}
\expandafter\def\csname cdnum@oo.compgem.gcovlo\endcsname{0.739}
\expandafter\def\csname cdnum@oo.compgem.gfar\endcsname{0.170}
\expandafter\def\csname cdnum@oo.compgem.gfarhi\endcsname{0.307}
\expandafter\def\csname cdnum@oo.compgem.gfarlo\endcsname{0.041}
\expandafter\def\csname cdnum@oo.compgem.oosrecall\endcsname{0.715}
\expandafter\def\csname cdnum@oo.compgemll.accin\endcsname{0.937}
\expandafter\def\csname cdnum@oo.compgemll.falsereject\endcsname{0.005}
\expandafter\def\csname cdnum@oo.compgemll.gcov\endcsname{0.972}
\expandafter\def\csname cdnum@oo.compgemll.gcovhi\endcsname{0.997}
\expandafter\def\csname cdnum@oo.compgemll.gcovlo\endcsname{0.896}
\expandafter\def\csname cdnum@oo.compgemll.gfar\endcsname{0.210}
\expandafter\def\csname cdnum@oo.compgemll.gfarhi\endcsname{0.319}
\expandafter\def\csname cdnum@oo.compgemll.gfarlo\endcsname{0.103}
\expandafter\def\csname cdnum@oo.compgemll.oosrecall\endcsname{0.715}
\expandafter\def\csname cdnum@oo.compmis.accin\endcsname{0.765}
\expandafter\def\csname cdnum@oo.compmis.falsereject\endcsname{0.017}
\expandafter\def\csname cdnum@oo.compmis.gcov\endcsname{0.000}
\expandafter\def\csname cdnum@oo.compmis.gcovhi\endcsname{0.000}
\expandafter\def\csname cdnum@oo.compmis.gcovlo\endcsname{0.000}
\expandafter\def\csname cdnum@oo.compmis.gfar\endcsname{0.000}
\expandafter\def\csname cdnum@oo.compmis.gfarhi\endcsname{0.000}
\expandafter\def\csname cdnum@oo.compmis.gfarlo\endcsname{0.000}
\expandafter\def\csname cdnum@oo.compmis.oosrecall\endcsname{0.300}
\expandafter\def\csname cdnum@oo.compmisll.accin\endcsname{0.783}
\expandafter\def\csname cdnum@oo.compmisll.falsereject\endcsname{0.005}
\expandafter\def\csname cdnum@oo.compmisll.gcov\endcsname{0.442}
\expandafter\def\csname cdnum@oo.compmisll.gcovhi\endcsname{0.607}
\expandafter\def\csname cdnum@oo.compmisll.gcovlo\endcsname{0.329}
\expandafter\def\csname cdnum@oo.compmisll.gfar\endcsname{0.035}
\expandafter\def\csname cdnum@oo.compmisll.gfarhi\endcsname{0.103}
\expandafter\def\csname cdnum@oo.compmisll.gfarlo\endcsname{0.005}
\expandafter\def\csname cdnum@oo.compmisll.oosrecall\endcsname{0.135}
\expandafter\def\csname cdnum@oo.compopen.accin\endcsname{0.732}
\expandafter\def\csname cdnum@oo.compopen.falsereject\endcsname{0.015}
\expandafter\def\csname cdnum@oo.compopen.gcov\endcsname{0.000}
\expandafter\def\csname cdnum@oo.compopen.gcovhi\endcsname{0.023}
\expandafter\def\csname cdnum@oo.compopen.gcovlo\endcsname{0.000}
\expandafter\def\csname cdnum@oo.compopen.gfar\endcsname{0.000}
\expandafter\def\csname cdnum@oo.compopen.gfarhi\endcsname{0.015}
\expandafter\def\csname cdnum@oo.compopen.gfarlo\endcsname{0.000}
\expandafter\def\csname cdnum@oo.compopen.oosrecall\endcsname{0.280}
\expandafter\def\csname cdnum@oo.comptwo.accin\endcsname{0.648}
\expandafter\def\csname cdnum@oo.comptwo.falsereject\endcsname{0.042}
\expandafter\def\csname cdnum@oo.comptwo.gcov\endcsname{0.653}
\expandafter\def\csname cdnum@oo.comptwo.gcovhi\endcsname{0.704}
\expandafter\def\csname cdnum@oo.comptwo.gcovlo\endcsname{0.014}
\expandafter\def\csname cdnum@oo.comptwo.gfar\endcsname{0.055}
\expandafter\def\csname cdnum@oo.comptwo.gfarhi\endcsname{0.102}
\expandafter\def\csname cdnum@oo.comptwo.gfarlo\endcsname{0.000}
\expandafter\def\csname cdnum@oo.comptwo.oosrecall\endcsname{0.880}
\expandafter\def\csname cdnum@oo.comptwoll.accin\endcsname{0.885}
\expandafter\def\csname cdnum@oo.comptwoll.falsereject\endcsname{0.040}
\expandafter\def\csname cdnum@oo.comptwoll.gcov\endcsname{0.837}
\expandafter\def\csname cdnum@oo.comptwoll.gcovhi\endcsname{0.914}
\expandafter\def\csname cdnum@oo.comptwoll.gcovlo\endcsname{0.770}
\expandafter\def\csname cdnum@oo.comptwoll.gfar\endcsname{0.035}
\expandafter\def\csname cdnum@oo.comptwoll.gfarhi\endcsname{0.080}
\expandafter\def\csname cdnum@oo.comptwoll.gfarlo\endcsname{0.013}
\expandafter\def\csname cdnum@oo.comptwoll.oosrecall\endcsname{0.860}
\expandafter\def\csname cdnum@oo.comptwothink.accin\endcsname{0.925}
\expandafter\def\csname cdnum@oo.comptwothink.falsereject\endcsname{0.007}
\expandafter\def\csname cdnum@oo.comptwothink.gcov\endcsname{0.890}
\expandafter\def\csname cdnum@oo.comptwothink.gcovhi\endcsname{0.983}
\expandafter\def\csname cdnum@oo.comptwothink.gcovlo\endcsname{0.691}
\expandafter\def\csname cdnum@oo.comptwothink.gfar\endcsname{0.165}
\expandafter\def\csname cdnum@oo.comptwothink.gfarhi\endcsname{0.310}
\expandafter\def\csname cdnum@oo.comptwothink.gfarlo\endcsname{0.066}
\expandafter\def\csname cdnum@oo.comptwothink.oosrecall\endcsname{0.665}
\expandafter\def\csname cdnum@oo.decider.accin\endcsname{0.855}
\expandafter\def\csname cdnum@oo.decider.falsereject\endcsname{0.000}
\expandafter\def\csname cdnum@oo.decider.gcov\endcsname{0.762}
\expandafter\def\csname cdnum@oo.decider.gcovhi\endcsname{0.853}
\expandafter\def\csname cdnum@oo.decider.gcovlo\endcsname{0.651}
\expandafter\def\csname cdnum@oo.decider.gfar\endcsname{0.140}
\expandafter\def\csname cdnum@oo.decider.gfarhi\endcsname{0.250}
\expandafter\def\csname cdnum@oo.decider.gfarlo\endcsname{0.070}
\expandafter\def\csname cdnum@oo.decider.oosrecall\endcsname{0.130}
\expandafter\def\csname cdnum@oo.jev.accin\endcsname{0.908}
\expandafter\def\csname cdnum@oo.jev.falsereject\endcsname{0.003}
\expandafter\def\csname cdnum@oo.jev.gcov\endcsname{0.925}
\expandafter\def\csname cdnum@oo.jev.gcovhi\endcsname{0.973}
\expandafter\def\csname cdnum@oo.jev.gcovlo\endcsname{0.794}
\expandafter\def\csname cdnum@oo.jev.gfar\endcsname{0.195}
\expandafter\def\csname cdnum@oo.jev.gfarhi\endcsname{0.310}
\expandafter\def\csname cdnum@oo.jev.gfarlo\endcsname{0.070}
\expandafter\def\csname cdnum@oo.jev.oosrecall\endcsname{0.585}
\expandafter\def\csname cdnum@oo.kevlarge.accin\endcsname{0.830}
\expandafter\def\csname cdnum@oo.kevlarge.falsereject\endcsname{0.047}
\expandafter\def\csname cdnum@oo.kevlarge.gcov\endcsname{0.610}
\expandafter\def\csname cdnum@oo.kevlarge.gcovhi\endcsname{0.714}
\expandafter\def\csname cdnum@oo.kevlarge.gcovlo\endcsname{0.454}
\expandafter\def\csname cdnum@oo.kevlarge.gfar\endcsname{0.015}
\expandafter\def\csname cdnum@oo.kevlarge.gfarhi\endcsname{0.055}
\expandafter\def\csname cdnum@oo.kevlarge.gfarlo\endcsname{0.000}
\expandafter\def\csname cdnum@oo.kevlarge.oosrecall\endcsname{0.840}
\expandafter\def\csname cdnum@oo.kevsmall.accin\endcsname{0.682}
\expandafter\def\csname cdnum@oo.kevsmall.falsereject\endcsname{0.013}
\expandafter\def\csname cdnum@oo.kevsmall.gcov\endcsname{0.223}
\expandafter\def\csname cdnum@oo.kevsmall.gcovhi\endcsname{0.395}
\expandafter\def\csname cdnum@oo.kevsmall.gcovlo\endcsname{0.080}
\expandafter\def\csname cdnum@oo.kevsmall.gfar\endcsname{0.010}
\expandafter\def\csname cdnum@oo.kevsmall.gfarhi\endcsname{0.054}
\expandafter\def\csname cdnum@oo.kevsmall.gfarlo\endcsname{0.000}
\expandafter\def\csname cdnum@oo.kevsmall.oosrecall\endcsname{0.110}
\expandafter\def\csname cdnum@oo.layaml.accin\endcsname{0.008}
\expandafter\def\csname cdnum@oo.layaml.falsereject\endcsname{0.003}
\expandafter\def\csname cdnum@oo.layaml.gcov\endcsname{0.000}
\expandafter\def\csname cdnum@oo.layaml.gcovhi\endcsname{0.000}
\expandafter\def\csname cdnum@oo.layaml.gcovlo\endcsname{0.000}
\expandafter\def\csname cdnum@oo.layaml.gfar\endcsname{0.000}
\expandafter\def\csname cdnum@oo.layaml.gfarhi\endcsname{0.000}
\expandafter\def\csname cdnum@oo.layaml.gfarlo\endcsname{0.000}
\expandafter\def\csname cdnum@oo.layaml.oosrecall\endcsname{0.005}
\expandafter\def\csname cdnum@oo.nimble.accin\endcsname{0.860}
\expandafter\def\csname cdnum@oo.nimble.falsereject\endcsname{0.012}
\expandafter\def\csname cdnum@oo.nimble.gcov\endcsname{0.657}
\expandafter\def\csname cdnum@oo.nimble.gcovhi\endcsname{0.805}
\expandafter\def\csname cdnum@oo.nimble.gcovlo\endcsname{0.475}
\expandafter\def\csname cdnum@oo.nimble.gfar\endcsname{0.030}
\expandafter\def\csname cdnum@oo.nimble.gfarhi\endcsname{0.079}
\expandafter\def\csname cdnum@oo.nimble.gfarlo\endcsname{0.005}
\expandafter\def\csname cdnum@oo.nimble.oosrecall\endcsname{0.645}
\expandafter\def\csname cdnum@oo.thisthat.accin\endcsname{0.862}
\expandafter\def\csname cdnum@oo.thisthat.falsereject\endcsname{0.002}
\expandafter\def\csname cdnum@oo.thisthat.gcov\endcsname{0.780}
\expandafter\def\csname cdnum@oo.thisthat.gcovhi\endcsname{0.858}
\expandafter\def\csname cdnum@oo.thisthat.gcovlo\endcsname{0.717}
\expandafter\def\csname cdnum@oo.thisthat.gfar\endcsname{0.185}
\expandafter\def\csname cdnum@oo.thisthat.gfarhi\endcsname{0.277}
\expandafter\def\csname cdnum@oo.thisthat.gfarlo\endcsname{0.124}
\expandafter\def\csname cdnum@oo.thisthat.oosrecall\endcsname{0.165}
\expandafter\def\csname cdnum@ov.clinc.deciderfit.exact\endcsname{201}
\expandafter\def\csname cdnum@ov.clinc.deciderfit.nitems\endcsname{800}
\expandafter\def\csname cdnum@ov.clinc.full.train.neardupchar\endcsname{20}
\expandafter\def\csname cdnum@ov.clinc.full.train.nitems\endcsname{800}
\expandafter\def\csname cdnum@ov.clinc.full.train.normalized\endcsname{5}
\expandafter\def\csname cdnum@p.anchor.ceiling\endcsname{0.735}
\expandafter\def\csname cdnum@p.anchor.jev.acc\endcsname{0.727}
\expandafter\def\csname cdnum@p.anchor.jev.brier\endcsname{0.148}
\expandafter\def\csname cdnum@p.anchor.jev.softacc\endcsname{0.580}
\expandafter\def\csname cdnum@p.anchor.package.emo\endcsname{0.494}
\expandafter\def\csname cdnum@p.anchor.package.pwr\endcsname{0.610}
\expandafter\def\csname cdnum@p.anchor.package.tox\endcsname{0.572}
\expandafter\def\csname cdnum@p.anchor.prior\endcsname{0.470}
\expandafter\def\csname cdnum@p.banking.perintent\endcsname{8}
\expandafter\def\csname cdnum@p.bins\endcsname{10}
\expandafter\def\csname cdnum@p.bone.requests\endcsname{40}
\expandafter\def\csname cdnum@p.boot.level\endcsname{95}
\expandafter\def\csname cdnum@p.boot.pairedreps\endcsname{10000}
\expandafter\def\csname cdnum@p.boot.refitreps\endcsname{1000}
\expandafter\def\csname cdnum@p.calibperworkflow\endcsname{75}
\expandafter\def\csname cdnum@p.comp.fullk\endcsname{20}
\expandafter\def\csname cdnum@p.comp.maxlen\endcsname{12288}
\expandafter\def\csname cdnum@p.comp.maxtokens\endcsname{768}
\expandafter\def\csname cdnum@p.comp.topn\endcsname{5}
\expandafter\def\csname cdnum@p.d1kmax\endcsname{5}
\expandafter\def\csname cdnum@p.d1kmin\endcsname{2}
\expandafter\def\csname cdnum@p.domains\endcsname{10}
\expandafter\def\csname cdnum@p.folds\endcsname{5}
\expandafter\def\csname cdnum@p.goemo.emotions\endcsname{27}
\expandafter\def\csname cdnum@p.gpuhourcost\endcsname{0.48}
\expandafter\def\csname cdnum@p.gpumem\endcsname{80}
\expandafter\def\csname cdnum@p.intentsperdomain\endcsname{15}
\expandafter\def\csname cdnum@p.jevreps\endcsname{3}
\expandafter\def\csname cdnum@p.jevretestreps\endcsname{5}
\expandafter\def\csname cdnum@p.kall\endcsname{150}
\expandafter\def\csname cdnum@p.kfifty\endcsname{50}
\expandafter\def\csname cdnum@p.kfive\endcsname{5}
\expandafter\def\csname cdnum@p.ktwenty\endcsname{20}
\expandafter\def\csname cdnum@p.oos\endcsname{200}
\expandafter\def\csname cdnum@p.perintent\endcsname{4}
\expandafter\def\csname cdnum@p.pninetyfive\endcsname{95}
\expandafter\def\csname cdnum@p.price\endcsname{0.042}
\expandafter\def\csname cdnum@p.questionsperstate\endcsname{5}
\expandafter\def\csname cdnum@p.riskfive\endcsname{5}
\expandafter\def\csname cdnum@p.riskone\endcsname{1}
\expandafter\def\csname cdnum@p.satlevel\endcsname{0.999}
\expandafter\def\csname cdnum@p.seed\endcsname{20260924}
\expandafter\def\csname cdnum@p.tgrid.hi\endcsname{20}
\expandafter\def\csname cdnum@p.tgrid.lo\endcsname{0.2}
\expandafter\def\csname cdnum@p.tgrid.n\endcsname{120}
\expandafter\def\csname cdnum@pd.layaml.politeness.absdiff\endcsname{29.1}
\expandafter\def\csname cdnum@pd.nimble.politeness.absdiff\endcsname{14.1}
\expandafter\def\csname cdnum@pd.thisthat.doneneutral.absdiff\endcsname{13.5}
\expandafter\def\csname cdnum@pd.thisthat.dtwo.absdiff\endcsname{5.0}
\expandafter\def\csname cdnum@pos.calibration.full\endcsname{5}
\expandafter\def\csname cdnum@pos.calibration.partial\endcsname{6}
\expandafter\def\csname cdnum@pos.cost.full\endcsname{7}
\expandafter\def\csname cdnum@pos.cost.partial\endcsname{9}
\expandafter\def\csname cdnum@pos.families.full\endcsname{5}
\expandafter\def\csname cdnum@pos.families.partial\endcsname{10}
\expandafter\def\csname cdnum@pos.identical.full\endcsname{6}
\expandafter\def\csname cdnum@pos.identical.partial\endcsname{11}
\expandafter\def\csname cdnum@pos.llm.full\endcsname{13}
\expandafter\def\csname cdnum@pos.llm.partial\endcsname{2}
\expandafter\def\csname cdnum@pos.models.full\endcsname{4}
\expandafter\def\csname cdnum@pos.models.partial\endcsname{4}
\expandafter\def\csname cdnum@pos.nstudies\endcsname{17}
\expandafter\def\csname cdnum@pos.outofscope.full\endcsname{3}
\expandafter\def\csname cdnum@pos.outofscope.partial\endcsname{4}
\expandafter\def\csname cdnum@pos.selective.full\endcsname{3}
\expandafter\def\csname cdnum@pos.selective.partial\endcsname{4}
\expandafter\def\csname cdnum@pos.stress.full\endcsname{0}
\expandafter\def\csname cdnum@pos.stress.partial\endcsname{10}
\expandafter\def\csname cdnum@pos.trained.full\endcsname{2}
\expandafter\def\csname cdnum@pos.trained.partial\endcsname{3}
\expandafter\def\csname cdnum@pos.uncertainty.full\endcsname{8}
\expandafter\def\csname cdnum@pos.uncertainty.partial\endcsname{4}
\expandafter\def\csname cdnum@rc.alpha\endcsname{5}
\expandafter\def\csname cdnum@rc.compgem.doneneutral.accepted\endcsname{0}
\expandafter\def\csname cdnum@rc.compgem.doneneutral.cov\endcsname{0.000}
\expandafter\def\csname cdnum@rc.compgem.dtwo.accepted\endcsname{304}
\expandafter\def\csname cdnum@rc.compgem.dtwo.cov\endcsname{0.507}
\expandafter\def\csname cdnum@rc.compgem.dtwo.risk\endcsname{0.016}
\expandafter\def\csname cdnum@rc.compgemll.doneneutral.accepted\endcsname{0}
\expandafter\def\csname cdnum@rc.compgemll.doneneutral.cov\endcsname{0.000}
\expandafter\def\csname cdnum@rc.compgemll.dtwo.accepted\endcsname{0}
\expandafter\def\csname cdnum@rc.compgemll.dtwo.cov\endcsname{0.000}
\expandafter\def\csname cdnum@rc.compmis.doneneutral.accepted\endcsname{0}
\expandafter\def\csname cdnum@rc.compmis.doneneutral.cov\endcsname{0.000}
\expandafter\def\csname cdnum@rc.compmis.dtwo.accepted\endcsname{0}
\expandafter\def\csname cdnum@rc.compmis.dtwo.cov\endcsname{0.000}
\expandafter\def\csname cdnum@rc.compmisll.doneneutral.accepted\endcsname{0}
\expandafter\def\csname cdnum@rc.compmisll.doneneutral.cov\endcsname{0.000}
\expandafter\def\csname cdnum@rc.compmisll.dtwo.accepted\endcsname{0}
\expandafter\def\csname cdnum@rc.compmisll.dtwo.cov\endcsname{0.000}
\expandafter\def\csname cdnum@rc.compopen.doneneutral.accepted\endcsname{0}
\expandafter\def\csname cdnum@rc.compopen.doneneutral.cov\endcsname{0.000}
\expandafter\def\csname cdnum@rc.compopen.dtwo.accepted\endcsname{0}
\expandafter\def\csname cdnum@rc.compopen.dtwo.cov\endcsname{0.000}
\expandafter\def\csname cdnum@rc.comptwo.doneneutral.accepted\endcsname{0}
\expandafter\def\csname cdnum@rc.comptwo.doneneutral.cov\endcsname{0.000}
\expandafter\def\csname cdnum@rc.comptwo.dtwo.accepted\endcsname{0}
\expandafter\def\csname cdnum@rc.comptwo.dtwo.cov\endcsname{0.000}
\expandafter\def\csname cdnum@rc.comptwoll.doneneutral.accepted\endcsname{326}
\expandafter\def\csname cdnum@rc.comptwoll.doneneutral.cov\endcsname{0.163}
\expandafter\def\csname cdnum@rc.comptwoll.doneneutral.risk\endcsname{0.006}
\expandafter\def\csname cdnum@rc.comptwoll.dtwo.accepted\endcsname{152}
\expandafter\def\csname cdnum@rc.comptwoll.dtwo.cov\endcsname{0.253}
\expandafter\def\csname cdnum@rc.comptwoll.dtwo.risk\endcsname{0.039}
\expandafter\def\csname cdnum@rc.comptwothink.doneneutral.accepted\endcsname{0}
\expandafter\def\csname cdnum@rc.comptwothink.doneneutral.cov\endcsname{0.000}
\expandafter\def\csname cdnum@rc.comptwothink.dtwo.accepted\endcsname{47}
\expandafter\def\csname cdnum@rc.comptwothink.dtwo.cov\endcsname{0.078}
\expandafter\def\csname cdnum@rc.comptwothink.dtwo.risk\endcsname{0.021}
\expandafter\def\csname cdnum@rc.confidence\endcsname{90}
\expandafter\def\csname cdnum@rc.decider.doneneutral.accepted\endcsname{0}
\expandafter\def\csname cdnum@rc.decider.doneneutral.cov\endcsname{0.000}
\expandafter\def\csname cdnum@rc.decider.dtwo.accepted\endcsname{108}
\expandafter\def\csname cdnum@rc.decider.dtwo.cov\endcsname{0.180}
\expandafter\def\csname cdnum@rc.decider.dtwo.risk\endcsname{0.028}
\expandafter\def\csname cdnum@rc.gridsize\endcsname{50}
\expandafter\def\csname cdnum@rc.jev.doneneutral.accepted\endcsname{166}
\expandafter\def\csname cdnum@rc.jev.doneneutral.cov\endcsname{0.083}
\expandafter\def\csname cdnum@rc.jev.doneneutral.risk\endcsname{0.024}
\expandafter\def\csname cdnum@rc.jev.dtwo.accepted\endcsname{317}
\expandafter\def\csname cdnum@rc.jev.dtwo.cov\endcsname{0.528}
\expandafter\def\csname cdnum@rc.jev.dtwo.risk\endcsname{0.009}
\expandafter\def\csname cdnum@rc.kevlarge.doneneutral.accepted\endcsname{0}
\expandafter\def\csname cdnum@rc.kevlarge.doneneutral.cov\endcsname{0.000}
\expandafter\def\csname cdnum@rc.kevlarge.dtwo.accepted\endcsname{0}
\expandafter\def\csname cdnum@rc.kevlarge.dtwo.cov\endcsname{0.000}
\expandafter\def\csname cdnum@rc.kevsmall.doneneutral.accepted\endcsname{0}
\expandafter\def\csname cdnum@rc.kevsmall.doneneutral.cov\endcsname{0.000}
\expandafter\def\csname cdnum@rc.kevsmall.dtwo.accepted\endcsname{0}
\expandafter\def\csname cdnum@rc.kevsmall.dtwo.cov\endcsname{0.000}
\expandafter\def\csname cdnum@rc.layaen.doneneutral.accepted\endcsname{0}
\expandafter\def\csname cdnum@rc.layaen.doneneutral.cov\endcsname{0.000}
\expandafter\def\csname cdnum@rc.layaml.doneneutral.accepted\endcsname{0}
\expandafter\def\csname cdnum@rc.layaml.doneneutral.cov\endcsname{0.000}
\expandafter\def\csname cdnum@rc.layaml.dtwo.accepted\endcsname{0}
\expandafter\def\csname cdnum@rc.layaml.dtwo.cov\endcsname{0.000}
\expandafter\def\csname cdnum@rc.nimble.doneneutral.accepted\endcsname{179}
\expandafter\def\csname cdnum@rc.nimble.doneneutral.cov\endcsname{0.089}
\expandafter\def\csname cdnum@rc.nimble.doneneutral.risk\endcsname{0.022}
\expandafter\def\csname cdnum@rc.nimble.dtwo.accepted\endcsname{0}
\expandafter\def\csname cdnum@rc.nimble.dtwo.cov\endcsname{0.000}
\expandafter\def\csname cdnum@rc.thisthat.doneneutral.accepted\endcsname{0}
\expandafter\def\csname cdnum@rc.thisthat.doneneutral.cov\endcsname{0.000}
\expandafter\def\csname cdnum@rc.thisthat.dtwo.accepted\endcsname{43}
\expandafter\def\csname cdnum@rc.thisthat.dtwo.cov\endcsname{0.072}
\expandafter\def\csname cdnum@rc.thisthat.dtwo.risk\endcsname{0.047}
\expandafter\def\csname cdnum@rp.done.descvsoriginal.diff\endcsname{\ensuremath{-}1.1}
\expandafter\def\csname cdnum@rp.dtwo.descvsjev.diff\endcsname{1.8}
\expandafter\def\csname cdnum@rp.dtwo.descvsjev.p\endcsname{0.13}
\expandafter\def\csname cdnum@rp.dtwo.descvsoriginal.acc\endcsname{0.932}
\expandafter\def\csname cdnum@rp.dtwo.descvsoriginal.diff\endcsname{6.8}
\expandafter\def\csname cdnum@rp.dtwo.descvsoriginal.hi\endcsname{9.3}
\expandafter\def\csname cdnum@rp.dtwo.descvsoriginal.lo\endcsname{4.5}
\expandafter\def\csname cdnum@rr.comptwoll.doneneutral.covrounded\endcsname{0.42}
\expandafter\def\csname cdnum@rv.card.layaen.fifty.predqone\endcsname{0.837}
\expandafter\def\csname cdnum@rv.cas.decider.compgem.doneneutral.acc\endcsname{0.740}
\expandafter\def\csname cdnum@rv.cas.decider.compgem.doneneutral.acchi\endcsname{0.760}
\expandafter\def\csname cdnum@rv.cas.decider.compgem.doneneutral.acclo\endcsname{0.718}
\expandafter\def\csname cdnum@rv.cas.decider.compgem.doneneutral.costfrac\endcsname{1.11}
\expandafter\def\csname cdnum@rv.cas.decider.compgem.doneneutral.esc\endcsname{0.94}
\expandafter\def\csname cdnum@rv.cas.decider.compgem.doneneutral.gain\endcsname{\ensuremath{-}0.1}
\expandafter\def\csname cdnum@rv.cas.decider.compgem.doneneutral.gainhi\endcsname{0.1}
\expandafter\def\csname cdnum@rv.cas.decider.compgem.doneneutral.gainlo\endcsname{\ensuremath{-}0.4}
\expandafter\def\csname cdnum@rv.cas.decider.compgem.dtwo.acc\endcsname{0.940}
\expandafter\def\csname cdnum@rv.cas.decider.compgem.dtwo.acchi\endcsname{0.960}
\expandafter\def\csname cdnum@rv.cas.decider.compgem.dtwo.acclo\endcsname{0.920}
\expandafter\def\csname cdnum@rv.cas.decider.compgem.dtwo.costfrac\endcsname{0.47}
\expandafter\def\csname cdnum@rv.cas.decider.compgem.dtwo.esc\endcsname{0.40}
\expandafter\def\csname cdnum@rv.cas.decider.compgem.dtwo.gain\endcsname{\ensuremath{-}0.2}
\expandafter\def\csname cdnum@rv.cas.decider.compgem.dtwo.gainhi\endcsname{1.2}
\expandafter\def\csname cdnum@rv.cas.decider.compgem.dtwo.gainlo\endcsname{\ensuremath{-}1.3}
\expandafter\def\csname cdnum@rv.cas.decider.compopen.doneneutral.acc\endcsname{0.679}
\expandafter\def\csname cdnum@rv.cas.decider.compopen.doneneutral.acchi\endcsname{0.703}
\expandafter\def\csname cdnum@rv.cas.decider.compopen.doneneutral.acclo\endcsname{0.654}
\expandafter\def\csname cdnum@rv.cas.decider.compopen.doneneutral.costfrac\endcsname{1.34}
\expandafter\def\csname cdnum@rv.cas.decider.compopen.doneneutral.esc\endcsname{0.93}
\expandafter\def\csname cdnum@rv.cas.decider.compopen.doneneutral.gain\endcsname{\ensuremath{-}0.1}
\expandafter\def\csname cdnum@rv.cas.decider.compopen.doneneutral.gainhi\endcsname{0.2}
\expandafter\def\csname cdnum@rv.cas.decider.compopen.doneneutral.gainlo\endcsname{\ensuremath{-}0.4}
\expandafter\def\csname cdnum@rv.cas.decider.compopen.dtwo.acc\endcsname{0.853}
\expandafter\def\csname cdnum@rv.cas.decider.compopen.dtwo.acchi\endcsname{0.887}
\expandafter\def\csname cdnum@rv.cas.decider.compopen.dtwo.acclo\endcsname{0.825}
\expandafter\def\csname cdnum@rv.cas.decider.compopen.dtwo.costfrac\endcsname{0.24}
\expandafter\def\csname cdnum@rv.cas.decider.compopen.dtwo.esc\endcsname{0.11}
\expandafter\def\csname cdnum@rv.cas.decider.compopen.dtwo.gain\endcsname{\ensuremath{-}0.7}
\expandafter\def\csname cdnum@rv.cas.decider.compopen.dtwo.gainhi\endcsname{1.8}
\expandafter\def\csname cdnum@rv.cas.decider.compopen.dtwo.gainlo\endcsname{\ensuremath{-}2.3}
\expandafter\def\csname cdnum@rv.cas.decider.comptwo.doneneutral.acc\endcsname{0.724}
\expandafter\def\csname cdnum@rv.cas.decider.comptwo.doneneutral.acchi\endcsname{0.744}
\expandafter\def\csname cdnum@rv.cas.decider.comptwo.doneneutral.acclo\endcsname{0.701}
\expandafter\def\csname cdnum@rv.cas.decider.comptwo.doneneutral.costfrac\endcsname{1.09}
\expandafter\def\csname cdnum@rv.cas.decider.comptwo.doneneutral.esc\endcsname{0.94}
\expandafter\def\csname cdnum@rv.cas.decider.comptwo.doneneutral.gain\endcsname{\ensuremath{-}0.1}
\expandafter\def\csname cdnum@rv.cas.decider.comptwo.doneneutral.gainhi\endcsname{0.0}
\expandafter\def\csname cdnum@rv.cas.decider.comptwo.doneneutral.gainlo\endcsname{\ensuremath{-}0.6}
\expandafter\def\csname cdnum@rv.cas.decider.comptwo.dtwo.acc\endcsname{0.862}
\expandafter\def\csname cdnum@rv.cas.decider.comptwo.dtwo.acchi\endcsname{0.892}
\expandafter\def\csname cdnum@rv.cas.decider.comptwo.dtwo.acclo\endcsname{0.830}
\expandafter\def\csname cdnum@rv.cas.decider.comptwo.dtwo.costfrac\endcsname{0.20}
\expandafter\def\csname cdnum@rv.cas.decider.comptwo.dtwo.esc\endcsname{0.09}
\expandafter\def\csname cdnum@rv.cas.decider.comptwo.dtwo.gain\endcsname{0.2}
\expandafter\def\csname cdnum@rv.cas.decider.comptwo.dtwo.gainhi\endcsname{2.7}
\expandafter\def\csname cdnum@rv.cas.decider.comptwo.dtwo.gainlo\endcsname{\ensuremath{-}2.2}
\expandafter\def\csname cdnum@rv.cas.decider.comptwoll.doneneutral.acc\endcsname{0.732}
\expandafter\def\csname cdnum@rv.cas.decider.comptwoll.doneneutral.acchi\endcsname{0.751}
\expandafter\def\csname cdnum@rv.cas.decider.comptwoll.doneneutral.acclo\endcsname{0.710}
\expandafter\def\csname cdnum@rv.cas.decider.comptwoll.doneneutral.costfrac\endcsname{1.13}
\expandafter\def\csname cdnum@rv.cas.decider.comptwoll.doneneutral.esc\endcsname{0.94}
\expandafter\def\csname cdnum@rv.cas.decider.comptwoll.doneneutral.gain\endcsname{\ensuremath{-}0.1}
\expandafter\def\csname cdnum@rv.cas.decider.comptwoll.doneneutral.gainhi\endcsname{0.0}
\expandafter\def\csname cdnum@rv.cas.decider.comptwoll.doneneutral.gainlo\endcsname{\ensuremath{-}0.8}
\expandafter\def\csname cdnum@rv.cas.decider.comptwoll.dtwo.acc\endcsname{0.922}
\expandafter\def\csname cdnum@rv.cas.decider.comptwoll.dtwo.acchi\endcsname{0.943}
\expandafter\def\csname cdnum@rv.cas.decider.comptwoll.dtwo.acclo\endcsname{0.893}
\expandafter\def\csname cdnum@rv.cas.decider.comptwoll.dtwo.costfrac\endcsname{0.37}
\expandafter\def\csname cdnum@rv.cas.decider.comptwoll.dtwo.esc\endcsname{0.35}
\expandafter\def\csname cdnum@rv.cas.decider.comptwoll.dtwo.gain\endcsname{1.0}
\expandafter\def\csname cdnum@rv.cas.decider.comptwoll.dtwo.gainhi\endcsname{2.5}
\expandafter\def\csname cdnum@rv.cas.decider.comptwoll.dtwo.gainlo\endcsname{\ensuremath{-}1.0}
\expandafter\def\csname cdnum@rv.cas.decider.jev.doneneutral.acc\endcsname{0.731}
\expandafter\def\csname cdnum@rv.cas.decider.jev.doneneutral.acchi\endcsname{0.751}
\expandafter\def\csname cdnum@rv.cas.decider.jev.doneneutral.acclo\endcsname{0.707}
\expandafter\def\csname cdnum@rv.cas.decider.jev.doneneutral.costfrac\endcsname{2.31}
\expandafter\def\csname cdnum@rv.cas.decider.jev.doneneutral.esc\endcsname{0.94}
\expandafter\def\csname cdnum@rv.cas.decider.jev.doneneutral.gain\endcsname{\ensuremath{-}0.1}
\expandafter\def\csname cdnum@rv.cas.decider.jev.doneneutral.gainhi\endcsname{0.0}
\expandafter\def\csname cdnum@rv.cas.decider.jev.doneneutral.gainlo\endcsname{\ensuremath{-}0.7}
\expandafter\def\csname cdnum@rv.cas.decider.jev.dtwo.acc\endcsname{0.925}
\expandafter\def\csname cdnum@rv.cas.decider.jev.dtwo.acchi\endcsname{0.942}
\expandafter\def\csname cdnum@rv.cas.decider.jev.dtwo.acclo\endcsname{0.892}
\expandafter\def\csname cdnum@rv.cas.decider.jev.dtwo.costfrac\endcsname{0.58}
\expandafter\def\csname cdnum@rv.cas.decider.jev.dtwo.esc\endcsname{0.40}
\expandafter\def\csname cdnum@rv.cas.decider.jev.dtwo.gain\endcsname{1.2}
\expandafter\def\csname cdnum@rv.cas.decider.jev.dtwo.gainhi\endcsname{2.0}
\expandafter\def\csname cdnum@rv.cas.decider.jev.dtwo.gainlo\endcsname{\ensuremath{-}1.2}
\expandafter\def\csname cdnum@rv.cas.jev.compgem.doneneutral.acc\endcsname{0.749}
\expandafter\def\csname cdnum@rv.cas.jev.compgem.doneneutral.acchi\endcsname{0.768}
\expandafter\def\csname cdnum@rv.cas.jev.compgem.doneneutral.acclo\endcsname{0.725}
\expandafter\def\csname cdnum@rv.cas.jev.compgem.doneneutral.costfrac\endcsname{0.52}
\expandafter\def\csname cdnum@rv.cas.jev.compgem.doneneutral.esc\endcsname{0.39}
\expandafter\def\csname cdnum@rv.cas.jev.compgem.doneneutral.gain\endcsname{0.8}
\expandafter\def\csname cdnum@rv.cas.jev.compgem.doneneutral.gainhi\endcsname{1.7}
\expandafter\def\csname cdnum@rv.cas.jev.compgem.doneneutral.gainlo\endcsname{\ensuremath{-}0.5}
\expandafter\def\csname cdnum@rv.cas.jev.compgem.dtwo.acc\endcsname{0.942}
\expandafter\def\csname cdnum@rv.cas.jev.compgem.dtwo.acchi\endcsname{0.957}
\expandafter\def\csname cdnum@rv.cas.jev.compgem.dtwo.acclo\endcsname{0.917}
\expandafter\def\csname cdnum@rv.cas.jev.compgem.dtwo.costfrac\endcsname{0.64}
\expandafter\def\csname cdnum@rv.cas.jev.compgem.dtwo.esc\endcsname{0.26}
\expandafter\def\csname cdnum@rv.cas.jev.compgem.dtwo.gain\endcsname{0.0}
\expandafter\def\csname cdnum@rv.cas.jev.compgem.dtwo.gainhi\endcsname{0.5}
\expandafter\def\csname cdnum@rv.cas.jev.compgem.dtwo.gainlo\endcsname{\ensuremath{-}1.5}
\expandafter\def\csname cdnum@rv.cas.jev.compgem.dtwo.usd\endcsname{0.186}
\expandafter\def\csname cdnum@rv.cas.jev.compopen.doneneutral.acc\endcsname{0.732}
\expandafter\def\csname cdnum@rv.cas.jev.compopen.doneneutral.acchi\endcsname{0.762}
\expandafter\def\csname cdnum@rv.cas.jev.compopen.doneneutral.acclo\endcsname{0.714}
\expandafter\def\csname cdnum@rv.cas.jev.compopen.doneneutral.costfrac\endcsname{0.31}
\expandafter\def\csname cdnum@rv.cas.jev.compopen.doneneutral.esc\endcsname{0.01}
\expandafter\def\csname cdnum@rv.cas.jev.compopen.doneneutral.gain\endcsname{0.0}
\expandafter\def\csname cdnum@rv.cas.jev.compopen.doneneutral.gainhi\endcsname{2.5}
\expandafter\def\csname cdnum@rv.cas.jev.compopen.doneneutral.gainlo\endcsname{\ensuremath{-}0.4}
\expandafter\def\csname cdnum@rv.cas.jev.compopen.dtwo.acc\endcsname{0.915}
\expandafter\def\csname cdnum@rv.cas.jev.compopen.dtwo.acchi\endcsname{0.935}
\expandafter\def\csname cdnum@rv.cas.jev.compopen.dtwo.acclo\endcsname{0.887}
\expandafter\def\csname cdnum@rv.cas.jev.compopen.dtwo.costfrac\endcsname{0.73}
\expandafter\def\csname cdnum@rv.cas.jev.compopen.dtwo.esc\endcsname{0.01}
\expandafter\def\csname cdnum@rv.cas.jev.compopen.dtwo.gain\endcsname{0.2}
\expandafter\def\csname cdnum@rv.cas.jev.compopen.dtwo.gainhi\endcsname{0.5}
\expandafter\def\csname cdnum@rv.cas.jev.compopen.dtwo.gainlo\endcsname{\ensuremath{-}1.2}
\expandafter\def\csname cdnum@rv.cas.jev.comptwo.doneneutral.acc\endcsname{0.733}
\expandafter\def\csname cdnum@rv.cas.jev.comptwo.doneneutral.acchi\endcsname{0.754}
\expandafter\def\csname cdnum@rv.cas.jev.comptwo.doneneutral.acclo\endcsname{0.709}
\expandafter\def\csname cdnum@rv.cas.jev.comptwo.doneneutral.costfrac\endcsname{0.34}
\expandafter\def\csname cdnum@rv.cas.jev.comptwo.doneneutral.esc\endcsname{0.23}
\expandafter\def\csname cdnum@rv.cas.jev.comptwo.doneneutral.gain\endcsname{0.1}
\expandafter\def\csname cdnum@rv.cas.jev.comptwo.doneneutral.gainhi\endcsname{1.2}
\expandafter\def\csname cdnum@rv.cas.jev.comptwo.doneneutral.gainlo\endcsname{\ensuremath{-}1.6}
\expandafter\def\csname cdnum@rv.cas.jev.comptwo.dtwo.acc\endcsname{0.913}
\expandafter\def\csname cdnum@rv.cas.jev.comptwo.dtwo.acchi\endcsname{0.937}
\expandafter\def\csname cdnum@rv.cas.jev.comptwo.dtwo.acclo\endcsname{0.892}
\expandafter\def\csname cdnum@rv.cas.jev.comptwo.dtwo.costfrac\endcsname{0.60}
\expandafter\def\csname cdnum@rv.cas.jev.comptwo.dtwo.esc\endcsname{0.02}
\expandafter\def\csname cdnum@rv.cas.jev.comptwo.dtwo.gain\endcsname{0.0}
\expandafter\def\csname cdnum@rv.cas.jev.comptwo.dtwo.gainhi\endcsname{1.0}
\expandafter\def\csname cdnum@rv.cas.jev.comptwo.dtwo.gainlo\endcsname{\ensuremath{-}0.5}
\expandafter\def\csname cdnum@rv.cas.jev.comptwoll.doneneutral.acc\endcsname{0.733}
\expandafter\def\csname cdnum@rv.cas.jev.comptwoll.doneneutral.acchi\endcsname{0.754}
\expandafter\def\csname cdnum@rv.cas.jev.comptwoll.doneneutral.acclo\endcsname{0.711}
\expandafter\def\csname cdnum@rv.cas.jev.comptwoll.doneneutral.costfrac\endcsname{0.35}
\expandafter\def\csname cdnum@rv.cas.jev.comptwoll.doneneutral.esc\endcsname{0.22}
\expandafter\def\csname cdnum@rv.cas.jev.comptwoll.doneneutral.gain\endcsname{\ensuremath{-}0.0}
\expandafter\def\csname cdnum@rv.cas.jev.comptwoll.doneneutral.gainhi\endcsname{0.9}
\expandafter\def\csname cdnum@rv.cas.jev.comptwoll.doneneutral.gainlo\endcsname{\ensuremath{-}1.8}
\expandafter\def\csname cdnum@rv.cas.jev.comptwoll.dtwo.acc\endcsname{0.922}
\expandafter\def\csname cdnum@rv.cas.jev.comptwoll.dtwo.acchi\endcsname{0.943}
\expandafter\def\csname cdnum@rv.cas.jev.comptwoll.dtwo.acclo\endcsname{0.895}
\expandafter\def\csname cdnum@rv.cas.jev.comptwoll.dtwo.costfrac\endcsname{0.15}
\expandafter\def\csname cdnum@rv.cas.jev.comptwoll.dtwo.esc\endcsname{0.06}
\expandafter\def\csname cdnum@rv.cas.jev.comptwoll.dtwo.gain\endcsname{0.8}
\expandafter\def\csname cdnum@rv.cas.jev.comptwoll.dtwo.gainhi\endcsname{2.0}
\expandafter\def\csname cdnum@rv.cas.jev.comptwoll.dtwo.gainlo\endcsname{\ensuremath{-}1.0}
\expandafter\def\csname cdnum@rv.cas.kevlarge.compgem.doneneutral.acc\endcsname{0.745}
\expandafter\def\csname cdnum@rv.cas.kevlarge.compgem.doneneutral.acchi\endcsname{0.765}
\expandafter\def\csname cdnum@rv.cas.kevlarge.compgem.doneneutral.acclo\endcsname{0.720}
\expandafter\def\csname cdnum@rv.cas.kevlarge.compgem.doneneutral.costfrac\endcsname{0.77}
\expandafter\def\csname cdnum@rv.cas.kevlarge.compgem.doneneutral.esc\endcsname{0.53}
\expandafter\def\csname cdnum@rv.cas.kevlarge.compgem.doneneutral.gain\endcsname{0.4}
\expandafter\def\csname cdnum@rv.cas.kevlarge.compgem.doneneutral.gainhi\endcsname{1.3}
\expandafter\def\csname cdnum@rv.cas.kevlarge.compgem.doneneutral.gainlo\endcsname{\ensuremath{-}0.7}
\expandafter\def\csname cdnum@rv.cas.kevlarge.compgem.dtwo.acc\endcsname{0.940}
\expandafter\def\csname cdnum@rv.cas.kevlarge.compgem.dtwo.acchi\endcsname{0.955}
\expandafter\def\csname cdnum@rv.cas.kevlarge.compgem.dtwo.acclo\endcsname{0.917}
\expandafter\def\csname cdnum@rv.cas.kevlarge.compgem.dtwo.costfrac\endcsname{0.94}
\expandafter\def\csname cdnum@rv.cas.kevlarge.compgem.dtwo.esc\endcsname{0.74}
\expandafter\def\csname cdnum@rv.cas.kevlarge.compgem.dtwo.gain\endcsname{\ensuremath{-}0.2}
\expandafter\def\csname cdnum@rv.cas.kevlarge.compgem.dtwo.gainhi\endcsname{0.0}
\expandafter\def\csname cdnum@rv.cas.kevlarge.compgem.dtwo.gainlo\endcsname{\ensuremath{-}1.2}
\expandafter\def\csname cdnum@rv.cas.kevlarge.compopen.doneneutral.acc\endcsname{0.718}
\expandafter\def\csname cdnum@rv.cas.kevlarge.compopen.doneneutral.acchi\endcsname{0.741}
\expandafter\def\csname cdnum@rv.cas.kevlarge.compopen.doneneutral.acclo\endcsname{0.690}
\expandafter\def\csname cdnum@rv.cas.kevlarge.compopen.doneneutral.costfrac\endcsname{0.60}
\expandafter\def\csname cdnum@rv.cas.kevlarge.compopen.doneneutral.esc\endcsname{0.04}
\expandafter\def\csname cdnum@rv.cas.kevlarge.compopen.doneneutral.gain\endcsname{0.6}
\expandafter\def\csname cdnum@rv.cas.kevlarge.compopen.doneneutral.gainhi\endcsname{1.4}
\expandafter\def\csname cdnum@rv.cas.kevlarge.compopen.doneneutral.gainlo\endcsname{\ensuremath{-}0.3}
\expandafter\def\csname cdnum@rv.cas.kevlarge.compopen.dtwo.acc\endcsname{0.850}
\expandafter\def\csname cdnum@rv.cas.kevlarge.compopen.dtwo.acchi\endcsname{0.873}
\expandafter\def\csname cdnum@rv.cas.kevlarge.compopen.dtwo.acclo\endcsname{0.813}
\expandafter\def\csname cdnum@rv.cas.kevlarge.compopen.dtwo.costfrac\endcsname{0.38}
\expandafter\def\csname cdnum@rv.cas.kevlarge.compopen.dtwo.esc\endcsname{0.00}
\expandafter\def\csname cdnum@rv.cas.kevlarge.compopen.dtwo.gain\endcsname{0.0}
\expandafter\def\csname cdnum@rv.cas.kevlarge.compopen.dtwo.gainhi\endcsname{0.0}
\expandafter\def\csname cdnum@rv.cas.kevlarge.compopen.dtwo.gainlo\endcsname{\ensuremath{-}1.7}
\expandafter\def\csname cdnum@rv.cas.kevlarge.comptwo.doneneutral.acc\endcsname{0.725}
\expandafter\def\csname cdnum@rv.cas.kevlarge.comptwo.doneneutral.acchi\endcsname{0.745}
\expandafter\def\csname cdnum@rv.cas.kevlarge.comptwo.doneneutral.acclo\endcsname{0.702}
\expandafter\def\csname cdnum@rv.cas.kevlarge.comptwo.doneneutral.costfrac\endcsname{1.18}
\expandafter\def\csname cdnum@rv.cas.kevlarge.comptwo.doneneutral.esc\endcsname{0.97}
\expandafter\def\csname cdnum@rv.cas.kevlarge.comptwo.doneneutral.gain\endcsname{0.0}
\expandafter\def\csname cdnum@rv.cas.kevlarge.comptwo.doneneutral.gainhi\endcsname{0.1}
\expandafter\def\csname cdnum@rv.cas.kevlarge.comptwo.doneneutral.gainlo\endcsname{\ensuremath{-}1.1}
\expandafter\def\csname cdnum@rv.cas.kevlarge.comptwo.dtwo.acc\endcsname{0.845}
\expandafter\def\csname cdnum@rv.cas.kevlarge.comptwo.dtwo.acchi\endcsname{0.875}
\expandafter\def\csname cdnum@rv.cas.kevlarge.comptwo.dtwo.acclo\endcsname{0.815}
\expandafter\def\csname cdnum@rv.cas.kevlarge.comptwo.dtwo.costfrac\endcsname{0.33}
\expandafter\def\csname cdnum@rv.cas.kevlarge.comptwo.dtwo.esc\endcsname{0.03}
\expandafter\def\csname cdnum@rv.cas.kevlarge.comptwo.dtwo.gain\endcsname{\ensuremath{-}0.5}
\expandafter\def\csname cdnum@rv.cas.kevlarge.comptwo.dtwo.gainhi\endcsname{1.2}
\expandafter\def\csname cdnum@rv.cas.kevlarge.comptwo.dtwo.gainlo\endcsname{\ensuremath{-}1.5}
\expandafter\def\csname cdnum@rv.cas.kevlarge.comptwoll.doneneutral.acc\endcsname{0.733}
\expandafter\def\csname cdnum@rv.cas.kevlarge.comptwoll.doneneutral.acchi\endcsname{0.753}
\expandafter\def\csname cdnum@rv.cas.kevlarge.comptwoll.doneneutral.acclo\endcsname{0.709}
\expandafter\def\csname cdnum@rv.cas.kevlarge.comptwoll.doneneutral.costfrac\endcsname{1.22}
\expandafter\def\csname cdnum@rv.cas.kevlarge.comptwoll.doneneutral.esc\endcsname{0.97}
\expandafter\def\csname cdnum@rv.cas.kevlarge.comptwoll.doneneutral.gain\endcsname{0.0}
\expandafter\def\csname cdnum@rv.cas.kevlarge.comptwoll.doneneutral.gainhi\endcsname{0.0}
\expandafter\def\csname cdnum@rv.cas.kevlarge.comptwoll.doneneutral.gainlo\endcsname{\ensuremath{-}1.4}
\expandafter\def\csname cdnum@rv.cas.kevlarge.comptwoll.dtwo.acc\endcsname{0.910}
\expandafter\def\csname cdnum@rv.cas.kevlarge.comptwoll.dtwo.acchi\endcsname{0.930}
\expandafter\def\csname cdnum@rv.cas.kevlarge.comptwoll.dtwo.acclo\endcsname{0.880}
\expandafter\def\csname cdnum@rv.cas.kevlarge.comptwoll.dtwo.costfrac\endcsname{0.80}
\expandafter\def\csname cdnum@rv.cas.kevlarge.comptwoll.dtwo.esc\endcsname{0.75}
\expandafter\def\csname cdnum@rv.cas.kevlarge.comptwoll.dtwo.gain\endcsname{\ensuremath{-}0.2}
\expandafter\def\csname cdnum@rv.cas.kevlarge.comptwoll.dtwo.gainhi\endcsname{\ensuremath{-}0.2}
\expandafter\def\csname cdnum@rv.cas.kevlarge.comptwoll.dtwo.gainlo\endcsname{\ensuremath{-}1.5}
\expandafter\def\csname cdnum@rv.cas.kevlarge.jev.doneneutral.acc\endcsname{0.738}
\expandafter\def\csname cdnum@rv.cas.kevlarge.jev.doneneutral.acchi\endcsname{0.759}
\expandafter\def\csname cdnum@rv.cas.kevlarge.jev.doneneutral.acclo\endcsname{0.714}
\expandafter\def\csname cdnum@rv.cas.kevlarge.jev.doneneutral.costfrac\endcsname{2.37}
\expandafter\def\csname cdnum@rv.cas.kevlarge.jev.doneneutral.esc\endcsname{0.53}
\expandafter\def\csname cdnum@rv.cas.kevlarge.jev.doneneutral.gain\endcsname{0.6}
\expandafter\def\csname cdnum@rv.cas.kevlarge.jev.doneneutral.gainhi\endcsname{1.5}
\expandafter\def\csname cdnum@rv.cas.kevlarge.jev.doneneutral.gainlo\endcsname{\ensuremath{-}0.4}
\expandafter\def\csname cdnum@rv.cas.kevlarge.jev.dtwo.acc\endcsname{0.910}
\expandafter\def\csname cdnum@rv.cas.kevlarge.jev.dtwo.acchi\endcsname{0.932}
\expandafter\def\csname cdnum@rv.cas.kevlarge.jev.dtwo.acclo\endcsname{0.880}
\expandafter\def\csname cdnum@rv.cas.kevlarge.jev.dtwo.costfrac\endcsname{1.08}
\expandafter\def\csname cdnum@rv.cas.kevlarge.jev.dtwo.esc\endcsname{0.56}
\expandafter\def\csname cdnum@rv.cas.kevlarge.jev.dtwo.gain\endcsname{\ensuremath{-}0.3}
\expandafter\def\csname cdnum@rv.cas.kevlarge.jev.dtwo.gainhi\endcsname{0.2}
\expandafter\def\csname cdnum@rv.cas.kevlarge.jev.dtwo.gainlo\endcsname{\ensuremath{-}1.8}
\expandafter\def\csname cdnum@rv.cas.kevsmall.compgem.doneneutral.acc\endcsname{0.741}
\expandafter\def\csname cdnum@rv.cas.kevsmall.compgem.doneneutral.acchi\endcsname{0.761}
\expandafter\def\csname cdnum@rv.cas.kevsmall.compgem.doneneutral.acclo\endcsname{0.718}
\expandafter\def\csname cdnum@rv.cas.kevsmall.compgem.doneneutral.costfrac\endcsname{1.03}
\expandafter\def\csname cdnum@rv.cas.kevsmall.compgem.doneneutral.esc\endcsname{0.99}
\expandafter\def\csname cdnum@rv.cas.kevsmall.compgem.doneneutral.gain\endcsname{0.0}
\expandafter\def\csname cdnum@rv.cas.kevsmall.compgem.doneneutral.gainhi\endcsname{0.0}
\expandafter\def\csname cdnum@rv.cas.kevsmall.compgem.doneneutral.gainlo\endcsname{\ensuremath{-}0.3}
\expandafter\def\csname cdnum@rv.cas.kevsmall.compgem.dtwo.acc\endcsname{0.940}
\expandafter\def\csname cdnum@rv.cas.kevsmall.compgem.dtwo.acchi\endcsname{0.957}
\expandafter\def\csname cdnum@rv.cas.kevsmall.compgem.dtwo.acclo\endcsname{0.920}
\expandafter\def\csname cdnum@rv.cas.kevsmall.compgem.dtwo.costfrac\endcsname{0.97}
\expandafter\def\csname cdnum@rv.cas.kevsmall.compgem.dtwo.esc\endcsname{0.94}
\expandafter\def\csname cdnum@rv.cas.kevsmall.compgem.dtwo.gain\endcsname{\ensuremath{-}0.2}
\expandafter\def\csname cdnum@rv.cas.kevsmall.compgem.dtwo.gainhi\endcsname{\ensuremath{-}0.2}
\expandafter\def\csname cdnum@rv.cas.kevsmall.compgem.dtwo.gainlo\endcsname{\ensuremath{-}0.8}
\expandafter\def\csname cdnum@rv.cas.kevsmall.compopen.doneneutral.acc\endcsname{0.679}
\expandafter\def\csname cdnum@rv.cas.kevsmall.compopen.doneneutral.acchi\endcsname{0.703}
\expandafter\def\csname cdnum@rv.cas.kevsmall.compopen.doneneutral.acclo\endcsname{0.652}
\expandafter\def\csname cdnum@rv.cas.kevsmall.compopen.doneneutral.costfrac\endcsname{1.06}
\expandafter\def\csname cdnum@rv.cas.kevsmall.compopen.doneneutral.esc\endcsname{0.96}
\expandafter\def\csname cdnum@rv.cas.kevsmall.compopen.doneneutral.gain\endcsname{\ensuremath{-}0.1}
\expandafter\def\csname cdnum@rv.cas.kevsmall.compopen.doneneutral.gainhi\endcsname{0.2}
\expandafter\def\csname cdnum@rv.cas.kevsmall.compopen.doneneutral.gainlo\endcsname{\ensuremath{-}1.5}
\expandafter\def\csname cdnum@rv.cas.kevsmall.compopen.dtwo.acc\endcsname{0.740}
\expandafter\def\csname cdnum@rv.cas.kevsmall.compopen.dtwo.acchi\endcsname{0.782}
\expandafter\def\csname cdnum@rv.cas.kevsmall.compopen.dtwo.acclo\endcsname{0.705}
\expandafter\def\csname cdnum@rv.cas.kevsmall.compopen.dtwo.costfrac\endcsname{0.54}
\expandafter\def\csname cdnum@rv.cas.kevsmall.compopen.dtwo.esc\endcsname{0.47}
\expandafter\def\csname cdnum@rv.cas.kevsmall.compopen.dtwo.gain\endcsname{1.2}
\expandafter\def\csname cdnum@rv.cas.kevsmall.compopen.dtwo.gainhi\endcsname{4.2}
\expandafter\def\csname cdnum@rv.cas.kevsmall.compopen.dtwo.gainlo\endcsname{\ensuremath{-}1.2}
\expandafter\def\csname cdnum@rv.cas.kevsmall.comptwo.doneneutral.acc\endcsname{0.723}
\expandafter\def\csname cdnum@rv.cas.kevsmall.comptwo.doneneutral.acchi\endcsname{0.744}
\expandafter\def\csname cdnum@rv.cas.kevsmall.comptwo.doneneutral.acclo\endcsname{0.702}
\expandafter\def\csname cdnum@rv.cas.kevsmall.comptwo.doneneutral.costfrac\endcsname{1.00}
\expandafter\def\csname cdnum@rv.cas.kevsmall.comptwo.doneneutral.esc\endcsname{0.96}
\expandafter\def\csname cdnum@rv.cas.kevsmall.comptwo.doneneutral.gain\endcsname{\ensuremath{-}0.2}
\expandafter\def\csname cdnum@rv.cas.kevsmall.comptwo.doneneutral.gainhi\endcsname{0.1}
\expandafter\def\csname cdnum@rv.cas.kevsmall.comptwo.doneneutral.gainlo\endcsname{\ensuremath{-}0.8}
\expandafter\def\csname cdnum@rv.cas.kevsmall.comptwo.dtwo.acc\endcsname{0.738}
\expandafter\def\csname cdnum@rv.cas.kevsmall.comptwo.dtwo.acchi\endcsname{0.783}
\expandafter\def\csname cdnum@rv.cas.kevsmall.comptwo.dtwo.acclo\endcsname{0.702}
\expandafter\def\csname cdnum@rv.cas.kevsmall.comptwo.dtwo.costfrac\endcsname{0.42}
\expandafter\def\csname cdnum@rv.cas.kevsmall.comptwo.dtwo.esc\endcsname{0.36}
\expandafter\def\csname cdnum@rv.cas.kevsmall.comptwo.dtwo.gain\endcsname{5.3}
\expandafter\def\csname cdnum@rv.cas.kevsmall.comptwo.dtwo.gainhi\endcsname{8.3}
\expandafter\def\csname cdnum@rv.cas.kevsmall.comptwo.dtwo.gainlo\endcsname{2.0}
\expandafter\def\csname cdnum@rv.cas.kevsmall.comptwoll.doneneutral.acc\endcsname{0.731}
\expandafter\def\csname cdnum@rv.cas.kevsmall.comptwoll.doneneutral.acchi\endcsname{0.752}
\expandafter\def\csname cdnum@rv.cas.kevsmall.comptwoll.doneneutral.acclo\endcsname{0.709}
\expandafter\def\csname cdnum@rv.cas.kevsmall.comptwoll.doneneutral.costfrac\endcsname{1.01}
\expandafter\def\csname cdnum@rv.cas.kevsmall.comptwoll.doneneutral.esc\endcsname{0.96}
\expandafter\def\csname cdnum@rv.cas.kevsmall.comptwoll.doneneutral.gain\endcsname{\ensuremath{-}0.2}
\expandafter\def\csname cdnum@rv.cas.kevsmall.comptwoll.doneneutral.gainhi\endcsname{0.1}
\expandafter\def\csname cdnum@rv.cas.kevsmall.comptwoll.doneneutral.gainlo\endcsname{\ensuremath{-}1.0}
\expandafter\def\csname cdnum@rv.cas.kevsmall.comptwoll.dtwo.acc\endcsname{0.910}
\expandafter\def\csname cdnum@rv.cas.kevsmall.comptwoll.dtwo.acchi\endcsname{0.930}
\expandafter\def\csname cdnum@rv.cas.kevsmall.comptwoll.dtwo.acclo\endcsname{0.883}
\expandafter\def\csname cdnum@rv.cas.kevsmall.comptwoll.dtwo.costfrac\endcsname{0.95}
\expandafter\def\csname cdnum@rv.cas.kevsmall.comptwoll.dtwo.esc\endcsname{0.94}
\expandafter\def\csname cdnum@rv.cas.kevsmall.comptwoll.dtwo.gain\endcsname{\ensuremath{-}0.2}
\expandafter\def\csname cdnum@rv.cas.kevsmall.comptwoll.dtwo.gainhi\endcsname{\ensuremath{-}0.2}
\expandafter\def\csname cdnum@rv.cas.kevsmall.comptwoll.dtwo.gainlo\endcsname{\ensuremath{-}1.0}
\expandafter\def\csname cdnum@rv.cas.kevsmall.jev.doneneutral.acc\endcsname{0.732}
\expandafter\def\csname cdnum@rv.cas.kevsmall.jev.doneneutral.acchi\endcsname{0.752}
\expandafter\def\csname cdnum@rv.cas.kevsmall.jev.doneneutral.acclo\endcsname{0.708}
\expandafter\def\csname cdnum@rv.cas.kevsmall.jev.doneneutral.costfrac\endcsname{1.30}
\expandafter\def\csname cdnum@rv.cas.kevsmall.jev.doneneutral.esc\endcsname{0.99}
\expandafter\def\csname cdnum@rv.cas.kevsmall.jev.doneneutral.gain\endcsname{0.0}
\expandafter\def\csname cdnum@rv.cas.kevsmall.jev.doneneutral.gainhi\endcsname{0.0}
\expandafter\def\csname cdnum@rv.cas.kevsmall.jev.doneneutral.gainlo\endcsname{\ensuremath{-}1.3}
\expandafter\def\csname cdnum@rv.cas.kevsmall.jev.dtwo.acc\endcsname{0.912}
\expandafter\def\csname cdnum@rv.cas.kevsmall.jev.dtwo.acchi\endcsname{0.932}
\expandafter\def\csname cdnum@rv.cas.kevsmall.jev.dtwo.acclo\endcsname{0.887}
\expandafter\def\csname cdnum@rv.cas.kevsmall.jev.dtwo.costfrac\endcsname{1.03}
\expandafter\def\csname cdnum@rv.cas.kevsmall.jev.dtwo.esc\endcsname{0.94}
\expandafter\def\csname cdnum@rv.cas.kevsmall.jev.dtwo.gain\endcsname{\ensuremath{-}0.2}
\expandafter\def\csname cdnum@rv.cas.kevsmall.jev.dtwo.gainhi\endcsname{\ensuremath{-}0.2}
\expandafter\def\csname cdnum@rv.cas.kevsmall.jev.dtwo.gainlo\endcsname{\ensuremath{-}1.2}
\expandafter\def\csname cdnum@rv.cas.layaen.compgem.doneneutral.acc\endcsname{0.741}
\expandafter\def\csname cdnum@rv.cas.layaen.compgem.doneneutral.acchi\endcsname{0.761}
\expandafter\def\csname cdnum@rv.cas.layaen.compgem.doneneutral.acclo\endcsname{0.719}
\expandafter\def\csname cdnum@rv.cas.layaen.compgem.doneneutral.costfrac\endcsname{1.02}
\expandafter\def\csname cdnum@rv.cas.layaen.compgem.doneneutral.esc\endcsname{0.99}
\expandafter\def\csname cdnum@rv.cas.layaen.compgem.doneneutral.gain\endcsname{0.0}
\expandafter\def\csname cdnum@rv.cas.layaen.compgem.doneneutral.gainhi\endcsname{0.0}
\expandafter\def\csname cdnum@rv.cas.layaen.compgem.doneneutral.gainlo\endcsname{0.0}
\expandafter\def\csname cdnum@rv.cas.layaen.compopen.doneneutral.acc\endcsname{0.681}
\expandafter\def\csname cdnum@rv.cas.layaen.compopen.doneneutral.acchi\endcsname{0.704}
\expandafter\def\csname cdnum@rv.cas.layaen.compopen.doneneutral.acclo\endcsname{0.656}
\expandafter\def\csname cdnum@rv.cas.layaen.compopen.doneneutral.costfrac\endcsname{1.05}
\expandafter\def\csname cdnum@rv.cas.layaen.compopen.doneneutral.esc\endcsname{0.99}
\expandafter\def\csname cdnum@rv.cas.layaen.compopen.doneneutral.gain\endcsname{0.1}
\expandafter\def\csname cdnum@rv.cas.layaen.compopen.doneneutral.gainhi\endcsname{0.4}
\expandafter\def\csname cdnum@rv.cas.layaen.compopen.doneneutral.gainlo\endcsname{\ensuremath{-}0.8}
\expandafter\def\csname cdnum@rv.cas.layaen.comptwo.doneneutral.acc\endcsname{0.726}
\expandafter\def\csname cdnum@rv.cas.layaen.comptwo.doneneutral.acchi\endcsname{0.746}
\expandafter\def\csname cdnum@rv.cas.layaen.comptwo.doneneutral.acclo\endcsname{0.702}
\expandafter\def\csname cdnum@rv.cas.layaen.comptwo.doneneutral.costfrac\endcsname{1.02}
\expandafter\def\csname cdnum@rv.cas.layaen.comptwo.doneneutral.esc\endcsname{0.99}
\expandafter\def\csname cdnum@rv.cas.layaen.comptwo.doneneutral.gain\endcsname{0.1}
\expandafter\def\csname cdnum@rv.cas.layaen.comptwo.doneneutral.gainhi\endcsname{0.4}
\expandafter\def\csname cdnum@rv.cas.layaen.comptwo.doneneutral.gainlo\endcsname{\ensuremath{-}1.1}
\expandafter\def\csname cdnum@rv.cas.layaen.comptwoll.doneneutral.acc\endcsname{0.734}
\expandafter\def\csname cdnum@rv.cas.layaen.comptwoll.doneneutral.acchi\endcsname{0.754}
\expandafter\def\csname cdnum@rv.cas.layaen.comptwoll.doneneutral.acclo\endcsname{0.710}
\expandafter\def\csname cdnum@rv.cas.layaen.comptwoll.doneneutral.costfrac\endcsname{1.02}
\expandafter\def\csname cdnum@rv.cas.layaen.comptwoll.doneneutral.esc\endcsname{0.99}
\expandafter\def\csname cdnum@rv.cas.layaen.comptwoll.doneneutral.gain\endcsname{0.1}
\expandafter\def\csname cdnum@rv.cas.layaen.comptwoll.doneneutral.gainhi\endcsname{0.4}
\expandafter\def\csname cdnum@rv.cas.layaen.comptwoll.doneneutral.gainlo\endcsname{\ensuremath{-}1.1}
\expandafter\def\csname cdnum@rv.cas.layaen.jev.doneneutral.acc\endcsname{0.732}
\expandafter\def\csname cdnum@rv.cas.layaen.jev.doneneutral.acchi\endcsname{0.754}
\expandafter\def\csname cdnum@rv.cas.layaen.jev.doneneutral.acclo\endcsname{0.710}
\expandafter\def\csname cdnum@rv.cas.layaen.jev.doneneutral.costfrac\endcsname{1.20}
\expandafter\def\csname cdnum@rv.cas.layaen.jev.doneneutral.esc\endcsname{1.00}
\expandafter\def\csname cdnum@rv.cas.layaen.jev.doneneutral.gain\endcsname{0.0}
\expandafter\def\csname cdnum@rv.cas.layaen.jev.doneneutral.gainhi\endcsname{0.3}
\expandafter\def\csname cdnum@rv.cas.layaen.jev.doneneutral.gainlo\endcsname{0.0}
\expandafter\def\csname cdnum@rv.cas.layaml.compgem.doneneutral.acc\endcsname{0.741}
\expandafter\def\csname cdnum@rv.cas.layaml.compgem.doneneutral.acchi\endcsname{0.761}
\expandafter\def\csname cdnum@rv.cas.layaml.compgem.doneneutral.acclo\endcsname{0.719}
\expandafter\def\csname cdnum@rv.cas.layaml.compgem.doneneutral.costfrac\endcsname{1.01}
\expandafter\def\csname cdnum@rv.cas.layaml.compgem.doneneutral.esc\endcsname{1.00}
\expandafter\def\csname cdnum@rv.cas.layaml.compgem.doneneutral.gain\endcsname{0.0}
\expandafter\def\csname cdnum@rv.cas.layaml.compgem.doneneutral.gainhi\endcsname{0.0}
\expandafter\def\csname cdnum@rv.cas.layaml.compgem.doneneutral.gainlo\endcsname{0.0}
\expandafter\def\csname cdnum@rv.cas.layaml.compgem.dtwo.acc\endcsname{0.942}
\expandafter\def\csname cdnum@rv.cas.layaml.compgem.dtwo.acchi\endcsname{0.960}
\expandafter\def\csname cdnum@rv.cas.layaml.compgem.dtwo.acclo\endcsname{0.923}
\expandafter\def\csname cdnum@rv.cas.layaml.compgem.dtwo.costfrac\endcsname{1.02}
\expandafter\def\csname cdnum@rv.cas.layaml.compgem.dtwo.esc\endcsname{1.00}
\expandafter\def\csname cdnum@rv.cas.layaml.compgem.dtwo.gain\endcsname{0.0}
\expandafter\def\csname cdnum@rv.cas.layaml.compgem.dtwo.gainhi\endcsname{0.0}
\expandafter\def\csname cdnum@rv.cas.layaml.compgem.dtwo.gainlo\endcsname{0.0}
\expandafter\def\csname cdnum@rv.cas.layaml.compopen.doneneutral.acc\endcsname{0.680}
\expandafter\def\csname cdnum@rv.cas.layaml.compopen.doneneutral.acchi\endcsname{0.704}
\expandafter\def\csname cdnum@rv.cas.layaml.compopen.doneneutral.acclo\endcsname{0.657}
\expandafter\def\csname cdnum@rv.cas.layaml.compopen.doneneutral.costfrac\endcsname{1.03}
\expandafter\def\csname cdnum@rv.cas.layaml.compopen.doneneutral.esc\endcsname{1.00}
\expandafter\def\csname cdnum@rv.cas.layaml.compopen.doneneutral.gain\endcsname{0.0}
\expandafter\def\csname cdnum@rv.cas.layaml.compopen.doneneutral.gainhi\endcsname{0.3}
\expandafter\def\csname cdnum@rv.cas.layaml.compopen.doneneutral.gainlo\endcsname{\ensuremath{-}0.2}
\expandafter\def\csname cdnum@rv.cas.layaml.compopen.dtwo.acc\endcsname{0.728}
\expandafter\def\csname cdnum@rv.cas.layaml.compopen.dtwo.acchi\endcsname{0.763}
\expandafter\def\csname cdnum@rv.cas.layaml.compopen.dtwo.acclo\endcsname{0.695}
\expandafter\def\csname cdnum@rv.cas.layaml.compopen.dtwo.costfrac\endcsname{1.03}
\expandafter\def\csname cdnum@rv.cas.layaml.compopen.dtwo.esc\endcsname{1.00}
\expandafter\def\csname cdnum@rv.cas.layaml.compopen.dtwo.gain\endcsname{0.0}
\expandafter\def\csname cdnum@rv.cas.layaml.compopen.dtwo.gainhi\endcsname{0.0}
\expandafter\def\csname cdnum@rv.cas.layaml.compopen.dtwo.gainlo\endcsname{0.0}
\expandafter\def\csname cdnum@rv.cas.layaml.comptwo.doneneutral.acc\endcsname{0.725}
\expandafter\def\csname cdnum@rv.cas.layaml.comptwo.doneneutral.acchi\endcsname{0.745}
\expandafter\def\csname cdnum@rv.cas.layaml.comptwo.doneneutral.acclo\endcsname{0.704}
\expandafter\def\csname cdnum@rv.cas.layaml.comptwo.doneneutral.costfrac\endcsname{1.01}
\expandafter\def\csname cdnum@rv.cas.layaml.comptwo.doneneutral.esc\endcsname{1.00}
\expandafter\def\csname cdnum@rv.cas.layaml.comptwo.doneneutral.gain\endcsname{0.0}
\expandafter\def\csname cdnum@rv.cas.layaml.comptwo.doneneutral.gainhi\endcsname{0.3}
\expandafter\def\csname cdnum@rv.cas.layaml.comptwo.doneneutral.gainlo\endcsname{\ensuremath{-}0.1}
\expandafter\def\csname cdnum@rv.cas.layaml.comptwo.dtwo.acc\endcsname{0.683}
\expandafter\def\csname cdnum@rv.cas.layaml.comptwo.dtwo.acchi\endcsname{0.720}
\expandafter\def\csname cdnum@rv.cas.layaml.comptwo.dtwo.acclo\endcsname{0.648}
\expandafter\def\csname cdnum@rv.cas.layaml.comptwo.dtwo.costfrac\endcsname{1.03}
\expandafter\def\csname cdnum@rv.cas.layaml.comptwo.dtwo.esc\endcsname{1.00}
\expandafter\def\csname cdnum@rv.cas.layaml.comptwo.dtwo.gain\endcsname{0.0}
\expandafter\def\csname cdnum@rv.cas.layaml.comptwo.dtwo.gainhi\endcsname{0.0}
\expandafter\def\csname cdnum@rv.cas.layaml.comptwo.dtwo.gainlo\endcsname{\ensuremath{-}0.5}
\expandafter\def\csname cdnum@rv.cas.layaml.comptwoll.doneneutral.acc\endcsname{0.733}
\expandafter\def\csname cdnum@rv.cas.layaml.comptwoll.doneneutral.acchi\endcsname{0.754}
\expandafter\def\csname cdnum@rv.cas.layaml.comptwoll.doneneutral.acclo\endcsname{0.712}
\expandafter\def\csname cdnum@rv.cas.layaml.comptwoll.doneneutral.costfrac\endcsname{1.01}
\expandafter\def\csname cdnum@rv.cas.layaml.comptwoll.doneneutral.esc\endcsname{1.00}
\expandafter\def\csname cdnum@rv.cas.layaml.comptwoll.doneneutral.gain\endcsname{0.0}
\expandafter\def\csname cdnum@rv.cas.layaml.comptwoll.doneneutral.gainhi\endcsname{0.3}
\expandafter\def\csname cdnum@rv.cas.layaml.comptwoll.doneneutral.gainlo\endcsname{\ensuremath{-}0.2}
\expandafter\def\csname cdnum@rv.cas.layaml.comptwoll.dtwo.acc\endcsname{0.912}
\expandafter\def\csname cdnum@rv.cas.layaml.comptwoll.dtwo.acchi\endcsname{0.932}
\expandafter\def\csname cdnum@rv.cas.layaml.comptwoll.dtwo.acclo\endcsname{0.888}
\expandafter\def\csname cdnum@rv.cas.layaml.comptwoll.dtwo.costfrac\endcsname{1.00}
\expandafter\def\csname cdnum@rv.cas.layaml.comptwoll.dtwo.esc\endcsname{1.00}
\expandafter\def\csname cdnum@rv.cas.layaml.comptwoll.dtwo.gain\endcsname{0.0}
\expandafter\def\csname cdnum@rv.cas.layaml.comptwoll.dtwo.gainhi\endcsname{0.0}
\expandafter\def\csname cdnum@rv.cas.layaml.comptwoll.dtwo.gainlo\endcsname{0.0}
\expandafter\def\csname cdnum@rv.cas.layaml.jev.doneneutral.acc\endcsname{0.732}
\expandafter\def\csname cdnum@rv.cas.layaml.jev.doneneutral.acchi\endcsname{0.752}
\expandafter\def\csname cdnum@rv.cas.layaml.jev.doneneutral.acclo\endcsname{0.710}
\expandafter\def\csname cdnum@rv.cas.layaml.jev.doneneutral.costfrac\endcsname{1.11}
\expandafter\def\csname cdnum@rv.cas.layaml.jev.doneneutral.esc\endcsname{1.00}
\expandafter\def\csname cdnum@rv.cas.layaml.jev.doneneutral.gain\endcsname{0.0}
\expandafter\def\csname cdnum@rv.cas.layaml.jev.doneneutral.gainhi\endcsname{0.1}
\expandafter\def\csname cdnum@rv.cas.layaml.jev.doneneutral.gainlo\endcsname{\ensuremath{-}0.1}
\expandafter\def\csname cdnum@rv.cas.layaml.jev.dtwo.acc\endcsname{0.913}
\expandafter\def\csname cdnum@rv.cas.layaml.jev.dtwo.acchi\endcsname{0.935}
\expandafter\def\csname cdnum@rv.cas.layaml.jev.dtwo.acclo\endcsname{0.888}
\expandafter\def\csname cdnum@rv.cas.layaml.jev.dtwo.costfrac\endcsname{1.04}
\expandafter\def\csname cdnum@rv.cas.layaml.jev.dtwo.esc\endcsname{1.00}
\expandafter\def\csname cdnum@rv.cas.layaml.jev.dtwo.gain\endcsname{0.0}
\expandafter\def\csname cdnum@rv.cas.layaml.jev.dtwo.gainhi\endcsname{0.0}
\expandafter\def\csname cdnum@rv.cas.layaml.jev.dtwo.gainlo\endcsname{\ensuremath{-}0.5}
\expandafter\def\csname cdnum@rv.cas.nimble.compgem.doneneutral.acc\endcsname{0.743}
\expandafter\def\csname cdnum@rv.cas.nimble.compgem.doneneutral.acchi\endcsname{0.764}
\expandafter\def\csname cdnum@rv.cas.nimble.compgem.doneneutral.acclo\endcsname{0.713}
\expandafter\def\csname cdnum@rv.cas.nimble.compgem.doneneutral.costfrac\endcsname{1.72}
\expandafter\def\csname cdnum@rv.cas.nimble.compgem.doneneutral.esc\endcsname{0.86}
\expandafter\def\csname cdnum@rv.cas.nimble.compgem.doneneutral.gain\endcsname{0.2}
\expandafter\def\csname cdnum@rv.cas.nimble.compgem.doneneutral.gainhi\endcsname{0.9}
\expandafter\def\csname cdnum@rv.cas.nimble.compgem.doneneutral.gainlo\endcsname{\ensuremath{-}2.0}
\expandafter\def\csname cdnum@rv.cas.nimble.compgem.dtwo.acc\endcsname{0.942}
\expandafter\def\csname cdnum@rv.cas.nimble.compgem.dtwo.acchi\endcsname{0.958}
\expandafter\def\csname cdnum@rv.cas.nimble.compgem.dtwo.acclo\endcsname{0.920}
\expandafter\def\csname cdnum@rv.cas.nimble.compgem.dtwo.costfrac\endcsname{1.20}
\expandafter\def\csname cdnum@rv.cas.nimble.compgem.dtwo.esc\endcsname{0.43}
\expandafter\def\csname cdnum@rv.cas.nimble.compgem.dtwo.gain\endcsname{0.0}
\expandafter\def\csname cdnum@rv.cas.nimble.compgem.dtwo.gainhi\endcsname{0.5}
\expandafter\def\csname cdnum@rv.cas.nimble.compgem.dtwo.gainlo\endcsname{\ensuremath{-}0.8}
\expandafter\def\csname cdnum@rv.cas.nimble.compopen.doneneutral.acc\endcsname{0.708}
\expandafter\def\csname cdnum@rv.cas.nimble.compopen.doneneutral.acchi\endcsname{0.730}
\expandafter\def\csname cdnum@rv.cas.nimble.compopen.doneneutral.acclo\endcsname{0.684}
\expandafter\def\csname cdnum@rv.cas.nimble.compopen.doneneutral.costfrac\endcsname{2.31}
\expandafter\def\csname cdnum@rv.cas.nimble.compopen.doneneutral.esc\endcsname{0.31}
\expandafter\def\csname cdnum@rv.cas.nimble.compopen.doneneutral.gain\endcsname{2.2}
\expandafter\def\csname cdnum@rv.cas.nimble.compopen.doneneutral.gainhi\endcsname{3.3}
\expandafter\def\csname cdnum@rv.cas.nimble.compopen.doneneutral.gainlo\endcsname{0.3}
\expandafter\def\csname cdnum@rv.cas.nimble.compopen.dtwo.acc\endcsname{0.860}
\expandafter\def\csname cdnum@rv.cas.nimble.compopen.dtwo.acchi\endcsname{0.885}
\expandafter\def\csname cdnum@rv.cas.nimble.compopen.dtwo.acclo\endcsname{0.827}
\expandafter\def\csname cdnum@rv.cas.nimble.compopen.dtwo.costfrac\endcsname{1.48}
\expandafter\def\csname cdnum@rv.cas.nimble.compopen.dtwo.esc\endcsname{0.02}
\expandafter\def\csname cdnum@rv.cas.nimble.compopen.dtwo.gain\endcsname{0.0}
\expandafter\def\csname cdnum@rv.cas.nimble.compopen.dtwo.gainhi\endcsname{1.5}
\expandafter\def\csname cdnum@rv.cas.nimble.compopen.dtwo.gainlo\endcsname{\ensuremath{-}1.5}
\expandafter\def\csname cdnum@rv.cas.nimble.comptwo.doneneutral.acc\endcsname{0.727}
\expandafter\def\csname cdnum@rv.cas.nimble.comptwo.doneneutral.acchi\endcsname{0.748}
\expandafter\def\csname cdnum@rv.cas.nimble.comptwo.doneneutral.acclo\endcsname{0.701}
\expandafter\def\csname cdnum@rv.cas.nimble.comptwo.doneneutral.costfrac\endcsname{1.29}
\expandafter\def\csname cdnum@rv.cas.nimble.comptwo.doneneutral.esc\endcsname{0.53}
\expandafter\def\csname cdnum@rv.cas.nimble.comptwo.doneneutral.gain\endcsname{0.2}
\expandafter\def\csname cdnum@rv.cas.nimble.comptwo.doneneutral.gainhi\endcsname{1.3}
\expandafter\def\csname cdnum@rv.cas.nimble.comptwo.doneneutral.gainlo\endcsname{\ensuremath{-}1.6}
\expandafter\def\csname cdnum@rv.cas.nimble.comptwo.dtwo.acc\endcsname{0.872}
\expandafter\def\csname cdnum@rv.cas.nimble.comptwo.dtwo.acchi\endcsname{0.895}
\expandafter\def\csname cdnum@rv.cas.nimble.comptwo.dtwo.acclo\endcsname{0.838}
\expandafter\def\csname cdnum@rv.cas.nimble.comptwo.dtwo.costfrac\endcsname{1.22}
\expandafter\def\csname cdnum@rv.cas.nimble.comptwo.dtwo.esc\endcsname{0.04}
\expandafter\def\csname cdnum@rv.cas.nimble.comptwo.dtwo.gain\endcsname{1.2}
\expandafter\def\csname cdnum@rv.cas.nimble.comptwo.dtwo.gainhi\endcsname{2.5}
\expandafter\def\csname cdnum@rv.cas.nimble.comptwo.dtwo.gainlo\endcsname{\ensuremath{-}0.5}
\expandafter\def\csname cdnum@rv.cas.nimble.comptwoll.doneneutral.acc\endcsname{0.736}
\expandafter\def\csname cdnum@rv.cas.nimble.comptwoll.doneneutral.acchi\endcsname{0.753}
\expandafter\def\csname cdnum@rv.cas.nimble.comptwoll.doneneutral.acclo\endcsname{0.703}
\expandafter\def\csname cdnum@rv.cas.nimble.comptwoll.doneneutral.costfrac\endcsname{1.59}
\expandafter\def\csname cdnum@rv.cas.nimble.comptwoll.doneneutral.esc\endcsname{0.67}
\expandafter\def\csname cdnum@rv.cas.nimble.comptwoll.doneneutral.gain\endcsname{0.3}
\expandafter\def\csname cdnum@rv.cas.nimble.comptwoll.doneneutral.gainhi\endcsname{0.9}
\expandafter\def\csname cdnum@rv.cas.nimble.comptwoll.doneneutral.gainlo\endcsname{\ensuremath{-}2.4}
\expandafter\def\csname cdnum@rv.cas.nimble.comptwoll.dtwo.acc\endcsname{0.908}
\expandafter\def\csname cdnum@rv.cas.nimble.comptwoll.dtwo.acchi\endcsname{0.930}
\expandafter\def\csname cdnum@rv.cas.nimble.comptwoll.dtwo.acclo\endcsname{0.883}
\expandafter\def\csname cdnum@rv.cas.nimble.comptwoll.dtwo.costfrac\endcsname{0.59}
\expandafter\def\csname cdnum@rv.cas.nimble.comptwoll.dtwo.esc\endcsname{0.40}
\expandafter\def\csname cdnum@rv.cas.nimble.comptwoll.dtwo.gain\endcsname{\ensuremath{-}0.3}
\expandafter\def\csname cdnum@rv.cas.nimble.comptwoll.dtwo.gainhi\endcsname{0.5}
\expandafter\def\csname cdnum@rv.cas.nimble.comptwoll.dtwo.gainlo\endcsname{\ensuremath{-}1.3}
\expandafter\def\csname cdnum@rv.cas.nimble.jev.doneneutral.acc\endcsname{0.732}
\expandafter\def\csname cdnum@rv.cas.nimble.jev.doneneutral.acchi\endcsname{0.756}
\expandafter\def\csname cdnum@rv.cas.nimble.jev.doneneutral.acclo\endcsname{0.711}
\expandafter\def\csname cdnum@rv.cas.nimble.jev.doneneutral.costfrac\endcsname{7.55}
\expandafter\def\csname cdnum@rv.cas.nimble.jev.doneneutral.esc\endcsname{0.86}
\expandafter\def\csname cdnum@rv.cas.nimble.jev.doneneutral.gain\endcsname{0.0}
\expandafter\def\csname cdnum@rv.cas.nimble.jev.doneneutral.gainhi\endcsname{1.1}
\expandafter\def\csname cdnum@rv.cas.nimble.jev.doneneutral.gainlo\endcsname{\ensuremath{-}1.2}
\expandafter\def\csname cdnum@rv.cas.nimble.jev.dtwo.acc\endcsname{0.910}
\expandafter\def\csname cdnum@rv.cas.nimble.jev.dtwo.acchi\endcsname{0.933}
\expandafter\def\csname cdnum@rv.cas.nimble.jev.dtwo.acclo\endcsname{0.882}
\expandafter\def\csname cdnum@rv.cas.nimble.jev.dtwo.costfrac\endcsname{2.33}
\expandafter\def\csname cdnum@rv.cas.nimble.jev.dtwo.esc\endcsname{0.28}
\expandafter\def\csname cdnum@rv.cas.nimble.jev.dtwo.gain\endcsname{\ensuremath{-}0.3}
\expandafter\def\csname cdnum@rv.cas.nimble.jev.dtwo.gainhi\endcsname{0.3}
\expandafter\def\csname cdnum@rv.cas.nimble.jev.dtwo.gainlo\endcsname{\ensuremath{-}1.7}
\expandafter\def\csname cdnum@rv.cas.thisthat.compgem.doneneutral.acc\endcsname{0.740}
\expandafter\def\csname cdnum@rv.cas.thisthat.compgem.doneneutral.acchi\endcsname{0.760}
\expandafter\def\csname cdnum@rv.cas.thisthat.compgem.doneneutral.acclo\endcsname{0.717}
\expandafter\def\csname cdnum@rv.cas.thisthat.compgem.doneneutral.costfrac\endcsname{0.97}
\expandafter\def\csname cdnum@rv.cas.thisthat.compgem.doneneutral.esc\endcsname{0.94}
\expandafter\def\csname cdnum@rv.cas.thisthat.compgem.doneneutral.gain\endcsname{\ensuremath{-}0.1}
\expandafter\def\csname cdnum@rv.cas.thisthat.compgem.doneneutral.gainhi\endcsname{0.0}
\expandafter\def\csname cdnum@rv.cas.thisthat.compgem.doneneutral.gainlo\endcsname{\ensuremath{-}1.2}
\expandafter\def\csname cdnum@rv.cas.thisthat.compgem.dtwo.acc\endcsname{0.942}
\expandafter\def\csname cdnum@rv.cas.thisthat.compgem.dtwo.acchi\endcsname{0.957}
\expandafter\def\csname cdnum@rv.cas.thisthat.compgem.dtwo.acclo\endcsname{0.917}
\expandafter\def\csname cdnum@rv.cas.thisthat.compgem.dtwo.costfrac\endcsname{0.37}
\expandafter\def\csname cdnum@rv.cas.thisthat.compgem.dtwo.esc\endcsname{0.30}
\expandafter\def\csname cdnum@rv.cas.thisthat.compgem.dtwo.eschi\endcsname{0.56}
\expandafter\def\csname cdnum@rv.cas.thisthat.compgem.dtwo.esclo\endcsname{0.23}
\expandafter\def\csname cdnum@rv.cas.thisthat.compgem.dtwo.gain\endcsname{0.0}
\expandafter\def\csname cdnum@rv.cas.thisthat.compgem.dtwo.gainhi\endcsname{1.3}
\expandafter\def\csname cdnum@rv.cas.thisthat.compgem.dtwo.gainlo\endcsname{\ensuremath{-}1.7}
\expandafter\def\csname cdnum@rv.cas.thisthat.compgem.dtwo.pfifty\endcsname{153}
\expandafter\def\csname cdnum@rv.cas.thisthat.compgem.dtwo.pninetyfive\endcsname{1039}
\expandafter\def\csname cdnum@rv.cas.thisthat.compgem.dtwo.usd\endcsname{0.108}
\expandafter\def\csname cdnum@rv.cas.thisthat.compopen.doneneutral.acc\endcsname{0.680}
\expandafter\def\csname cdnum@rv.cas.thisthat.compopen.doneneutral.acchi\endcsname{0.701}
\expandafter\def\csname cdnum@rv.cas.thisthat.compopen.doneneutral.acclo\endcsname{0.648}
\expandafter\def\csname cdnum@rv.cas.thisthat.compopen.doneneutral.costfrac\endcsname{0.62}
\expandafter\def\csname cdnum@rv.cas.thisthat.compopen.doneneutral.esc\endcsname{0.55}
\expandafter\def\csname cdnum@rv.cas.thisthat.compopen.doneneutral.gain\endcsname{0.0}
\expandafter\def\csname cdnum@rv.cas.thisthat.compopen.doneneutral.gainhi\endcsname{1.3}
\expandafter\def\csname cdnum@rv.cas.thisthat.compopen.doneneutral.gainlo\endcsname{\ensuremath{-}2.9}
\expandafter\def\csname cdnum@rv.cas.thisthat.compopen.dtwo.acc\endcsname{0.868}
\expandafter\def\csname cdnum@rv.cas.thisthat.compopen.dtwo.acchi\endcsname{0.892}
\expandafter\def\csname cdnum@rv.cas.thisthat.compopen.dtwo.acclo\endcsname{0.833}
\expandafter\def\csname cdnum@rv.cas.thisthat.compopen.dtwo.costfrac\endcsname{0.20}
\expandafter\def\csname cdnum@rv.cas.thisthat.compopen.dtwo.esc\endcsname{0.06}
\expandafter\def\csname cdnum@rv.cas.thisthat.compopen.dtwo.gain\endcsname{0.5}
\expandafter\def\csname cdnum@rv.cas.thisthat.compopen.dtwo.gainhi\endcsname{2.2}
\expandafter\def\csname cdnum@rv.cas.thisthat.compopen.dtwo.gainlo\endcsname{\ensuremath{-}1.8}
\expandafter\def\csname cdnum@rv.cas.thisthat.comptwo.doneneutral.acc\endcsname{0.724}
\expandafter\def\csname cdnum@rv.cas.thisthat.comptwo.doneneutral.acchi\endcsname{0.745}
\expandafter\def\csname cdnum@rv.cas.thisthat.comptwo.doneneutral.acclo\endcsname{0.702}
\expandafter\def\csname cdnum@rv.cas.thisthat.comptwo.doneneutral.costfrac\endcsname{0.97}
\expandafter\def\csname cdnum@rv.cas.thisthat.comptwo.doneneutral.esc\endcsname{0.95}
\expandafter\def\csname cdnum@rv.cas.thisthat.comptwo.doneneutral.gain\endcsname{\ensuremath{-}0.1}
\expandafter\def\csname cdnum@rv.cas.thisthat.comptwo.doneneutral.gainhi\endcsname{0.1}
\expandafter\def\csname cdnum@rv.cas.thisthat.comptwo.doneneutral.gainlo\endcsname{\ensuremath{-}0.4}
\expandafter\def\csname cdnum@rv.cas.thisthat.comptwo.dtwo.acc\endcsname{0.870}
\expandafter\def\csname cdnum@rv.cas.thisthat.comptwo.dtwo.acchi\endcsname{0.900}
\expandafter\def\csname cdnum@rv.cas.thisthat.comptwo.dtwo.acclo\endcsname{0.838}
\expandafter\def\csname cdnum@rv.cas.thisthat.comptwo.dtwo.costfrac\endcsname{0.17}
\expandafter\def\csname cdnum@rv.cas.thisthat.comptwo.dtwo.esc\endcsname{0.06}
\expandafter\def\csname cdnum@rv.cas.thisthat.comptwo.dtwo.gain\endcsname{0.7}
\expandafter\def\csname cdnum@rv.cas.thisthat.comptwo.dtwo.gainhi\endcsname{2.7}
\expandafter\def\csname cdnum@rv.cas.thisthat.comptwo.dtwo.gainlo\endcsname{\ensuremath{-}1.2}
\expandafter\def\csname cdnum@rv.cas.thisthat.comptwoll.doneneutral.acc\endcsname{0.732}
\expandafter\def\csname cdnum@rv.cas.thisthat.comptwoll.doneneutral.acchi\endcsname{0.753}
\expandafter\def\csname cdnum@rv.cas.thisthat.comptwoll.doneneutral.acclo\endcsname{0.710}
\expandafter\def\csname cdnum@rv.cas.thisthat.comptwoll.doneneutral.costfrac\endcsname{0.99}
\expandafter\def\csname cdnum@rv.cas.thisthat.comptwoll.doneneutral.esc\endcsname{0.95}
\expandafter\def\csname cdnum@rv.cas.thisthat.comptwoll.doneneutral.gain\endcsname{\ensuremath{-}0.1}
\expandafter\def\csname cdnum@rv.cas.thisthat.comptwoll.doneneutral.gainhi\endcsname{0.1}
\expandafter\def\csname cdnum@rv.cas.thisthat.comptwoll.doneneutral.gainlo\endcsname{\ensuremath{-}0.8}
\expandafter\def\csname cdnum@rv.cas.thisthat.comptwoll.dtwo.acc\endcsname{0.932}
\expandafter\def\csname cdnum@rv.cas.thisthat.comptwoll.dtwo.acchi\endcsname{0.948}
\expandafter\def\csname cdnum@rv.cas.thisthat.comptwoll.dtwo.acclo\endcsname{0.900}
\expandafter\def\csname cdnum@rv.cas.thisthat.comptwoll.dtwo.costfrac\endcsname{0.27}
\expandafter\def\csname cdnum@rv.cas.thisthat.comptwoll.dtwo.esc\endcsname{0.25}
\expandafter\def\csname cdnum@rv.cas.thisthat.comptwoll.dtwo.gain\endcsname{2.0}
\expandafter\def\csname cdnum@rv.cas.thisthat.comptwoll.dtwo.gainhi\endcsname{3.3}
\expandafter\def\csname cdnum@rv.cas.thisthat.comptwoll.dtwo.gainlo\endcsname{\ensuremath{-}0.5}
\expandafter\def\csname cdnum@rv.cas.thisthat.jev.doneneutral.acc\endcsname{0.731}
\expandafter\def\csname cdnum@rv.cas.thisthat.jev.doneneutral.acchi\endcsname{0.751}
\expandafter\def\csname cdnum@rv.cas.thisthat.jev.doneneutral.acclo\endcsname{0.708}
\expandafter\def\csname cdnum@rv.cas.thisthat.jev.doneneutral.costfrac\endcsname{1.19}
\expandafter\def\csname cdnum@rv.cas.thisthat.jev.doneneutral.esc\endcsname{0.94}
\expandafter\def\csname cdnum@rv.cas.thisthat.jev.doneneutral.gain\endcsname{\ensuremath{-}0.1}
\expandafter\def\csname cdnum@rv.cas.thisthat.jev.doneneutral.gainhi\endcsname{0.0}
\expandafter\def\csname cdnum@rv.cas.thisthat.jev.doneneutral.gainlo\endcsname{\ensuremath{-}0.9}
\expandafter\def\csname cdnum@rv.cas.thisthat.jev.dtwo.acc\endcsname{0.927}
\expandafter\def\csname cdnum@rv.cas.thisthat.jev.dtwo.acchi\endcsname{0.945}
\expandafter\def\csname cdnum@rv.cas.thisthat.jev.dtwo.acclo\endcsname{0.898}
\expandafter\def\csname cdnum@rv.cas.thisthat.jev.dtwo.costfrac\endcsname{0.43}
\expandafter\def\csname cdnum@rv.cas.thisthat.jev.dtwo.esc\endcsname{0.25}
\expandafter\def\csname cdnum@rv.cas.thisthat.jev.dtwo.eschi\endcsname{0.40}
\expandafter\def\csname cdnum@rv.cas.thisthat.jev.dtwo.esclo\endcsname{0.20}
\expandafter\def\csname cdnum@rv.cas.thisthat.jev.dtwo.gain\endcsname{1.3}
\expandafter\def\csname cdnum@rv.cas.thisthat.jev.dtwo.gaindec\endcsname{8}
\expandafter\def\csname cdnum@rv.cas.thisthat.jev.dtwo.gainhi\endcsname{2.8}
\expandafter\def\csname cdnum@rv.cas.thisthat.jev.dtwo.gainlo\endcsname{\ensuremath{-}0.7}
\expandafter\def\csname cdnum@rv.cas.thisthat.jev.dtwo.gainp\endcsname{0.16}
\expandafter\def\csname cdnum@rv.cas.thisthat.jev.dtwo.n\endcsname{600}
\expandafter\def\csname cdnum@rv.cas.thisthat.jev.dtwo.pfifty\endcsname{153}
\expandafter\def\csname cdnum@rv.cas.thisthat.jev.dtwo.pninetyfive\endcsname{320}
\expandafter\def\csname cdnum@rv.cas.thisthat.jev.dtwo.usd\endcsname{0.048}
\expandafter\def\csname cdnum@rv.cost.thisthat.doneneutral.breakeven\endcsname{1.90}
\expandafter\def\csname cdnum@rv.cost.thisthat.doneneutral.usdquarter\endcsname{0.0080}
\expandafter\def\csname cdnum@rv.fail.compopen.answeredquestions\endcsname{10,004}
\expandafter\def\csname cdnum@rv.fail.compopen.percentscale\endcsname{3,525}
\expandafter\def\csname cdnum@rv.fail.compopen.renormalized\endcsname{4,601}
\expandafter\def\csname cdnum@rv.fail.layaen.dtwo.refusals\endcsname{800}
\expandafter\def\csname cdnum@rv.gate.compgem.cov\endcsname{0.95}
\expandafter\def\csname cdnum@rv.gate.compgem.covhi\endcsname{1.00}
\expandafter\def\csname cdnum@rv.gate.compgem.covlo\endcsname{0.74}
\expandafter\def\csname cdnum@rv.gate.compgem.far\endcsname{0.330}
\expandafter\def\csname cdnum@rv.gate.compgem.farhi\endcsname{0.823}
\expandafter\def\csname cdnum@rv.gate.compgem.farlo\endcsname{0.077}
\expandafter\def\csname cdnum@rv.gate.compgem.risk\endcsname{0.051}
\expandafter\def\csname cdnum@rv.gate.compgem.riskub\endcsname{0.069}
\expandafter\def\csname cdnum@rv.gate.compgemll.far\endcsname{0.655}
\expandafter\def\csname cdnum@rv.gate.compmis.far\endcsname{0.000}
\expandafter\def\csname cdnum@rv.gate.compmisll.far\endcsname{0.035}
\expandafter\def\csname cdnum@rv.gate.compopen.far\endcsname{0.005}
\expandafter\def\csname cdnum@rv.gate.comptwo.far\endcsname{0.325}
\expandafter\def\csname cdnum@rv.gate.comptwoll.cov\endcsname{0.87}
\expandafter\def\csname cdnum@rv.gate.comptwoll.covhi\endcsname{0.94}
\expandafter\def\csname cdnum@rv.gate.comptwoll.covlo\endcsname{0.80}
\expandafter\def\csname cdnum@rv.gate.comptwoll.far\endcsname{0.165}
\expandafter\def\csname cdnum@rv.gate.comptwoll.farhi\endcsname{0.346}
\expandafter\def\csname cdnum@rv.gate.comptwoll.farlo\endcsname{0.081}
\expandafter\def\csname cdnum@rv.gate.comptwoll.risk\endcsname{0.052}
\expandafter\def\csname cdnum@rv.gate.comptwoll.riskub\endcsname{0.071}
\expandafter\def\csname cdnum@rv.gate.comptwothink.far\endcsname{0.290}
\expandafter\def\csname cdnum@rv.gate.decider.cov\endcsname{0.77}
\expandafter\def\csname cdnum@rv.gate.decider.covhi\endcsname{0.85}
\expandafter\def\csname cdnum@rv.gate.decider.covlo\endcsname{0.65}
\expandafter\def\csname cdnum@rv.gate.decider.far\endcsname{0.155}
\expandafter\def\csname cdnum@rv.gate.decider.farhi\endcsname{0.300}
\expandafter\def\csname cdnum@rv.gate.decider.farlo\endcsname{0.083}
\expandafter\def\csname cdnum@rv.gate.decider.risk\endcsname{0.059}
\expandafter\def\csname cdnum@rv.gate.decider.riskub\endcsname{0.080}
\expandafter\def\csname cdnum@rv.gate.jev.cov\endcsname{0.92}
\expandafter\def\csname cdnum@rv.gate.jev.covhi\endcsname{0.97}
\expandafter\def\csname cdnum@rv.gate.jev.covlo\endcsname{0.81}
\expandafter\def\csname cdnum@rv.gate.jev.far\endcsname{0.310}
\expandafter\def\csname cdnum@rv.gate.jev.farhi\endcsname{0.502}
\expandafter\def\csname cdnum@rv.gate.jev.farlo\endcsname{0.142}
\expandafter\def\csname cdnum@rv.gate.jev.risk\endcsname{0.047}
\expandafter\def\csname cdnum@rv.gate.jev.riskub\endcsname{0.065}
\expandafter\def\csname cdnum@rv.gate.kevlarge.cov\endcsname{0.61}
\expandafter\def\csname cdnum@rv.gate.kevlarge.covhi\endcsname{0.72}
\expandafter\def\csname cdnum@rv.gate.kevlarge.covlo\endcsname{0.47}
\expandafter\def\csname cdnum@rv.gate.kevlarge.far\endcsname{0.015}
\expandafter\def\csname cdnum@rv.gate.kevlarge.farhi\endcsname{0.064}
\expandafter\def\csname cdnum@rv.gate.kevlarge.farlo\endcsname{0.000}
\expandafter\def\csname cdnum@rv.gate.kevlarge.risk\endcsname{0.051}
\expandafter\def\csname cdnum@rv.gate.kevlarge.riskub\endcsname{0.075}
\expandafter\def\csname cdnum@rv.gate.kevsmall.far\endcsname{0.010}
\expandafter\def\csname cdnum@rv.gate.layaml.far\endcsname{0.000}
\expandafter\def\csname cdnum@rv.gate.nimble.cov\endcsname{0.66}
\expandafter\def\csname cdnum@rv.gate.nimble.far\endcsname{0.030}
\expandafter\def\csname cdnum@rv.gate.thisthat.far\endcsname{0.255}
\expandafter\def\csname cdnum@rv.ho.compgem.doneneutral.five.cov\endcsname{0.12}
\expandafter\def\csname cdnum@rv.ho.compgem.doneneutral.five.risk\endcsname{0.016}
\expandafter\def\csname cdnum@rv.ho.compgem.doneneutral.five.riskub\endcsname{0.036}
\expandafter\def\csname cdnum@rv.ho.compgem.dtwo.five.cov\endcsname{0.95}
\expandafter\def\csname cdnum@rv.ho.compgem.dtwo.five.risk\endcsname{0.053}
\expandafter\def\csname cdnum@rv.ho.compgem.dtwo.five.riskub\endcsname{0.071}
\expandafter\def\csname cdnum@rv.ho.compgemll.doneneutral.five.cov\endcsname{0.38}
\expandafter\def\csname cdnum@rv.ho.compgemll.doneneutral.five.risk\endcsname{0.072}
\expandafter\def\csname cdnum@rv.ho.compgemll.doneneutral.five.riskub\endcsname{0.089}
\expandafter\def\csname cdnum@rv.ho.compgemll.dtwo.five.cov\endcsname{0.98}
\expandafter\def\csname cdnum@rv.ho.compgemll.dtwo.five.risk\endcsname{0.051}
\expandafter\def\csname cdnum@rv.ho.compgemll.dtwo.five.riskub\endcsname{0.069}
\expandafter\def\csname cdnum@rv.ho.compmis.doneneutral.five.cov\endcsname{0.00}
\expandafter\def\csname cdnum@rv.ho.compmis.dtwo.five.cov\endcsname{0.00}
\expandafter\def\csname cdnum@rv.ho.compmisll.doneneutral.five.cov\endcsname{0.04}
\expandafter\def\csname cdnum@rv.ho.compmisll.doneneutral.five.risk\endcsname{0.038}
\expandafter\def\csname cdnum@rv.ho.compmisll.doneneutral.five.riskub\endcsname{0.095}
\expandafter\def\csname cdnum@rv.ho.compmisll.dtwo.five.cov\endcsname{0.43}
\expandafter\def\csname cdnum@rv.ho.compmisll.dtwo.five.risk\endcsname{0.051}
\expandafter\def\csname cdnum@rv.ho.compmisll.dtwo.five.riskub\endcsname{0.080}
\expandafter\def\csname cdnum@rv.ho.compopen.doneneutral.five.cov\endcsname{0.00}
\expandafter\def\csname cdnum@rv.ho.compopen.doneneutral.five.risk\endcsname{0.000}
\expandafter\def\csname cdnum@rv.ho.compopen.doneneutral.five.riskub\endcsname{0.776}
\expandafter\def\csname cdnum@rv.ho.compopen.dtwo.five.cov\endcsname{0.02}
\expandafter\def\csname cdnum@rv.ho.compopen.dtwo.five.risk\endcsname{0.091}
\expandafter\def\csname cdnum@rv.ho.compopen.dtwo.five.riskub\endcsname{0.364}
\expandafter\def\csname cdnum@rv.ho.comptwo.doneneutral.five.cov\endcsname{0.00}
\expandafter\def\csname cdnum@rv.ho.comptwo.dtwo.five.cov\endcsname{0.69}
\expandafter\def\csname cdnum@rv.ho.comptwo.dtwo.five.risk\endcsname{0.048}
\expandafter\def\csname cdnum@rv.ho.comptwo.dtwo.five.riskub\endcsname{0.069}
\expandafter\def\csname cdnum@rv.ho.comptwoll.doneneutral.five.cov\endcsname{0.41}
\expandafter\def\csname cdnum@rv.ho.comptwoll.doneneutral.five.risk\endcsname{0.049}
\expandafter\def\csname cdnum@rv.ho.comptwoll.doneneutral.five.riskub\endcsname{0.064}
\expandafter\def\csname cdnum@rv.ho.comptwoll.dtwo.five.cov\endcsname{0.85}
\expandafter\def\csname cdnum@rv.ho.comptwoll.dtwo.five.risk\endcsname{0.049}
\expandafter\def\csname cdnum@rv.ho.comptwoll.dtwo.five.riskub\endcsname{0.067}
\expandafter\def\csname cdnum@rv.ho.comptwothink.doneneutral.five.cov\endcsname{0.00}
\expandafter\def\csname cdnum@rv.ho.comptwothink.doneneutral.five.risk\endcsname{0.000}
\expandafter\def\csname cdnum@rv.ho.comptwothink.doneneutral.five.riskub\endcsname{0.527}
\expandafter\def\csname cdnum@rv.ho.comptwothink.dtwo.five.cov\endcsname{0.87}
\expandafter\def\csname cdnum@rv.ho.comptwothink.dtwo.five.risk\endcsname{0.044}
\expandafter\def\csname cdnum@rv.ho.comptwothink.dtwo.five.riskub\endcsname{0.062}
\expandafter\def\csname cdnum@rv.ho.decider.doneneutral.five.cov\endcsname{0.04}
\expandafter\def\csname cdnum@rv.ho.decider.doneneutral.five.risk\endcsname{0.011}
\expandafter\def\csname cdnum@rv.ho.decider.doneneutral.five.riskub\endcsname{0.053}
\expandafter\def\csname cdnum@rv.ho.decider.dtwo.five.cov\endcsname{0.76}
\expandafter\def\csname cdnum@rv.ho.decider.dtwo.five.risk\endcsname{0.053}
\expandafter\def\csname cdnum@rv.ho.decider.dtwo.five.riskub\endcsname{0.074}
\expandafter\def\csname cdnum@rv.ho.jev.doneneutral.five.cov\endcsname{0.17}
\expandafter\def\csname cdnum@rv.ho.jev.doneneutral.five.risk\endcsname{0.058}
\expandafter\def\csname cdnum@rv.ho.jev.doneneutral.five.riskub\endcsname{0.084}
\expandafter\def\csname cdnum@rv.ho.jev.dtwo.five.cov\endcsname{0.93}
\expandafter\def\csname cdnum@rv.ho.jev.dtwo.five.risk\endcsname{0.049}
\expandafter\def\csname cdnum@rv.ho.jev.dtwo.five.riskub\endcsname{0.066}
\expandafter\def\csname cdnum@rv.ho.kevlarge.doneneutral.five.cov\endcsname{0.02}
\expandafter\def\csname cdnum@rv.ho.kevlarge.doneneutral.five.risk\endcsname{0.000}
\expandafter\def\csname cdnum@rv.ho.kevlarge.doneneutral.five.riskub\endcsname{0.066}
\expandafter\def\csname cdnum@rv.ho.kevlarge.dtwo.five.cov\endcsname{0.61}
\expandafter\def\csname cdnum@rv.ho.kevlarge.dtwo.five.risk\endcsname{0.049}
\expandafter\def\csname cdnum@rv.ho.kevlarge.dtwo.five.riskub\endcsname{0.072}
\expandafter\def\csname cdnum@rv.ho.kevsmall.doneneutral.five.cov\endcsname{0.01}
\expandafter\def\csname cdnum@rv.ho.kevsmall.doneneutral.five.risk\endcsname{0.000}
\expandafter\def\csname cdnum@rv.ho.kevsmall.doneneutral.five.riskub\endcsname{0.133}
\expandafter\def\csname cdnum@rv.ho.kevsmall.dtwo.five.cov\endcsname{0.23}
\expandafter\def\csname cdnum@rv.ho.kevsmall.dtwo.five.risk\endcsname{0.051}
\expandafter\def\csname cdnum@rv.ho.kevsmall.dtwo.five.riskub\endcsname{0.094}
\expandafter\def\csname cdnum@rv.ho.layaen.doneneutral.five.cov\endcsname{0.00}
\expandafter\def\csname cdnum@rv.ho.layaen.doneneutral.five.risk\endcsname{0.500}
\expandafter\def\csname cdnum@rv.ho.layaen.doneneutral.five.riskub\endcsname{0.975}
\expandafter\def\csname cdnum@rv.ho.layaml.doneneutral.five.cov\endcsname{0.00}
\expandafter\def\csname cdnum@rv.ho.layaml.doneneutral.five.risk\endcsname{0.000}
\expandafter\def\csname cdnum@rv.ho.layaml.doneneutral.five.riskub\endcsname{0.950}
\expandafter\def\csname cdnum@rv.ho.layaml.dtwo.five.cov\endcsname{0.00}
\expandafter\def\csname cdnum@rv.ho.nimble.doneneutral.five.cov\endcsname{0.16}
\expandafter\def\csname cdnum@rv.ho.nimble.doneneutral.five.risk\endcsname{0.047}
\expandafter\def\csname cdnum@rv.ho.nimble.doneneutral.five.riskub\endcsname{0.072}
\expandafter\def\csname cdnum@rv.ho.nimble.dtwo.five.cov\endcsname{0.66}
\expandafter\def\csname cdnum@rv.ho.nimble.dtwo.five.risk\endcsname{0.051}
\expandafter\def\csname cdnum@rv.ho.nimble.dtwo.five.riskub\endcsname{0.073}
\expandafter\def\csname cdnum@rv.ho.thisthat.doneneutral.five.cov\endcsname{0.05}
\expandafter\def\csname cdnum@rv.ho.thisthat.doneneutral.five.risk\endcsname{0.062}
\expandafter\def\csname cdnum@rv.ho.thisthat.doneneutral.five.riskub\endcsname{0.118}
\expandafter\def\csname cdnum@rv.ho.thisthat.dtwo.five.cov\endcsname{0.79}
\expandafter\def\csname cdnum@rv.ho.thisthat.dtwo.five.risk\endcsname{0.048}
\expandafter\def\csname cdnum@rv.ho.thisthat.dtwo.five.riskub\endcsname{0.068}
\expandafter\def\csname cdnum@rv.nm.jev.done.kswap.dauchi\endcsname{\ensuremath{-}0.52}
\expandafter\def\csname cdnum@rv.nm.jev.done.kswap.dauclo\endcsname{\ensuremath{-}0.62}
\expandafter\def\csname cdnum@rv.nm.jev.done.kswap.flipshi\endcsname{54.7}
\expandafter\def\csname cdnum@rv.nm.jev.done.kswap.flipslo\endcsname{46.2}
\expandafter\def\csname cdnum@rv.nm.jev.power.floor\endcsname{0.8}
\expandafter\def\csname cdnum@rv.nm.jev.toxicity.floor\endcsname{0.4}
\expandafter\def\csname cdnum@rv.pd.compgem.doneneutral.diff\endcsname{0.9}
\expandafter\def\csname cdnum@rv.pd.compgem.doneneutral.diffhi\endcsname{2.9}
\expandafter\def\csname cdnum@rv.pd.compgem.doneneutral.difflo\endcsname{\ensuremath{-}1.1}
\expandafter\def\csname cdnum@rv.pd.compgem.dtwo.diff\endcsname{2.8}
\expandafter\def\csname cdnum@rv.pd.compgem.dtwo.diffhi\endcsname{4.7}
\expandafter\def\csname cdnum@rv.pd.compgem.dtwo.difflo\endcsname{1.0}
\expandafter\def\csname cdnum@rv.pd.compgem.dtwo.holm\endcsname{0.004}
\expandafter\def\csname cdnum@rv.pd.compgem.emotion.diff\endcsname{\ensuremath{-}3.0}
\expandafter\def\csname cdnum@rv.pd.compgem.emotion.diffhi\endcsname{\ensuremath{-}0.2}
\expandafter\def\csname cdnum@rv.pd.compgem.emotion.difflo\endcsname{\ensuremath{-}6.0}
\expandafter\def\csname cdnum@rv.pd.compgem.politeness.diff\endcsname{0.4}
\expandafter\def\csname cdnum@rv.pd.compgem.politeness.diffhi\endcsname{3.8}
\expandafter\def\csname cdnum@rv.pd.compgem.politeness.difflo\endcsname{\ensuremath{-}3.0}
\expandafter\def\csname cdnum@rv.pd.compgem.power.diff\endcsname{0.6}
\expandafter\def\csname cdnum@rv.pd.compgem.power.diffhi\endcsname{3.8}
\expandafter\def\csname cdnum@rv.pd.compgem.power.difflo\endcsname{\ensuremath{-}2.6}
\expandafter\def\csname cdnum@rv.pd.compgem.toxicity.diff\endcsname{3.0}
\expandafter\def\csname cdnum@rv.pd.compgem.toxicity.diffhi\endcsname{5.2}
\expandafter\def\csname cdnum@rv.pd.compgem.toxicity.difflo\endcsname{0.8}
\expandafter\def\csname cdnum@rv.pd.compgemll.doneneutral.diff\endcsname{1.1}
\expandafter\def\csname cdnum@rv.pd.compgemll.doneneutral.diffhi\endcsname{3.0}
\expandafter\def\csname cdnum@rv.pd.compgemll.doneneutral.difflo\endcsname{\ensuremath{-}1.0}
\expandafter\def\csname cdnum@rv.pd.compgemll.dtwo.diff\endcsname{2.7}
\expandafter\def\csname cdnum@rv.pd.compgemll.dtwo.diffhi\endcsname{4.5}
\expandafter\def\csname cdnum@rv.pd.compgemll.dtwo.difflo\endcsname{1.0}
\expandafter\def\csname cdnum@rv.pd.compgemll.emotion.diff\endcsname{\ensuremath{-}3.0}
\expandafter\def\csname cdnum@rv.pd.compgemll.emotion.diffhi\endcsname{\ensuremath{-}0.2}
\expandafter\def\csname cdnum@rv.pd.compgemll.emotion.difflo\endcsname{\ensuremath{-}6.0}
\expandafter\def\csname cdnum@rv.pd.compgemll.politeness.diff\endcsname{0.4}
\expandafter\def\csname cdnum@rv.pd.compgemll.politeness.diffhi\endcsname{3.8}
\expandafter\def\csname cdnum@rv.pd.compgemll.politeness.difflo\endcsname{\ensuremath{-}3.0}
\expandafter\def\csname cdnum@rv.pd.compgemll.power.diff\endcsname{0.4}
\expandafter\def\csname cdnum@rv.pd.compgemll.power.diffhi\endcsname{3.4}
\expandafter\def\csname cdnum@rv.pd.compgemll.power.difflo\endcsname{\ensuremath{-}2.8}
\expandafter\def\csname cdnum@rv.pd.compgemll.toxicity.diff\endcsname{3.0}
\expandafter\def\csname cdnum@rv.pd.compgemll.toxicity.diffhi\endcsname{5.2}
\expandafter\def\csname cdnum@rv.pd.compgemll.toxicity.difflo\endcsname{0.8}
\expandafter\def\csname cdnum@rv.pd.compmis.doneneutral.diff\endcsname{\ensuremath{-}12.6}
\expandafter\def\csname cdnum@rv.pd.compmis.doneneutral.diffhi\endcsname{\ensuremath{-}9.7}
\expandafter\def\csname cdnum@rv.pd.compmis.doneneutral.difflo\endcsname{\ensuremath{-}15.6}
\expandafter\def\csname cdnum@rv.pd.compmis.dtwo.diff\endcsname{\ensuremath{-}15.7}
\expandafter\def\csname cdnum@rv.pd.compmis.dtwo.diffhi\endcsname{\ensuremath{-}12.3}
\expandafter\def\csname cdnum@rv.pd.compmis.dtwo.difflo\endcsname{\ensuremath{-}19.2}
\expandafter\def\csname cdnum@rv.pd.compmis.emotion.diff\endcsname{\ensuremath{-}26.1}
\expandafter\def\csname cdnum@rv.pd.compmis.emotion.diffhi\endcsname{\ensuremath{-}21.1}
\expandafter\def\csname cdnum@rv.pd.compmis.emotion.difflo\endcsname{\ensuremath{-}30.9}
\expandafter\def\csname cdnum@rv.pd.compmis.politeness.diff\endcsname{\ensuremath{-}0.6}
\expandafter\def\csname cdnum@rv.pd.compmis.politeness.diffhi\endcsname{3.8}
\expandafter\def\csname cdnum@rv.pd.compmis.politeness.difflo\endcsname{\ensuremath{-}5.0}
\expandafter\def\csname cdnum@rv.pd.compmis.power.diff\endcsname{\ensuremath{-}5.0}
\expandafter\def\csname cdnum@rv.pd.compmis.power.diffhi\endcsname{\ensuremath{-}1.2}
\expandafter\def\csname cdnum@rv.pd.compmis.power.difflo\endcsname{\ensuremath{-}8.8}
\expandafter\def\csname cdnum@rv.pd.compmis.toxicity.diff\endcsname{2.8}
\expandafter\def\csname cdnum@rv.pd.compmis.toxicity.diffhi\endcsname{5.6}
\expandafter\def\csname cdnum@rv.pd.compmis.toxicity.difflo\endcsname{0.2}
\expandafter\def\csname cdnum@rv.pd.compmisll.doneneutral.diff\endcsname{\ensuremath{-}7.6}
\expandafter\def\csname cdnum@rv.pd.compmisll.doneneutral.diffhi\endcsname{\ensuremath{-}5.5}
\expandafter\def\csname cdnum@rv.pd.compmisll.doneneutral.difflo\endcsname{\ensuremath{-}9.8}
\expandafter\def\csname cdnum@rv.pd.compmisll.dtwo.absdiff\endcsname{13.2}
\expandafter\def\csname cdnum@rv.pd.compmisll.dtwo.diff\endcsname{\ensuremath{-}13.2}
\expandafter\def\csname cdnum@rv.pd.compmisll.dtwo.diffhi\endcsname{\ensuremath{-}9.8}
\expandafter\def\csname cdnum@rv.pd.compmisll.dtwo.difflo\endcsname{\ensuremath{-}16.5}
\expandafter\def\csname cdnum@rv.pd.compmisll.emotion.diff\endcsname{\ensuremath{-}5.0}
\expandafter\def\csname cdnum@rv.pd.compmisll.emotion.diffhi\endcsname{\ensuremath{-}1.2}
\expandafter\def\csname cdnum@rv.pd.compmisll.emotion.difflo\endcsname{\ensuremath{-}8.8}
\expandafter\def\csname cdnum@rv.pd.compmisll.politeness.diff\endcsname{3.2}
\expandafter\def\csname cdnum@rv.pd.compmisll.politeness.diffhi\endcsname{7.4}
\expandafter\def\csname cdnum@rv.pd.compmisll.politeness.difflo\endcsname{\ensuremath{-}1.0}
\expandafter\def\csname cdnum@rv.pd.compmisll.power.diff\endcsname{\ensuremath{-}2.8}
\expandafter\def\csname cdnum@rv.pd.compmisll.power.diffhi\endcsname{0.4}
\expandafter\def\csname cdnum@rv.pd.compmisll.power.difflo\endcsname{\ensuremath{-}6.0}
\expandafter\def\csname cdnum@rv.pd.compmisll.toxicity.diff\endcsname{1.6}
\expandafter\def\csname cdnum@rv.pd.compmisll.toxicity.diffhi\endcsname{3.6}
\expandafter\def\csname cdnum@rv.pd.compmisll.toxicity.difflo\endcsname{\ensuremath{-}0.4}
\expandafter\def\csname cdnum@rv.pd.compopen.doneneutral.diff\endcsname{\ensuremath{-}5.1}
\expandafter\def\csname cdnum@rv.pd.compopen.doneneutral.diffhi\endcsname{\ensuremath{-}2.6}
\expandafter\def\csname cdnum@rv.pd.compopen.doneneutral.difflo\endcsname{\ensuremath{-}7.7}
\expandafter\def\csname cdnum@rv.pd.compopen.dtwo.diff\endcsname{\ensuremath{-}18.5}
\expandafter\def\csname cdnum@rv.pd.compopen.dtwo.diffhi\endcsname{\ensuremath{-}15.2}
\expandafter\def\csname cdnum@rv.pd.compopen.dtwo.difflo\endcsname{\ensuremath{-}22.0}
\expandafter\def\csname cdnum@rv.pd.compopen.emotion.diff\endcsname{\ensuremath{-}4.0}
\expandafter\def\csname cdnum@rv.pd.compopen.emotion.diffhi\endcsname{\ensuremath{-}0.6}
\expandafter\def\csname cdnum@rv.pd.compopen.emotion.difflo\endcsname{\ensuremath{-}7.4}
\expandafter\def\csname cdnum@rv.pd.compopen.politeness.diff\endcsname{\ensuremath{-}0.4}
\expandafter\def\csname cdnum@rv.pd.compopen.politeness.diffhi\endcsname{4.2}
\expandafter\def\csname cdnum@rv.pd.compopen.politeness.difflo\endcsname{\ensuremath{-}5.0}
\expandafter\def\csname cdnum@rv.pd.compopen.power.diff\endcsname{\ensuremath{-}3.2}
\expandafter\def\csname cdnum@rv.pd.compopen.power.diffhi\endcsname{0.6}
\expandafter\def\csname cdnum@rv.pd.compopen.power.difflo\endcsname{\ensuremath{-}7.0}
\expandafter\def\csname cdnum@rv.pd.compopen.toxicity.diff\endcsname{\ensuremath{-}0.4}
\expandafter\def\csname cdnum@rv.pd.compopen.toxicity.diffhi\endcsname{1.6}
\expandafter\def\csname cdnum@rv.pd.compopen.toxicity.difflo\endcsname{\ensuremath{-}2.4}
\expandafter\def\csname cdnum@rv.pd.comptwo.doneneutral.diff\endcsname{\ensuremath{-}0.7}
\expandafter\def\csname cdnum@rv.pd.comptwo.doneneutral.diffhi\endcsname{1.2}
\expandafter\def\csname cdnum@rv.pd.comptwo.doneneutral.difflo\endcsname{\ensuremath{-}2.6}
\expandafter\def\csname cdnum@rv.pd.comptwo.dtwo.diff\endcsname{\ensuremath{-}23.0}
\expandafter\def\csname cdnum@rv.pd.comptwo.dtwo.diffhi\endcsname{\ensuremath{-}19.5}
\expandafter\def\csname cdnum@rv.pd.comptwo.dtwo.difflo\endcsname{\ensuremath{-}26.5}
\expandafter\def\csname cdnum@rv.pd.comptwo.emotion.diff\endcsname{\ensuremath{-}10.6}
\expandafter\def\csname cdnum@rv.pd.comptwo.emotion.diffhi\endcsname{\ensuremath{-}6.8}
\expandafter\def\csname cdnum@rv.pd.comptwo.emotion.difflo\endcsname{\ensuremath{-}14.5}
\expandafter\def\csname cdnum@rv.pd.comptwo.politeness.diff\endcsname{\ensuremath{-}0.2}
\expandafter\def\csname cdnum@rv.pd.comptwo.politeness.diffhi\endcsname{2.6}
\expandafter\def\csname cdnum@rv.pd.comptwo.politeness.difflo\endcsname{\ensuremath{-}3.2}
\expandafter\def\csname cdnum@rv.pd.comptwo.power.diff\endcsname{0.8}
\expandafter\def\csname cdnum@rv.pd.comptwo.power.diffhi\endcsname{3.8}
\expandafter\def\csname cdnum@rv.pd.comptwo.power.difflo\endcsname{\ensuremath{-}2.2}
\expandafter\def\csname cdnum@rv.pd.comptwo.toxicity.diff\endcsname{\ensuremath{-}2.4}
\expandafter\def\csname cdnum@rv.pd.comptwo.toxicity.diffhi\endcsname{\ensuremath{-}0.2}
\expandafter\def\csname cdnum@rv.pd.comptwo.toxicity.difflo\endcsname{\ensuremath{-}4.8}
\expandafter\def\csname cdnum@rv.pd.comptwoll.doneneutral.diff\endcsname{0.1}
\expandafter\def\csname cdnum@rv.pd.comptwoll.doneneutral.diffhi\endcsname{2.1}
\expandafter\def\csname cdnum@rv.pd.comptwoll.doneneutral.difflo\endcsname{\ensuremath{-}1.8}
\expandafter\def\csname cdnum@rv.pd.comptwoll.dtwo.diff\endcsname{\ensuremath{-}0.2}
\expandafter\def\csname cdnum@rv.pd.comptwoll.dtwo.diffhi\endcsname{1.7}
\expandafter\def\csname cdnum@rv.pd.comptwoll.dtwo.difflo\endcsname{\ensuremath{-}2.0}
\expandafter\def\csname cdnum@rv.pd.comptwoll.emotion.diff\endcsname{\ensuremath{-}4.0}
\expandafter\def\csname cdnum@rv.pd.comptwoll.emotion.diffhi\endcsname{\ensuremath{-}1.0}
\expandafter\def\csname cdnum@rv.pd.comptwoll.emotion.difflo\endcsname{\ensuremath{-}7.2}
\expandafter\def\csname cdnum@rv.pd.comptwoll.politeness.diff\endcsname{0.2}
\expandafter\def\csname cdnum@rv.pd.comptwoll.politeness.diffhi\endcsname{3.0}
\expandafter\def\csname cdnum@rv.pd.comptwoll.politeness.difflo\endcsname{\ensuremath{-}2.6}
\expandafter\def\csname cdnum@rv.pd.comptwoll.power.diff\endcsname{1.0}
\expandafter\def\csname cdnum@rv.pd.comptwoll.power.diffhi\endcsname{4.0}
\expandafter\def\csname cdnum@rv.pd.comptwoll.power.difflo\endcsname{\ensuremath{-}2.0}
\expandafter\def\csname cdnum@rv.pd.comptwoll.toxicity.diff\endcsname{\ensuremath{-}2.4}
\expandafter\def\csname cdnum@rv.pd.comptwoll.toxicity.diffhi\endcsname{\ensuremath{-}0.2}
\expandafter\def\csname cdnum@rv.pd.comptwoll.toxicity.difflo\endcsname{\ensuremath{-}4.8}
\expandafter\def\csname cdnum@rv.pd.comptwothink.doneneutral.diff\endcsname{\ensuremath{-}1.1}
\expandafter\def\csname cdnum@rv.pd.comptwothink.doneneutral.diffhi\endcsname{1.1}
\expandafter\def\csname cdnum@rv.pd.comptwothink.doneneutral.difflo\endcsname{\ensuremath{-}3.3}
\expandafter\def\csname cdnum@rv.pd.comptwothink.dtwo.diff\endcsname{1.5}
\expandafter\def\csname cdnum@rv.pd.comptwothink.dtwo.diffhi\endcsname{3.5}
\expandafter\def\csname cdnum@rv.pd.comptwothink.dtwo.difflo\endcsname{\ensuremath{-}0.3}
\expandafter\def\csname cdnum@rv.pd.comptwothink.emotion.diff\endcsname{1.4}
\expandafter\def\csname cdnum@rv.pd.comptwothink.emotion.diffhi\endcsname{4.0}
\expandafter\def\csname cdnum@rv.pd.comptwothink.emotion.difflo\endcsname{\ensuremath{-}1.2}
\expandafter\def\csname cdnum@rv.pd.comptwothink.politeness.diff\endcsname{\ensuremath{-}1.2}
\expandafter\def\csname cdnum@rv.pd.comptwothink.politeness.diffhi\endcsname{2.2}
\expandafter\def\csname cdnum@rv.pd.comptwothink.politeness.difflo\endcsname{\ensuremath{-}4.6}
\expandafter\def\csname cdnum@rv.pd.comptwothink.power.diff\endcsname{2.6}
\expandafter\def\csname cdnum@rv.pd.comptwothink.power.diffhi\endcsname{5.6}
\expandafter\def\csname cdnum@rv.pd.comptwothink.power.difflo\endcsname{\ensuremath{-}0.4}
\expandafter\def\csname cdnum@rv.pd.comptwothink.toxicity.diff\endcsname{\ensuremath{-}1.6}
\expandafter\def\csname cdnum@rv.pd.comptwothink.toxicity.diffhi\endcsname{0.4}
\expandafter\def\csname cdnum@rv.pd.comptwothink.toxicity.difflo\endcsname{\ensuremath{-}3.6}
\expandafter\def\csname cdnum@rv.pd.decider.doneneutral.diff\endcsname{\ensuremath{-}14.4}
\expandafter\def\csname cdnum@rv.pd.decider.doneneutral.diffhi\endcsname{\ensuremath{-}11.8}
\expandafter\def\csname cdnum@rv.pd.decider.doneneutral.difflo\endcsname{\ensuremath{-}17.1}
\expandafter\def\csname cdnum@rv.pd.decider.dtwo.diff\endcsname{\ensuremath{-}5.3}
\expandafter\def\csname cdnum@rv.pd.decider.dtwo.diffhi\endcsname{\ensuremath{-}2.5}
\expandafter\def\csname cdnum@rv.pd.decider.dtwo.difflo\endcsname{\ensuremath{-}8.2}
\expandafter\def\csname cdnum@rv.pd.decider.emotion.diff\endcsname{29.9}
\expandafter\def\csname cdnum@rv.pd.decider.emotion.diffhi\endcsname{34.5}
\expandafter\def\csname cdnum@rv.pd.decider.emotion.difflo\endcsname{25.3}
\expandafter\def\csname cdnum@rv.pd.decider.politeness.diff\endcsname{\ensuremath{-}19.3}
\expandafter\def\csname cdnum@rv.pd.decider.politeness.diffhi\endcsname{\ensuremath{-}13.1}
\expandafter\def\csname cdnum@rv.pd.decider.politeness.difflo\endcsname{\ensuremath{-}25.7}
\expandafter\def\csname cdnum@rv.pd.decider.power.diff\endcsname{\ensuremath{-}7.4}
\expandafter\def\csname cdnum@rv.pd.decider.power.diffhi\endcsname{\ensuremath{-}3.4}
\expandafter\def\csname cdnum@rv.pd.decider.power.difflo\endcsname{\ensuremath{-}11.4}
\expandafter\def\csname cdnum@rv.pd.decider.toxicity.diff\endcsname{\ensuremath{-}4.4}
\expandafter\def\csname cdnum@rv.pd.decider.toxicity.diffhi\endcsname{\ensuremath{-}1.8}
\expandafter\def\csname cdnum@rv.pd.decider.toxicity.difflo\endcsname{\ensuremath{-}7.2}
\expandafter\def\csname cdnum@rv.pd.kevlarge.doneneutral.diff\endcsname{\ensuremath{-}1.9}
\expandafter\def\csname cdnum@rv.pd.kevlarge.doneneutral.diffhi\endcsname{0.1}
\expandafter\def\csname cdnum@rv.pd.kevlarge.doneneutral.difflo\endcsname{\ensuremath{-}4.0}
\expandafter\def\csname cdnum@rv.pd.kevlarge.doneneutral.holm\endcsname{0.37}
\expandafter\def\csname cdnum@rv.pd.kevlarge.dtwo.diff\endcsname{\ensuremath{-}6.3}
\expandafter\def\csname cdnum@rv.pd.kevlarge.dtwo.diffhi\endcsname{\ensuremath{-}3.8}
\expandafter\def\csname cdnum@rv.pd.kevlarge.dtwo.difflo\endcsname{\ensuremath{-}8.8}
\expandafter\def\csname cdnum@rv.pd.kevlarge.emotion.diff\endcsname{2.4}
\expandafter\def\csname cdnum@rv.pd.kevlarge.emotion.diffhi\endcsname{6.0}
\expandafter\def\csname cdnum@rv.pd.kevlarge.emotion.difflo\endcsname{\ensuremath{-}1.2}
\expandafter\def\csname cdnum@rv.pd.kevlarge.politeness.diff\endcsname{\ensuremath{-}18.1}
\expandafter\def\csname cdnum@rv.pd.kevlarge.politeness.diffhi\endcsname{\ensuremath{-}12.2}
\expandafter\def\csname cdnum@rv.pd.kevlarge.politeness.difflo\endcsname{\ensuremath{-}23.9}
\expandafter\def\csname cdnum@rv.pd.kevlarge.power.diff\endcsname{\ensuremath{-}1.4}
\expandafter\def\csname cdnum@rv.pd.kevlarge.power.diffhi\endcsname{2.0}
\expandafter\def\csname cdnum@rv.pd.kevlarge.power.difflo\endcsname{\ensuremath{-}4.6}
\expandafter\def\csname cdnum@rv.pd.kevlarge.toxicity.diff\endcsname{\ensuremath{-}2.8}
\expandafter\def\csname cdnum@rv.pd.kevlarge.toxicity.diffhi\endcsname{\ensuremath{-}0.4}
\expandafter\def\csname cdnum@rv.pd.kevlarge.toxicity.difflo\endcsname{\ensuremath{-}5.2}
\expandafter\def\csname cdnum@rv.pd.kevsmall.doneneutral.diff\endcsname{\ensuremath{-}28.1}
\expandafter\def\csname cdnum@rv.pd.kevsmall.doneneutral.diffhi\endcsname{\ensuremath{-}25.1}
\expandafter\def\csname cdnum@rv.pd.kevsmall.doneneutral.difflo\endcsname{\ensuremath{-}31.1}
\expandafter\def\csname cdnum@rv.pd.kevsmall.dtwo.diff\endcsname{\ensuremath{-}22.8}
\expandafter\def\csname cdnum@rv.pd.kevsmall.dtwo.diffhi\endcsname{\ensuremath{-}19.2}
\expandafter\def\csname cdnum@rv.pd.kevsmall.dtwo.difflo\endcsname{\ensuremath{-}26.5}
\expandafter\def\csname cdnum@rv.pd.kevsmall.emotion.diff\endcsname{1.2}
\expandafter\def\csname cdnum@rv.pd.kevsmall.emotion.diffhi\endcsname{5.4}
\expandafter\def\csname cdnum@rv.pd.kevsmall.emotion.difflo\endcsname{\ensuremath{-}2.8}
\expandafter\def\csname cdnum@rv.pd.kevsmall.politeness.diff\endcsname{\ensuremath{-}18.1}
\expandafter\def\csname cdnum@rv.pd.kevsmall.politeness.diffhi\endcsname{\ensuremath{-}11.6}
\expandafter\def\csname cdnum@rv.pd.kevsmall.politeness.difflo\endcsname{\ensuremath{-}24.5}
\expandafter\def\csname cdnum@rv.pd.kevsmall.power.diff\endcsname{\ensuremath{-}8.8}
\expandafter\def\csname cdnum@rv.pd.kevsmall.power.diffhi\endcsname{\ensuremath{-}4.4}
\expandafter\def\csname cdnum@rv.pd.kevsmall.power.difflo\endcsname{\ensuremath{-}13.2}
\expandafter\def\csname cdnum@rv.pd.kevsmall.toxicity.diff\endcsname{\ensuremath{-}2.8}
\expandafter\def\csname cdnum@rv.pd.kevsmall.toxicity.diffhi\endcsname{0.2}
\expandafter\def\csname cdnum@rv.pd.kevsmall.toxicity.difflo\endcsname{\ensuremath{-}5.8}
\expandafter\def\csname cdnum@rv.pd.layaen.doneneutral.diff\endcsname{\ensuremath{-}36.5}
\expandafter\def\csname cdnum@rv.pd.layaen.doneneutral.diffhi\endcsname{\ensuremath{-}33.2}
\expandafter\def\csname cdnum@rv.pd.layaen.doneneutral.difflo\endcsname{\ensuremath{-}39.9}
\expandafter\def\csname cdnum@rv.pd.layaen.emotion.diff\endcsname{0.0}
\expandafter\def\csname cdnum@rv.pd.layaen.emotion.diffhi\endcsname{4.6}
\expandafter\def\csname cdnum@rv.pd.layaen.emotion.difflo\endcsname{\ensuremath{-}4.6}
\expandafter\def\csname cdnum@rv.pd.layaen.politeness.diff\endcsname{\ensuremath{-}28.3}
\expandafter\def\csname cdnum@rv.pd.layaen.politeness.diffhi\endcsname{\ensuremath{-}21.9}
\expandafter\def\csname cdnum@rv.pd.layaen.politeness.difflo\endcsname{\ensuremath{-}34.7}
\expandafter\def\csname cdnum@rv.pd.layaen.power.diff\endcsname{\ensuremath{-}10.0}
\expandafter\def\csname cdnum@rv.pd.layaen.power.diffhi\endcsname{\ensuremath{-}3.8}
\expandafter\def\csname cdnum@rv.pd.layaen.power.difflo\endcsname{\ensuremath{-}16.2}
\expandafter\def\csname cdnum@rv.pd.layaen.toxicity.diff\endcsname{1.0}
\expandafter\def\csname cdnum@rv.pd.layaen.toxicity.diffhi\endcsname{5.4}
\expandafter\def\csname cdnum@rv.pd.layaen.toxicity.difflo\endcsname{\ensuremath{-}3.4}
\expandafter\def\csname cdnum@rv.pd.layaml.doneneutral.diff\endcsname{\ensuremath{-}38.2}
\expandafter\def\csname cdnum@rv.pd.layaml.doneneutral.diffhi\endcsname{\ensuremath{-}35.3}
\expandafter\def\csname cdnum@rv.pd.layaml.doneneutral.difflo\endcsname{\ensuremath{-}41.2}
\expandafter\def\csname cdnum@rv.pd.layaml.dtwo.diff\endcsname{\ensuremath{-}90.8}
\expandafter\def\csname cdnum@rv.pd.layaml.dtwo.diffhi\endcsname{\ensuremath{-}88.5}
\expandafter\def\csname cdnum@rv.pd.layaml.dtwo.difflo\endcsname{\ensuremath{-}93.0}
\expandafter\def\csname cdnum@rv.pd.layaml.emotion.diff\endcsname{\ensuremath{-}10.8}
\expandafter\def\csname cdnum@rv.pd.layaml.emotion.diffhi\endcsname{\ensuremath{-}5.8}
\expandafter\def\csname cdnum@rv.pd.layaml.emotion.difflo\endcsname{\ensuremath{-}15.7}
\expandafter\def\csname cdnum@rv.pd.layaml.politeness.diff\endcsname{\ensuremath{-}29.1}
\expandafter\def\csname cdnum@rv.pd.layaml.politeness.diffhi\endcsname{\ensuremath{-}22.5}
\expandafter\def\csname cdnum@rv.pd.layaml.politeness.difflo\endcsname{\ensuremath{-}35.7}
\expandafter\def\csname cdnum@rv.pd.layaml.power.diff\endcsname{\ensuremath{-}12.6}
\expandafter\def\csname cdnum@rv.pd.layaml.power.diffhi\endcsname{\ensuremath{-}7.6}
\expandafter\def\csname cdnum@rv.pd.layaml.power.difflo\endcsname{\ensuremath{-}17.6}
\expandafter\def\csname cdnum@rv.pd.layaml.toxicity.diff\endcsname{\ensuremath{-}0.6}
\expandafter\def\csname cdnum@rv.pd.layaml.toxicity.diffhi\endcsname{4.2}
\expandafter\def\csname cdnum@rv.pd.layaml.toxicity.difflo\endcsname{\ensuremath{-}5.4}
\expandafter\def\csname cdnum@rv.pd.nimble.doneneutral.diff\endcsname{\ensuremath{-}4.6}
\expandafter\def\csname cdnum@rv.pd.nimble.doneneutral.diffhi\endcsname{\ensuremath{-}2.5}
\expandafter\def\csname cdnum@rv.pd.nimble.doneneutral.difflo\endcsname{\ensuremath{-}6.8}
\expandafter\def\csname cdnum@rv.pd.nimble.dtwo.diff\endcsname{\ensuremath{-}5.3}
\expandafter\def\csname cdnum@rv.pd.nimble.dtwo.diffhi\endcsname{\ensuremath{-}3.0}
\expandafter\def\csname cdnum@rv.pd.nimble.dtwo.difflo\endcsname{\ensuremath{-}7.7}
\expandafter\def\csname cdnum@rv.pd.nimble.emotion.diff\endcsname{1.0}
\expandafter\def\csname cdnum@rv.pd.nimble.emotion.diffhi\endcsname{4.2}
\expandafter\def\csname cdnum@rv.pd.nimble.emotion.difflo\endcsname{\ensuremath{-}2.0}
\expandafter\def\csname cdnum@rv.pd.nimble.politeness.diff\endcsname{\ensuremath{-}14.1}
\expandafter\def\csname cdnum@rv.pd.nimble.politeness.diffhi\endcsname{\ensuremath{-}9.8}
\expandafter\def\csname cdnum@rv.pd.nimble.politeness.difflo\endcsname{\ensuremath{-}18.3}
\expandafter\def\csname cdnum@rv.pd.nimble.power.diff\endcsname{\ensuremath{-}5.4}
\expandafter\def\csname cdnum@rv.pd.nimble.power.diffhi\endcsname{\ensuremath{-}1.8}
\expandafter\def\csname cdnum@rv.pd.nimble.power.difflo\endcsname{\ensuremath{-}9.0}
\expandafter\def\csname cdnum@rv.pd.nimble.toxicity.diff\endcsname{\ensuremath{-}0.2}
\expandafter\def\csname cdnum@rv.pd.nimble.toxicity.diffhi\endcsname{1.6}
\expandafter\def\csname cdnum@rv.pd.nimble.toxicity.difflo\endcsname{\ensuremath{-}2.0}
\expandafter\def\csname cdnum@rv.pd.thisthat.doneneutral.diff\endcsname{\ensuremath{-}13.5}
\expandafter\def\csname cdnum@rv.pd.thisthat.doneneutral.diffhi\endcsname{\ensuremath{-}10.9}
\expandafter\def\csname cdnum@rv.pd.thisthat.doneneutral.difflo\endcsname{\ensuremath{-}16.0}
\expandafter\def\csname cdnum@rv.pd.thisthat.dtwo.diff\endcsname{\ensuremath{-}5.0}
\expandafter\def\csname cdnum@rv.pd.thisthat.dtwo.diffhi\endcsname{\ensuremath{-}2.3}
\expandafter\def\csname cdnum@rv.pd.thisthat.dtwo.difflo\endcsname{\ensuremath{-}7.8}
\expandafter\def\csname cdnum@rv.pd.thisthat.emotion.diff\endcsname{30.1}
\expandafter\def\csname cdnum@rv.pd.thisthat.emotion.diffhi\endcsname{34.9}
\expandafter\def\csname cdnum@rv.pd.thisthat.emotion.difflo\endcsname{25.5}
\expandafter\def\csname cdnum@rv.pd.thisthat.politeness.diff\endcsname{\ensuremath{-}24.3}
\expandafter\def\csname cdnum@rv.pd.thisthat.politeness.diffhi\endcsname{\ensuremath{-}17.7}
\expandafter\def\csname cdnum@rv.pd.thisthat.politeness.difflo\endcsname{\ensuremath{-}30.9}
\expandafter\def\csname cdnum@rv.pd.thisthat.power.diff\endcsname{\ensuremath{-}7.8}
\expandafter\def\csname cdnum@rv.pd.thisthat.power.diffhi\endcsname{\ensuremath{-}3.4}
\expandafter\def\csname cdnum@rv.pd.thisthat.power.difflo\endcsname{\ensuremath{-}12.0}
\expandafter\def\csname cdnum@rv.pd.thisthat.toxicity.diff\endcsname{\ensuremath{-}4.2}
\expandafter\def\csname cdnum@rv.pd.thisthat.toxicity.diffhi\endcsname{\ensuremath{-}1.6}
\expandafter\def\csname cdnum@rv.pd.thisthat.toxicity.difflo\endcsname{\ensuremath{-}6.8}
\expandafter\def\csname cdnum@rv.sel.compgem.doneneutral.covfive\endcsname{0.12}
\expandafter\def\csname cdnum@rv.sel.compgem.doneneutral.covfivehi\endcsname{0.15}
\expandafter\def\csname cdnum@rv.sel.compgem.doneneutral.covfivelo\endcsname{0.10}
\expandafter\def\csname cdnum@rv.sel.compgem.dtwo.covfive\endcsname{0.97}
\expandafter\def\csname cdnum@rv.sel.compgem.dtwo.covfivehi\endcsname{1.00}
\expandafter\def\csname cdnum@rv.sel.compgem.dtwo.covfivelo\endcsname{0.65}
\expandafter\def\csname cdnum@rv.sel.compgemll.doneneutral.covfive\endcsname{0.28}
\expandafter\def\csname cdnum@rv.sel.compgemll.doneneutral.covfivehi\endcsname{0.37}
\expandafter\def\csname cdnum@rv.sel.compgemll.doneneutral.covfivelo\endcsname{0.21}
\expandafter\def\csname cdnum@rv.sel.compgemll.dtwo.covfive\endcsname{0.98}
\expandafter\def\csname cdnum@rv.sel.compgemll.dtwo.covfivehi\endcsname{1.00}
\expandafter\def\csname cdnum@rv.sel.compgemll.dtwo.covfivelo\endcsname{0.93}
\expandafter\def\csname cdnum@rv.sel.compmis.doneneutral.covfive\endcsname{0.00}
\expandafter\def\csname cdnum@rv.sel.compmis.doneneutral.covfivehi\endcsname{0.00}
\expandafter\def\csname cdnum@rv.sel.compmis.doneneutral.covfivelo\endcsname{0.00}
\expandafter\def\csname cdnum@rv.sel.compmis.dtwo.covfive\endcsname{0.00}
\expandafter\def\csname cdnum@rv.sel.compmis.dtwo.covfivehi\endcsname{0.00}
\expandafter\def\csname cdnum@rv.sel.compmis.dtwo.covfivelo\endcsname{0.00}
\expandafter\def\csname cdnum@rv.sel.compmisll.doneneutral.covfive\endcsname{0.08}
\expandafter\def\csname cdnum@rv.sel.compmisll.doneneutral.covfivehi\endcsname{0.14}
\expandafter\def\csname cdnum@rv.sel.compmisll.doneneutral.covfivelo\endcsname{0.03}
\expandafter\def\csname cdnum@rv.sel.compmisll.dtwo.covfive\endcsname{0.41}
\expandafter\def\csname cdnum@rv.sel.compmisll.dtwo.covfivehi\endcsname{0.62}
\expandafter\def\csname cdnum@rv.sel.compmisll.dtwo.covfivelo\endcsname{0.32}
\expandafter\def\csname cdnum@rv.sel.compopen.doneneutral.covfive\endcsname{0.00}
\expandafter\def\csname cdnum@rv.sel.compopen.doneneutral.covfivehi\endcsname{0.00}
\expandafter\def\csname cdnum@rv.sel.compopen.doneneutral.covfivelo\endcsname{0.00}
\expandafter\def\csname cdnum@rv.sel.compopen.dtwo.covfive\endcsname{0.01}
\expandafter\def\csname cdnum@rv.sel.compopen.dtwo.covfivehi\endcsname{0.03}
\expandafter\def\csname cdnum@rv.sel.compopen.dtwo.covfivelo\endcsname{0.01}
\expandafter\def\csname cdnum@rv.sel.comptwo.doneneutral.covfive\endcsname{0.04}
\expandafter\def\csname cdnum@rv.sel.comptwo.doneneutral.covfivehi\endcsname{0.05}
\expandafter\def\csname cdnum@rv.sel.comptwo.doneneutral.covfivelo\endcsname{0.03}
\expandafter\def\csname cdnum@rv.sel.comptwo.dtwo.covfive\endcsname{0.69}
\expandafter\def\csname cdnum@rv.sel.comptwo.dtwo.covfivehi\endcsname{0.75}
\expandafter\def\csname cdnum@rv.sel.comptwo.dtwo.covfivelo\endcsname{0.01}
\expandafter\def\csname cdnum@rv.sel.comptwoll.doneneutral.covfive\endcsname{0.42}
\expandafter\def\csname cdnum@rv.sel.comptwoll.doneneutral.covfivehi\endcsname{0.47}
\expandafter\def\csname cdnum@rv.sel.comptwoll.doneneutral.covfivelo\endcsname{0.36}
\expandafter\def\csname cdnum@rv.sel.comptwoll.dtwo.covfive\endcsname{0.86}
\expandafter\def\csname cdnum@rv.sel.comptwoll.dtwo.covfivehi\endcsname{0.93}
\expandafter\def\csname cdnum@rv.sel.comptwoll.dtwo.covfivelo\endcsname{0.80}
\expandafter\def\csname cdnum@rv.sel.comptwothink.doneneutral.covfive\endcsname{0.10}
\expandafter\def\csname cdnum@rv.sel.comptwothink.doneneutral.covfivehi\endcsname{0.28}
\expandafter\def\csname cdnum@rv.sel.comptwothink.doneneutral.covfivelo\endcsname{0.00}
\expandafter\def\csname cdnum@rv.sel.comptwothink.dtwo.covfive\endcsname{0.85}
\expandafter\def\csname cdnum@rv.sel.comptwothink.dtwo.covfivehi\endcsname{1.00}
\expandafter\def\csname cdnum@rv.sel.comptwothink.dtwo.covfivelo\endcsname{0.45}
\expandafter\def\csname cdnum@rv.sel.decider.doneneutral.covfive\endcsname{0.06}
\expandafter\def\csname cdnum@rv.sel.decider.doneneutral.covfivehi\endcsname{0.08}
\expandafter\def\csname cdnum@rv.sel.decider.doneneutral.covfivelo\endcsname{0.04}
\expandafter\def\csname cdnum@rv.sel.decider.dtwo.covfive\endcsname{0.75}
\expandafter\def\csname cdnum@rv.sel.decider.dtwo.covfivehi\endcsname{0.85}
\expandafter\def\csname cdnum@rv.sel.decider.dtwo.covfivelo\endcsname{0.64}
\expandafter\def\csname cdnum@rv.sel.jev.doneneutral.covfive\endcsname{0.13}
\expandafter\def\csname cdnum@rv.sel.jev.doneneutral.covfivehi\endcsname{0.24}
\expandafter\def\csname cdnum@rv.sel.jev.doneneutral.covfivelo\endcsname{0.10}
\expandafter\def\csname cdnum@rv.sel.jev.dtwo.covfive\endcsname{0.93}
\expandafter\def\csname cdnum@rv.sel.jev.dtwo.covfivehi\endcsname{0.97}
\expandafter\def\csname cdnum@rv.sel.jev.dtwo.covfivelo\endcsname{0.80}
\expandafter\def\csname cdnum@rv.sel.jev.oos.aurochi\endcsname{0.940}
\expandafter\def\csname cdnum@rv.sel.jev.oos.auroclo\endcsname{0.901}
\expandafter\def\csname cdnum@rv.sel.kevlarge.doneneutral.covfive\endcsname{0.12}
\expandafter\def\csname cdnum@rv.sel.kevlarge.doneneutral.covfivehi\endcsname{0.21}
\expandafter\def\csname cdnum@rv.sel.kevlarge.doneneutral.covfivelo\endcsname{0.09}
\expandafter\def\csname cdnum@rv.sel.kevlarge.dtwo.covfive\endcsname{0.61}
\expandafter\def\csname cdnum@rv.sel.kevlarge.dtwo.covfivehi\endcsname{0.72}
\expandafter\def\csname cdnum@rv.sel.kevlarge.dtwo.covfivelo\endcsname{0.48}
\expandafter\def\csname cdnum@rv.sel.kevsmall.doneneutral.covfive\endcsname{0.02}
\expandafter\def\csname cdnum@rv.sel.kevsmall.doneneutral.covfivehi\endcsname{0.03}
\expandafter\def\csname cdnum@rv.sel.kevsmall.doneneutral.covfivelo\endcsname{0.02}
\expandafter\def\csname cdnum@rv.sel.kevsmall.dtwo.covfive\endcsname{0.22}
\expandafter\def\csname cdnum@rv.sel.kevsmall.dtwo.covfivehi\endcsname{0.42}
\expandafter\def\csname cdnum@rv.sel.kevsmall.dtwo.covfivelo\endcsname{0.04}
\expandafter\def\csname cdnum@rv.sel.layaen.doneneutral.covfive\endcsname{0.00}
\expandafter\def\csname cdnum@rv.sel.layaen.doneneutral.covfivehi\endcsname{0.00}
\expandafter\def\csname cdnum@rv.sel.layaen.doneneutral.covfivelo\endcsname{0.00}
\expandafter\def\csname cdnum@rv.sel.layaml.doneneutral.covfive\endcsname{0.01}
\expandafter\def\csname cdnum@rv.sel.layaml.doneneutral.covfivehi\endcsname{0.02}
\expandafter\def\csname cdnum@rv.sel.layaml.doneneutral.covfivelo\endcsname{0.01}
\expandafter\def\csname cdnum@rv.sel.layaml.dtwo.covfive\endcsname{0.00}
\expandafter\def\csname cdnum@rv.sel.layaml.dtwo.covfivehi\endcsname{0.00}
\expandafter\def\csname cdnum@rv.sel.layaml.dtwo.covfivelo\endcsname{0.00}
\expandafter\def\csname cdnum@rv.sel.nimble.doneneutral.covfive\endcsname{0.20}
\expandafter\def\csname cdnum@rv.sel.nimble.doneneutral.covfivehi\endcsname{0.26}
\expandafter\def\csname cdnum@rv.sel.nimble.doneneutral.covfivelo\endcsname{0.09}
\expandafter\def\csname cdnum@rv.sel.nimble.dtwo.covfive\endcsname{0.66}
\expandafter\def\csname cdnum@rv.sel.nimble.dtwo.covfivehi\endcsname{0.84}
\expandafter\def\csname cdnum@rv.sel.nimble.dtwo.covfivelo\endcsname{0.49}
\expandafter\def\csname cdnum@rv.sel.thisthat.doneneutral.covfive\endcsname{0.04}
\expandafter\def\csname cdnum@rv.sel.thisthat.doneneutral.covfivehi\endcsname{0.07}
\expandafter\def\csname cdnum@rv.sel.thisthat.doneneutral.covfivelo\endcsname{0.03}
\expandafter\def\csname cdnum@rv.sel.thisthat.dtwo.covfive\endcsname{0.78}
\expandafter\def\csname cdnum@rv.sel.thisthat.dtwo.covfivehi\endcsname{0.86}
\expandafter\def\csname cdnum@rv.sel.thisthat.dtwo.covfivelo\endcsname{0.73}
\expandafter\def\csname cdnum@rv.ts.compgem.doneneutral.ece\endcsname{0.106}
\expandafter\def\csname cdnum@rv.ts.compgem.doneneutral.ecescaled\endcsname{0.043}
\expandafter\def\csname cdnum@rv.ts.compgem.doneneutral.ecescaledhi\endcsname{0.074}
\expandafter\def\csname cdnum@rv.ts.compgem.doneneutral.ecescaledlo\endcsname{0.035}
\expandafter\def\csname cdnum@rv.ts.compgem.dtwo.ece\endcsname{0.022}
\expandafter\def\csname cdnum@rv.ts.compgem.dtwo.ecescaled\endcsname{0.026}
\expandafter\def\csname cdnum@rv.ts.compgem.dtwo.ecescaledhi\endcsname{0.054}
\expandafter\def\csname cdnum@rv.ts.compgem.dtwo.ecescaledlo\endcsname{0.017}
\expandafter\def\csname cdnum@rv.ts.compgem.emotion.ece\endcsname{0.274}
\expandafter\def\csname cdnum@rv.ts.compgem.emotion.ecescaled\endcsname{0.075}
\expandafter\def\csname cdnum@rv.ts.compgem.emotion.ecescaledhi\endcsname{0.119}
\expandafter\def\csname cdnum@rv.ts.compgem.emotion.ecescaledlo\endcsname{0.051}
\expandafter\def\csname cdnum@rv.ts.compgem.politeness.ece\endcsname{0.138}
\expandafter\def\csname cdnum@rv.ts.compgem.politeness.ecescaled\endcsname{0.067}
\expandafter\def\csname cdnum@rv.ts.compgem.politeness.ecescaledhi\endcsname{0.122}
\expandafter\def\csname cdnum@rv.ts.compgem.politeness.ecescaledlo\endcsname{0.055}
\expandafter\def\csname cdnum@rv.ts.compgem.power.ece\endcsname{0.312}
\expandafter\def\csname cdnum@rv.ts.compgem.power.ecescaled\endcsname{0.069}
\expandafter\def\csname cdnum@rv.ts.compgem.power.ecescaledhi\endcsname{0.117}
\expandafter\def\csname cdnum@rv.ts.compgem.power.ecescaledlo\endcsname{0.041}
\expandafter\def\csname cdnum@rv.ts.compgem.toxicity.ece\endcsname{0.311}
\expandafter\def\csname cdnum@rv.ts.compgem.toxicity.ecescaled\endcsname{0.057}
\expandafter\def\csname cdnum@rv.ts.compgem.toxicity.ecescaledhi\endcsname{0.114}
\expandafter\def\csname cdnum@rv.ts.compgem.toxicity.ecescaledlo\endcsname{0.045}
\expandafter\def\csname cdnum@rv.ts.compgemll.doneneutral.ece\endcsname{0.244}
\expandafter\def\csname cdnum@rv.ts.compgemll.doneneutral.ecescaled\endcsname{0.064}
\expandafter\def\csname cdnum@rv.ts.compgemll.doneneutral.ecescaledhi\endcsname{0.089}
\expandafter\def\csname cdnum@rv.ts.compgemll.doneneutral.ecescaledlo\endcsname{0.047}
\expandafter\def\csname cdnum@rv.ts.compgemll.dtwo.ece\endcsname{0.058}
\expandafter\def\csname cdnum@rv.ts.compgemll.dtwo.ecescaled\endcsname{0.034}
\expandafter\def\csname cdnum@rv.ts.compgemll.dtwo.ecescaledhi\endcsname{0.067}
\expandafter\def\csname cdnum@rv.ts.compgemll.dtwo.ecescaledlo\endcsname{0.031}
\expandafter\def\csname cdnum@rv.ts.compgemll.emotion.ece\endcsname{0.517}
\expandafter\def\csname cdnum@rv.ts.compgemll.emotion.ecescaled\endcsname{0.106}
\expandafter\def\csname cdnum@rv.ts.compgemll.emotion.ecescaledhi\endcsname{0.163}
\expandafter\def\csname cdnum@rv.ts.compgemll.emotion.ecescaledlo\endcsname{0.072}
\expandafter\def\csname cdnum@rv.ts.compgemll.politeness.ece\endcsname{0.329}
\expandafter\def\csname cdnum@rv.ts.compgemll.politeness.ecescaled\endcsname{0.101}
\expandafter\def\csname cdnum@rv.ts.compgemll.politeness.ecescaledhi\endcsname{0.139}
\expandafter\def\csname cdnum@rv.ts.compgemll.politeness.ecescaledlo\endcsname{0.064}
\expandafter\def\csname cdnum@rv.ts.compgemll.power.ece\endcsname{0.384}
\expandafter\def\csname cdnum@rv.ts.compgemll.power.ecescaled\endcsname{0.047}
\expandafter\def\csname cdnum@rv.ts.compgemll.power.ecescaledhi\endcsname{0.108}
\expandafter\def\csname cdnum@rv.ts.compgemll.power.ecescaledlo\endcsname{0.045}
\expandafter\def\csname cdnum@rv.ts.compgemll.toxicity.ece\endcsname{0.408}
\expandafter\def\csname cdnum@rv.ts.compgemll.toxicity.ecescaled\endcsname{0.066}
\expandafter\def\csname cdnum@rv.ts.compgemll.toxicity.ecescaledhi\endcsname{0.128}
\expandafter\def\csname cdnum@rv.ts.compgemll.toxicity.ecescaledlo\endcsname{0.059}
\expandafter\def\csname cdnum@rv.ts.compmis.doneneutral.ece\endcsname{0.151}
\expandafter\def\csname cdnum@rv.ts.compmis.doneneutral.ecescaled\endcsname{0.065}
\expandafter\def\csname cdnum@rv.ts.compmis.doneneutral.ecescaledhi\endcsname{0.085}
\expandafter\def\csname cdnum@rv.ts.compmis.doneneutral.ecescaledlo\endcsname{0.038}
\expandafter\def\csname cdnum@rv.ts.compmis.dtwo.ece\endcsname{0.135}
\expandafter\def\csname cdnum@rv.ts.compmis.dtwo.ecescaled\endcsname{0.186}
\expandafter\def\csname cdnum@rv.ts.compmis.dtwo.ecescaledhi\endcsname{0.230}
\expandafter\def\csname cdnum@rv.ts.compmis.dtwo.ecescaledlo\endcsname{0.158}
\expandafter\def\csname cdnum@rv.ts.compmis.emotion.ece\endcsname{0.071}
\expandafter\def\csname cdnum@rv.ts.compmis.emotion.ecescaled\endcsname{0.045}
\expandafter\def\csname cdnum@rv.ts.compmis.emotion.ecescaledhi\endcsname{0.088}
\expandafter\def\csname cdnum@rv.ts.compmis.emotion.ecescaledlo\endcsname{0.031}
\expandafter\def\csname cdnum@rv.ts.compmis.politeness.ece\endcsname{0.200}
\expandafter\def\csname cdnum@rv.ts.compmis.politeness.ecescaled\endcsname{0.109}
\expandafter\def\csname cdnum@rv.ts.compmis.politeness.ecescaledhi\endcsname{0.145}
\expandafter\def\csname cdnum@rv.ts.compmis.politeness.ecescaledlo\endcsname{0.068}
\expandafter\def\csname cdnum@rv.ts.compmis.power.ece\endcsname{0.316}
\expandafter\def\csname cdnum@rv.ts.compmis.power.ecescaled\endcsname{0.061}
\expandafter\def\csname cdnum@rv.ts.compmis.power.ecescaledhi\endcsname{0.125}
\expandafter\def\csname cdnum@rv.ts.compmis.power.ecescaledlo\endcsname{0.049}
\expandafter\def\csname cdnum@rv.ts.compmis.toxicity.ece\endcsname{0.298}
\expandafter\def\csname cdnum@rv.ts.compmis.toxicity.ecescaled\endcsname{0.057}
\expandafter\def\csname cdnum@rv.ts.compmis.toxicity.ecescaledhi\endcsname{0.127}
\expandafter\def\csname cdnum@rv.ts.compmis.toxicity.ecescaledlo\endcsname{0.049}
\expandafter\def\csname cdnum@rv.ts.compmisll.doneneutral.ece\endcsname{0.112}
\expandafter\def\csname cdnum@rv.ts.compmisll.doneneutral.ecescaled\endcsname{0.045}
\expandafter\def\csname cdnum@rv.ts.compmisll.doneneutral.ecescaledhi\endcsname{0.066}
\expandafter\def\csname cdnum@rv.ts.compmisll.doneneutral.ecescaledlo\endcsname{0.027}
\expandafter\def\csname cdnum@rv.ts.compmisll.dtwo.ece\endcsname{0.055}
\expandafter\def\csname cdnum@rv.ts.compmisll.dtwo.ecescaled\endcsname{0.053}
\expandafter\def\csname cdnum@rv.ts.compmisll.dtwo.ecescaledhi\endcsname{0.081}
\expandafter\def\csname cdnum@rv.ts.compmisll.dtwo.ecescaledlo\endcsname{0.033}
\expandafter\def\csname cdnum@rv.ts.compmisll.emotion.ece\endcsname{0.370}
\expandafter\def\csname cdnum@rv.ts.compmisll.emotion.ecescaled\endcsname{0.072}
\expandafter\def\csname cdnum@rv.ts.compmisll.emotion.ecescaledhi\endcsname{0.118}
\expandafter\def\csname cdnum@rv.ts.compmisll.emotion.ecescaledlo\endcsname{0.050}
\expandafter\def\csname cdnum@rv.ts.compmisll.politeness.ece\endcsname{0.082}
\expandafter\def\csname cdnum@rv.ts.compmisll.politeness.ecescaled\endcsname{0.047}
\expandafter\def\csname cdnum@rv.ts.compmisll.politeness.ecescaledhi\endcsname{0.099}
\expandafter\def\csname cdnum@rv.ts.compmisll.politeness.ecescaledlo\endcsname{0.038}
\expandafter\def\csname cdnum@rv.ts.compmisll.power.ece\endcsname{0.359}
\expandafter\def\csname cdnum@rv.ts.compmisll.power.ecescaled\endcsname{0.048}
\expandafter\def\csname cdnum@rv.ts.compmisll.power.ecescaledhi\endcsname{0.128}
\expandafter\def\csname cdnum@rv.ts.compmisll.power.ecescaledlo\endcsname{0.051}
\expandafter\def\csname cdnum@rv.ts.compmisll.toxicity.ece\endcsname{0.366}
\expandafter\def\csname cdnum@rv.ts.compmisll.toxicity.ecescaled\endcsname{0.061}
\expandafter\def\csname cdnum@rv.ts.compmisll.toxicity.ecescaledhi\endcsname{0.127}
\expandafter\def\csname cdnum@rv.ts.compmisll.toxicity.ecescaledlo\endcsname{0.050}
\expandafter\def\csname cdnum@rv.ts.compopen.doneneutral.ece\endcsname{0.117}
\expandafter\def\csname cdnum@rv.ts.compopen.doneneutral.ecescaled\endcsname{0.091}
\expandafter\def\csname cdnum@rv.ts.compopen.doneneutral.ecescaledhi\endcsname{0.135}
\expandafter\def\csname cdnum@rv.ts.compopen.doneneutral.ecescaledlo\endcsname{0.062}
\expandafter\def\csname cdnum@rv.ts.compopen.dtwo.ece\endcsname{0.204}
\expandafter\def\csname cdnum@rv.ts.compopen.dtwo.ecescaled\endcsname{0.096}
\expandafter\def\csname cdnum@rv.ts.compopen.dtwo.ecescaledhi\endcsname{0.126}
\expandafter\def\csname cdnum@rv.ts.compopen.dtwo.ecescaledlo\endcsname{0.066}
\expandafter\def\csname cdnum@rv.ts.compopen.emotion.ece\endcsname{0.136}
\expandafter\def\csname cdnum@rv.ts.compopen.emotion.ecescaled\endcsname{0.070}
\expandafter\def\csname cdnum@rv.ts.compopen.emotion.ecescaledhi\endcsname{0.125}
\expandafter\def\csname cdnum@rv.ts.compopen.emotion.ecescaledlo\endcsname{0.054}
\expandafter\def\csname cdnum@rv.ts.compopen.politeness.ece\endcsname{0.117}
\expandafter\def\csname cdnum@rv.ts.compopen.politeness.ecescaled\endcsname{0.098}
\expandafter\def\csname cdnum@rv.ts.compopen.politeness.ecescaledhi\endcsname{0.145}
\expandafter\def\csname cdnum@rv.ts.compopen.politeness.ecescaledlo\endcsname{0.066}
\expandafter\def\csname cdnum@rv.ts.compopen.power.ece\endcsname{0.284}
\expandafter\def\csname cdnum@rv.ts.compopen.power.ecescaled\endcsname{0.064}
\expandafter\def\csname cdnum@rv.ts.compopen.power.ecescaledhi\endcsname{0.118}
\expandafter\def\csname cdnum@rv.ts.compopen.power.ecescaledlo\endcsname{0.046}
\expandafter\def\csname cdnum@rv.ts.compopen.toxicity.ece\endcsname{0.292}
\expandafter\def\csname cdnum@rv.ts.compopen.toxicity.ecescaled\endcsname{0.061}
\expandafter\def\csname cdnum@rv.ts.compopen.toxicity.ecescaledhi\endcsname{0.125}
\expandafter\def\csname cdnum@rv.ts.compopen.toxicity.ecescaledlo\endcsname{0.046}
\expandafter\def\csname cdnum@rv.ts.comptwo.doneneutral.ece\endcsname{0.126}
\expandafter\def\csname cdnum@rv.ts.comptwo.doneneutral.ecescaled\endcsname{0.063}
\expandafter\def\csname cdnum@rv.ts.comptwo.doneneutral.ecescaledhi\endcsname{0.088}
\expandafter\def\csname cdnum@rv.ts.comptwo.doneneutral.ecescaledlo\endcsname{0.046}
\expandafter\def\csname cdnum@rv.ts.comptwo.dtwo.ece\endcsname{0.023}
\expandafter\def\csname cdnum@rv.ts.comptwo.dtwo.ecescaled\endcsname{0.027}
\expandafter\def\csname cdnum@rv.ts.comptwo.dtwo.ecescaledhi\endcsname{0.059}
\expandafter\def\csname cdnum@rv.ts.comptwo.dtwo.ecescaledlo\endcsname{0.021}
\expandafter\def\csname cdnum@rv.ts.comptwo.emotion.ece\endcsname{0.290}
\expandafter\def\csname cdnum@rv.ts.comptwo.emotion.ecescaled\endcsname{0.067}
\expandafter\def\csname cdnum@rv.ts.comptwo.emotion.ecescaledhi\endcsname{0.114}
\expandafter\def\csname cdnum@rv.ts.comptwo.emotion.ecescaledlo\endcsname{0.051}
\expandafter\def\csname cdnum@rv.ts.comptwo.politeness.ece\endcsname{0.197}
\expandafter\def\csname cdnum@rv.ts.comptwo.politeness.ecescaled\endcsname{0.089}
\expandafter\def\csname cdnum@rv.ts.comptwo.politeness.ecescaledhi\endcsname{0.145}
\expandafter\def\csname cdnum@rv.ts.comptwo.politeness.ecescaledlo\endcsname{0.074}
\expandafter\def\csname cdnum@rv.ts.comptwo.power.ece\endcsname{0.295}
\expandafter\def\csname cdnum@rv.ts.comptwo.power.ecescaled\endcsname{0.081}
\expandafter\def\csname cdnum@rv.ts.comptwo.power.ecescaledhi\endcsname{0.127}
\expandafter\def\csname cdnum@rv.ts.comptwo.power.ecescaledlo\endcsname{0.058}
\expandafter\def\csname cdnum@rv.ts.comptwo.toxicity.ece\endcsname{0.391}
\expandafter\def\csname cdnum@rv.ts.comptwo.toxicity.ecescaled\endcsname{0.083}
\expandafter\def\csname cdnum@rv.ts.comptwo.toxicity.ecescaledhi\endcsname{0.134}
\expandafter\def\csname cdnum@rv.ts.comptwo.toxicity.ecescaledlo\endcsname{0.055}
\expandafter\def\csname cdnum@rv.ts.comptwoll.doneneutral.ece\endcsname{0.061}
\expandafter\def\csname cdnum@rv.ts.comptwoll.doneneutral.ecescaled\endcsname{0.029}
\expandafter\def\csname cdnum@rv.ts.comptwoll.doneneutral.ecescaledhi\endcsname{0.049}
\expandafter\def\csname cdnum@rv.ts.comptwoll.doneneutral.ecescaledlo\endcsname{0.019}
\expandafter\def\csname cdnum@rv.ts.comptwoll.dtwo.ece\endcsname{0.044}
\expandafter\def\csname cdnum@rv.ts.comptwoll.dtwo.ecescaled\endcsname{0.035}
\expandafter\def\csname cdnum@rv.ts.comptwoll.dtwo.ecescaledhi\endcsname{0.060}
\expandafter\def\csname cdnum@rv.ts.comptwoll.dtwo.ecescaledlo\endcsname{0.021}
\expandafter\def\csname cdnum@rv.ts.comptwoll.emotion.ece\endcsname{0.371}
\expandafter\def\csname cdnum@rv.ts.comptwoll.emotion.ecescaled\endcsname{0.055}
\expandafter\def\csname cdnum@rv.ts.comptwoll.emotion.ecescaledhi\endcsname{0.104}
\expandafter\def\csname cdnum@rv.ts.comptwoll.emotion.ecescaledlo\endcsname{0.043}
\expandafter\def\csname cdnum@rv.ts.comptwoll.politeness.ece\endcsname{0.161}
\expandafter\def\csname cdnum@rv.ts.comptwoll.politeness.ecescaled\endcsname{0.039}
\expandafter\def\csname cdnum@rv.ts.comptwoll.politeness.ecescaledhi\endcsname{0.097}
\expandafter\def\csname cdnum@rv.ts.comptwoll.politeness.ecescaledlo\endcsname{0.037}
\expandafter\def\csname cdnum@rv.ts.comptwoll.power.ece\endcsname{0.294}
\expandafter\def\csname cdnum@rv.ts.comptwoll.power.ecescaled\endcsname{0.047}
\expandafter\def\csname cdnum@rv.ts.comptwoll.power.ecescaledhi\endcsname{0.115}
\expandafter\def\csname cdnum@rv.ts.comptwoll.power.ecescaledlo\endcsname{0.040}
\expandafter\def\csname cdnum@rv.ts.comptwoll.toxicity.ece\endcsname{0.427}
\expandafter\def\csname cdnum@rv.ts.comptwoll.toxicity.ecescaled\endcsname{0.053}
\expandafter\def\csname cdnum@rv.ts.comptwoll.toxicity.ecescaledhi\endcsname{0.117}
\expandafter\def\csname cdnum@rv.ts.comptwoll.toxicity.ecescaledlo\endcsname{0.051}
\expandafter\def\csname cdnum@rv.ts.comptwothink.doneneutral.ece\endcsname{0.129}
\expandafter\def\csname cdnum@rv.ts.comptwothink.doneneutral.ecescaled\endcsname{0.045}
\expandafter\def\csname cdnum@rv.ts.comptwothink.doneneutral.ecescaledhi\endcsname{0.077}
\expandafter\def\csname cdnum@rv.ts.comptwothink.doneneutral.ecescaledlo\endcsname{0.036}
\expandafter\def\csname cdnum@rv.ts.comptwothink.dtwo.ece\endcsname{0.038}
\expandafter\def\csname cdnum@rv.ts.comptwothink.dtwo.ecescaled\endcsname{0.032}
\expandafter\def\csname cdnum@rv.ts.comptwothink.dtwo.ecescaledhi\endcsname{0.056}
\expandafter\def\csname cdnum@rv.ts.comptwothink.dtwo.ecescaledlo\endcsname{0.023}
\expandafter\def\csname cdnum@rv.ts.comptwothink.emotion.ece\endcsname{0.263}
\expandafter\def\csname cdnum@rv.ts.comptwothink.emotion.ecescaled\endcsname{0.067}
\expandafter\def\csname cdnum@rv.ts.comptwothink.emotion.ecescaledhi\endcsname{0.121}
\expandafter\def\csname cdnum@rv.ts.comptwothink.emotion.ecescaledlo\endcsname{0.049}
\expandafter\def\csname cdnum@rv.ts.comptwothink.politeness.ece\endcsname{0.180}
\expandafter\def\csname cdnum@rv.ts.comptwothink.politeness.ecescaled\endcsname{0.049}
\expandafter\def\csname cdnum@rv.ts.comptwothink.politeness.ecescaledhi\endcsname{0.111}
\expandafter\def\csname cdnum@rv.ts.comptwothink.politeness.ecescaledlo\endcsname{0.043}
\expandafter\def\csname cdnum@rv.ts.comptwothink.power.ece\endcsname{0.288}
\expandafter\def\csname cdnum@rv.ts.comptwothink.power.ecescaled\endcsname{0.043}
\expandafter\def\csname cdnum@rv.ts.comptwothink.power.ecescaledhi\endcsname{0.106}
\expandafter\def\csname cdnum@rv.ts.comptwothink.power.ecescaledlo\endcsname{0.039}
\expandafter\def\csname cdnum@rv.ts.comptwothink.toxicity.ece\endcsname{0.352}
\expandafter\def\csname cdnum@rv.ts.comptwothink.toxicity.ecescaled\endcsname{0.058}
\expandafter\def\csname cdnum@rv.ts.comptwothink.toxicity.ecescaledhi\endcsname{0.123}
\expandafter\def\csname cdnum@rv.ts.comptwothink.toxicity.ecescaledlo\endcsname{0.050}
\expandafter\def\csname cdnum@rv.ts.decider.doneneutral.ece\endcsname{0.094}
\expandafter\def\csname cdnum@rv.ts.decider.doneneutral.ecescaled\endcsname{0.036}
\expandafter\def\csname cdnum@rv.ts.decider.doneneutral.ecescaledhi\endcsname{0.064}
\expandafter\def\csname cdnum@rv.ts.decider.doneneutral.ecescaledlo\endcsname{0.028}
\expandafter\def\csname cdnum@rv.ts.decider.dtwo.ece\endcsname{0.035}
\expandafter\def\csname cdnum@rv.ts.decider.dtwo.ecescaled\endcsname{0.025}
\expandafter\def\csname cdnum@rv.ts.decider.dtwo.ecescaledhi\endcsname{0.055}
\expandafter\def\csname cdnum@rv.ts.decider.dtwo.ecescaledlo\endcsname{0.017}
\expandafter\def\csname cdnum@rv.ts.decider.emotion.ece\endcsname{0.043}
\expandafter\def\csname cdnum@rv.ts.decider.emotion.ecescaled\endcsname{0.034}
\expandafter\def\csname cdnum@rv.ts.decider.emotion.ecescaledhi\endcsname{0.078}
\expandafter\def\csname cdnum@rv.ts.decider.emotion.ecescaledlo\endcsname{0.029}
\expandafter\def\csname cdnum@rv.ts.decider.politeness.ece\endcsname{0.236}
\expandafter\def\csname cdnum@rv.ts.decider.politeness.ecescaled\endcsname{0.065}
\expandafter\def\csname cdnum@rv.ts.decider.politeness.ecescaledhi\endcsname{0.113}
\expandafter\def\csname cdnum@rv.ts.decider.politeness.ecescaledlo\endcsname{0.044}
\expandafter\def\csname cdnum@rv.ts.decider.power.ece\endcsname{0.228}
\expandafter\def\csname cdnum@rv.ts.decider.power.ecescaled\endcsname{0.055}
\expandafter\def\csname cdnum@rv.ts.decider.power.ecescaledhi\endcsname{0.126}
\expandafter\def\csname cdnum@rv.ts.decider.power.ecescaledlo\endcsname{0.048}
\expandafter\def\csname cdnum@rv.ts.decider.toxicity.ece\endcsname{0.444}
\expandafter\def\csname cdnum@rv.ts.decider.toxicity.ecescaled\endcsname{0.081}
\expandafter\def\csname cdnum@rv.ts.decider.toxicity.ecescaledhi\endcsname{0.138}
\expandafter\def\csname cdnum@rv.ts.decider.toxicity.ecescaledlo\endcsname{0.056}
\expandafter\def\csname cdnum@rv.ts.jev.doneneutral.ece\endcsname{0.033}
\expandafter\def\csname cdnum@rv.ts.jev.doneneutral.ecescaled\endcsname{0.032}
\expandafter\def\csname cdnum@rv.ts.jev.doneneutral.ecescaledhi\endcsname{0.055}
\expandafter\def\csname cdnum@rv.ts.jev.doneneutral.ecescaledlo\endcsname{0.023}
\expandafter\def\csname cdnum@rv.ts.jev.dtwo.ece\endcsname{0.032}
\expandafter\def\csname cdnum@rv.ts.jev.dtwo.ecescaled\endcsname{0.032}
\expandafter\def\csname cdnum@rv.ts.jev.dtwo.ecescaledhi\endcsname{0.058}
\expandafter\def\csname cdnum@rv.ts.jev.dtwo.ecescaledlo\endcsname{0.023}
\expandafter\def\csname cdnum@rv.ts.jev.emotion.ece\endcsname{0.314}
\expandafter\def\csname cdnum@rv.ts.jev.emotion.ecescaled\endcsname{0.073}
\expandafter\def\csname cdnum@rv.ts.jev.emotion.ecescaledhi\endcsname{0.118}
\expandafter\def\csname cdnum@rv.ts.jev.emotion.ecescaledlo\endcsname{0.051}
\expandafter\def\csname cdnum@rv.ts.jev.politeness.ece\endcsname{0.085}
\expandafter\def\csname cdnum@rv.ts.jev.politeness.ecescaled\endcsname{0.048}
\expandafter\def\csname cdnum@rv.ts.jev.politeness.ecescaledhi\endcsname{0.097}
\expandafter\def\csname cdnum@rv.ts.jev.politeness.ecescaledlo\endcsname{0.036}
\expandafter\def\csname cdnum@rv.ts.jev.power.ece\endcsname{0.247}
\expandafter\def\csname cdnum@rv.ts.jev.power.ecescaled\endcsname{0.069}
\expandafter\def\csname cdnum@rv.ts.jev.power.ecescaledhi\endcsname{0.136}
\expandafter\def\csname cdnum@rv.ts.jev.power.ecescaledlo\endcsname{0.061}
\expandafter\def\csname cdnum@rv.ts.jev.toxicity.ece\endcsname{0.365}
\expandafter\def\csname cdnum@rv.ts.jev.toxicity.ecescaled\endcsname{0.084}
\expandafter\def\csname cdnum@rv.ts.jev.toxicity.ecescaledhi\endcsname{0.133}
\expandafter\def\csname cdnum@rv.ts.jev.toxicity.ecescaledlo\endcsname{0.058}
\expandafter\def\csname cdnum@rv.ts.kevlarge.doneneutral.ece\endcsname{0.087}
\expandafter\def\csname cdnum@rv.ts.kevlarge.doneneutral.ecescaled\endcsname{0.047}
\expandafter\def\csname cdnum@rv.ts.kevlarge.doneneutral.ecescaledhi\endcsname{0.072}
\expandafter\def\csname cdnum@rv.ts.kevlarge.doneneutral.ecescaledlo\endcsname{0.035}
\expandafter\def\csname cdnum@rv.ts.kevlarge.dtwo.ece\endcsname{0.377}
\expandafter\def\csname cdnum@rv.ts.kevlarge.dtwo.ecescaled\endcsname{0.046}
\expandafter\def\csname cdnum@rv.ts.kevlarge.dtwo.ecescaledhi\endcsname{0.079}
\expandafter\def\csname cdnum@rv.ts.kevlarge.dtwo.ecescaledlo\endcsname{0.033}
\expandafter\def\csname cdnum@rv.ts.kevlarge.emotion.ece\endcsname{0.057}
\expandafter\def\csname cdnum@rv.ts.kevlarge.emotion.ecescaled\endcsname{0.051}
\expandafter\def\csname cdnum@rv.ts.kevlarge.emotion.ecescaledhi\endcsname{0.106}
\expandafter\def\csname cdnum@rv.ts.kevlarge.emotion.ecescaledlo\endcsname{0.043}
\expandafter\def\csname cdnum@rv.ts.kevlarge.politeness.ece\endcsname{0.150}
\expandafter\def\csname cdnum@rv.ts.kevlarge.politeness.ecescaled\endcsname{0.094}
\expandafter\def\csname cdnum@rv.ts.kevlarge.politeness.ecescaledhi\endcsname{0.149}
\expandafter\def\csname cdnum@rv.ts.kevlarge.politeness.ecescaledlo\endcsname{0.069}
\expandafter\def\csname cdnum@rv.ts.kevlarge.power.ece\endcsname{0.120}
\expandafter\def\csname cdnum@rv.ts.kevlarge.power.ecescaled\endcsname{0.037}
\expandafter\def\csname cdnum@rv.ts.kevlarge.power.ecescaledhi\endcsname{0.113}
\expandafter\def\csname cdnum@rv.ts.kevlarge.power.ecescaledlo\endcsname{0.042}
\expandafter\def\csname cdnum@rv.ts.kevlarge.toxicity.ece\endcsname{0.270}
\expandafter\def\csname cdnum@rv.ts.kevlarge.toxicity.ecescaled\endcsname{0.075}
\expandafter\def\csname cdnum@rv.ts.kevlarge.toxicity.ecescaledhi\endcsname{0.125}
\expandafter\def\csname cdnum@rv.ts.kevlarge.toxicity.ecescaledlo\endcsname{0.049}
\expandafter\def\csname cdnum@rv.ts.kevsmall.doneneutral.ece\endcsname{0.060}
\expandafter\def\csname cdnum@rv.ts.kevsmall.doneneutral.ecescaled\endcsname{0.070}
\expandafter\def\csname cdnum@rv.ts.kevsmall.doneneutral.ecescaledhi\endcsname{0.094}
\expandafter\def\csname cdnum@rv.ts.kevsmall.doneneutral.ecescaledlo\endcsname{0.051}
\expandafter\def\csname cdnum@rv.ts.kevsmall.dtwo.ece\endcsname{0.254}
\expandafter\def\csname cdnum@rv.ts.kevsmall.dtwo.ecescaled\endcsname{0.087}
\expandafter\def\csname cdnum@rv.ts.kevsmall.dtwo.ecescaledhi\endcsname{0.118}
\expandafter\def\csname cdnum@rv.ts.kevsmall.dtwo.ecescaledlo\endcsname{0.066}
\expandafter\def\csname cdnum@rv.ts.kevsmall.emotion.ece\endcsname{0.073}
\expandafter\def\csname cdnum@rv.ts.kevsmall.emotion.ecescaled\endcsname{0.074}
\expandafter\def\csname cdnum@rv.ts.kevsmall.emotion.ecescaledhi\endcsname{0.128}
\expandafter\def\csname cdnum@rv.ts.kevsmall.emotion.ecescaledlo\endcsname{0.053}
\expandafter\def\csname cdnum@rv.ts.kevsmall.politeness.ece\endcsname{0.125}
\expandafter\def\csname cdnum@rv.ts.kevsmall.politeness.ecescaled\endcsname{0.042}
\expandafter\def\csname cdnum@rv.ts.kevsmall.politeness.ecescaledhi\endcsname{0.112}
\expandafter\def\csname cdnum@rv.ts.kevsmall.politeness.ecescaledlo\endcsname{0.045}
\expandafter\def\csname cdnum@rv.ts.kevsmall.power.ece\endcsname{0.153}
\expandafter\def\csname cdnum@rv.ts.kevsmall.power.ecescaled\endcsname{0.064}
\expandafter\def\csname cdnum@rv.ts.kevsmall.power.ecescaledhi\endcsname{0.119}
\expandafter\def\csname cdnum@rv.ts.kevsmall.power.ecescaledlo\endcsname{0.045}
\expandafter\def\csname cdnum@rv.ts.kevsmall.toxicity.ece\endcsname{0.183}
\expandafter\def\csname cdnum@rv.ts.kevsmall.toxicity.ecescaled\endcsname{0.043}
\expandafter\def\csname cdnum@rv.ts.kevsmall.toxicity.ecescaledhi\endcsname{0.126}
\expandafter\def\csname cdnum@rv.ts.kevsmall.toxicity.ecescaledlo\endcsname{0.045}
\expandafter\def\csname cdnum@rv.ts.layaen.doneneutral.ece\endcsname{0.156}
\expandafter\def\csname cdnum@rv.ts.layaen.doneneutral.ecescaled\endcsname{0.068}
\expandafter\def\csname cdnum@rv.ts.layaen.doneneutral.ecescaledhi\endcsname{0.099}
\expandafter\def\csname cdnum@rv.ts.layaen.doneneutral.ecescaledlo\endcsname{0.058}
\expandafter\def\csname cdnum@rv.ts.layaen.emotion.ece\endcsname{0.327}
\expandafter\def\csname cdnum@rv.ts.layaen.emotion.ecescaled\endcsname{0.069}
\expandafter\def\csname cdnum@rv.ts.layaen.emotion.ecescaledhi\endcsname{0.106}
\expandafter\def\csname cdnum@rv.ts.layaen.emotion.ecescaledlo\endcsname{0.044}
\expandafter\def\csname cdnum@rv.ts.layaen.politeness.ece\endcsname{0.190}
\expandafter\def\csname cdnum@rv.ts.layaen.politeness.ecescaled\endcsname{0.066}
\expandafter\def\csname cdnum@rv.ts.layaen.politeness.ecescaledhi\endcsname{0.114}
\expandafter\def\csname cdnum@rv.ts.layaen.politeness.ecescaledlo\endcsname{0.045}
\expandafter\def\csname cdnum@rv.ts.layaen.power.ece\endcsname{0.090}
\expandafter\def\csname cdnum@rv.ts.layaen.power.ecescaled\endcsname{0.049}
\expandafter\def\csname cdnum@rv.ts.layaen.power.ecescaledhi\endcsname{0.118}
\expandafter\def\csname cdnum@rv.ts.layaen.power.ecescaledlo\endcsname{0.045}
\expandafter\def\csname cdnum@rv.ts.layaen.toxicity.ece\endcsname{0.065}
\expandafter\def\csname cdnum@rv.ts.layaen.toxicity.ecescaled\endcsname{0.042}
\expandafter\def\csname cdnum@rv.ts.layaen.toxicity.ecescaledhi\endcsname{0.123}
\expandafter\def\csname cdnum@rv.ts.layaen.toxicity.ecescaledlo\endcsname{0.043}
\expandafter\def\csname cdnum@rv.ts.layaml.doneneutral.ece\endcsname{0.279}
\expandafter\def\csname cdnum@rv.ts.layaml.doneneutral.ecescaled\endcsname{0.044}
\expandafter\def\csname cdnum@rv.ts.layaml.doneneutral.ecescaledhi\endcsname{0.069}
\expandafter\def\csname cdnum@rv.ts.layaml.doneneutral.ecescaledlo\endcsname{0.033}
\expandafter\def\csname cdnum@rv.ts.layaml.dtwo.ece\endcsname{0.031}
\expandafter\def\csname cdnum@rv.ts.layaml.dtwo.ecescaled\endcsname{0.010}
\expandafter\def\csname cdnum@rv.ts.layaml.dtwo.ecescaledhi\endcsname{0.016}
\expandafter\def\csname cdnum@rv.ts.layaml.dtwo.ecescaledlo\endcsname{0.007}
\expandafter\def\csname cdnum@rv.ts.layaml.emotion.ece\endcsname{0.269}
\expandafter\def\csname cdnum@rv.ts.layaml.emotion.ecescaled\endcsname{0.071}
\expandafter\def\csname cdnum@rv.ts.layaml.emotion.ecescaledhi\endcsname{0.132}
\expandafter\def\csname cdnum@rv.ts.layaml.emotion.ecescaledlo\endcsname{0.060}
\expandafter\def\csname cdnum@rv.ts.layaml.politeness.ece\endcsname{0.163}
\expandafter\def\csname cdnum@rv.ts.layaml.politeness.ecescaled\endcsname{0.031}
\expandafter\def\csname cdnum@rv.ts.layaml.politeness.ecescaledhi\endcsname{0.123}
\expandafter\def\csname cdnum@rv.ts.layaml.politeness.ecescaledlo\endcsname{0.048}
\expandafter\def\csname cdnum@rv.ts.layaml.power.ece\endcsname{0.217}
\expandafter\def\csname cdnum@rv.ts.layaml.power.ecescaled\endcsname{0.040}
\expandafter\def\csname cdnum@rv.ts.layaml.power.ecescaledhi\endcsname{0.111}
\expandafter\def\csname cdnum@rv.ts.layaml.power.ecescaledlo\endcsname{0.045}
\expandafter\def\csname cdnum@rv.ts.layaml.toxicity.ece\endcsname{0.173}
\expandafter\def\csname cdnum@rv.ts.layaml.toxicity.ecescaled\endcsname{0.080}
\expandafter\def\csname cdnum@rv.ts.layaml.toxicity.ecescaledhi\endcsname{0.117}
\expandafter\def\csname cdnum@rv.ts.layaml.toxicity.ecescaledlo\endcsname{0.046}
\expandafter\def\csname cdnum@rv.ts.nimble.doneneutral.ece\endcsname{0.057}
\expandafter\def\csname cdnum@rv.ts.nimble.doneneutral.ecescaled\endcsname{0.026}
\expandafter\def\csname cdnum@rv.ts.nimble.doneneutral.ecescaledhi\endcsname{0.058}
\expandafter\def\csname cdnum@rv.ts.nimble.doneneutral.ecescaledlo\endcsname{0.020}
\expandafter\def\csname cdnum@rv.ts.nimble.dtwo.ece\endcsname{0.069}
\expandafter\def\csname cdnum@rv.ts.nimble.dtwo.ecescaled\endcsname{0.037}
\expandafter\def\csname cdnum@rv.ts.nimble.dtwo.ecescaledhi\endcsname{0.065}
\expandafter\def\csname cdnum@rv.ts.nimble.dtwo.ecescaledlo\endcsname{0.026}
\expandafter\def\csname cdnum@rv.ts.nimble.emotion.ece\endcsname{0.287}
\expandafter\def\csname cdnum@rv.ts.nimble.emotion.ecescaled\endcsname{0.050}
\expandafter\def\csname cdnum@rv.ts.nimble.emotion.ecescaledhi\endcsname{0.116}
\expandafter\def\csname cdnum@rv.ts.nimble.emotion.ecescaledlo\endcsname{0.047}
\expandafter\def\csname cdnum@rv.ts.nimble.politeness.ece\endcsname{0.170}
\expandafter\def\csname cdnum@rv.ts.nimble.politeness.ecescaled\endcsname{0.077}
\expandafter\def\csname cdnum@rv.ts.nimble.politeness.ecescaledhi\endcsname{0.130}
\expandafter\def\csname cdnum@rv.ts.nimble.politeness.ecescaledlo\endcsname{0.057}
\expandafter\def\csname cdnum@rv.ts.nimble.power.ece\endcsname{0.337}
\expandafter\def\csname cdnum@rv.ts.nimble.power.ecescaled\endcsname{0.064}
\expandafter\def\csname cdnum@rv.ts.nimble.power.ecescaledhi\endcsname{0.126}
\expandafter\def\csname cdnum@rv.ts.nimble.power.ecescaledlo\endcsname{0.051}
\expandafter\def\csname cdnum@rv.ts.nimble.toxicity.ece\endcsname{0.345}
\expandafter\def\csname cdnum@rv.ts.nimble.toxicity.ecescaled\endcsname{0.050}
\expandafter\def\csname cdnum@rv.ts.nimble.toxicity.ecescaledhi\endcsname{0.111}
\expandafter\def\csname cdnum@rv.ts.nimble.toxicity.ecescaledlo\endcsname{0.041}
\expandafter\def\csname cdnum@rv.ts.thisthat.doneneutral.ece\endcsname{0.196}
\expandafter\def\csname cdnum@rv.ts.thisthat.doneneutral.ecescaled\endcsname{0.040}
\expandafter\def\csname cdnum@rv.ts.thisthat.doneneutral.ecescaledhi\endcsname{0.064}
\expandafter\def\csname cdnum@rv.ts.thisthat.doneneutral.ecescaledlo\endcsname{0.026}
\expandafter\def\csname cdnum@rv.ts.thisthat.dtwo.ece\endcsname{0.055}
\expandafter\def\csname cdnum@rv.ts.thisthat.dtwo.ecescaled\endcsname{0.023}
\expandafter\def\csname cdnum@rv.ts.thisthat.dtwo.ecescaledhi\endcsname{0.056}
\expandafter\def\csname cdnum@rv.ts.thisthat.dtwo.ecescaledlo\endcsname{0.017}
\expandafter\def\csname cdnum@rv.ts.thisthat.emotion.ece\endcsname{0.111}
\expandafter\def\csname cdnum@rv.ts.thisthat.emotion.ecescaled\endcsname{0.039}
\expandafter\def\csname cdnum@rv.ts.thisthat.emotion.ecescaledhi\endcsname{0.080}
\expandafter\def\csname cdnum@rv.ts.thisthat.emotion.ecescaledlo\endcsname{0.033}
\expandafter\def\csname cdnum@rv.ts.thisthat.politeness.ece\endcsname{0.452}
\expandafter\def\csname cdnum@rv.ts.thisthat.politeness.ecescaled\endcsname{0.070}
\expandafter\def\csname cdnum@rv.ts.thisthat.politeness.ecescaledhi\endcsname{0.128}
\expandafter\def\csname cdnum@rv.ts.thisthat.politeness.ecescaledlo\endcsname{0.058}
\expandafter\def\csname cdnum@rv.ts.thisthat.power.ece\endcsname{0.296}
\expandafter\def\csname cdnum@rv.ts.thisthat.power.ecescaled\endcsname{0.037}
\expandafter\def\csname cdnum@rv.ts.thisthat.power.ecescaledhi\endcsname{0.117}
\expandafter\def\csname cdnum@rv.ts.thisthat.power.ecescaledlo\endcsname{0.042}
\expandafter\def\csname cdnum@rv.ts.thisthat.toxicity.ece\endcsname{0.466}
\expandafter\def\csname cdnum@rv.ts.thisthat.toxicity.ecescaled\endcsname{0.105}
\expandafter\def\csname cdnum@rv.ts.thisthat.toxicity.ecescaledhi\endcsname{0.152}
\expandafter\def\csname cdnum@rv.ts.thisthat.toxicity.ecescaledlo\endcsname{0.075}
\expandafter\def\csname cdnum@sat.compgemll.doneneutral\endcsname{0.78}
\expandafter\def\csname cdnum@sat.compgemll.dtwo\endcsname{0.96}
\expandafter\def\csname cdnum@vr.compgem.banking.acclenient\endcsname{0.821}
\expandafter\def\csname cdnum@vr.compgem.banking.accstrict\endcsname{0.821}
\expandafter\def\csname cdnum@vr.compgem.banking.failshare\endcsname{0.000}
\expandafter\def\csname cdnum@vr.compgem.doneneutral.acclenient\endcsname{0.741}
\expandafter\def\csname cdnum@vr.compgem.doneneutral.accstrict\endcsname{0.741}
\expandafter\def\csname cdnum@vr.compgem.doneneutral.failshare\endcsname{0.001}
\expandafter\def\csname cdnum@vr.compgem.dtwo.acclenient\endcsname{0.942}
\expandafter\def\csname cdnum@vr.compgem.dtwo.accstrict\endcsname{0.942}
\expandafter\def\csname cdnum@vr.compgem.dtwo.failshare\endcsname{0.000}
\expandafter\def\csname cdnum@vr.compgem.emotion.acclenient\endcsname{0.474}
\expandafter\def\csname cdnum@vr.compgem.emotion.accstrict\endcsname{0.474}
\expandafter\def\csname cdnum@vr.compgem.emotion.failshare\endcsname{0.000}
\expandafter\def\csname cdnum@vr.compgem.goemo.acclenient\endcsname{0.573}
\expandafter\def\csname cdnum@vr.compgem.goemo.accstrict\endcsname{0.573}
\expandafter\def\csname cdnum@vr.compgem.goemo.failshare\endcsname{0.000}
\expandafter\def\csname cdnum@vr.compmis.banking.acclenient\endcsname{0.679}
\expandafter\def\csname cdnum@vr.compmis.banking.accstrict\endcsname{0.679}
\expandafter\def\csname cdnum@vr.compmis.banking.failshare\endcsname{0.000}
\expandafter\def\csname cdnum@vr.compmis.doneneutral.acclenient\endcsname{0.606}
\expandafter\def\csname cdnum@vr.compmis.doneneutral.accstrict\endcsname{0.606}
\expandafter\def\csname cdnum@vr.compmis.doneneutral.failshare\endcsname{0.119}
\expandafter\def\csname cdnum@vr.compmis.doneneutral.parsefailed\endcsname{235}
\expandafter\def\csname cdnum@vr.compmis.dtwo.acclenient\endcsname{0.757}
\expandafter\def\csname cdnum@vr.compmis.dtwo.accstrict\endcsname{0.757}
\expandafter\def\csname cdnum@vr.compmis.dtwo.failshare\endcsname{0.000}
\expandafter\def\csname cdnum@vr.compmis.emotion.acclenient\endcsname{0.464}
\expandafter\def\csname cdnum@vr.compmis.emotion.accstrict\endcsname{0.243}
\expandafter\def\csname cdnum@vr.compmis.emotion.failshare\endcsname{0.767}
\expandafter\def\csname cdnum@vr.compmis.emotion.n\endcsname{498}
\expandafter\def\csname cdnum@vr.compmis.emotion.recovered\endcsname{382}
\expandafter\def\csname cdnum@vr.compmis.goemo.acclenient\endcsname{0.567}
\expandafter\def\csname cdnum@vr.compmis.goemo.accstrict\endcsname{0.226}
\expandafter\def\csname cdnum@vr.compmis.goemo.failshare\endcsname{0.695}
\expandafter\def\csname cdnum@vr.compopen.banking.acclenient\endcsname{0.677}
\expandafter\def\csname cdnum@vr.compopen.banking.accstrict\endcsname{0.659}
\expandafter\def\csname cdnum@vr.compopen.banking.failshare\endcsname{0.045}
\expandafter\def\csname cdnum@vr.compopen.doneneutral.acclenient\endcsname{0.680}
\expandafter\def\csname cdnum@vr.compopen.doneneutral.accstrict\endcsname{0.680}
\expandafter\def\csname cdnum@vr.compopen.doneneutral.failshare\endcsname{0.000}
\expandafter\def\csname cdnum@vr.compopen.dtwo.acclenient\endcsname{0.735}
\expandafter\def\csname cdnum@vr.compopen.dtwo.accstrict\endcsname{0.728}
\expandafter\def\csname cdnum@vr.compopen.dtwo.failshare\endcsname{0.015}
\expandafter\def\csname cdnum@vr.compopen.emotion.acclenient\endcsname{0.466}
\expandafter\def\csname cdnum@vr.compopen.emotion.accstrict\endcsname{0.464}
\expandafter\def\csname cdnum@vr.compopen.emotion.failshare\endcsname{0.004}
\expandafter\def\csname cdnum@vr.compopen.goemo.acclenient\endcsname{0.585}
\expandafter\def\csname cdnum@vr.compopen.goemo.accstrict\endcsname{0.585}
\expandafter\def\csname cdnum@vr.compopen.goemo.failshare\endcsname{0.000}
\expandafter\def\csname cdnum@vr.comptwo.banking.acclenient\endcsname{0.794}
\expandafter\def\csname cdnum@vr.comptwo.banking.accstrict\endcsname{0.765}
\expandafter\def\csname cdnum@vr.comptwo.banking.failshare\endcsname{0.036}
\expandafter\def\csname cdnum@vr.comptwo.doneneutral.acclenient\endcsname{0.725}
\expandafter\def\csname cdnum@vr.comptwo.doneneutral.accstrict\endcsname{0.725}
\expandafter\def\csname cdnum@vr.comptwo.doneneutral.failshare\endcsname{0.000}
\expandafter\def\csname cdnum@vr.comptwo.dtwo.acclenient\endcsname{0.907}
\expandafter\def\csname cdnum@vr.comptwo.dtwo.accstrict\endcsname{0.683}
\expandafter\def\csname cdnum@vr.comptwo.dtwo.failshare\endcsname{0.267}
\expandafter\def\csname cdnum@vr.comptwo.dtwo.n\endcsname{600}
\expandafter\def\csname cdnum@vr.comptwo.dtwo.recovered\endcsname{160}
\expandafter\def\csname cdnum@vr.comptwo.emotion.acclenient\endcsname{0.466}
\expandafter\def\csname cdnum@vr.comptwo.emotion.accstrict\endcsname{0.398}
\expandafter\def\csname cdnum@vr.comptwo.emotion.failshare\endcsname{0.151}
\expandafter\def\csname cdnum@vr.comptwo.goemo.acclenient\endcsname{0.583}
\expandafter\def\csname cdnum@vr.comptwo.goemo.accstrict\endcsname{0.530}
\expandafter\def\csname cdnum@vr.comptwo.goemo.failshare\endcsname{0.063}
\expandafter\def\csname cdnum@vr.comptwothink.banking.acclenient\endcsname{0.825}
\expandafter\def\csname cdnum@vr.comptwothink.banking.accstrict\endcsname{0.825}
\expandafter\def\csname cdnum@vr.comptwothink.banking.failshare\endcsname{0.000}
\expandafter\def\csname cdnum@vr.comptwothink.doneneutral.acclenient\endcsname{0.721}
\expandafter\def\csname cdnum@vr.comptwothink.doneneutral.accstrict\endcsname{0.721}
\expandafter\def\csname cdnum@vr.comptwothink.doneneutral.failshare\endcsname{0.000}
\expandafter\def\csname cdnum@vr.comptwothink.dtwo.acclenient\endcsname{0.928}
\expandafter\def\csname cdnum@vr.comptwothink.dtwo.accstrict\endcsname{0.928}
\expandafter\def\csname cdnum@vr.comptwothink.dtwo.failshare\endcsname{0.007}
\expandafter\def\csname cdnum@vr.comptwothink.emotion.acclenient\endcsname{0.518}
\expandafter\def\csname cdnum@vr.comptwothink.emotion.accstrict\endcsname{0.518}
\expandafter\def\csname cdnum@vr.comptwothink.emotion.failshare\endcsname{0.002}
\expandafter\def\csname cdnum@vr.comptwothink.goemo.acclenient\endcsname{0.606}
\expandafter\def\csname cdnum@vr.comptwothink.goemo.accstrict\endcsname{0.606}
\expandafter\def\csname cdnum@vr.comptwothink.goemo.failshare\endcsname{0.002}
\expandafter\def\csname cdnum@x.decider.emotion.accunseen\endcsname{0.803}
\expandafter\def\csname cdnum@x.emotion.unseen\endcsname{468}

%% file: sections/abstract.tex
\begin{abstract}
Software that hands branching decisions to a model needs a declared option and a probability it can threshold. Typed decision models, also called System One models, return such probabilities without generating text, while supervised classifiers and generative language models are the established alternatives. One harness sends eight decision-model checkpoints from six families, including the hosted model Jev, and four open generative models from three developers the same semantic requests, and scores trained and zero-shot classifiers on the same workflow, intent, emotion and social-science items. With task labels, a fine-tuned DeBERTa-v3-large has the highest observed accuracy on every labeled benchmark but one. Without labels, no decision model is significantly more accurate than Jev on workflows or intents, but Gemma-4-31B matches it on workflows and exceeds it on CLINC-150 at higher cost and latency. Stated probabilities of generative models become unreadable when replies miss the key format, whereas key likelihoods avoid this but can saturate. A guaranteed \num{rc.alpha} percent risk leaves Jev \num{rc.jev.dtwo.cov} of the intent decisions, and an in-scope threshold still accepts \num{rv.gate.jev.far} of out-of-scope requests. On typed-decisions, swapping yes and no flips \num{e2.jev.done.kswap.flips} answers per hundred for Jev and at least \num{e2.compgemll.done.kswap.flips} for every generative model tested, against at most \num{e2.kevlarge.done.kswap.flips} for four fine-tuned decision checkpoints. Exposure to a benchmark's training data explains the largest lead of an open checkpoint, which vanishes on rater-labeled emotions. An intent-trained first stage escalating to Gemma-4-31B reaches that model's accuracy at about Jev's price. The results yield condition-dependent design rules for automated decision gates.
\end{abstract}

%% file: sections/introduction.tex
\section{Introduction}
\label{sec:introduction}

Software systems increasingly delegate routine branching decisions to machine learning models, and most of these decisions have a small set of admissible outcomes. A support system decides whether to escalate a reply, an accounts-payable system decides whether to pay, hold or reject an invoice, an incident system assigns an urgency level, and a virtual assistant routes a request to one of many handlers or to a fallback. A program that hands such a branch to a model needs one of the declared options and a number it can compare with a threshold. Generative language models answer in text, and the confidence they state in words is often poorly aligned with their accuracy \citep{xiong2024can}, while their token probabilities can be worse still after preference tuning \citep{tian2023just}. Constrained decoding guarantees that the text matches a schema \citep{geng2023grammar,willard2023efficient,dong2025xgrammar}, but it does not turn the answer into a calibrated probability over the declared options.

Several kinds of model can fill this role. Supervised classifiers trained on the task's own labels remain a standard choice where such labels exist \citep{devlin2019bert,tunstall2022setfit}. Zero-shot classifiers score label descriptions without task training \citep{yin2019benchmarking}. A newer class, called decision models here, reads a state and a set of typed questions and returns for every question a probability distribution over the declared options in one forward pass without generating text \citep{typesafe2026launch,typesafe2026api}. Their vendor calls them System One models \citep{typesafe2026systemone}. The hosted model Jev introduced the interface with a yes-or-no probability, a choice among declared options and a score on an ordered scale. Open implementations of the same contract followed on public backbones with different decision heads, among them Laya \citep{laya2026card}, Kev \citep{kev9b2026card}, decider-2b \citep{decider2b2026card}, this-that-model-1.0 \citep{thisthat2026card} and Nimble-9B \citep{nimble2026card}. An analysis of \num{lit.ling2026jev.projects} public software projects that call the hosted decision model Jev finds it used as a reusable decision component, for attribute judgment and scoring, action selection, content filtering, and the selection of models and tools \citep{ling2026jev}.

A practitioner who must choose among these options has model cards that report numbers on different items and a growing set of independent evaluations. Each evaluation stays inside one task family, namely computational social science \citep{ibrahim2026evaluating}, option naming \citep{sun2026typesafe}, judging \citep{li2026jevasajudge}, radiology report checking \citep{huang2026can} or call screening \citep{ren2026openjev}. Their prompts, option texts and scoring code differ, so their numbers cannot be placed side by side, and where trained classifiers appear they are quoted from earlier work or confined to one task family.

The engineering setting of this study is the automated decision gate, the point in a software system where a program asks a model for one of a few declared options and acts on the answer without a person in the loop. Beyond accuracy, a gate needs a confidence threshold with a known error rate and stability when the option schema grows or its names change. It also needs a way to turn away requests outside its scope, a predictable cost and latency, and a way to escalate the decisions the model doubts. Each of these properties depends on conditions the engineer controls or inherits, such as whether labeled data exist, how many options a question has, how options are named and rendered, and what a second stage costs.

This paper compares these model classes under matched conditions. One harness sends six families of decision models, eight checkpoints in all, and four open generative language models from three developers the same semantic requests, and supervised and zero-shot classifiers are scored on the same test items. The tasks are a synthetic workflow benchmark, an intent benchmark with \num{p.kall} intents and out-of-scope inputs, and four social-science tasks. Two further benchmarks, emotions labeled by agreeing raters and a second intent set, change which open checkpoints have seen the task in training. The study asks four questions. Research question 1 (RQ1) asks how the relative accuracy of the model classes depends on whether labels for the task are available. RQ2 asks how well calibration transfers across tasks and option counts, what error rate a threshold chosen on held-out data actually delivers, and what coverage remains when that error rate must be guaranteed. RQ3 asks how option count, input budget, option naming and option rendering change the behavior of each model. RQ4 asks under which accuracy and cost conditions a confidence-gated cascade of two models helps.

The paper contributes a controlled, model-agnostic evaluation methodology rather than a new model. It consists of a harness that normalizes one typed request format across hosted and local checkpoints and three propositions that make temperature recovery, hierarchical accuracy and cascade accuracy computable from the recorded answers. It also includes stress tests that vary option count across repeated distractor permutations and the names and rendering of options, and controls that score each decoder-based checkpoint's untuned backbone and read every larger generative model in two ways. Each such model states its probabilities in a reply and is also read through the likelihood of each option key. Its practical contribution is a set of condition-dependent design rules for automated decision gates, each tied to a measured quantity. The harness, the frozen inputs with their digests, the per-request answers and the analysis code are released, so each number here can be regenerated.

The remainder of the paper is organized as follows. Section~\ref{sec:related} reviews the related literature, Section~\ref{sec:methods} states the problem and describes the harness and measures, and Section~\ref{sec:models} describes the models and datasets. Section~\ref{sec:results} reports the results, Section~\ref{sec:engineering} derives the design rules, Section~\ref{sec:discussion} relates the findings to prior studies, and Section~\ref{sec:conclusion} concludes.

%% file: sections/related.tex
\section{Related work}
\label{sec:related}

The models compared here sit at the meeting point of text classification with pretrained language models, structured generation, calibration, selective prediction and model cascades, and this section reviews these lines before stating the gap that the benchmark addresses.

Text classification with pretrained models began with fine-tuned encoders that attach a task-specific head to a bidirectional transformer \citep{devlin2019bert,liu2019roberta}. Later work removed the need for a trained head by recasting classification as a cloze task whose answer words, the verbalizers, stand for the labels \citep{schick2021exploiting,gao2021making}, or as an entailment problem between the input and a description of each label \citep{yin2019benchmarking}. Efficient few-shot fine-tuning reduced the labeled data that a new label set needs \citep{tunstall2022setfit}, and generalist encoders that score labels given in the input handle new label sets without retraining \citep{zaratiana2023gliner,stepanov2025gliclass}. Instruction-tuned generative models now classify text from a prompt alone, and recent comparisons weigh them against fine-tuned classifiers on accuracy, cost and language coverage \citep{chaedavidson2026llmclassifiers,llmclassifiers2025era}. Scoring label words with a language model is sensitive to how the label is phrased, because different surface forms of the same answer compete for probability \citep{holtzman2021surface}. Decision models inherit this idea of scoring declared options, and they add a typed request format and a head trained to produce one distribution per question.

Structured generation takes the opposite route and constrains a generative model so that its text matches a grammar or a schema. Grammar-constrained decoding \citep{geng2023grammar,scholak2021picard}, query languages for prompting \citep{beurerkellner2023prompting}, finite-state guidance \citep{willard2023efficient} and fast grammar engines inside serving runtimes \citep{dong2025xgrammar,zheng2024sglang} make a well-formed answer certain. They do not make the probability of that answer meaningful, and a generative model asked to state its confidence in words tends to be over-confident \citep{xiong2024can}, although verbalized confidence can be better calibrated than token probabilities after preference tuning \citep{tian2023just}. The generative comparators in this study follow this route, returning a JSON object with a probability per option under structured decoding, which makes them the natural baseline for a decision model that returns the same object without generating text. Reading the larger one also through the likelihood of each option key separates the readout from the model.

Independent evaluations of Jev have grown quickly since its launch, and most stay inside one task family. \citet{ibrahim2026evaluating} score Jev and several open decision models against nineteen large language models on eighteen computational social science tasks. In their results Jev trails the best language model for each task on most tasks but costs less and calibrates better than most comparators. A cascade that routes low-confidence items to a language model matches or exceeds that model alone at a fraction of its cost. \citet{li2026jevasajudge} evaluate Jev as a judge and find that a frozen threshold-gated cascade to a stronger judge keeps nearly all of that judge's accuracy at a little over half of its fee. Jev's gap to the stronger judge widens on judgments that require checking a multi-step derivation.

\citet{sun2026typesafe} show that a constrained decision head can track the name assigned to an option rather than its rubric. \citet{cheng2026thisthatmodel} introduce this-that-model-1.0, adapted from decider-2b, and compare it on a spatial reasoning benchmark and a third-party question set with Jev results obtained from third parties. \citet{ren2026openjev} train a small open scam-call screener that reads its probability the way Jev does and compare its calibration and latency with a generative judge and a fine-tuned encoder. \citet{huang2026can} find that Jev agrees more closely with expert error counts in radiology reports than an open inference judge under matched decomposition, while a local finding-matching pipeline agrees more closely on clinically significant errors.

A second group of studies uses a decision model inside an application or a domain benchmark and reports the outcome that the application cares about. These include legal document understanding \citep{zhang2026same}, source-grounded choices in the scientific literature \citep{deng2026jev} and intent-driven service orchestration at the network edge \citep{li2026fast,li2026replacing}. Others address selective control and memory in language-model agents \citep{wu2026reflex,jiang2026jevmem}, game control \citep{ma2026jevstar}, decisions from shared visual context \citep{yu2026visual} and video quality prediction \citep{robitza2026jevqa}. One of them observed that accuracy fell as the number of options grew, mainly on authorization and risk decisions \citep{wu2026reflex}, which is the cardinality effect measured here for several decision models on the same items. Across both groups, most studies fix one decision model or one pair of models, and each writes its own prompts and option texts, so a difference between two studies mixes the model with everything else that changed.

The calibration of probabilistic predictions has a long record of its own. Proper scoring rules such as the Brier score are optimized in expectation only when a forecaster reports its actual belief \citep{brier1950verification,gneiting2007strictly}, and the reliability and resolution terms of the Brier decomposition separate calibration from the ability to discriminate \citep{murphy1973new}. Expected calibration error summarizes reliability over bins of confidence \citep{naeini2015obtaining}, and its estimate depends on the binning scheme, which motivates equal-mass bins and bias-aware estimators \citep{nixon2019measuring,kumar2019verified,roelofs2022mitigating}. Temperature scaling remains a common post-hoc baseline for calibrating modern neural classifiers \citep{guo2017calibration}. Pretrained transformers are well calibrated in domain and less so out of domain relative to their in-domain level \citep{desai2020calibration}, and the calibration of prompted language models, including whether their stated confidence tracks correctness, has its own track \citep{jiang2021how,kadavath2022language}. The scaled arm of this study follows that tradition, and the stored temperatures of several decision models raise the question of whether a temperature fitted on one option count transfers to another.

The sensitivity of language models to the surface of a prompt is well documented. Their predictions shift with the wording and ordering of in-context examples, which contextual calibration partly corrects \citep{zhao2021calibrate}, with the order of the options \citep{pezeshkpour2024large} and with formatting choices that carry no meaning \citep{sclar2024quantifying}. Multiple-choice answers also favor particular option identifiers regardless of content \citep{zheng2024large}. The option-naming test asks the same question of typed decision models, whose answer space is defined by a rubric, and the cardinality test asks how a fixed decision head scales as the option set grows.

Selective prediction and escalation supply the decision rule that acts on a probability. Chow's rule balances error against abstention \citep{chow1970optimum}, the theory of selective classification characterizes the trade-off between risk and coverage \citep{elyaniv2010foundations}, and deep networks can be given a reject option with a guaranteed risk \citep{geifman2017selective}. The maximum softmax probability is a strong baseline for detecting errors and out-of-distribution inputs \citep{hendrycks2017baseline}, and no selective method consistently beats it across in-domain, out-of-domain and adversarial language tasks \citep{varshney2022investigating}. Learning to defer trains a model together with a rule for handing a case to an expert \citep{madras2018predict,mozannar2020consistent,strong2025learning}. Cascades descend from the boosted detectors of \citet{viola2001rapid} and now route language-model calls under a budget \citep{chen2023frugalgpt}, through self-verification \citep{madaan2023automix} or through learned routers \citep{ong2025routellm}, and recent surveys organize this body of work \citep{varangotreille2026routing,moslem2026cascading}. The cascades here gate on a decision model's own top-label probability, which is the simplest rule a software gate can apply.

Benchmarking practice for machine learning models closes the review. Standardized harnesses that evaluate many models under one protocol made language models comparable across tasks \citep{liang2023holistic}. Reporting guidelines ask for the budget and variance behind each number \citep{dodge2019show}, and leakage between training and test data is a documented source of inflated results \citep{kapoor2023leakage}. Table~\ref{tab:positioning} codes the \num{pos.nstudies} closest prior evaluations against the conditions a deployed decision gate depends on, each from the study's own text. Besides the studies above, a leaderboard card scores the hosted model and an open decision model with trained classifiers on the benchmark used here as D1 \citep{typeddecisions2026dataset}, and an open reproduction sends identical requests to the hosted model and three open checkpoints \citep{robbalian2026rev}. Three public sources that appeared after that sweep are also close to this design. A preprint evaluates the hosted model zero-shot on \num{lit.deusser2026evaluating.datasets} datasets with one frozen template each and compares it with Qwen3.8-27B and Gemma-4-E4B, read through their option probabilities on identical requests \citep{deusser2026evaluating}. It reports bootstrap intervals, calibration, selective prediction and cost, but it evaluates no other decision model and no classifier trained on the same items. A community leaderboard, the Decision Index, runs \num{lit.webDecisionIndex.engines} open decision models and reproductions of Jev beside the hosted model on \num{lit.webDecisionIndex.requests} frozen requests from a panel of \num{lit.webDecisionIndex.panel} benchmarks, each model with its own native prompts \citep{web_decision_index}. It reports accuracy, calibration error and latency for far more models and tasks than this study, but it scores no classifier trained on the same items and no held-out selective risk. A developer launch of the open Clef decision models reports point scores against Jev, Kev-9B and Laya on benchmarks that include CLINC-150 and Banking77, for models post-trained on internal synthetic data \citep{web_cloudflare_clef}.

A generative comparator is the most common design, run fully in \num{pos.llm.full} of the \num{pos.nstudies} studies. None of the studies reviewed here compares several hosted and open decision models in full while also scoring a classifier trained on the same items. None measures held-out selective risk together with the rejection of out-of-scope requests, and none varies both the number and the naming of options. The benchmark reported here addresses these criteria on the same items across three task families and audits exposure to the evaluation data wherever the training data are public. It complements broad leaderboards by holding the request fixed while one condition changes at a time, so that a difference between models can be traced to labels, readout, option schema or second stage.

\begin{table}
\centering
\caption{Coverage of the evaluation criteria in prior studies of typed decision models and in this study, coded from each study's own text, with a row counting the prior studies that meet each criterion.}
\label{tab:positioning}
\input{tables/t15_positioning}
\end{table}

%% file: tables/t15_positioning.tex
\tablestyle
\setlength{\tabcolsep}{2pt}%
\providecommand{\poshead}[2]{\rotatebox{90}{\shortstack[l]{#1\\#2}\hspace{3pt}}}%
\providecommand{\posbar}[1]{\begin{tikzpicture}[baseline=-0.2ex]\pgfmathsetlengthmacro{\posw}{0.78\linewidth}\pgfmathsetmacro{\posf}{\csname cdnum@pos.#1.full\endcsname/\csname cdnum@pos.nstudies\endcsname}\pgfmathsetmacro{\posp}{\csname cdnum@pos.#1.partial\endcsname/\csname cdnum@pos.nstudies\endcsname}\fill[accent] (0,0) rectangle ({\posf*\posw},1.2ex);\fill[accent!35] ({\posf*\posw},0) rectangle ({(\posf+\posp)*\posw},1.2ex);\draw[accent!55, line width=0.3pt] (0,0) rectangle (\posw,1.2ex);\end{tikzpicture}}%
\providecommand{\poscount}[1]{\posbar{#1}\newline\num{pos.#1.full}\,/\,\num{pos.#1.partial}}%
\begin{tabular}{P{0.2080}Q{0.0720}Q{0.0720}Q{0.0720}Q{0.0720}Q{0.0720}Q{0.0720}Q{0.0720}Q{0.0720}Q{0.0720}Q{0.0720}Q{0.0720}}
\toprule
\headrow \hdtop{} & \hdgroup{two}{Scope} & \hdgroup{two}{Baselines} & \hdgroup{four}{Measures} & \hdtop{Stress} & \hdgroup{two}{Design} \\
\bandrules{2-3,4-5,6-9,10-10,11-12}
\headrow \hdsub{Study} & \hdsub{\poshead{Several models,}{hosted and open}} & \hdsub{\poshead{Several}{task families}} & \hdsub{\poshead{Trained}{classifier}} & \hdsub{\poshead{Generative}{LLM}} & \hdsub{\poshead{Held-out}{calibration}} & \hdsub{\poshead{Held-out}{selective risk}} & \hdsub{\poshead{Out-of-scope}{rejection}} & \hdsub{\poshead{Cost with}{hardware}} & \hdsub{\poshead{Option count}{and naming}} & \hdsub{\poshead{Identical}{requests}} & \hdsub{\poshead{Paired or}{clustered intervals}} \\
\midrule
\groupspan{twelve}{Multi-model evaluations}
\citet{ibrahim2026evaluating} & \symfull & \symhalf & \symhalf & \symfull & \symfull & \symhalf & \symnone & \symhalf & \symhalf & \symhalf & \symfull \\
\citet{sun2026typesafe} & \symfull & \symhalf & \symnone & \symnone & \symnone & \symnone & \symnone & \symnone & \symhalf & \symfull & \symfull \\
\citet{typeddecisions2026dataset} & \symhalf & \symhalf & \symfull & \symnone & \symhalf & \symnone & \symnone & \symhalf & \symnone & \symhalf & \symnone \\
\citet{robbalian2026rev} & \symfull & \symfull & \symnone & \symhalf & \symnone & \symnone & \symnone & \symfull & \symnone & \symfull & \symhalf \\
\citet{web_decision_index} & \symfull & \symfull & \symnone & \symhalf & \symhalf & \symnone & \symhalf & \symfull & \symhalf & \symfull & \symhalf \\
\citet{deusser2026evaluating} & \symnone & \symfull & \symnone & \symfull & \symhalf & \symhalf & \symhalf & \symfull & \symhalf & \symfull & \symhalf \\
\groupspan{twelve}{Single decision model against language models}
\citet{li2026jevasajudge} & \symnone & \symhalf & \symhalf & \symfull & \symfull & \symfull & \symhalf & \symhalf & \symhalf & \symhalf & \symfull \\
\citet{huang2026can} & \symnone & \symhalf & \symhalf & \symfull & \symfull & \symnone & \symnone & \symhalf & \symhalf & \symhalf & \symfull \\
\citet{cheng2026thisthatmodel} & \symhalf & \symfull & \symnone & \symfull & \symhalf & \symnone & \symnone & \symhalf & \symnone & \symfull & \symnone \\
\citet{ren2026openjev} & \symhalf & \symhalf & \symfull & \symfull & \symfull & \symfull & \symnone & \symfull & \symhalf & \symhalf & \symfull \\
\groupspan{twelve}{Applications}
\citet{zhang2026same} & \symnone & \symnone & \symnone & \symfull & \symnone & \symnone & \symnone & \symfull & \symnone & \symhalf & \symfull \\
\citet{deng2026jev} & \symnone & \symhalf & \symnone & \symfull & \symnone & \symnone & \symnone & \symhalf & \symnone & \symhalf & \symnone \\
\citet{rafe2026calibrated} & \symnone & \symhalf & \symnone & \symfull & \symfull & \symfull & \symhalf & \symhalf & \symnone & \symhalf & \symfull \\
\citet{li2026fast,li2026replacing} & \symnone & \symhalf & \symnone & \symfull & \symnone & \symnone & \symfull & \symfull & \symhalf & \symhalf & \symnone \\
\citet{wu2026reflex} & \symnone & \symhalf & \symnone & \symfull & \symhalf & \symhalf & \symfull & \symhalf & \symhalf & \symhalf & \symfull \\
\citet{yu2026visual} & \symhalf & \symfull & \symnone & \symfull & \symhalf & \symhalf & \symfull & \symfull & \symhalf & \symfull & \symhalf \\
\citet{jiang2026jevmem} & \symnone & \symnone & \symnone & \symfull & \symnone & \symnone & \symnone & \symhalf & \symnone & \symhalf & \symnone \\
\midrule
Prior studies, full\,/\,partial & \poscount{models} & \poscount{families} & \poscount{trained} & \poscount{llm} & \poscount{calibration} & \poscount{selective} & \poscount{outofscope} & \poscount{cost} & \poscount{stress} & \poscount{identical} & \poscount{uncertainty} \\
\midrule
This study & \symfull & \symfull & \symfull & \symfull & \symfull & \symfull & \symfull & \symfull & \symfull & \symhalf & \symfull \\
\bottomrule
\end{tabular}
\tablenote{\symfull\ full, \symhalf\ partial and \symnone\ absent, each coded from the study's own text. The first column combines several decision models with hosted and open models in one study, and the stress column combines option count with option naming or rendering. Each combined cell is full when both criteria are full, empty when both are absent and half otherwise. The two orchestration papers of Li et al. share one row with the higher code of each criterion. The coverage row counts the \num{pos.nstudies} prior rows that meet each criterion fully and partly. The supplementary evidence file gives the criteria and, for every cell, a supporting quotation or, for an absent cell without one, the search terms that found nothing.}

%% file: sections/methods.tex
\section{Methods}
\label{sec:methods}

\subsection{Problem formulation and notation}
\label{sec:formulation}

A decision model receives a state \(s\), which is text or a JSON object, together with a set of typed questions, and it returns one probability distribution per question over the options that the caller declared. A question \(q\) carries an instruction, a finite option set \(\mathcal{O}_q\) of size \(K_q\) and, for each option, a rubric sentence that describes when the option applies. Three primitive types exist. A Choice question has \(K_q\) named options, a Score question has ordered levels, and a Noul question is a yes-or-no question whose answer is the probability of yes. For every question the model returns a vector \(\mathbf{p}_q = (p_{q,1}, \dots, p_{q,K_q})\) on the probability simplex, and a program acting on the answer reads the top option \(\hat{y}_q = \arg\max_k p_{q,k}\) and its top-label probability \(c_q = \max_k p_{q,k}\). The engineering use studied here is a decision gate, in which a program accepts \(\hat{y}_q\) when \(c_q\) reaches a threshold \(\tau\) and otherwise escalates the decision to a stronger model or to a person. Table~\ref{tab:notation} lists the symbols used throughout.

\begin{table}
\centering
\caption{Mathematical notation and symbols.}
\label{tab:notation}
\tablestyle
\begin{tabular}{P{0.2}P{0.8}}
\toprule
\headrow \hd{Symbol} & \hd{Meaning} \\
\midrule
\groupspan{two}{Inputs and outputs}
\(s\) & state, the input shared by all questions of one request \\
\(q\), \(\mathcal{O}_q\), \(K_q\) & question, its declared option set, and the number of options \\
\(\mathbf{p}_q\), \(p_{q,k}\) & returned distribution and the probability of option \(k\) \\
\(\hat{y}_q\), \(c_q\) & top option and top-label probability (confidence) \\
\(y_q\) & reference label of question \(q\) \\
\(\mathbf{z}_q\) & logits behind \(\mathbf{p}_q\) for models that expose them \\
\groupspan{two}{Calibration}
\(T\), \(T_s\) & temperature, and the temperature a model applies by default \\
\(B\), \(n_b\), \(\bar c_b\), \(\bar a_b\) & number of bins, bin size, mean confidence and mean accuracy of bin \(b\) \\
\groupspan{two}{Gating and cost}
\(\tau\) & confidence threshold of a gate or cascade \\
\(e(\tau)\) & share of decisions escalated at threshold \(\tau\) \\
\(\bar k_1\), \(\bar k_2\), \(\bar k_2^{E}\) & mean cost per decision of the first and the second stage, and of the second stage on the escalated decisions \\
\(\phi(t)\), \(r(t)\) & coverage and empirical risk of the decisions accepted at threshold \(t\) \\
\(\mathrm{cov}(\alpha)\) & largest coverage whose accepted decisions have empirical risk at most \(\alpha\) \\
\bottomrule
\end{tabular}
\end{table}

Three quantities that the analysis computes follow from short identities, which are stated here and proved in Appendix~\ref{sec:proofs}. The first allows the probabilities of a model that serves a stored temperature to be compared with its probabilities at a temperature of one without running the model twice.

\begin{proposition}[Temperature recovery]
\label{prop:temperature}
Let \(\mathbf{p}^{(T)} = \mathrm{softmax}(\mathbf{z}/T)\) for logits \(\mathbf{z}\) and a temperature \(T>0\). For any other temperature \(T'>0\), \(\mathbf{p}^{(T')}\) is obtained from \(\mathbf{p}^{(T)}\) alone by raising each component to the power \(T/T'\) and renormalizing to sum to one.
\end{proposition}

The second shows that a two-stage question, which first asks for a domain and then for the intent within the gold domain, yields the accuracy of the full hierarchical policy even though the second stage is never asked within a wrongly predicted domain. The identity is exact for a model that returns the same distribution for the same request, and it is an offline estimate for a model whose repeated answers vary.

\begin{proposition}[Hierarchical accuracy]
\label{prop:hierarchical}
Let every intent belong to exactly one domain, and let a hierarchical policy answer the intent question only among the intents of the domain it predicts. If the model returns the same answer whenever it receives the same request, then the policy's answer is correct if and only if its domain answer and its intent answer within the gold domain are both correct. Its accuracy is then the mean of the product of these two indicators.
\end{proposition}

The third expresses the accuracy of a confidence-gated cascade through the two stages and states when escalation can help, which Section~\ref{sec:engineering} uses to turn the measurements into a rule for choosing a second stage.

\begin{proposition}[Cascade accuracy]
\label{prop:cascade}
Let a first stage answer each of \(N\) decisions with confidence \(c_i\) and let the decisions in \(E(\tau)=\{i : c_i<\tau\}\) be escalated to a second stage. Write \(a_{1,i}\) and \(a_{2,i}\) for the correctness of the two stages on decision \(i\), \(A_1\) and \(A_2\) for their accuracies alone and \(e(\tau)=|E(\tau)|/N\) for the escalated share. Then the cascade accuracy equals \(A_1 + N^{-1}\sum_{i\in E(\tau)}(a_{2,i}-a_{1,i})\). When \(E(\tau)\) is not empty this equals \(A_1 + e(\tau)\,\Delta_{\mathrm{esc}}(\tau)\), where \(\Delta_{\mathrm{esc}}(\tau)\) is the accuracy of the second stage minus that of the first on the escalated decisions, and the cascade exceeds \(A_1\) only if \(\Delta_{\mathrm{esc}}(\tau)>0\). When some decisions are kept, the cascade exceeds \(A_2\) only if the first stage is more accurate than the second on the kept decisions.
\end{proposition}

\subsection{Harness and experimental protocol}
\label{sec:protocol}

One harness sends every model the same request, which is the TypeSafe wire body of a shared state and a set of typed questions. The hosted model receives the request through the vendor's software development kit, and every local checkpoint receives it through an in-process adapter that implements the same route. Every model therefore receives the same semantic request, with the same state, question and option texts in the same order. The rendering into tokens is the model's own, and one adapter departs from the common option text, since this-that-model-1.0 exposes no field for an option description and receives each option key joined to its description by a colon (Section~\ref{sec:models}).

Every returned distribution is normalized into one shape before any metric is computed, so a difference in the reported numbers reflects a difference between models rather than between parsing paths. Each condition is frozen once to a JSON Lines file whose SHA-256 digest and random seed, \num{p.seed}, are recorded in a manifest, and every model reads that same file. The frozen files also let an interrupted run resume without repeating work. The harness never shortens a request, and each model applies its own input limits.

A transient service error is retried a bounded number of times with only the final outcome recorded, and a request that a model rejects is recorded as a refusal and scored as missing rather than repaired. A generative reply from which no distribution over the declared options can be read, because it holds no parseable JSON or names no declared key, is scored as a uniform distribution, so an unreadable reply counts against the model rather than leaving the scored set. The only refusals in the study are the \num{rv.fail.layaen.dtwo.refusals} requests on which Laya (English) found the full intent list longer than its option budget, and Table~\ref{tab:manifest} in Appendix~\ref{sec:appendix-per-task} lists the resolved revision, served temperature and runtime of every model. The study measures the checkpoints available on the run dates in that table, and the released harness accepts any further model that implements the same request format.

Choice questions carry neutral option identifiers \(o_1\) through \(o_K\) with every rubric sentence unchanged. A model that reads an identifier such as \texttt{billing} rather than \texttt{o1} can shift its answer toward what the word suggests, independently of the rubric. A comparison across models that lean on surface wording to different degrees is confounded unless the identifiers carry no meaning. The dataset's native option names are kept in one condition, which validates the harness against the published leaderboard row for Jev, with accuracy \num{p.anchor.jev.acc} and Brier score \num{p.anchor.jev.brier}. The dataset card also reports a teacher self-agreement of \num{p.anchor.ceiling}, which serves as a reference level for agreement with the teacher rather than as an upper bound on the accuracy of another model.

Repeated runs separate a model's own variability from a difference between models. The hosted model answers each condition \num{p.jevreps} times. The first run is the one analyzed in every table, and the other runs measure service variability. It also answers \num{p.jevretestreps} times on a subset of \num{n.retest.states} neutral workflow states, which gives a test-retest floor against which any option-naming effect is read. Every open model is checked for determinism on the same subset, since a checkpoint served with fixed weights and no sampling should return the same distribution twice.

Three probability arms are compared for every scored condition. The shipped arm is what the service or checkpoint returns by default, including any temperature \(T_s\) that it applies itself. The raw arm is the same distribution at a temperature of one. It exists only for checkpoints whose served temperature is known and whose adapter stores the unrounded distribution at a temperature of one, which is recovered by Proposition~\ref{prop:temperature} without a second forward pass. The hosted model serves two-decimal probabilities at an undisclosed temperature and the verbal readouts report verbalized probabilities, so none of them has a raw arm, and their raw cells are left empty. The scaled arm applies one further temperature fitted on data that the scored items never touch. The temperature minimizes the negative log-likelihood of the reference labels over \num{p.tgrid.n} log-spaced values from \num{p.tgrid.lo} to \num{p.tgrid.hi}. For D1 that temperature is one value per primitive type, fitted on \num{p.calibperworkflow} training-split states per workflow. For D2 and for each D3 task separately it comes from \num{p.folds}-fold cross-fitting that is disjoint by item, so every scored decision is evaluated under a temperature that never saw it.

\subsection{Measures and stress tests}
\label{sec:measures}

Accuracy is the share of decisions whose top option equals the dataset's reference label, which makes it agreement with that reference wherever the reference comes from a teacher model or from hashtags rather than from people. Accuracy is reported with a percentile bootstrap interval at the \num{p.boot.level} percent level that resamples whole states, because the questions asked about one state are not independent draws. Paired differences between models use a paired cluster bootstrap with \num{p.boot.pairedreps} replicates, a two-sided bootstrap p-value and a Holm correction across the models compared on each dataset. Calibration errors, coverages and cascade gains carry cluster-bootstrap intervals from \num{p.boot.refitreps} replicates that refit every fitted temperature or threshold inside each replicate. The multi-class Brier score and the negative log-likelihood use the full option distribution. Calibration is reported as top-label expected calibration error with \num{p.bins} equal-mass bins,
\begin{equation}
  \mathrm{ECE} = \sum_{b=1}^{B} \frac{n_b}{N}\,\bigl|\bar c_b - \bar a_b\bigr| ,
  \label{eq:ece}
\end{equation}
and the multi-class Brier score and the negative log-likelihood of the reference label are reported in Appendix~\ref{sec:appendix-per-task}. The multi-class Brier score is
\begin{equation}
  \mathrm{Brier} = \frac{1}{N}\sum_{i=1}^{N}\sum_{k=1}^{K}\bigl(p_{i,k} - y_{i,k}\bigr)^2 ,
  \label{eq:brier}
\end{equation}
where \(y_{i,k}\) is one for the reference option and zero otherwise. On D1, whose reference is a hard label together with a soft distribution \(\mathbf{s}_i\) that the dataset derives from the teacher samples, soft accuracy is the mean teacher probability of the option the model selects,
\begin{equation}
  \mathrm{SoftAcc} = \frac{1}{N}\sum_{i=1}^{N} s_{i,\hat{y}_i} .
  \label{eq:softacc}
\end{equation}
This measure rewards a choice the teacher also found plausible, whereas the Brier score compares the whole returned distribution with the hard label. Selective prediction accepts the decisions whose confidence reaches a threshold \(t\), with ties accepted together. For the accepted set \(\mathcal{A}(t)=\{i : c_i \ge t\}\), the coverage is \(\phi(t)=|\mathcal{A}(t)|/N\) and the empirical risk \(r(t)\) is the share of errors in \(\mathcal{A}(t)\). The risk-coverage curve traces \(r(t)\) against \(\phi(t)\) over the observed confidences, and the coverage at \num{p.riskone} and \num{p.riskfive} percent risk is
\begin{equation}
  \mathrm{cov}(\alpha) = \max_{t}\,\{\, \phi(t) \mid r(t) \le \alpha \,\} .
  \label{eq:covrisk}
\end{equation}
The review budget is the share of decisions that fall outside that accepted set. Computed on the scored items, \(\mathrm{cov}(\alpha)\) traces the frontier of those items. A deployed gate fixes its threshold beforehand, so the analysis also chooses the threshold on held-out data, applies it to the scored items, and reports the realized coverage and risk with a one-sided upper bound on the risk. A held-out threshold still gives no guarantee, so a third rule controls the risk by Learn-then-Test \citep{angelopoulos2021learn}. Each of \num{rc.gridsize} candidate thresholds fixed before any data are seen is tested with an exact binomial test of the hypothesis that the error rate of the decisions it accepts on the calibration data exceeds \num{rc.alpha} percent. The lowest threshold whose hypothesis is rejected after a Bonferroni correction over the candidates is applied to the scored items, so with probability at least \num{rc.confidence} percent its risk on new exchangeable decisions stays at or below \num{rc.alpha} percent. Out-of-scope detection on D2 is scored as the area under the receiver operating characteristic curve of the top-label probability.

The option-naming experiment recasts every binary question under four identifier pairs with the rubric text and order fixed. The pairs are the digits zero and one, an aligned \emph{no} and \emph{yes}, the same pair swapped against the rubric, and a pair of random five-letter strings. Each condition is scored by the share of decisions whose chosen rubric flips relative to the aligned condition and by the area under the curve of the positive-rubric probability against the reference label, with its paired-bootstrap change. All three quantities are read against the test-retest floor.

The cardinality experiment asks the same \num{n.d2k5.requests} in-scope intent utterances with \num{p.kfive}, \num{p.ktwenty}, \num{p.kfifty} and \num{p.kall} options. The option sets are nested prefixes of a fixed random permutation per item and always contain the gold intent, so the option count changes together with the particular distractors, their positions and the length of the input. Two further independent permutations separate the effect of the option count from that of one draw of distractors. A two-stage condition asks the domain and then the intent within the gold domain, and Proposition~\ref{prop:hierarchical} turns the two answers into the accuracy of the hierarchical policy, exactly for the deterministic checkpoints and as an offline estimate for the hosted model.

Latency is reported for each platform separately and is never expressed as a ratio between platforms, because the network path and the serving stack differ. The decision models answer one request at a time, and each comparator is timed on \num{p.bone.requests} requests sent one at a time, while its cost comes from the batched run that produced its answers. Costs on different platforms are compared at the stated hourly rates and with the graphics processing unit (GPU) fully occupied, with any lower occupancy stated as a sensitivity, and the cascade cost fraction expresses each cascade relative to its second stage. The hosted model's cost is its recorded input-token usage at the published price, and an open model's cost is its measured GPU time at a stated hourly rate. The cascade experiment pairs each decision model as a first stage with a second stage, sweeps the threshold \(\tau\) over the first stage's observed confidences, and reports cascade accuracy against the cost fraction of the pair,
\begin{equation}
  \mathrm{cost\ fraction}(\tau) = \frac{\bar k_1 + e(\tau)\,\bar k_2^{E}}{\bar k_2} .
  \label{eq:costfraction}
\end{equation}
Here \(\bar k_1\) and \(\bar k_2\) are the mean costs per decision of the two stages over all decisions and \(\bar k_2^{E}\) is the mean second-stage cost over the escalated decisions, all computed item by item from recorded tokens or GPU time.
The threshold that maximizes accuracy on the scored items overstates what a deployed cascade would achieve, so every reported cascade uses a threshold chosen on held-out data, namely the calibration slice for D1 and item-disjoint cross-fitting for D2 and D3.

%% file: sections/models_datasets.tex
\section{Models and datasets}
\label{sec:models}
\label{sec:datasets}

Six decision-model families are measured, one hosted and five open with eight checkpoints in all, alongside four generative language models from three developers read in eight ways, and Table~\ref{tab:models} summarizes each. Jev is accessed through the vendor's typesafe-sdk pinned to \texttt{\num{model.version}}, since the alias \texttt{jev-latest} moves without notice. It bills \num{p.price} United States dollars per million input tokens, with output tokens free, and returns probabilities on a two-decimal grid \citep{typesafe2026api,typesafe2026launch}. Laya is an encoder that scores each option at a marker token and is trained with reinforcement learning against proper scoring rules, a recipe its developers call RLCD. It has a \num{m.laya.en.params}-parameter English ModernBERT-large checkpoint with a sequence budget of \num{m.laya.en.context} tokens per question, of which \num{lb.layaen.fifty.head} hold the option text. The multilingual mmBERT-base checkpoint has \num{m.laya.ml.params} parameters and a sequence budget of \num{m.laya.ml.context} tokens, of which \num{lb.layaml.fifty.head} hold the option text \citep{laya2026card,layamultilingual2026card}. Laya applies its own temperature per question type and option count, and the table lists its nominal value.

Kev-0.8B and Kev-9B add a LoRA adapter of rank \num{m.kev.lorarank} and a pointer head to Qwen3.5, trained with cross-entropy. They serve at stored temperatures of \num{m.kev9b.temp} for Kev-9B and \num{m.kev08b.temp} for Kev-0.8B, the latter read from its checkpoint file on the day of the run \citep{kev08b2026card,kev9b2026card}. The decider-2b model fine-tunes the \num{m.decider.baseparams}-parameter Qwen3.5-2B-Base to softmax the letter logits at each answer slot, accepts about \num{m.decider.context}K tokens, and serves at a stored temperature of \num{m.decider.temp} \citep{decider2b2026card}.

The this-that-model-1.0 checkpoint, a \num{m.thisthat.params}-parameter model adapted from decider-2b, scores a hidden state against each option's label-token embedding without a built-in temperature \citep{thisthat2026card}. It exposes no request format matching the wire body, so the harness joins each option's key and description with a colon, the one place where a model receives option text that no other model sees. Laya, Kev, decider-2b and the comparator prompts also join each key to its description with a colon inside their own rendering, so the difference lies in the request field rather than in the text the model reads. A sensitivity run with the description alone measures how much this rendering matters, and Section~\ref{sec:results-accuracy} reports it. Nimble-9B, a LoRA adapter on Qwen3.5-9B, reads a single-token answer code per question. Its main branch was pinned by commit hash on the day it was updated to serve at \num{m.nimble.temp} with a context limit of \num{m.nimble.context} tokens, and the checkpoint measured here is that pinned revision \citep{nimble2026card}.

The first comparator is Qwen3-14B in its official \num{m.comparator.quantbits}-bit AWQ release, run in non-thinking mode with its chat template under vLLM \citep{yang2025qwen3,kwon2023efficient}. It returns one JSON object with its chosen option and a probability per option under structured decoding at a sampling temperature of \num{m.comparator.temp}, so its probabilities are verbalized rather than token probabilities. A larger comparator, Qwen3.6-27B in its official FP8 release, answers the neutral-key requests of all three dataset families on a Colab A100 GPU \citep{web_qwen36_27b_fp8}. It is read out in two ways after the same chat prompt, once through the JSON protocol and once through the likelihood of each option key, normalized over the options at a temperature of one.

The same checkpoint also answers every main condition in its thinking mode, which samples its reasoning at a temperature of \num{m.think.temp} with a fixed seed and up to \num{m.think.maxtokens} generated tokens before the JSON reply. Two further open models from other developers answer the same conditions with both readouts, Gemma-4-31B-it \citep{web_gemma4_31b} and Mistral-Small-24B-Instruct-2501 \citep{web_mistral_small_2501}. Both run in bfloat16 on a Colab A100 GPU with \num{p.gpumem} gigabytes of memory, without graph compilation, because the online FP8 path of the serving stack does not run on that GPU. Appendix~\ref{sec:appendix-question-texts} gives the prompts, the reporting rule and the parser.

Four conventional classifiers serve as baselines outside the decision-model interface. A BERT-base encoder is fine-tuned on the CLINC-150 training split, the same split decider-2b was trained on, and a logistic regression is fitted on sentence embeddings from BGE-small for the same split and for each further benchmark. A DeBERTa-v3-large encoder \citep{he2023debertav3} is fine-tuned on the training split of every benchmark with labels, as a cross-encoder with soft targets on typed-decisions and as a sequence classifier elsewhere, with early stopping on held-out training data. On the typed-decisions training split the same embeddings, with added word n-gram features, feed a soft-label regression per question. Training texts that match a test item after normalization are removed first, and hyperparameters are chosen on the validation split. A DeBERTa-v3 entailment model scores each option description as a hypothesis without task training. Each baseline receives one temperature fitted on held-out data.

Untuned backbones separate what fine-tuning adds to each decoder-based checkpoint from what its base model already does. Each decoder-based checkpoint is paired with the Qwen3.5 checkpoint it starts from \citep{web_qwen35_base}, a base model for Kev and decider-2b, the post-trained Qwen3.5-9B for Nimble-9B, and for this-that-model-1.0 the base of decider-2b, from which it is adapted. Each backbone reads the same content as a plain prompt without a chat template, and the likelihood of each option key gives its probabilities. All open checkpoints are pinned by resolved snapshot commit. The decision models and backbones ran on a Colab L4 GPU in eager PyTorch mode with CUDA graphs and fused kernels off, so no served temperature is compiled into a kernel. The comparators ran under vLLM, Qwen3-14B on the L4 and the three larger models on an A100.

\begin{table}
\centering
\caption{Access route, backbone, decision head, served temperature and hardware of every model. The vendor does not disclose the backbone or the head of Jev, and they are not inferred here.}
\label{tab:models}
\tablestyle
\begin{tabular}{P{0.27}P{0.1}P{0.21}P{0.17}N{0.1}P{0.15}}
\toprule
\headrow \hd{Model} & \hd{Access} & \hd{Backbone} & \hd{Head} & \hd{Served \(T\)} & \hd{Hardware} \\
\midrule
\groupspan{six}{Hosted}
\msw{jev}Jev 1.13.0 & hosted API & \cellnote{not disclosed} & \cellnote{not disclosed} & \cellnote{not disclosed} & vendor infra. \\
\groupspan{six}{Open, encoder head}
\msw{layaen}Laya (English) & local & ModernBERT-large & option-marker scorer & \num{m.laya.temp} & Colab L4 \\
\msw{layaml}Laya (multilingual) & local & mmBERT-base & option-marker scorer & \num{m.laya.temp} & Colab L4 \\
\groupspan{six}{Open, decoder head}
\msw{kevsmall}Kev-0.8B & local & Qwen3.5-0.8B-Base + LoRA & pointer head & \num{m.kev08b.temp} & Colab L4 \\
\msw{kevlarge}Kev-9B & local & Qwen3.5-9B-Base + LoRA & pointer head & \num{m.kev9b.temp} & Colab L4 \\
\msw{decider}decider-2b & local & Qwen3.5-2B-Base & letter logits & \num{m.decider.temp} & Colab L4 \\
\msw{thisthat}this-that-model-1.0 & local & decider-2b & hidden-state readout & \num{m.thisthat.temp} & Colab L4 \\
\msw{nimble}Nimble-9B & local & Qwen3.5-9B + LoRA & single-token code & \num{m.nimble.temp} & Colab L4 \\
\groupspan{six}{Generative comparators}
\msw{compopen}Qwen3-14B, verbal & local, vLLM & Qwen3-14B, \num{m.comparator.quantbits}-bit AWQ & structured JSON & \num{m.comparator.temp} & Colab L4 \\
\msw{comptwo}Qwen3.6-27B, verbal & local, vLLM & Qwen3.6-27B, FP8 & structured JSON & \num{m.comparator.temp} & Colab A100 \\
\msw{comptwoll}Qwen3.6-27B, likelihood & local, vLLM & Qwen3.6-27B, FP8 & option-key likelihood & \num{m.likelihood.temp} & Colab A100 \\
\msw{comptwothink}Qwen3.6-27B, thinking & local, vLLM & Qwen3.6-27B, FP8 & JSON after reasoning & \num{m.think.temp} & Colab A100 \\
\addlinespace[2.5pt]
\msw{compgem}Gemma-4-31B, verbal & local, vLLM & Gemma-4-31B-it, bf16 & structured JSON & \num{m.comparator.temp} & Colab A100 \\
\msw{compgemll}Gemma-4-31B, likelihood & local, vLLM & Gemma-4-31B-it, bf16 & option-key likelihood & \num{m.likelihood.temp} & Colab A100 \\
\addlinespace[2.5pt]
\msw{compmis}Mistral-Small-24B, verbal & local, vLLM & Mistral-Small-24B, bf16 & structured JSON & \num{m.comparator.temp} & Colab A100 \\
\msw{compmisll}Mistral-Small-24B, likelihood & local, vLLM & Mistral-Small-24B, bf16 & option-key likelihood & \num{m.likelihood.temp} & Colab A100 \\
\bottomrule
\end{tabular}

\end{table}

Three dataset families supply the questions, with sizes, option counts, reference labels and licenses in Table~\ref{tab:datasets}. D1, typed-decisions, is a synthetic benchmark of four workflows, agent-trace observability, customer service, invoice processing and security incidents, each posing a shared state to \num{p.questionsperstate} typed questions at once \citep{typeddecisions2026dataset}. Its test split holds \num{n.d1neutral.requests} states and \num{n.d1neutral.decisions} decisions, and its gold label is the mean of three samples from a teacher endpoint of roughly four-billion-parameter class. The card reports a teacher self-agreement of \num{p.anchor.ceiling}, the accuracy of a fresh teacher sample against a reference built from the other samples on the card's own evaluation set. The soft gold defines soft accuracy, the training split supplies the calibration slice, and the license is \num{d.d1.license}. Its option count varies by question type, from a binary Noul question to the widest Score rubric.

D2 is CLINC-150, an intent-classification benchmark with an explicit out-of-scope class \citep{larson2019evaluation}, under a \num{d.d2.license} license. The test condition draws \num{p.perintent} utterances for each of the \num{p.kall} intents, for \num{n.d2k5.decisions} in-scope decisions, plus \num{p.oos} out-of-scope utterances under the same option set, so confidence in and out of scope share one scale. The hierarchical condition asks the domain, one of \num{p.domains}, and then the intent, one of \num{p.intentsperdomain}. A further condition appends an option named out of scope to the \num{p.kall} intents, and that option becomes the reference label of the out-of-scope utterances. The reference label is otherwise the intent tag of the released dataset.

D3 draws four text-classification tasks from the computational social science collection of \citet{ziems2024can}, worded as in the replication package of a prior decision-model evaluation on that collection \citep{ibrahim2026evaluating}. The two binary tasks ask whether a Wikipedia talk-page exchange will derail into a personal attack (conversations gone awry, license \num{d.d3convgoawry.license}) and whether a named participant holds a position of power (Wikipedia corpus power, \num{d.d3wikicorpus.license}). The two multi-class tasks ask for one of six basic emotions in a social media post, with the dataset's distant hashtag label as reference (emotion, \num{d.d3emotion.license}), and which of three politeness levels a workplace request expresses (Wikipedia politeness, \num{d.d3wikipoliteness.license}). Every D3 label is the reference label of the original collection, and accuracy is measured against it.

Two further benchmarks test whether the results hold beyond these reference labels and beyond the training data of the open checkpoints. GoEmotions labels Reddit comments with \num{p.goemo.emotions} emotions by several raters each \citep{demszky2020goemotions}. The test items are the \num{n.d3goemotions.decisions} comments whose rater-agreed labels map to one of the six Ekman emotions of the dataset's own mapping, stratified by emotion, and they are asked with the question of the D3 emotion task (license \num{d.goemo.license}). It replaces the distant hashtag label of the D3 emotion task with a label people agreed on, and none of the decision models lists it among its training sets. Banking77 holds \num{d.banking.k} human-assigned intents of online-banking queries \citep{casanueva2020efficient}, of which \num{n.d2banking77.decisions} test queries, \num{p.banking.perintent} per intent, are each asked with all \num{d.banking.k} intents under neutral keys (license \num{d.banking.license}). The Kev cards list Banking77 among their training sets, whereas decider-2b and this-that-model-1.0 were trained on CLINC-150, so the two intent benchmarks reverse which checkpoints have seen the task.

\begin{table}
\centering
\caption{Size, option count, reference-label type and license of every dataset, where $n$ counts scored decisions.}
\label{tab:datasets}
\tablestyle
\begin{tabular}{P{0.28}N{0.1}N{0.08}P{0.27}P{0.27}}
\toprule
\headrow \hd{Dataset} & \hd{\(n\)} & \hd{\(K\)} & \hd{Reference} & \hd{License} \\
\midrule
\groupspan{five}{Workflow and intent}
D1 typed-decisions & \num{n.d1neutral.decisions} & \num{p.d1kmin} to \num{p.d1kmax} & teacher consensus & \num{d.d1.license} \\
D2 CLINC-150 & \num{n.d2k150.decisions} & \num{p.kall} & administrative tag & \num{d.d2.license} \\
Banking77 & \num{n.d2banking77.decisions} & \num{d.banking.k} & human-assigned intent & \num{d.banking.license} \\
\groupspan{five}{Computational social science}
D3 conv.\ gone awry & \num{n.d3convgoawry.decisions} & \num{d.d3convgoawry.k} & original collection & \num{d.d3convgoawry.license} \\
D3 wiki corpus power & \num{n.d3wikicorpus.decisions} & \num{d.d3wikicorpus.k} & original collection & \num{d.d3wikicorpus.license} \\
D3 emotion & \num{n.d3emotion.decisions} & \num{d.d3emotion.k} & distant (hashtag) & \num{d.d3emotion.license} \\
D3 wiki politeness & \num{n.d3wikipoliteness.decisions} & \num{d.d3wikipoliteness.k} & original collection & \num{d.d3wikipoliteness.license} \\
GoEmotions, Ekman & \num{n.d3goemotions.decisions} & \num{d.goemo.k} & rater agreement & \num{d.goemo.license} \\
\bottomrule
\end{tabular}

\end{table}

%% file: sections/results.tex
\section{Results}
\label{sec:results}

The results are reported in the order of the research questions. Section~\ref{sec:results-accuracy} answers RQ1 and RQ2 on accuracy, calibration and selective prediction, Section~\ref{sec:results-stress} answers RQ3 on option count and option names, and Section~\ref{sec:results-cascade} answers RQ4 on cost, latency and escalation.

\subsection{Accuracy, calibration and selective prediction}
\label{sec:results-accuracy}

The harness reproduces the published reference points before any comparison is read. On the native typed-decisions condition the hosted model reaches an accuracy of \num{e1.jev.donenative.acc} and a soft accuracy of \num{e1.jev.donenative.softacc} against leaderboard values of \num{p.anchor.jev.acc} and \num{p.anchor.jev.softacc}. The hosted model's accuracy of \num{e1.jev.toxicity.acc} on conversation derailment, \num{e1.jev.power.acc} on power and \num{e1.jev.emotion.acc} on emotion lies close to the \num{p.anchor.package.tox}, \num{p.anchor.package.pwr} and \num{p.anchor.package.emo} of the replication package of \citet{ibrahim2026evaluating}. All results below use the neutral-identifier conditions unless stated otherwise. In Table~\ref{tab:headline}, D1 is typed-decisions with neutral option identifiers, D2 is CLINC-150 with all \num{p.kall} intents, and the next four columns are the D3 tasks of conversation derailment, power, emotion and politeness. D3 F1 is the mean macro-F1 of the four tasks in percent, D1 soft is agreement with the teacher distribution, and D2 AUROC separates in-scope from out-of-scope inputs by the top probability.

\begin{table}
\centering
\caption{Accuracy and related measures of every model class under the same harness. A dash marks a condition the model refused or a condition not run.}
\label{tab:headline}
\input{tables/t03_headline}

\end{table}

Without training on the task, accuracy differs widely between the decision models (Table~\ref{tab:headline}), and among them the hosted model has the highest observed accuracy on typed-decisions and CLINC-150. On typed-decisions it reaches \num{e1.jev.doneneutral.acc}, and the nearest open decision model, Kev-9B, reaches \num{e1.kevlarge.doneneutral.acc}. Their paired difference of \num{rv.pd.kevlarge.doneneutral.diff} points has a state-clustered interval from \num{rv.pd.kevlarge.doneneutral.difflo} to \num{rv.pd.kevlarge.doneneutral.diffhi} and a Holm-adjusted p-value of \num{rv.pd.kevlarge.doneneutral.holm}, so the two are not distinguishable on this benchmark. On CLINC-150 the difference of every other decision model from the hosted model remains significant after the same correction (Table~\ref{tab:paired} in Appendix~\ref{sec:appendix-per-task}). Nimble-9B follows at \num{e1.nimble.doneneutral.acc}, the two 2B-class models at \num{e1.thisthat.doneneutral.acc} and \num{e1.decider.doneneutral.acc}, Kev-0.8B at \num{e1.kevsmall.doneneutral.acc}, and both Laya checkpoints near \num{e1.layaen.doneneutral.acc}, below the input-blind prior of \num{p.anchor.prior}. On CLINC-150 the hosted model reaches \num{e1.jev.dtwo.acc}, and decider-2b, this-that-model-1.0, Nimble-9B and Kev-9B cluster between \num{e1.kevlarge.dtwo.acc} and \num{e1.thisthat.dtwo.acc}.

The decider-2b model was trained on the training split of CLINC-150, and this-that-model-1.0 is adapted from it, so both are task-trained models on the intent benchmark rather than zero-shot comparisons. None of the \num{ov.clinc.full.train.nitems} intent test utterances occurs verbatim in that training split, \num{ov.clinc.full.train.normalized} occur after normalizing case and punctuation, and \num{ov.clinc.full.train.neardupchar} are near duplicates by character overlap. The stored temperature of decider-2b was fitted on in-task evaluation sets, and rebuilding them from the published loader places \num{ov.clinc.deciderfit.exact} of the \num{ov.clinc.deciderfit.nitems} test items in that pool \citep{decider2b2026card}. This exposure reaches one stored temperature, not the accuracy, and the scaled arm refits the temperature on the benchmark items themselves.

The rendering exception of this-that-model-1.0 matters on this benchmark. Given each option's description alone instead of its key joined to the description, the checkpoint gains \num{rp.dtwo.descvsoriginal.diff} points on CLINC-150, with an interval from \num{rp.dtwo.descvsoriginal.lo} to \num{rp.dtwo.descvsoriginal.hi}, and reaches \num{rp.dtwo.descvsoriginal.acc}. That is not significantly different from the hosted model, with a difference of \num{rp.dtwo.descvsjev.diff} points and a p-value of \num{rp.dtwo.descvsjev.p}, whereas on typed-decisions the same change moves its accuracy by \num{rp.done.descvsoriginal.diff} points. The tables keep the rendering that was fixed before any result was seen.

The generative comparators show that the hosted model's lead among the decision models does not extend to every language model without task training. Gemma-4-31B reaches \num{e1.compgem.doneneutral.acc} on typed-decisions and \num{e1.compgem.dtwo.acc} on CLINC-150 with its stated probabilities, and \num{e1.compgemll.doneneutral.acc} and \num{e1.compgemll.dtwo.acc} through its key likelihoods, as the two readouts choose the same option for nearly every decision. On typed-decisions its difference from the hosted model of \num{rv.pd.compgem.doneneutral.diff} points is not significant, with an interval from \num{rv.pd.compgem.doneneutral.difflo} to \num{rv.pd.compgem.doneneutral.diffhi}. On CLINC-150 it leads by \num{rv.pd.compgem.dtwo.diff} points, with an interval from \num{rv.pd.compgem.dtwo.difflo} to \num{rv.pd.compgem.dtwo.diffhi} and a Holm-adjusted p-value of \num{rv.pd.compgem.dtwo.holm}, and its likelihood readout leads by \num{rv.pd.compgemll.dtwo.diff} points. Qwen3.6-27B read through its key likelihoods reaches \num{e1.comptwoll.doneneutral.acc} and \num{e1.comptwoll.dtwo.acc}, level with the hosted model on both benchmarks, and in its thinking mode it reaches \num{e1.comptwothink.doneneutral.acc} and \num{e1.comptwothink.dtwo.acc}, also not significantly different. Qwen3-14B and both readouts of Mistral-Small-24B stay significantly below the hosted model on both benchmarks, Mistral-Small-24B by \num{rv.pd.compmisll.dtwo.absdiff} points on CLINC-150 even through its key likelihoods. Whether an open generative model matches a decision model therefore depends on the particular checkpoint rather than on the model class.

The verbal readout adds a source of error that the likelihood readout cannot have, because the reply has to name the declared keys in parseable JSON (Table~\ref{tab:readouts} in Appendix~\ref{sec:appendix-per-task}). Every unreadable reply is scored as a uniform distribution. Qwen3.6-27B wrote each option key followed by its description in \num{vr.comptwo.dtwo.recovered} of its \num{vr.comptwo.dtwo.n} intent replies, so its verbal accuracy of \num{e1.comptwo.dtwo.acc} on CLINC-150 rises to \num{vr.comptwo.dtwo.acclenient} when such keys are accepted. Mistral-Small-24B returned no parseable JSON for \num{vr.compmis.doneneutral.parsefailed} typed-decisions decisions and wrote keys with descriptions for \num{vr.compmis.emotion.recovered} of the \num{vr.compmis.emotion.n} emotion decisions, which leaves it at \num{e1.compmis.emotion.acc} on emotion. Gemma-4-31B and the thinking mode of Qwen3.6-27B left at most \num{vr.comptwothink.dtwo.failshare} of the decisions on any benchmark unusable. Format compliance is therefore a property of the checkpoint that has to be measured, and every table keeps the fixed parser for every model.

The social-science tasks narrow the spread, because no model separates the classes well on three of the four. Conversation derailment stays between \num{e1.decider.toxicity.acc} and \num{e1.compgem.toxicity.acc} for every model without its training labels on a balanced binary task, and power stays between \num{e1.layaml.power.acc} and \num{e1.jev.power.acc} for every decision model. Politeness favors the hosted model, which leads every open decision model by between \num{pd.nimble.politeness.absdiff} and \num{pd.layaml.politeness.absdiff} points. On emotion two open models lead by a wide margin, at \num{e1.decider.emotion.acc} for decider-2b and \num{e1.thisthat.emotion.acc} for this-that-model-1.0 against \num{e1.jev.emotion.acc} for the hosted model. The published training mixture of decider-2b, from which this-that-model-1.0 is adapted, includes the training split of the same emotion corpus. The lead persists on the \num{x.emotion.unseen} test items whose text does not occur in that split, at \num{x.decider.emotion.accunseen}, so it reflects exposure to the corpus and its labeling procedure rather than recall of items. On GoEmotions, whose items and labels these two models did not see, the lead disappears (Table~\ref{tab:newtasks}).

\begin{table}
\centering
\caption{Accuracy, macro-F1, calibration error and paired difference from Jev 1.13.0 on the two further benchmarks, GoEmotions with rater-agreed labels and Banking77 with all \num{d.banking.k} intents.}
\label{tab:newtasks}
\input{tables/t19_new_tasks}
\end{table}

The two further benchmarks test these findings on labels that people assigned and with the exposure of the intent benchmarks reversed (Table~\ref{tab:newtasks}). On GoEmotions the hosted model reaches \num{nt.jev.goemo.acc}, and decider-2b and this-that-model-1.0 reach \num{nt.decider.goemo.acc} and \num{nt.thisthat.goemo.acc}, not significantly different from it. Their lead on the D3 emotion task therefore reflects the training split of that corpus rather than a general skill with emotion. No decision model or comparator is significantly more accurate than the hosted model on GoEmotions. On Banking77, whose training split the Kev checkpoints list among their training data, Kev-9B reaches \num{nt.kevlarge.banking.acc} against \num{nt.jev.banking.acc} for the hosted model, a difference of \num{nt.kevlarge.banking.diff} points with a Holm-adjusted p-value of \num{nt.kevlarge.banking.holm}. The two models trained on CLINC-150 reach \num{nt.decider.banking.acc} and Gemma-4-31B equals the hosted model at \num{nt.compgem.banking.acc}, so on this intent set no model without its labels is significantly more accurate than the hosted model. Both Laya checkpoints stay near chance on Banking77, because its option text again exceeds their input budget. The trained classifiers keep their lead on both benchmarks, the fine-tuned DeBERTa-v3-large at \num{nt.debgoemo.goemo.acc} on GoEmotions and \num{nt.debbanking.banking.acc} on Banking77 and the embedding classifier at \num{nt.bgegoemo.goemo.acc} and \num{nt.bgebanking.banking.acc}.

When the task's own labels are available, conventional classifiers trained on them form a further model class (Table~\ref{tab:headline}, with calibration and latency in Table~\ref{tab:baselines} of Appendix~\ref{sec:appendix-per-task}). A DeBERTa-v3-large encoder fine-tuned on each training split has the highest observed accuracy on every task of the three main dataset families except power. It reaches \num{bl.debertalarged1.d1neutral.acc} on typed-decisions and \num{bl.debertalargeclinc.d2k150.acc} on CLINC-150, against \num{e1.jev.doneneutral.acc} and \num{e1.jev.dtwo.acc} for the hosted model, and separates out-of-scope requests with an area of \num{bl.debertalargeclinc.d2k150.aurocoos}. Its leads over the hosted model are significant after Holm correction on typed-decisions, CLINC-150, emotion and politeness, and on derailment and power the two do not differ significantly. Gemma-4-31B, which uses no task labels, is not significantly different from it on CLINC-150, with a difference of \num{bp.dtwo.trainedlarge.compgem.diff} points and a Holm-adjusted p-value of \num{bp.dtwo.trainedlarge.compgem.holm}.

The smaller trained classifiers are weaker but still competitive. A BERT-base encoder fine-tuned on the CLINC-150 training split reaches \num{bl.bertbaseclinc.d2k150.acc} on the intent benchmark and a logistic regression on small sentence embeddings reaches \num{bl.bgesmalllrclinc.d2k150.acc}, both above the hosted model, and the embedding classifier is better calibrated as shipped, at \num{bl.bgesmalllrclinc.d2k150.eceshipped}. On typed-decisions a set of soft-label regressions per question on the same embeddings with word n-gram features reaches \num{bl.bgesmalllrd1.d1neutral.acc}, from which the hosted model, Kev-9B and Nimble-9B do not differ significantly and which Gemma-4-31B exceeds by \num{bp.doneneutral.trained.compgem.diff} points.

Without task training the picture reverses, since a zero-shot entailment classifier reaches only \num{bl.nlidebertav3base.d2k150.acc} on the intent benchmark and \num{bl.nlidebertav3base.d1neutral.acc} on typed-decisions. Every decision model except the two Laya checkpoints is significantly more accurate than the zero-shot classifier on typed-decisions and CLINC-150, and on emotion the zero-shot classifier, at \num{bl.nlidebertav3base.d3emotion.acc}, is significantly ahead of every decision model not trained on that corpus. The ranking of the model classes therefore depends on whether task labels exist, and with them a fine-tuned encoder of under half a billion parameters has the highest observed accuracy on most tasks.

The untuned backbones show what fine-tuning adds to the base models (Table~\ref{tab:backbones} in Appendix~\ref{sec:appendix-per-task}). Scored by the likelihood of each option key, Qwen3.5-9B-Base reaches \num{bb.bbninebase.doneneutral.acc} on typed-decisions and \num{bb.bbninebase.dtwo.acc} on CLINC-150, and Kev-9B, trained from it, gains \num{bb.kevlarge.bbninebase.doneneutral.acc.diff} and \num{bb.kevlarge.bbninebase.dtwo.acc.diff} points. Nimble-9B gains \num{bb.nimble.bbnine.doneneutral.acc.diff} and \num{bb.nimble.bbnine.dtwo.acc.diff} points over the post-trained Qwen3.5-9B, and every such gain on the two benchmarks is significant after Holm correction except that of Kev-0.8B on typed-decisions. On derailment and power no fine-tuned checkpoint changes accuracy significantly, so fine-tuning adds accuracy on workflows and intents but not on the tasks where no model separates the classes well. Calibration moves in both directions, since every fine-tuned checkpoint is better calibrated as shipped than its backbone on typed-decisions and worse on derailment, and both Kev checkpoints are also worse on CLINC-150. Scoring each option's description instead of its key lowers the accuracy of all four backbones on typed-decisions and CLINC-150, for Qwen3.5-9B on CLINC-150 from \num{bd.bbnine.dtwo.acckey} to \num{bd.bbnine.dtwo.accdesc}, so the key readout does not understate them there.

\begin{table}
\centering
\caption{Top-label expected calibration error as shipped, at a served temperature of one (raw), and after one temperature fitted on a disjoint slice (scaled), with coverage at \num{p.riskfive} percent risk as shipped.}
\label{tab:calibration}
\input{tables/t04_calibration}

\end{table}

Calibration as shipped depends on the pairing of model and task more than on the model alone (Table~\ref{tab:calibration}). The hosted model is close to calibrated on typed-decisions and CLINC-150, with errors of \num{e1.jev.doneneutral.ece} and \num{e1.jev.dtwo.ece}, and far from it on the social-science tasks, where the pooled error is \num{e1.jev.dthree.ece}. The same pattern holds for most open models, and the scaled arm reduces the pooled social-science error of every decision model to between \num{e1.kevsmall.dthree.ecescaled} and \num{e1.thisthat.dthree.ecescaled}. Pooling averages tasks on which a model is over-confident with tasks on which it is under-confident, and per task the scaled error of the decision models lies between \num{rv.ts.layaml.politeness.ecescaled} and \num{rv.ts.thisthat.toxicity.ecescaled} (Table~\ref{tab:calibtasks} in Appendix~\ref{sec:appendix-per-task}). Among the decision models scaling therefore lowers the error on nearly every task, the exception being small increases for Kev-0.8B on typed-decisions and emotion, without bringing every task close to calibration.

The stored temperatures of Kev and decider-2b lower the error relative to their raw logits on the workflow and social-science tasks, but those of both Kev checkpoints raise it markedly on CLINC-150, for Kev-9B from \num{e1.kevlarge.dtwo.eceraw} at a temperature of one to \num{e1.kevlarge.dtwo.ece} as shipped. A temperature fitted on development data with few options therefore does not transfer to \num{p.kall} options. Scaling cannot create resolution, and the Laya checkpoints, whose binary answers rank the positive class below chance on typed-decisions, reach low scaled error only by moving toward uniform probabilities. The likelihood readouts of the comparators are over-confident as shipped where a model is sure of most answers, most clearly that of Gemma-4-31B. Its key likelihoods put at least \num{p.satlevel} on the top option for \num{sat.compgemll.doneneutral} of the typed-decisions decisions and \num{sat.compgemll.dtwo} of the intent decisions, and its error of \num{e1.compgemll.doneneutral.ece} on typed-decisions falls to \num{e1.compgemll.doneneutral.ecescaled} after scaling.

Selective prediction separates the models more sharply than accuracy does. At \num{p.riskfive} percent risk the hosted model accepts \num{e1.jev.dtwo.covfive} of the CLINC-150 decisions, against \num{e1.thisthat.dtwo.covfive} for this-that-model-1.0, \num{e1.kevlarge.dtwo.covfive} for Kev-9B and \num{e1.compgem.dtwo.covfive} for Gemma-4-31B, and on typed-decisions it accepts \num{e1.jev.doneneutral.covfive} against \num{e1.nimble.doneneutral.covfive} for Nimble-9B. The likelihood readout of Qwen3.6-27B accepts \num{e1.comptwoll.doneneutral.covfive} of the typed-decisions decisions at that risk and \num{rv.ho.comptwoll.doneneutral.five.cov} with a held-out threshold, against \num{rv.ho.jev.doneneutral.five.cov} for the hosted model, and rounding its probabilities to the two-decimal grid of the hosted model leaves \num{rr.comptwoll.doneneutral.covrounded}. On the social-science tasks no model holds that risk level beyond a coverage of \num{e1.compgemll.politeness.covfive}, apart from the emotion task of the two exposed models, so a review budget at that risk covers nearly every item.

\begin{figure}
\centering
\includegraphics{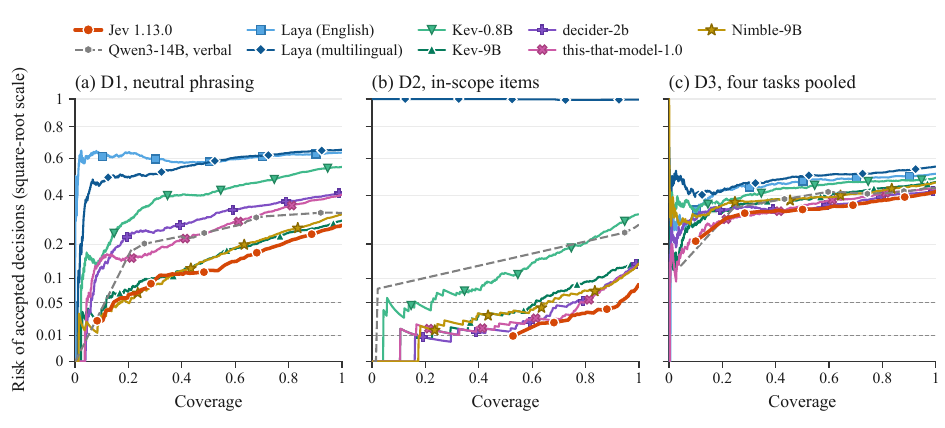}
\caption{Risk-coverage curves on D1, D2 and the pooled D3 tasks for every decision model and Qwen3-14B, with the probabilities as shipped. Horizontal lines mark \num{p.riskone} and \num{p.riskfive} percent risk.}
\label{fig:riskcov}
\end{figure}

These in-sample coverages describe the frontier of the scored items rather than a deployed gate. With the threshold chosen on held-out data, the hosted model accepts \num{rv.ho.jev.dtwo.five.cov} of the CLINC-150 decisions at a realized risk of \num{rv.ho.jev.dtwo.five.risk}, with a one-sided upper bound of \num{rv.ho.jev.dtwo.five.riskub}, and \num{rv.ho.jev.doneneutral.five.cov} of the typed-decisions decisions at \num{rv.ho.jev.doneneutral.five.risk}. The one-sided upper bound on the realized risk lies above the five percent target for every model except the stated probabilities of Gemma-4-31B on typed-decisions, whose held-out threshold accepts only \num{rv.ho.compgem.doneneutral.five.cov} of the decisions (Table~\ref{tab:selective} in Appendix~\ref{sec:appendix-per-task}). The risk-coverage curves of Figure~\ref{fig:riskcov} show where the models separate. On the in-scope CLINC-150 items the hosted model, Kev-9B, decider-2b, this-that-model-1.0 and Nimble-9B stay below ten percent risk until at least three quarters of the decisions are accepted. On the pooled social-science tasks, by contrast, every decision model's curve reaches a fifth or more of errors by a tenth of coverage, so ordering decisions by confidence buys little there.

A threshold chosen by Learn-then-Test turns the target into a guarantee and shows what the guarantee costs (Table~\ref{tab:riskcontrol} in Appendix~\ref{sec:appendix-per-task}). With a guarantee at \num{rc.confidence} percent confidence that the risk stays at or below \num{rc.alpha} percent, the hosted model accepts \num{rc.jev.dtwo.cov} of the CLINC-150 decisions, at a realized risk of \num{rc.jev.dtwo.risk}, and \num{rc.jev.doneneutral.cov} of the typed-decisions decisions. Gemma-4-31B accepts \num{rc.compgem.dtwo.cov} of the intent decisions with its stated probabilities, and the likelihood readout of Qwen3.6-27B accepts the most typed-decisions decisions, \num{rc.comptwoll.doneneutral.cov}. Most other models receive no guaranteed coverage on typed-decisions, and the saturated likelihoods of Gemma-4-31B receive none on either benchmark, because no threshold separates decisions whose top probabilities all lie near one.

The top probability separates out-of-scope intent requests from in-scope ones with an area of \num{e1.jev.oos.auroc} for the hosted model, with an interval from \num{rv.sel.jev.oos.auroclo} to \num{rv.sel.jev.oos.aurochi}, but a good ranking does not make one threshold serve both purposes. The gate analysis draws its folds over all \num{ov.clinc.full.train.nitems} utterances and therefore finds a slightly different threshold. At that held-out threshold, which keeps the in-scope risk near five percent, the hosted model accepts \num{rv.gate.jev.far} of the out-of-scope requests, against \num{rv.gate.kevlarge.far} for Kev-9B and \num{rv.gate.nimble.far} for Nimble-9B, which in turn accept fewer in-scope requests, \num{rv.gate.kevlarge.cov} and \num{rv.gate.nimble.cov} against \num{rv.gate.jev.cov}. The likelihood readout of Qwen3.6-27B accepts \num{rv.gate.comptwoll.far} of the out-of-scope requests at an in-scope coverage of \num{rv.gate.comptwoll.cov}, and Gemma-4-31B accepts \num{rv.gate.compgem.far} with its stated probabilities and \num{rv.gate.compgemll.far} with its saturated key likelihoods.

An explicit out-of-scope option changes this picture (Table~\ref{tab:oosoption} in Appendix~\ref{sec:appendix-per-task}). The hosted model chooses it for \num{oo.jev.oosrecall} of the out-of-scope requests and \num{oo.jev.falsereject} of the in-scope ones, and with a threshold chosen by the same held-out rule its false acceptance falls to \num{oo.jev.gfar} while it still routes \num{oo.jev.gcov} of the in-scope requests. Qwen3.6-27B chooses the option most often, for \num{oo.comptwo.oosrecall} of the out-of-scope requests with its stated probabilities and \num{oo.comptwoll.oosrecall} through its key likelihoods, whose gated false acceptance falls to \num{oo.comptwoll.gfar}. Gemma-4-31B and the thinking mode of Qwen3.6-27B choose it for \num{oo.compgem.oosrecall} and \num{oo.comptwothink.oosrecall}, whereas Laya (multilingual), Kev-0.8B, decider-2b and this-that-model-1.0 choose it for at most \num{oo.thisthat.oosrecall} of the out-of-scope requests and Qwen3-14B and Mistral-Small-24B for at most \num{oo.compmis.oosrecall}.

\subsection{Stress tests on option count and option names}
\label{sec:results-stress}

\begin{figure}
\centering
\includegraphics{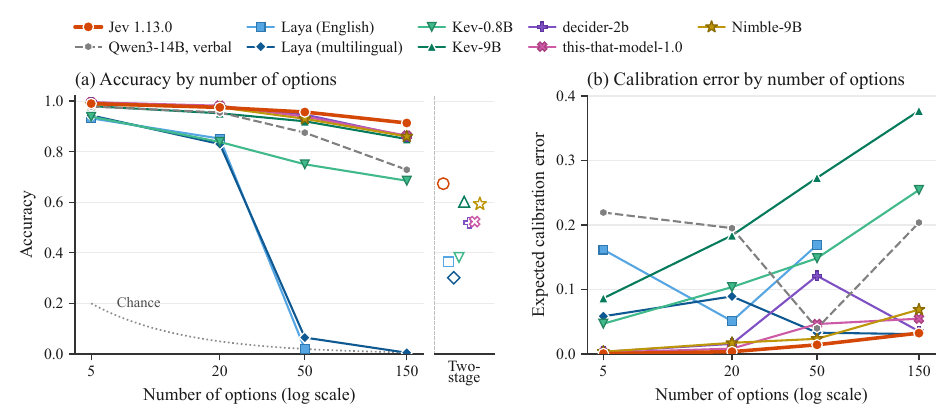}
\caption{Accuracy (a) and top-label calibration error (b) on the same \num{n.d2k5.requests} CLINC-150 utterances as the number of nested options grows. Hollow markers show the accuracy of the two-stage condition.}
\label{fig:cardinality}
\end{figure}

\begin{table}
\centering
\caption{Cardinality sweep on the same CLINC-150 utterances with the two-stage accuracy and the drop from the smallest to the largest option set. A dash marks a refused condition or one not run.}
\label{tab:cardinality}
\input{tables/t05_cardinality}
\end{table}

The cardinality sweep varies the number of options on the same utterances and shows two behaviors (Figure~\ref{fig:cardinality}). Six of the eight checkpoints lose gradually, the hosted model from \num{e3.jev.five.acc} with \num{p.kfive} options to \num{e3.jev.all.acc} with \num{p.kall}, and Kev-9B from \num{e3.kevlarge.five.acc} to \num{e3.kevlarge.all.acc}. Both Laya checkpoints collapse between \num{p.ktwenty} and \num{p.kfifty} options, the English one from \num{e3.layaen.twenty.acc} to \num{e3.layaen.fifty.acc}, and the English checkpoint rejects the full label set outright. Replaying Laya's own sequence builder with its tokenizer shows why. With \num{p.kfifty} options the option text of the English checkpoint needs \num{lb.layaen.fifty.opttokens} tokens against a head budget of \num{lb.layaen.fifty.head}, so every option is cut to its key and no description survives. With \num{p.kall} options \num{lb.layaen.all.dropped} options fall past the limit of \num{lb.layaen.all.maxlen} tokens and the checkpoint refuses the question. Without the descriptions both checkpoints choose largely by list position, and the English one picks from the first fifth of the list for \num{rv.card.layaen.fifty.predqone} of the items.

The two-stage alternative costs accuracy for every model that accepts the full label set except the Laya checkpoints, because the domain question is harder, and the hosted model reaches \num{e3.jev.hier.acc} with a domain accuracy of \num{e3.jev.hier.domacc}. For Laya the two stages restore part of the loss, at \num{e3.layaml.hier.acc} for the multilingual checkpoint against \num{e3.layaml.all.acc} with all intents at once. The calibration error with all intents in Table~\ref{tab:cardinality} repeats the effect of the stored temperatures noted above, at \num{e3.kevsmall.all.ece} for Kev-0.8B and \num{e3.kevlarge.all.ece} for Kev-9B against \num{e3.jev.all.ece} for the hosted model. A model that keeps its accuracy with many options can therefore still lose the meaning of its confidence. Two further random permutations of the distractors leave the hosted model's accuracy within \num{e3p.jev.fifty.accrange} of the original at every option count, and they change the accuracy of every other model by at most \num{e3p.kevsmall.twenty.accrange}, apart from Laya (English) with \num{p.ktwenty} options at \num{e3p.layaen.twenty.accrange}. Among the comparators Gemma-4-31B loses least, from \num{e3.compgem.five.acc} to \num{e3.compgem.all.acc}, and Qwen3.6-27B read through its key likelihoods falls from \num{e3.comptwoll.five.acc} to \num{e3.comptwoll.all.acc}.

\begin{figure}
\centering
\includegraphics{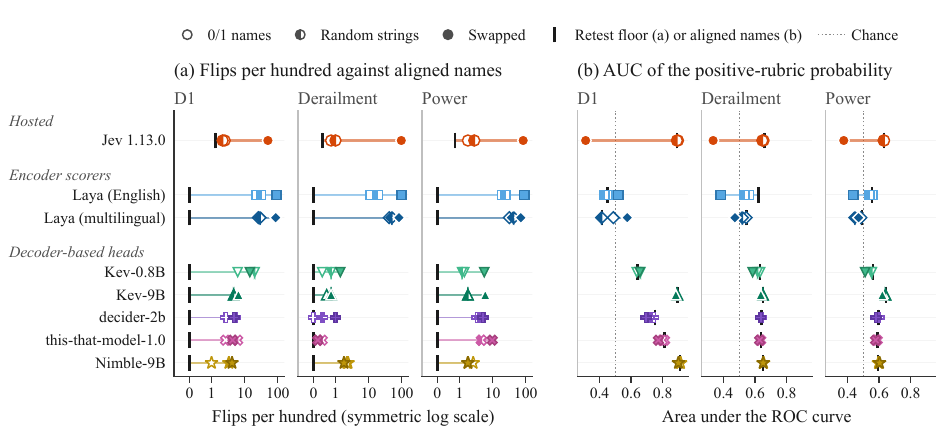}
\caption{Option-name conditions on the binary questions of D1 and the two binary D3 tasks with every rubric held fixed, as flips per hundred decisions (a) and the area under the curve (b).}
\label{fig:names}
\end{figure}

Attaching the words yes and no to the opposite rubrics moves the hosted model and both Laya checkpoints strongly, Kev-0.8B moderately, and the four larger decoder-based checkpoints little (Figure~\ref{fig:names}). On the binary questions of typed-decisions the swapped names flip \num{e2.jev.done.kswap.flips} answers per hundred for the hosted model, with a state-clustered interval from \num{rv.nm.jev.done.kswap.flipslo} to \num{rv.nm.jev.done.kswap.flipshi} and a test-retest floor of \num{e2.jev.done.floor}. Its area under the curve falls from \num{e2.jev.done.kny.auc} to \num{e2.jev.done.kswap.auc}, a change with an interval from \num{rv.nm.jev.done.kswap.dauclo} to \num{rv.nm.jev.done.kswap.dauchi}. Its flip rate reaches \num{e2.jev.toxicity.kswap.flips} on conversation derailment and \num{e2.jev.power.kswap.flips} on power, against measured floors of \num{rv.nm.jev.toxicity.floor} and \num{rv.nm.jev.power.floor}. The open checkpoints returned identical distributions for repeated requests on all three binary sets, so their floor is zero.

Laya flips \num{e2.layaen.done.kswap.flips} answers per hundred on the same questions and also moves under the digit and random-string conditions, at \num{e2.layaen.done.k01.flips} and \num{e2.layaen.done.krand.flips}. Kev-0.8B flips \num{e2.kevsmall.done.kswap.flips} answers per hundred under the swap and \num{e2.kevsmall.done.krand.flips} under random strings. Kev-9B, Nimble-9B, decider-2b and this-that-model-1.0 flip between \num{e2.nimble.done.kswap.flips} and \num{e2.kevlarge.done.kswap.flips} answers per hundred under the swap, and their areas under the curve change by at most \num{e2.decider.done.kswap.absdauc}. For the hosted model, digits and random strings stay within \num{e2.jev.neutralnames.maxflips} flips per hundred of the aligned condition on all three binary sets, so neutral identifiers reduce the measured effect for the hosted model to near its retest floor.

On the binary social-science tasks the untuned backbones show that this robustness is acquired in training. Under the swap the base models flip up to \num{bb.bbtwobase.toxicity.kswap.flips} derailment answers per hundred and the post-trained Qwen3.5-9B only \num{bb.bbnine.toxicity.kswap.flips}, so general post-training supplies part of it, and every decoder-based checkpoint flips significantly fewer than the backbone it adapts on both binary social-science tasks. On typed-decisions Kev-9B and Nimble-9B change the flip rate of their backbones by \num{bb.kevlarge.bbninebase.donekswap.flips.diff} and \num{bb.nimble.bbnine.donekswap.flips.diff} per hundred, whereas decider-2b and this-that-model-1.0 stay at the low rate of their backbone and Kev-0.8B flips more than its own under digits and random strings.

The generative comparators are more sensitive to the names than the decoder-based decision models. Under the swap Gemma-4-31B flips \num{e2.compgem.done.kswap.flips} typed-decisions answers per hundred, and the readouts of Qwen3.6-27B and Mistral-Small-24B flip between \num{e2.compmis.done.kswap.flips} and \num{e2.comptwoll.done.kswap.flips}, against at most \num{e2.decoderheads.kswap.maxflips} for the four larger decoder-based checkpoints on any binary set. Instruction tuning alone therefore does not supply the robustness that the decision heads acquire. Table~\ref{tab:names} in Appendix~\ref{sec:appendix-per-task} lists every condition on each binary set.

\subsection{Cost, latency and escalation}
\label{sec:results-cascade}

\begin{table}
\centering
\caption{Latency per request and cost per thousand decisions on typed-decisions, with the hosted model measured through its API and the open models on Colab GPUs at assumed hourly prices.}
\label{tab:cost}
\input{tables/t07_cost}
\end{table}

Latency and cost are measured on each platform separately (Table~\ref{tab:cost} and Figure~\ref{fig:cost}). The hosted model answers a typed-decisions request at a median of \num{e5.jev.d1neutral.pfifty} milliseconds at the client and costs \num{e5.jev.d1neutral.usd} US dollars per thousand decisions at its list price. From a cloud runtime as a second client location, the first \num{e5.jev.cloud.n} of these requests took a median of \num{e5.jev.cloud.pfifty} milliseconds against \num{e5.jev.cloudlocal.pfifty} for the same requests from the first location, so the hosted latency depends on the network path. On a Colab L4 with one request at a time, the median ranges from \num{e5.layaml.d1neutral.pfifty} milliseconds for Laya (multilingual) to \num{e5.nimble.d1neutral.pfifty} for Nimble-9B, which scores each question in its own pass.

At an assumed \num{p.gpuhourcost} US dollars per GPU hour, the cost per thousand decisions ranges from \num{e5.layaml.d1neutral.usd} to \num{e5.nimble.d1neutral.usd} US dollars, so the 9B checkpoints and decider-2b cost more per decision than the hosted list price at this batch size. The Laya checkpoints, Kev-0.8B and this-that-model-1.0 have the lowest cost and median latency among the open models, with this-that-model-1.0 at \num{e5.thisthat.d1neutral.pfifty} milliseconds and \num{e5.thisthat.d1neutral.usd} US dollars per thousand decisions, below the hosted list price. The this-that-model-1.0 cost stays below the hosted list price up to a price of \num{rv.cost.thisthat.doneneutral.breakeven} US dollars per GPU hour with the GPU fully occupied, and at a quarter of that occupancy it rises to \num{rv.cost.thisthat.doneneutral.usdquarter} US dollars per thousand decisions, about the hosted price.

The comparators cost more per decision than the hosted list price on typed-decisions. With batched scoring on the A100, the key likelihoods of Mistral-Small-24B cost \num{e5.compmisll.d1neutral.usd} and those of Gemma-4-31B \num{e5.compgemll.d1neutral.usd} US dollars per thousand decisions, the stated probabilities of Gemma-4-31B \num{e5.compgem.d1neutral.usd}, and the thinking mode of Qwen3.6-27B \num{e5.comptwothink.d1neutral.usd}. On CLINC-150, whose \num{p.kall} options lengthen every request, the key likelihoods of Mistral-Small-24B cost \num{e5.compmisll.d2k150.usd} against \num{e5.jev.d2k150.usd} for the hosted model, at an accuracy of \num{e1.compmisll.dtwo.acc}, and Gemma-4-31B costs \num{e5.compgem.d2k150.usd}. One request at a time and without graph compilation, the key likelihoods answer a typed-decisions request at a median of \num{e5.compmisllbone.d1neutral.pfifty} milliseconds for Mistral-Small-24B and \num{e5.compgemllbone.d1neutral.pfifty} for Gemma-4-31B. The stated probabilities of Gemma-4-31B take \num{e5.compgembone.d1neutral.pfifty} milliseconds, because every question of a request is a separate generated reply.

\begin{table}
\centering
\caption{Confidence-gated cascades on D1 and D2 with held-out thresholds and the cost fraction of Eq.~\eqref{eq:costfraction}. The upper block of each panel escalates to Gemma-4-31B and the lower block to Jev 1.13.0.}
\label{tab:cascade}
\input{tables/t08_cascade}
\end{table}

\begin{figure}
\centering
\includegraphics{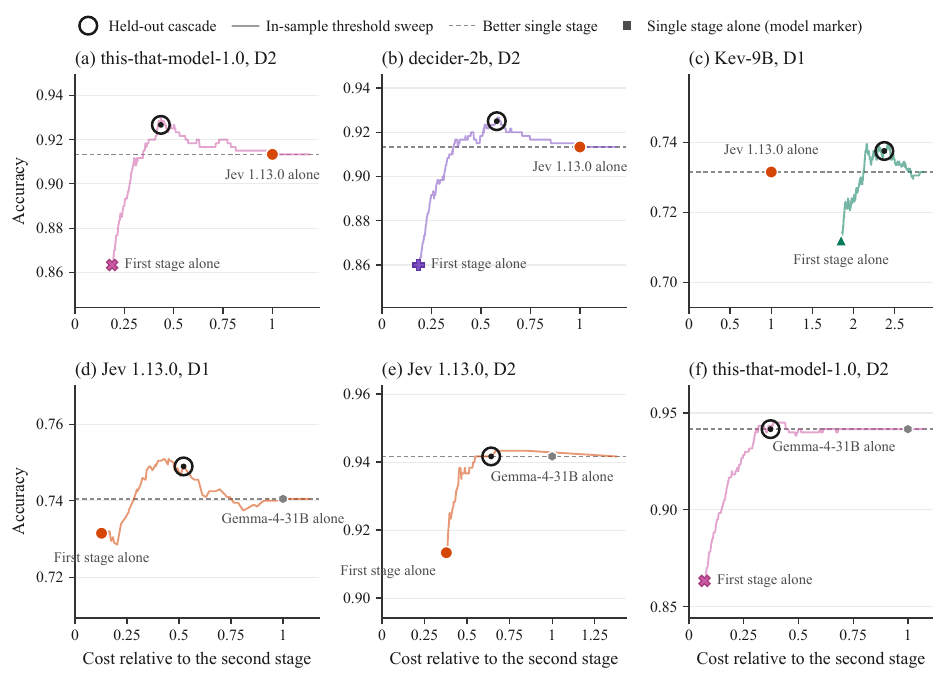}
\caption{Accuracy against cost fraction for six confidence-gated cascades, with the in-sample threshold sweep as a line, the out-of-fold cascade ringed and the better single stage dashed.}
\label{fig:cascade}
\end{figure}

Confidence-gated escalation adds accuracy only where the two stages differ in what they get right, and its cost depends on the price of the second stage (Table~\ref{tab:cascade} and Figure~\ref{fig:cascade}). Each threshold is chosen out of fold as Section~\ref{sec:methods} describes, with ties broken toward less escalation, and costs are computed per item from the recorded tokens and GPU time of each stage. On CLINC-150, this-that-model-1.0 escalates \num{rv.cas.thisthat.jev.dtwo.esc} of its decisions to the hosted model and reaches \num{rv.cas.thisthat.jev.dtwo.acc}, against \num{e1.jev.dtwo.acc} for the hosted model alone, at \num{rv.cas.thisthat.jev.dtwo.costfrac} of the hosted cost. Its gain of \num{rv.cas.thisthat.jev.dtwo.gain} points, or \num{rv.cas.thisthat.jev.dtwo.gaindec} more correct decisions out of \num{rv.cas.thisthat.jev.dtwo.n}, has an interval from \num{rv.cas.thisthat.jev.dtwo.gainlo} to \num{rv.cas.thisthat.jev.dtwo.gainhi} and a one-sided bootstrap p-value of \num{rv.cas.thisthat.jev.dtwo.gainp}, so the cascade matches the hosted model at under half its cost rather than exceeding it. The decider-2b model behaves alike, with a gain of \num{rv.cas.decider.jev.dtwo.gain} points from \num{rv.cas.decider.jev.dtwo.gainlo} to \num{rv.cas.decider.jev.dtwo.gainhi} at \num{rv.cas.decider.jev.dtwo.costfrac} of the hosted cost, whereas Kev-9B, which was not trained on the benchmark, escalates \num{rv.cas.kevlarge.jev.dtwo.esc} of its decisions and costs \num{rv.cas.kevlarge.jev.dtwo.costfrac} times the hosted model without gaining accuracy.

Escalating to Gemma-4-31B, the most accurate model on CLINC-150 among those without task labels, moves the same trade-off to a higher accuracy. The same this-that-model-1.0 escalates \num{rv.cas.thisthat.compgem.dtwo.esc} of its intent decisions to it and reaches \num{rv.cas.thisthat.compgem.dtwo.acc}, the accuracy of Gemma-4-31B alone, at \num{rv.cas.thisthat.compgem.dtwo.costfrac} of its cost. That is \num{rv.cas.thisthat.compgem.dtwo.usd} US dollars per thousand decisions, against \num{e5.jev.d2k150.usd} for the hosted model alone and \num{e5.compgem.d2k150.usd} for Gemma-4-31B alone. With the hosted model as first stage, the cascade to Gemma-4-31B escalates \num{rv.cas.jev.compgem.dtwo.esc} of the intent decisions and also reaches \num{rv.cas.jev.compgem.dtwo.acc}, at \num{rv.cas.jev.compgem.dtwo.usd} US dollars per thousand decisions. On typed-decisions the same cascade reaches \num{rv.cas.jev.compgem.doneneutral.acc}, a gain of \num{rv.cas.jev.compgem.doneneutral.gain} points over Gemma-4-31B alone whose interval from \num{rv.cas.jev.compgem.doneneutral.gainlo} to \num{rv.cas.jev.compgem.doneneutral.gainhi} includes zero.

The cascades to the two Qwen comparators are listed in Table~\ref{tab:cascadeqwen} of Appendix~\ref{sec:appendix-per-task}. Among them the only positive gain whose interval excludes zero on typed-decisions is that of Nimble-9B escalating to Qwen3-14B, \num{rv.cas.nimble.compopen.doneneutral.gain} points at \num{rv.cas.nimble.compopen.doneneutral.costfrac} times the cost of Qwen3-14B alone, and its cascade accuracy of \num{rv.cas.nimble.compopen.doneneutral.acc} remains below that of the hosted model alone. On CLINC-150 the one such gain joins two weak stages, Kev-0.8B and the verbal readout of Qwen3.6-27B, which together reach \num{rv.cas.kevsmall.comptwo.dtwo.acc}, \num{rv.cas.kevsmall.comptwo.dtwo.gain} points above the better of the two and still far below the hosted model. Proposition~\ref{prop:cascade} explains these outcomes, because the gain of a cascade over its first stage is the escalated share times the accuracy difference on the escalated decisions. The intent-trained models escalate between a quarter and two fifths of their decisions, and a second stage that is more accurate on those decisions corrects enough of them to reach its own accuracy while answering only the escalated share.

%% file: tables/t03_headline.tex
\tablestyle
\begin{tabular}{P{0.2850}N{0.0794}N{0.0794}N{0.0794}N{0.0794}N{0.0794}N{0.0794}N{0.0794}N{0.0794}N{0.0794}}
\toprule
\headrow \hdtop{} & \hdgroup{two}{D1 typed-decisions} & \hdgroup{two}{D2 CLINC-150} & \hdgroup{five}{D3 social science tasks} \\
\bandrules{2-3,4-5,6-10}
\headrow \hdsub{Model} & \hdsub{Acc.} & \hdsub{Soft} & \hdsub{Acc.} & \hdsub{AUROC} & \hdsub{Derail.} & \hdsub{Power} & \hdsub{Emotion} & \hdsub{Polite.} & \hdsub{F1} \\
\midrule
\tabhead
\groupspan{ten}{Hosted}
\msw{jev}Jev 1.13.0 & \num{e1.jev.doneneutral.acc} & \num{e1.jev.doneneutral.softacc} & \num{e1.jev.dtwo.acc} & \num{e1.jev.oos.auroc} & \num{e1.jev.toxicity.acc} & \num{e1.jev.power.acc} & \num{e1.jev.emotion.acc} & \num{e1.jev.politeness.acc} & \num{e1.jev.dthree.macrofone} \\
\groupspan{ten}{Open, encoder head}
\msw{layaen}Laya (English) & \num{e1.layaen.doneneutral.acc} & \num{e1.layaen.doneneutral.softacc} & -- & -- & \num{e1.layaen.toxicity.acc} & \num{e1.layaen.power.acc} & \num{e1.layaen.emotion.acc} & \num{e1.layaen.politeness.acc} & \num{e1.layaen.dthree.macrofone} \\
\msw{layaml}Laya (multilingual) & \num{e1.layaml.doneneutral.acc} & \num{e1.layaml.doneneutral.softacc} & \num{e1.layaml.dtwo.acc} & \num{e1.layaml.oos.auroc} & \num{e1.layaml.toxicity.acc} & \num{e1.layaml.power.acc} & \num{e1.layaml.emotion.acc} & \num{e1.layaml.politeness.acc} & \num{e1.layaml.dthree.macrofone} \\
\groupspan{ten}{Open, decoder head}
\msw{kevsmall}Kev-0.8B & \num{e1.kevsmall.doneneutral.acc} & \num{e1.kevsmall.doneneutral.softacc} & \num{e1.kevsmall.dtwo.acc} & \num{e1.kevsmall.oos.auroc} & \num{e1.kevsmall.toxicity.acc} & \num{e1.kevsmall.power.acc} & \num{e1.kevsmall.emotion.acc} & \num{e1.kevsmall.politeness.acc} & \num{e1.kevsmall.dthree.macrofone} \\
\msw{kevlarge}Kev-9B & \num{e1.kevlarge.doneneutral.acc} & \num{e1.kevlarge.doneneutral.softacc} & \num{e1.kevlarge.dtwo.acc} & \num{e1.kevlarge.oos.auroc} & \num{e1.kevlarge.toxicity.acc} & \num{e1.kevlarge.power.acc} & \num{e1.kevlarge.emotion.acc} & \num{e1.kevlarge.politeness.acc} & \num{e1.kevlarge.dthree.macrofone} \\
\msw{decider}decider-2b & \num{e1.decider.doneneutral.acc} & \num{e1.decider.doneneutral.softacc} & \num{e1.decider.dtwo.acc} & \num{e1.decider.oos.auroc} & \num{e1.decider.toxicity.acc} & \num{e1.decider.power.acc} & \num{e1.decider.emotion.acc} & \num{e1.decider.politeness.acc} & \num{e1.decider.dthree.macrofone} \\
\msw{thisthat}this-that-model-1.0 & \num{e1.thisthat.doneneutral.acc} & \num{e1.thisthat.doneneutral.softacc} & \num{e1.thisthat.dtwo.acc} & \num{e1.thisthat.oos.auroc} & \num{e1.thisthat.toxicity.acc} & \num{e1.thisthat.power.acc} & \num{e1.thisthat.emotion.acc} & \num{e1.thisthat.politeness.acc} & \num{e1.thisthat.dthree.macrofone} \\
\msw{nimble}Nimble-9B & \num{e1.nimble.doneneutral.acc} & \num{e1.nimble.doneneutral.softacc} & \num{e1.nimble.dtwo.acc} & \num{e1.nimble.oos.auroc} & \num{e1.nimble.toxicity.acc} & \num{e1.nimble.power.acc} & \num{e1.nimble.emotion.acc} & \num{e1.nimble.politeness.acc} & \num{e1.nimble.dthree.macrofone} \\
\groupspan{ten}{Generative comparators}
\msw{compopen}Qwen3-14B, verbal & \num{e1.compopen.doneneutral.acc} & \num{e1.compopen.doneneutral.softacc} & \num{e1.compopen.dtwo.acc} & \num{e1.compopen.oos.auroc} & \num{e1.compopen.toxicity.acc} & \num{e1.compopen.power.acc} & \num{e1.compopen.emotion.acc} & \num{e1.compopen.politeness.acc} & \num{e1.compopen.dthree.macrofone} \\
\addlinespace[2.5pt]
\msw{comptwo}Qwen3.6-27B, verbal & \num{e1.comptwo.doneneutral.acc} & \num{e1.comptwo.doneneutral.softacc} & \num{e1.comptwo.dtwo.acc} & \num{e1.comptwo.oos.auroc} & \num{e1.comptwo.toxicity.acc} & \num{e1.comptwo.power.acc} & \num{e1.comptwo.emotion.acc} & \num{e1.comptwo.politeness.acc} & \num{e1.comptwo.dthree.macrofone} \\
\msw{comptwoll}Qwen3.6-27B, likelihood & \num{e1.comptwoll.doneneutral.acc} & \num{e1.comptwoll.doneneutral.softacc} & \num{e1.comptwoll.dtwo.acc} & \num{e1.comptwoll.oos.auroc} & \num{e1.comptwoll.toxicity.acc} & \num{e1.comptwoll.power.acc} & \num{e1.comptwoll.emotion.acc} & \num{e1.comptwoll.politeness.acc} & \num{e1.comptwoll.dthree.macrofone} \\
\msw{comptwothink}Qwen3.6-27B, thinking & \num{e1.comptwothink.doneneutral.acc} & \num{e1.comptwothink.doneneutral.softacc} & \num{e1.comptwothink.dtwo.acc} & \num{e1.comptwothink.oos.auroc} & \num{e1.comptwothink.toxicity.acc} & \best{\num{e1.comptwothink.power.acc}} & \num{e1.comptwothink.emotion.acc} & \num{e1.comptwothink.politeness.acc} & \num{e1.comptwothink.dthree.macrofone} \\
\addlinespace[2.5pt]
\msw{compgem}Gemma-4-31B, verbal & \num{e1.compgem.doneneutral.acc} & \num{e1.compgem.doneneutral.softacc} & \num{e1.compgem.dtwo.acc} & \num{e1.compgem.oos.auroc} & \num{e1.compgem.toxicity.acc} & \num{e1.compgem.power.acc} & \num{e1.compgem.emotion.acc} & \num{e1.compgem.politeness.acc} & \num{e1.compgem.dthree.macrofone} \\
\msw{compgemll}Gemma-4-31B, likelihood & \num{e1.compgemll.doneneutral.acc} & \num{e1.compgemll.doneneutral.softacc} & \num{e1.compgemll.dtwo.acc} & \num{e1.compgemll.oos.auroc} & \num{e1.compgemll.toxicity.acc} & \num{e1.compgemll.power.acc} & \num{e1.compgemll.emotion.acc} & \num{e1.compgemll.politeness.acc} & \num{e1.compgemll.dthree.macrofone} \\
\addlinespace[2.5pt]
\msw{compmis}Mistral-Small-24B, verbal & \num{e1.compmis.doneneutral.acc} & \num{e1.compmis.doneneutral.softacc} & \num{e1.compmis.dtwo.acc} & \num{e1.compmis.oos.auroc} & \num{e1.compmis.toxicity.acc} & \num{e1.compmis.power.acc} & \num{e1.compmis.emotion.acc} & \num{e1.compmis.politeness.acc} & \num{e1.compmis.dthree.macrofone} \\
\msw{compmisll}Mistral-Small-24B, likelihood & \num{e1.compmisll.doneneutral.acc} & \num{e1.compmisll.doneneutral.softacc} & \num{e1.compmisll.dtwo.acc} & \num{e1.compmisll.oos.auroc} & \num{e1.compmisll.toxicity.acc} & \num{e1.compmisll.power.acc} & \num{e1.compmisll.emotion.acc} & \num{e1.compmisll.politeness.acc} & \num{e1.compmisll.dthree.macrofone} \\
\groupspan{ten}{Classifier trained on each task's own labels}
BGE-small, trained per task & \num{bl.bgesmalllrd1.d1neutral.acc} & \num{bl.bgesmalllrd1.d1neutral.softacc} & \num{bl.bgesmalllrclinc.d2k150.acc} & \num{bl.bgesmalllrclinc.d2k150.aurocoos} & \num{bl.bgesmalllrd3.d3convgoawry.acc} & \num{bl.bgesmalllrd3.d3wikicorpus.acc} & \num{bl.bgesmalllrd3.d3emotion.acc} & \num{bl.bgesmalllrd3.d3wikipoliteness.acc} & \num{bl.bgesmalllrd3.dthree.macrofone} \\
BERT-base, fine-tuned & -- & -- & \num{bl.bertbaseclinc.d2k150.acc} & \num{bl.bertbaseclinc.d2k150.aurocoos} & -- & -- & -- & -- & -- \\*
DeBERTa-v3-large, fine-tuned & \best{\num{bl.debertalarged1.d1neutral.acc}} & \best{\num{bl.debertalarged1.d1neutral.softacc}} & \best{\num{bl.debertalargeclinc.d2k150.acc}} & \best{\num{bl.debertalargeclinc.d2k150.aurocoos}} & \best{\num{bl.debertalarged3.d3convgoawry.acc}} & \num{bl.debertalarged3.d3wikicorpus.acc} & \best{\num{bl.debertalarged3.d3emotion.acc}} & \best{\num{bl.debertalarged3.d3wikipoliteness.acc}} & \best{\num{bl.debertalarged3.dthree.macrofone}} \\*
\groupspan{ten}{Zero-shot classifier}
DeBERTa-v3 NLI, zero-shot & \num{bl.nlidebertav3base.d1neutral.acc} & \num{bl.nlidebertav3base.d1neutral.softacc} & \num{bl.nlidebertav3base.d2k150.acc} & \num{bl.nlidebertav3base.d2k150.aurocoos} & \num{bl.nlidebertav3base.d3convgoawry.acc} & \num{bl.nlidebertav3base.d3wikicorpus.acc} & \num{bl.nlidebertav3base.d3emotion.acc} & \num{bl.nlidebertav3base.d3wikipoliteness.acc} & \num{bl.nlidebertav3base.dthree.macrofone} \\*
\bottomrule
\end{tabular}
\tablenote{Acc.\ is accuracy, Soft is agreement with the teacher distribution, AUROC separates in-scope from out-of-scope inputs, and F1 is the mean macro-F1 of the four D3 tasks in percent. The trained classifiers use labeled data of each task disjoint from its test items, and the decision models and the comparators use no task training except where Section~\ref{sec:models} states it. Bold marks the best value in each column over all rows, with ties bolded.}

%% file: tables/t19_new_tasks.tex
\tablestyle
\begin{tabular}{P{0.2850}N{0.0894}N{0.0894}N{0.0894}N{0.0894}N{0.0894}N{0.0894}N{0.0894}N{0.0894}}
\toprule
\headrow \hdtop{} & \hdgroup{four}{GoEmotions, Ekman emotions} & \hdgroup{four}{Banking77 intents} \\
\bandrules{2-5,6-9}
\headrow \hdsub{Model} & \hdsub{Acc.} & \hdsub{F1} & \hdsub{ECE} & \hdsub{vs Jev} & \hdsub{Acc.} & \hdsub{F1} & \hdsub{ECE} & \hdsub{vs Jev} \\
\midrule
\tabhead
\groupspan{nine}{Hosted}
\msw{jev}Jev 1.13.0 & \num{nt.jev.goemo.acc} & \num{nt.jev.goemo.fone} & \num{nt.jev.goemo.ece} & -- & \num{nt.jev.banking.acc} & \num{nt.jev.banking.fone} & \num{nt.jev.banking.ece} & -- \\
\groupspan{nine}{Open, encoder head}
\msw{layaen}Laya (English) & \num{nt.layaen.goemo.acc} & \num{nt.layaen.goemo.fone} & \num{nt.layaen.goemo.ece} & \best{\num{nt.layaen.goemo.diff}} & \num{nt.layaen.banking.acc} & \num{nt.layaen.banking.fone} & \num{nt.layaen.banking.ece} & \best{\num{nt.layaen.banking.diff}} \\
\msw{layaml}Laya (multilingual) & \num{nt.layaml.goemo.acc} & \num{nt.layaml.goemo.fone} & \best{\num{nt.layaml.goemo.ece}} & \best{\num{nt.layaml.goemo.diff}} & \num{nt.layaml.banking.acc} & \num{nt.layaml.banking.fone} & \num{nt.layaml.banking.ece} & \best{\num{nt.layaml.banking.diff}} \\
\groupspan{nine}{Open, decoder head}
\msw{kevsmall}Kev-0.8B & \num{nt.kevsmall.goemo.acc} & \num{nt.kevsmall.goemo.fone} & \num{nt.kevsmall.goemo.ece} & \num{nt.kevsmall.goemo.diff} & \num{nt.kevsmall.banking.acc} & \num{nt.kevsmall.banking.fone} & \num{nt.kevsmall.banking.ece} & \num{nt.kevsmall.banking.diff} \\
\msw{kevlarge}Kev-9B & \num{nt.kevlarge.goemo.acc} & \num{nt.kevlarge.goemo.fone} & \num{nt.kevlarge.goemo.ece} & \num{nt.kevlarge.goemo.diff} & \num{nt.kevlarge.banking.acc} & \num{nt.kevlarge.banking.fone} & \num{nt.kevlarge.banking.ece} & \num{nt.kevlarge.banking.diff} \\
\msw{decider}decider-2b & \num{nt.decider.goemo.acc} & \num{nt.decider.goemo.fone} & \num{nt.decider.goemo.ece} & \num{nt.decider.goemo.diff} & \num{nt.decider.banking.acc} & \num{nt.decider.banking.fone} & \num{nt.decider.banking.ece} & \num{nt.decider.banking.diff} \\
\msw{thisthat}this-that-model-1.0 & \num{nt.thisthat.goemo.acc} & \num{nt.thisthat.goemo.fone} & \num{nt.thisthat.goemo.ece} & \num{nt.thisthat.goemo.diff} & \num{nt.thisthat.banking.acc} & \num{nt.thisthat.banking.fone} & \num{nt.thisthat.banking.ece} & \num{nt.thisthat.banking.diff} \\
\msw{nimble}Nimble-9B & \num{nt.nimble.goemo.acc} & \num{nt.nimble.goemo.fone} & \num{nt.nimble.goemo.ece} & \num{nt.nimble.goemo.diff} & \num{nt.nimble.banking.acc} & \num{nt.nimble.banking.fone} & \num{nt.nimble.banking.ece} & \num{nt.nimble.banking.diff} \\
\groupspan{nine}{Generative comparators}
\msw{compopen}Qwen3-14B, verbal & \num{nt.compopen.goemo.acc} & \num{nt.compopen.goemo.fone} & \num{nt.compopen.goemo.ece} & \num{nt.compopen.goemo.diff} & \num{nt.compopen.banking.acc} & \num{nt.compopen.banking.fone} & \num{nt.compopen.banking.ece} & \best{\num{nt.compopen.banking.diff}} \\
\addlinespace[2.5pt]
\msw{comptwo}Qwen3.6-27B, verbal & \num{nt.comptwo.goemo.acc} & \num{nt.comptwo.goemo.fone} & \num{nt.comptwo.goemo.ece} & \best{\num{nt.comptwo.goemo.diff}} & \num{nt.comptwo.banking.acc} & \num{nt.comptwo.banking.fone} & \num{nt.comptwo.banking.ece} & \best{\num{nt.comptwo.banking.diff}} \\
\msw{comptwoll}Qwen3.6-27B, likelihood & \num{nt.comptwoll.goemo.acc} & \num{nt.comptwoll.goemo.fone} & \num{nt.comptwoll.goemo.ece} & \num{nt.comptwoll.goemo.diff} & \num{nt.comptwoll.banking.acc} & \num{nt.comptwoll.banking.fone} & \num{nt.comptwoll.banking.ece} & \num{nt.comptwoll.banking.diff} \\
\msw{comptwothink}Qwen3.6-27B, thinking & \num{nt.comptwothink.goemo.acc} & \num{nt.comptwothink.goemo.fone} & \num{nt.comptwothink.goemo.ece} & \num{nt.comptwothink.goemo.diff} & \num{nt.comptwothink.banking.acc} & \num{nt.comptwothink.banking.fone} & \num{nt.comptwothink.banking.ece} & \num{nt.comptwothink.banking.diff} \\
\addlinespace[2.5pt]
\msw{compgem}Gemma-4-31B, verbal & \num{nt.compgem.goemo.acc} & \num{nt.compgem.goemo.fone} & \num{nt.compgem.goemo.ece} & \num{nt.compgem.goemo.diff} & \num{nt.compgem.banking.acc} & \num{nt.compgem.banking.fone} & \num{nt.compgem.banking.ece} & \num{nt.compgem.banking.diff} \\
\msw{compgemll}Gemma-4-31B, likelihood & \num{nt.compgemll.goemo.acc} & \num{nt.compgemll.goemo.fone} & \num{nt.compgemll.goemo.ece} & \num{nt.compgemll.goemo.diff} & \num{nt.compgemll.banking.acc} & \num{nt.compgemll.banking.fone} & \num{nt.compgemll.banking.ece} & \num{nt.compgemll.banking.diff} \\
\addlinespace[2.5pt]
\msw{compmis}Mistral-Small-24B, verbal & \num{nt.compmis.goemo.acc} & \num{nt.compmis.goemo.fone} & \num{nt.compmis.goemo.ece} & \best{\num{nt.compmis.goemo.diff}} & \num{nt.compmis.banking.acc} & \num{nt.compmis.banking.fone} & \num{nt.compmis.banking.ece} & \best{\num{nt.compmis.banking.diff}} \\
\msw{compmisll}Mistral-Small-24B, likelihood & \num{nt.compmisll.goemo.acc} & \num{nt.compmisll.goemo.fone} & \num{nt.compmisll.goemo.ece} & \num{nt.compmisll.goemo.diff} & \num{nt.compmisll.banking.acc} & \num{nt.compmisll.banking.fone} & \num{nt.compmisll.banking.ece} & \best{\num{nt.compmisll.banking.diff}} \\
\groupspan{nine}{Classifier trained on each benchmark's own labels}
BGE-small, trained per task & \num{nt.bgegoemo.goemo.acc} & \num{nt.bgegoemo.goemo.fone} & \num{nt.bgegoemo.goemo.ece} & \best{\num{nt.bgegoemo.goemo.diff}} & \num{nt.bgebanking.banking.acc} & \num{nt.bgebanking.banking.fone} & \num{nt.bgebanking.banking.ece} & \best{\num{nt.bgebanking.banking.diff}} \\*
DeBERTa-v3-large, fine-tuned & \best{\num{nt.debgoemo.goemo.acc}} & \best{\num{nt.debgoemo.goemo.fone}} & \num{nt.debgoemo.goemo.ece} & \best{\num{nt.debgoemo.goemo.diff}} & \best{\num{nt.debbanking.banking.acc}} & \best{\num{nt.debbanking.banking.fone}} & \best{\num{nt.debbanking.banking.ece}} & \best{\num{nt.debbanking.banking.diff}} \\*
\groupspan{nine}{Zero-shot classifier}
DeBERTa-v3 NLI, zero-shot & \num{nt.nlizs.goemo.acc} & \num{nt.nlizs.goemo.fone} & \num{nt.nlizs.goemo.ece} & \best{\num{nt.nlizs.goemo.diff}} & \num{nt.nlizs.banking.acc} & \num{nt.nlizs.banking.fone} & \num{nt.nlizs.banking.ece} & \best{\num{nt.nlizs.banking.diff}} \\*
\bottomrule
\end{tabular}
\tablenote{GoEmotions has \num{n.d3goemotions.requests} Reddit comments whose rater-agreed labels map to one of six Ekman emotions, and Banking77 has \num{n.d2banking77.requests} queries, each asked with all 77 intents. F1 is macro-F1 in percent, ECE the calibration error as shipped, and vs Jev the paired accuracy difference from Jev 1.13.0 in points. The Kev checkpoints list Banking77 among their training sets. Bold marks the best value in each of the first three columns of a benchmark and a difference significant after Holm correction at the five percent level.}

%% file: tables/t04_calibration.tex
\tablestylewide
\begin{tabular}{P{0.2850}N{0.0680}N{0.0568}N{0.0568}N{0.0568}N{0.0680}N{0.0568}N{0.0568}N{0.0568}N{0.0680}N{0.0568}N{0.0568}N{0.0568}}
\toprule
\headrow \hdtop{} & \hdgroup{four}{D1 typed-decisions} & \hdgroup{four}{D2 CLINC-150} & \hdgroup{four}{D3 pooled} \\
\bandrules{2-5,6-9,10-13}
\headrow \hdsub{Model} & \hdsub{Ship.\mk{}} & \hdsub{Raw} & \hdsub{Scal.} & \hdsub{Cov.} & \hdsub{Ship.\mk{}} & \hdsub{Raw} & \hdsub{Scal.} & \hdsub{Cov.} & \hdsub{Ship.\mk{}} & \hdsub{Raw} & \hdsub{Scal.} & \hdsub{Cov.} \\
\midrule
\tabhead
\groupspan{thirteen}{Hosted}
\msw{jev}Jev 1.13.0 & \best{\num{e1.jev.doneneutral.ece}}\mk{} & -- & \num{e1.jev.doneneutral.ecescaled} & \num{e1.jev.doneneutral.covfive} & \num{e1.jev.dtwo.ece}\mk{} & -- & \num{e1.jev.dtwo.ecescaled} & \num{e1.jev.dtwo.covfive} & \num{e1.jev.dthree.ece}\mk{} & -- & \num{e1.jev.dthree.ecescaled} & \num{e1.jev.dthree.covfive} \\
\groupspan{thirteen}{Open, encoder head}
\msw{layaen}Laya (English) & \num{e1.layaen.doneneutral.ece}\mk{} & \num{e1.layaen.doneneutral.eceraw} & \num{e1.layaen.doneneutral.ecescaled} & \num{e1.layaen.doneneutral.covfive} & --\mk{} & -- & -- & -- & \num{e1.layaen.dthree.ece}\mk{} & \num{e1.layaen.dthree.eceraw} & \num{e1.layaen.dthree.ecescaled} & \best{\num{e1.layaen.dthree.covfive}} \\
\msw{layaml}Laya (multilingual) & \num{e1.layaml.doneneutral.ece}\mk{} & \num{e1.layaml.doneneutral.eceraw} & \num{e1.layaml.doneneutral.ecescaled} & \num{e1.layaml.doneneutral.covfive} & \num{e1.layaml.dtwo.ece}\mk{} & \num{e1.layaml.dtwo.eceraw} & \num{e1.layaml.dtwo.ecescaled} & \num{e1.layaml.dtwo.covfive} & \num{e1.layaml.dthree.ece}\mk{} & \best{\num{e1.layaml.dthree.eceraw}} & \num{e1.layaml.dthree.ecescaled} & \num{e1.layaml.dthree.covfive} \\
\groupspan{thirteen}{Open, decoder head}
\msw{kevsmall}Kev-0.8B & \num{e1.kevsmall.doneneutral.ece}\mk{} & \num{e1.kevsmall.doneneutral.eceraw} & \num{e1.kevsmall.doneneutral.ecescaled} & \num{e1.kevsmall.doneneutral.covfive} & \num{e1.kevsmall.dtwo.ece}\mk{\mkup} & \num{e1.kevsmall.dtwo.eceraw} & \num{e1.kevsmall.dtwo.ecescaled} & \num{e1.kevsmall.dtwo.covfive} & \best{\num{e1.kevsmall.dthree.ece}}\mk{} & \num{e1.kevsmall.dthree.eceraw} & \best{\num{e1.kevsmall.dthree.ecescaled}} & \num{e1.kevsmall.dthree.covfive} \\
\msw{kevlarge}Kev-9B & \num{e1.kevlarge.doneneutral.ece}\mk{} & \num{e1.kevlarge.doneneutral.eceraw} & \num{e1.kevlarge.doneneutral.ecescaled} & \num{e1.kevlarge.doneneutral.covfive} & \num{e1.kevlarge.dtwo.ece}\mk{\mkup} & \num{e1.kevlarge.dtwo.eceraw} & \num{e1.kevlarge.dtwo.ecescaled} & \num{e1.kevlarge.dtwo.covfive} & \num{e1.kevlarge.dthree.ece}\mk{} & \num{e1.kevlarge.dthree.eceraw} & \num{e1.kevlarge.dthree.ecescaled} & \num{e1.kevlarge.dthree.covfive} \\
\msw{decider}decider-2b & \num{e1.decider.doneneutral.ece}\mk{} & \num{e1.decider.doneneutral.eceraw} & \num{e1.decider.doneneutral.ecescaled} & \num{e1.decider.doneneutral.covfive} & \num{e1.decider.dtwo.ece}\mk{\mkup} & \best{\num{e1.decider.dtwo.eceraw}} & \num{e1.decider.dtwo.ecescaled} & \num{e1.decider.dtwo.covfive} & \num{e1.decider.dthree.ece}\mk{} & \num{e1.decider.dthree.eceraw} & \num{e1.decider.dthree.ecescaled} & \num{e1.decider.dthree.covfive} \\
\msw{thisthat}this-that-model-1.0 & \num{e1.thisthat.doneneutral.ece}\mk{} & \num{e1.thisthat.doneneutral.eceraw} & \num{e1.thisthat.doneneutral.ecescaled} & \num{e1.thisthat.doneneutral.covfive} & \num{e1.thisthat.dtwo.ece}\mk{} & \num{e1.thisthat.dtwo.eceraw} & \best{\num{e1.thisthat.dtwo.ecescaled}} & \num{e1.thisthat.dtwo.covfive} & \num{e1.thisthat.dthree.ece}\mk{} & \num{e1.thisthat.dthree.eceraw} & \num{e1.thisthat.dthree.ecescaled} & \num{e1.thisthat.dthree.covfive} \\
\msw{nimble}Nimble-9B & \num{e1.nimble.doneneutral.ece}\mk{} & \best{\num{e1.nimble.doneneutral.eceraw}} & \best{\num{e1.nimble.doneneutral.ecescaled}} & \num{e1.nimble.doneneutral.covfive} & \num{e1.nimble.dtwo.ece}\mk{} & \num{e1.nimble.dtwo.eceraw} & \num{e1.nimble.dtwo.ecescaled} & \num{e1.nimble.dtwo.covfive} & \num{e1.nimble.dthree.ece}\mk{} & \num{e1.nimble.dthree.eceraw} & \num{e1.nimble.dthree.ecescaled} & \num{e1.nimble.dthree.covfive} \\
\groupspan{thirteen}{Generative comparators}
\msw{compopen}Qwen3-14B, verbal & \num{e1.compopen.doneneutral.ece}\mk{} & -- & \num{e1.compopen.doneneutral.ecescaled} & \num{e1.compopen.doneneutral.covfive} & \num{e1.compopen.dtwo.ece}\mk{} & -- & \num{e1.compopen.dtwo.ecescaled} & \num{e1.compopen.dtwo.covfive} & \num{e1.compopen.dthree.ece}\mk{} & -- & \num{e1.compopen.dthree.ecescaled} & \num{e1.compopen.dthree.covfive} \\
\addlinespace[2.5pt]
\msw{comptwo}Qwen3.6-27B, verbal & \num{e1.comptwo.doneneutral.ece}\mk{} & -- & \num{e1.comptwo.doneneutral.ecescaled} & \num{e1.comptwo.doneneutral.covfive} & \num{e1.comptwo.dtwo.ece}\mk{} & -- & \num{e1.comptwo.dtwo.ecescaled} & \num{e1.comptwo.dtwo.covfive} & \num{e1.comptwo.dthree.ece}\mk{} & -- & \num{e1.comptwo.dthree.ecescaled} & \num{e1.comptwo.dthree.covfive} \\
\msw{comptwoll}Qwen3.6-27B, likelihood & \num{e1.comptwoll.doneneutral.ece}\mk{} & \num{e1.comptwoll.doneneutral.eceraw} & \num{e1.comptwoll.doneneutral.ecescaled} & \best{\num{e1.comptwoll.doneneutral.covfive}} & \num{e1.comptwoll.dtwo.ece}\mk{} & \num{e1.comptwoll.dtwo.eceraw} & \num{e1.comptwoll.dtwo.ecescaled} & \num{e1.comptwoll.dtwo.covfive} & \num{e1.comptwoll.dthree.ece}\mk{} & \num{e1.comptwoll.dthree.eceraw} & \num{e1.comptwoll.dthree.ecescaled} & \num{e1.comptwoll.dthree.covfive} \\
\msw{comptwothink}Qwen3.6-27B, thinking & \num{e1.comptwothink.doneneutral.ece}\mk{} & -- & \num{e1.comptwothink.doneneutral.ecescaled} & \num{e1.comptwothink.doneneutral.covfive} & \num{e1.comptwothink.dtwo.ece}\mk{} & -- & \num{e1.comptwothink.dtwo.ecescaled} & \num{e1.comptwothink.dtwo.covfive} & \num{e1.comptwothink.dthree.ece}\mk{} & -- & \best{\num{e1.comptwothink.dthree.ecescaled}} & \num{e1.comptwothink.dthree.covfive} \\
\addlinespace[2.5pt]
\msw{compgem}Gemma-4-31B, verbal & \num{e1.compgem.doneneutral.ece}\mk{} & -- & \num{e1.compgem.doneneutral.ecescaled} & \num{e1.compgem.doneneutral.covfive} & \best{\num{e1.compgem.dtwo.ece}}\mk{} & -- & \num{e1.compgem.dtwo.ecescaled} & \num{e1.compgem.dtwo.covfive} & \num{e1.compgem.dthree.ece}\mk{} & -- & \num{e1.compgem.dthree.ecescaled} & \num{e1.compgem.dthree.covfive} \\
\msw{compgemll}Gemma-4-31B, likelihood & \num{e1.compgemll.doneneutral.ece}\mk{} & \num{e1.compgemll.doneneutral.eceraw} & \num{e1.compgemll.doneneutral.ecescaled} & \num{e1.compgemll.doneneutral.covfive} & \num{e1.compgemll.dtwo.ece}\mk{} & \num{e1.compgemll.dtwo.eceraw} & \num{e1.compgemll.dtwo.ecescaled} & \best{\num{e1.compgemll.dtwo.covfive}} & \num{e1.compgemll.dthree.ece}\mk{} & \num{e1.compgemll.dthree.eceraw} & \num{e1.compgemll.dthree.ecescaled} & \num{e1.compgemll.dthree.covfive} \\*
\addlinespace[2.5pt]
\msw{compmis}Mistral-Small-24B, verbal & \num{e1.compmis.doneneutral.ece}\mk{} & -- & \num{e1.compmis.doneneutral.ecescaled} & \num{e1.compmis.doneneutral.covfive} & \num{e1.compmis.dtwo.ece}\mk{} & -- & \num{e1.compmis.dtwo.ecescaled} & \num{e1.compmis.dtwo.covfive} & \num{e1.compmis.dthree.ece}\mk{} & -- & \num{e1.compmis.dthree.ecescaled} & \num{e1.compmis.dthree.covfive} \\*
\msw{compmisll}Mistral-Small-24B, likelihood & \num{e1.compmisll.doneneutral.ece}\mk{} & \num{e1.compmisll.doneneutral.eceraw} & \num{e1.compmisll.doneneutral.ecescaled} & \num{e1.compmisll.doneneutral.covfive} & \num{e1.compmisll.dtwo.ece}\mk{} & \num{e1.compmisll.dtwo.eceraw} & \num{e1.compmisll.dtwo.ecescaled} & \num{e1.compmisll.dtwo.covfive} & \num{e1.compmisll.dthree.ece}\mk{} & \num{e1.compmisll.dthree.eceraw} & \best{\num{e1.compmisll.dthree.ecescaled}} & \best{\num{e1.compmisll.dthree.covfive}} \\*
\bottomrule
\end{tabular}
\tablenote{Ship., Raw and Scal.\ are the expected calibration error as shipped, at a temperature of one and after scaling, and Cov.\ is the coverage at \num{p.riskfive} percent risk as shipped. The raw arm exists only where the served temperature is known and the temperature-one distribution is stored, so Jev 1.13.0 and the verbal readouts have none, and the likelihood readout is itself read at a temperature of one. The Laya (multilingual) cells on D2, whose option text exceeds its input budget, are shown but not ranked. Bold marks the lowest error and the highest coverage in each column, with ties bolded, and \mkup\ marks a shipped error above the raw error before rounding, where the stored temperature raises the error.}

%% file: tables/t05_cardinality.tex
\tablestyle
\begin{tabular}{P{0.2850}N{0.1021}N{0.1021}N{0.1021}N{0.1021}N{0.1021}N{0.1021}N{0.1021}}
\toprule
\headrow \hdtop{} & \hdgroup{four}{One-stage accuracy by option count} & \hdgroup{three}{With all \num{p.kall} options} \\
\bandrules{2-5,6-8}
\headrow \hdsub{Model} & \hdsub{\num{p.kfive}} & \hdsub{\num{p.ktwenty}} & \hdsub{\num{p.kfifty}} & \hdsub{\num{p.kall}} & \hdsub{ECE} & \hdsub{Two-stage} & \hdsub{Drop} \\
\midrule
\tabhead
\groupspan{eight}{Hosted}
\msw{jev}Jev 1.13.0 & \num{e3.jev.five.acc} & \num{e3.jev.twenty.acc} & \num{e3.jev.fifty.acc} & \num{e3.jev.all.acc} & \num{e3.jev.all.ece} & \best{\num{e3.jev.hier.acc}} & \num{e3.jev.drop} \\
\groupspan{eight}{Open, encoder head}
\msw{layaen}Laya (English) & \num{e3.layaen.five.acc} & \num{e3.layaen.twenty.acc} & \num{e3.layaen.fifty.acc} & -- & -- & \num{e3.layaen.hier.acc} & -- \\
\msw{layaml}Laya (multilingual) & \num{e3.layaml.five.acc} & \num{e3.layaml.twenty.acc} & \num{e3.layaml.fifty.acc} & \num{e3.layaml.all.acc} & \num{e3.layaml.all.ece} & \num{e3.layaml.hier.acc} & \num{e3.layaml.drop} \\
\groupspan{eight}{Open, decoder head}
\msw{kevsmall}Kev-0.8B & \num{e3.kevsmall.five.acc} & \num{e3.kevsmall.twenty.acc} & \num{e3.kevsmall.fifty.acc} & \num{e3.kevsmall.all.acc} & \num{e3.kevsmall.all.ece} & \num{e3.kevsmall.hier.acc} & \num{e3.kevsmall.drop} \\
\msw{kevlarge}Kev-9B & \num{e3.kevlarge.five.acc} & \num{e3.kevlarge.twenty.acc} & \num{e3.kevlarge.fifty.acc} & \num{e3.kevlarge.all.acc} & \num{e3.kevlarge.all.ece} & \num{e3.kevlarge.hier.acc} & \num{e3.kevlarge.drop} \\
\msw{decider}decider-2b & \best{\num{e3.decider.five.acc}} & \num{e3.decider.twenty.acc} & \num{e3.decider.fifty.acc} & \num{e3.decider.all.acc} & \num{e3.decider.all.ece} & \num{e3.decider.hier.acc} & \num{e3.decider.drop} \\
\msw{thisthat}this-that-model-1.0 & \best{\num{e3.thisthat.five.acc}} & \best{\num{e3.thisthat.twenty.acc}} & \num{e3.thisthat.fifty.acc} & \num{e3.thisthat.all.acc} & \num{e3.thisthat.all.ece} & \num{e3.thisthat.hier.acc} & \num{e3.thisthat.drop} \\
\msw{nimble}Nimble-9B & \num{e3.nimble.five.acc} & \num{e3.nimble.twenty.acc} & \num{e3.nimble.fifty.acc} & \num{e3.nimble.all.acc} & \num{e3.nimble.all.ece} & \num{e3.nimble.hier.acc} & \num{e3.nimble.drop} \\
\groupspan{eight}{Generative comparators}
\msw{compopen}Qwen3-14B, verbal & \num{e3.compopen.five.acc} & \num{e3.compopen.twenty.acc} & \num{e3.compopen.fifty.acc} & \num{e3.compopen.all.acc} & \num{e3.compopen.all.ece} & -- & \num{e3.compopen.drop} \\
\addlinespace[2.5pt]
\msw{comptwo}Qwen3.6-27B, verbal & \num{e3.comptwo.five.acc} & \num{e3.comptwo.twenty.acc} & \num{e3.comptwo.fifty.acc} & \num{e3.comptwo.all.acc} & \num{e3.comptwo.all.ece} & -- & \num{e3.comptwo.drop} \\
\msw{comptwoll}Qwen3.6-27B, likelihood & \best{\num{e3.comptwoll.five.acc}} & \best{\num{e3.comptwoll.twenty.acc}} & \num{e3.comptwoll.fifty.acc} & \num{e3.comptwoll.all.acc} & \num{e3.comptwoll.all.ece} & -- & \num{e3.comptwoll.drop} \\
\addlinespace[2.5pt]
\msw{compgem}Gemma-4-31B, verbal & \best{\num{e3.compgem.five.acc}} & \num{e3.compgem.twenty.acc} & \num{e3.compgem.fifty.acc} & \best{\num{e3.compgem.all.acc}} & \best{\num{e3.compgem.all.ece}} & -- & \best{\num{e3.compgem.drop}} \\
\msw{compgemll}Gemma-4-31B, likelihood & \best{\num{e3.compgemll.five.acc}} & \num{e3.compgemll.twenty.acc} & \best{\num{e3.compgemll.fifty.acc}} & \num{e3.compgemll.all.acc} & \num{e3.compgemll.all.ece} & -- & \num{e3.compgemll.drop} \\*
\addlinespace[2.5pt]
\msw{compmis}Mistral-Small-24B, verbal & \num{e3.compmis.five.acc} & \num{e3.compmis.twenty.acc} & \num{e3.compmis.fifty.acc} & \num{e3.compmis.all.acc} & \num{e3.compmis.all.ece} & -- & \num{e3.compmis.drop} \\*
\msw{compmisll}Mistral-Small-24B, likelihood & \num{e3.compmisll.five.acc} & \num{e3.compmisll.twenty.acc} & \num{e3.compmisll.fifty.acc} & \num{e3.compmisll.all.acc} & \num{e3.compmisll.all.ece} & -- & \num{e3.compmisll.drop} \\*
\bottomrule
\end{tabular}
\tablenote{Two-stage is the accuracy of asking the domain and then the intent, and Drop is the one-stage accuracy lost from \num{p.kfive} to \num{p.kall} options. The two-stage condition was run for the decision models only. Bold marks the highest accuracy and the lowest error and drop in each column, with ties bolded.}

%% file: tables/t07_cost.tex
\tablestyle
\setlength{\valbarwidth}{\dimexpr0.2400\linewidth-2\tabcolsep\relax}
\begin{tabular}{P{0.2850}P{0.1300}N{0.1050}P{0.2400}N{0.1200}N{0.1200}}
\toprule
\headrow \hdtop{} & \hdtop{} & \hdgroup{two}{US dollars per thousand decisions} & \hdgroup{two}{Latency in ms} \\
\bandrules{3-4,5-6}
\headrow \hdsub{Model} & \hdsub{Measured on} & \hdsub{Cost} & \hdsub{} & \hdsub{Median} & \hdsub{\num{p.pninetyfive}th pct.} \\
\midrule
\tabhead
\groupspan{six}{Hosted}
\msw{jev}Jev 1.13.0 & API & \num{e5.jev.d1neutral.usd} & \valbar{e5.jev.d1neutral.usd}{e5.comptwo.d1neutral.usd}{e5.jev.d1neutral.usd} & \num{e5.jev.d1neutral.pfifty} & \num{e5.jev.d1neutral.pninetyfive} \\
\groupspan{six}{Open, encoder head}
\msw{layaen}Laya (English) & Colab L4 & \num{e5.layaen.d1neutral.usd} & \valbar{e5.layaen.d1neutral.usd}{e5.comptwo.d1neutral.usd}{e5.jev.d1neutral.usd} & \num{e5.layaen.d1neutral.pfifty} & \num{e5.layaen.d1neutral.pninetyfive} \\
\msw{layaml}Laya (multilingual) & Colab L4 & \best{\num{e5.layaml.d1neutral.usd}} & \valbar{e5.layaml.d1neutral.usd}{e5.comptwo.d1neutral.usd}{e5.jev.d1neutral.usd} & \best{\num{e5.layaml.d1neutral.pfifty}} & \best{\num{e5.layaml.d1neutral.pninetyfive}} \\
\groupspan{six}{Open, decoder head}
\msw{kevsmall}Kev-0.8B & Colab L4 & \num{e5.kevsmall.d1neutral.usd} & \valbar{e5.kevsmall.d1neutral.usd}{e5.comptwo.d1neutral.usd}{e5.jev.d1neutral.usd} & \num{e5.kevsmall.d1neutral.pfifty} & \num{e5.kevsmall.d1neutral.pninetyfive} \\
\msw{kevlarge}Kev-9B & Colab L4 & \num{e5.kevlarge.d1neutral.usd} & \valbar{e5.kevlarge.d1neutral.usd}{e5.comptwo.d1neutral.usd}{e5.jev.d1neutral.usd} & \num{e5.kevlarge.d1neutral.pfifty} & \num{e5.kevlarge.d1neutral.pninetyfive} \\
\msw{decider}decider-2b & Colab L4 & \num{e5.decider.d1neutral.usd} & \valbar{e5.decider.d1neutral.usd}{e5.comptwo.d1neutral.usd}{e5.jev.d1neutral.usd} & \num{e5.decider.d1neutral.pfifty} & \num{e5.decider.d1neutral.pninetyfive} \\
\msw{thisthat}this-that-model-1.0 & Colab L4 & \num{e5.thisthat.d1neutral.usd} & \valbar{e5.thisthat.d1neutral.usd}{e5.comptwo.d1neutral.usd}{e5.jev.d1neutral.usd} & \num{e5.thisthat.d1neutral.pfifty} & \num{e5.thisthat.d1neutral.pninetyfive} \\
\msw{nimble}Nimble-9B & Colab L4 & \num{e5.nimble.d1neutral.usd} & \valbar{e5.nimble.d1neutral.usd}{e5.comptwo.d1neutral.usd}{e5.jev.d1neutral.usd} & \num{e5.nimble.d1neutral.pfifty} & \num{e5.nimble.d1neutral.pninetyfive} \\
\groupspan{six}{Generative comparators}
\msw{compopen}Qwen3-14B, verbal & Colab L4 & \num{e5.compopen.d1neutral.usd} & \valbar{e5.compopen.d1neutral.usd}{e5.comptwo.d1neutral.usd}{e5.jev.d1neutral.usd} & \num{e5.compopenbone.d1neutral.pfifty} & \num{e5.compopenbone.d1neutral.pninetyfive} \\
\addlinespace[2.5pt]
\msw{comptwo}Qwen3.6-27B, verbal & Colab A100 & \num{e5.comptwo.d1neutral.usd} & \valbar{e5.comptwo.d1neutral.usd}{e5.comptwo.d1neutral.usd}{e5.jev.d1neutral.usd} & \num{e5.comptwobone.d1neutral.pfifty} & \num{e5.comptwobone.d1neutral.pninetyfive} \\
\msw{comptwoll}Qwen3.6-27B, likelihood & Colab A100 & \num{e5.comptwoll.d1neutral.usd} & \valbar{e5.comptwoll.d1neutral.usd}{e5.comptwo.d1neutral.usd}{e5.jev.d1neutral.usd} & \num{e5.comptwollbone.d1neutral.pfifty} & \num{e5.comptwollbone.d1neutral.pninetyfive} \\
\msw{comptwothink}Qwen3.6-27B, thinking & Colab A100 & \num{e5.comptwothink.d1neutral.usd} & \valbar{e5.comptwothink.d1neutral.usd}{e5.comptwo.d1neutral.usd}{e5.jev.d1neutral.usd} & -- & -- \\
\addlinespace[2.5pt]
\msw{compgem}Gemma-4-31B, verbal & Colab A100 & \num{e5.compgem.d1neutral.usd} & \valbar{e5.compgem.d1neutral.usd}{e5.comptwo.d1neutral.usd}{e5.jev.d1neutral.usd} & \num{e5.compgembone.d1neutral.pfifty} & \num{e5.compgembone.d1neutral.pninetyfive} \\
\msw{compgemll}Gemma-4-31B, likelihood & Colab A100 & \num{e5.compgemll.d1neutral.usd} & \valbar{e5.compgemll.d1neutral.usd}{e5.comptwo.d1neutral.usd}{e5.jev.d1neutral.usd} & \num{e5.compgemllbone.d1neutral.pfifty} & \num{e5.compgemllbone.d1neutral.pninetyfive} \\*
\addlinespace[2.5pt]
\msw{compmis}Mistral-Small-24B, verbal & Colab A100 & \num{e5.compmis.d1neutral.usd} & \valbar{e5.compmis.d1neutral.usd}{e5.comptwo.d1neutral.usd}{e5.jev.d1neutral.usd} & \num{e5.compmisbone.d1neutral.pfifty} & \num{e5.compmisbone.d1neutral.pninetyfive} \\*
\msw{compmisll}Mistral-Small-24B, likelihood & Colab A100 & \num{e5.compmisll.d1neutral.usd} & \valbar{e5.compmisll.d1neutral.usd}{e5.comptwo.d1neutral.usd}{e5.jev.d1neutral.usd} & \num{e5.compmisllbone.d1neutral.pfifty} & \num{e5.compmisllbone.d1neutral.pninetyfive} \\*
\bottomrule
\end{tabular}
\tablenote{Costs of the comparators come from batched runs and their latencies from \num{p.bone.requests} requests sent one at a time, with Gemma-4-31B and Mistral-Small-24B served without graph compilation and the thinking mode not timed alone. Bars share one linear scale, cut at its end for the thinking mode, and the orange tick marks the hosted model's list price. Bold marks the lowest value in each column, with ties bolded.}

%% file: tables/t08_cascade.tex
\tablestylewide
\begin{tabular}{P{0.2600}N{0.0850}N{0.0850}N{0.0850}P{0.1300}N{0.0750}P{0.1100}N{0.0850}N{0.0850}}
\toprule
\headrow \hdtop{D1 typed-decisions} & \hdgroup{two}{Single stage} & \hdgroup{four}{Cascade} & \hdgroup{two}{Escalation} \\
\bandrules{2-3,4-7,8-9}
\headrow \hdsub{First stage} & \hdsub{First} & \hdsub{Second} & \hdsub{Accuracy} & \hdsub{} & \hdsub{Gain\mk{}} & \hdsub{} & \hdsub{Esc.} & \hdsub{Cost} \\
\midrule
\tabhead
\groupspan{nine}{Escalation to Gemma-4-31B, verbal}
\msw{jev}Jev 1.13.0 & \num{e6.jev.compgem.doneneutral.accfirst} & \num{e6.jev.compgem.doneneutral.accsecond} & \num{rv.cas.jev.compgem.doneneutral.acc} & {\scriptsize\cellnote{[\num{rv.cas.jev.compgem.doneneutral.acclo}, \num{rv.cas.jev.compgem.doneneutral.acchi}]}} & \num{rv.cas.jev.compgem.doneneutral.gain}\mk{} & {\scriptsize\cellnote{[\num{rv.cas.jev.compgem.doneneutral.gainlo}, \num{rv.cas.jev.compgem.doneneutral.gainhi}]}} & \num{rv.cas.jev.compgem.doneneutral.esc} & \num{rv.cas.jev.compgem.doneneutral.costfrac} \\
\msw{layaen}Laya (English) & \num{e6.layaen.compgem.doneneutral.accfirst} & \num{e6.layaen.compgem.doneneutral.accsecond} & \num{rv.cas.layaen.compgem.doneneutral.acc} & {\scriptsize\cellnote{[\num{rv.cas.layaen.compgem.doneneutral.acclo}, \num{rv.cas.layaen.compgem.doneneutral.acchi}]}} & \num{rv.cas.layaen.compgem.doneneutral.gain}\mk{} & {\scriptsize\cellnote{[\num{rv.cas.layaen.compgem.doneneutral.gainlo}, \num{rv.cas.layaen.compgem.doneneutral.gainhi}]}} & \num{rv.cas.layaen.compgem.doneneutral.esc} & \num{rv.cas.layaen.compgem.doneneutral.costfrac} \\
\msw{layaml}Laya (multilingual) & \num{e6.layaml.compgem.doneneutral.accfirst} & \num{e6.layaml.compgem.doneneutral.accsecond} & \num{rv.cas.layaml.compgem.doneneutral.acc} & {\scriptsize\cellnote{[\num{rv.cas.layaml.compgem.doneneutral.acclo}, \num{rv.cas.layaml.compgem.doneneutral.acchi}]}} & \num{rv.cas.layaml.compgem.doneneutral.gain}\mk{} & {\scriptsize\cellnote{[\num{rv.cas.layaml.compgem.doneneutral.gainlo}, \num{rv.cas.layaml.compgem.doneneutral.gainhi}]}} & \num{rv.cas.layaml.compgem.doneneutral.esc} & \num{rv.cas.layaml.compgem.doneneutral.costfrac} \\
\msw{kevsmall}Kev-0.8B & \num{e6.kevsmall.compgem.doneneutral.accfirst} & \num{e6.kevsmall.compgem.doneneutral.accsecond} & \num{rv.cas.kevsmall.compgem.doneneutral.acc} & {\scriptsize\cellnote{[\num{rv.cas.kevsmall.compgem.doneneutral.acclo}, \num{rv.cas.kevsmall.compgem.doneneutral.acchi}]}} & \num{rv.cas.kevsmall.compgem.doneneutral.gain}\mk{} & {\scriptsize\cellnote{[\num{rv.cas.kevsmall.compgem.doneneutral.gainlo}, \num{rv.cas.kevsmall.compgem.doneneutral.gainhi}]}} & \num{rv.cas.kevsmall.compgem.doneneutral.esc} & \num{rv.cas.kevsmall.compgem.doneneutral.costfrac} \\
\msw{kevlarge}Kev-9B & \num{e6.kevlarge.compgem.doneneutral.accfirst} & \num{e6.kevlarge.compgem.doneneutral.accsecond} & \num{rv.cas.kevlarge.compgem.doneneutral.acc} & {\scriptsize\cellnote{[\num{rv.cas.kevlarge.compgem.doneneutral.acclo}, \num{rv.cas.kevlarge.compgem.doneneutral.acchi}]}} & \num{rv.cas.kevlarge.compgem.doneneutral.gain}\mk{} & {\scriptsize\cellnote{[\num{rv.cas.kevlarge.compgem.doneneutral.gainlo}, \num{rv.cas.kevlarge.compgem.doneneutral.gainhi}]}} & \num{rv.cas.kevlarge.compgem.doneneutral.esc} & \num{rv.cas.kevlarge.compgem.doneneutral.costfrac} \\
\msw{decider}decider-2b & \num{e6.decider.compgem.doneneutral.accfirst} & \num{e6.decider.compgem.doneneutral.accsecond} & \num{rv.cas.decider.compgem.doneneutral.acc} & {\scriptsize\cellnote{[\num{rv.cas.decider.compgem.doneneutral.acclo}, \num{rv.cas.decider.compgem.doneneutral.acchi}]}} & \num{rv.cas.decider.compgem.doneneutral.gain}\mk{} & {\scriptsize\cellnote{[\num{rv.cas.decider.compgem.doneneutral.gainlo}, \num{rv.cas.decider.compgem.doneneutral.gainhi}]}} & \num{rv.cas.decider.compgem.doneneutral.esc} & \num{rv.cas.decider.compgem.doneneutral.costfrac} \\
\msw{thisthat}this-that-model-1.0 & \num{e6.thisthat.compgem.doneneutral.accfirst} & \num{e6.thisthat.compgem.doneneutral.accsecond} & \num{rv.cas.thisthat.compgem.doneneutral.acc} & {\scriptsize\cellnote{[\num{rv.cas.thisthat.compgem.doneneutral.acclo}, \num{rv.cas.thisthat.compgem.doneneutral.acchi}]}} & \num{rv.cas.thisthat.compgem.doneneutral.gain}\mk{} & {\scriptsize\cellnote{[\num{rv.cas.thisthat.compgem.doneneutral.gainlo}, \num{rv.cas.thisthat.compgem.doneneutral.gainhi}]}} & \num{rv.cas.thisthat.compgem.doneneutral.esc} & \num{rv.cas.thisthat.compgem.doneneutral.costfrac} \\
\msw{nimble}Nimble-9B & \num{e6.nimble.compgem.doneneutral.accfirst} & \num{e6.nimble.compgem.doneneutral.accsecond} & \num{rv.cas.nimble.compgem.doneneutral.acc} & {\scriptsize\cellnote{[\num{rv.cas.nimble.compgem.doneneutral.acclo}, \num{rv.cas.nimble.compgem.doneneutral.acchi}]}} & \num{rv.cas.nimble.compgem.doneneutral.gain}\mk{} & {\scriptsize\cellnote{[\num{rv.cas.nimble.compgem.doneneutral.gainlo}, \num{rv.cas.nimble.compgem.doneneutral.gainhi}]}} & \num{rv.cas.nimble.compgem.doneneutral.esc} & \num{rv.cas.nimble.compgem.doneneutral.costfrac} \\
\groupspan{nine}{Escalation to Jev 1.13.0}
\msw{layaen}Laya (English) & \num{e6.layaen.jev.doneneutral.accfirst} & \num{e6.layaen.jev.doneneutral.accsecond} & \num{rv.cas.layaen.jev.doneneutral.acc} & {\scriptsize\cellnote{[\num{rv.cas.layaen.jev.doneneutral.acclo}, \num{rv.cas.layaen.jev.doneneutral.acchi}]}} & \num{rv.cas.layaen.jev.doneneutral.gain}\mk{} & {\scriptsize\cellnote{[\num{rv.cas.layaen.jev.doneneutral.gainlo}, \num{rv.cas.layaen.jev.doneneutral.gainhi}]}} & \num{rv.cas.layaen.jev.doneneutral.esc} & \num{rv.cas.layaen.jev.doneneutral.costfrac} \\
\msw{layaml}Laya (multilingual) & \num{e6.layaml.jev.doneneutral.accfirst} & \num{e6.layaml.jev.doneneutral.accsecond} & \num{rv.cas.layaml.jev.doneneutral.acc} & {\scriptsize\cellnote{[\num{rv.cas.layaml.jev.doneneutral.acclo}, \num{rv.cas.layaml.jev.doneneutral.acchi}]}} & \num{rv.cas.layaml.jev.doneneutral.gain}\mk{} & {\scriptsize\cellnote{[\num{rv.cas.layaml.jev.doneneutral.gainlo}, \num{rv.cas.layaml.jev.doneneutral.gainhi}]}} & \num{rv.cas.layaml.jev.doneneutral.esc} & \num{rv.cas.layaml.jev.doneneutral.costfrac} \\
\msw{kevsmall}Kev-0.8B & \num{e6.kevsmall.jev.doneneutral.accfirst} & \num{e6.kevsmall.jev.doneneutral.accsecond} & \num{rv.cas.kevsmall.jev.doneneutral.acc} & {\scriptsize\cellnote{[\num{rv.cas.kevsmall.jev.doneneutral.acclo}, \num{rv.cas.kevsmall.jev.doneneutral.acchi}]}} & \num{rv.cas.kevsmall.jev.doneneutral.gain}\mk{} & {\scriptsize\cellnote{[\num{rv.cas.kevsmall.jev.doneneutral.gainlo}, \num{rv.cas.kevsmall.jev.doneneutral.gainhi}]}} & \num{rv.cas.kevsmall.jev.doneneutral.esc} & \num{rv.cas.kevsmall.jev.doneneutral.costfrac} \\
\msw{kevlarge}Kev-9B & \num{e6.kevlarge.jev.doneneutral.accfirst} & \num{e6.kevlarge.jev.doneneutral.accsecond} & \num{rv.cas.kevlarge.jev.doneneutral.acc} & {\scriptsize\cellnote{[\num{rv.cas.kevlarge.jev.doneneutral.acclo}, \num{rv.cas.kevlarge.jev.doneneutral.acchi}]}} & \num{rv.cas.kevlarge.jev.doneneutral.gain}\mk{} & {\scriptsize\cellnote{[\num{rv.cas.kevlarge.jev.doneneutral.gainlo}, \num{rv.cas.kevlarge.jev.doneneutral.gainhi}]}} & \num{rv.cas.kevlarge.jev.doneneutral.esc} & \num{rv.cas.kevlarge.jev.doneneutral.costfrac} \\
\msw{decider}decider-2b & \num{e6.decider.jev.doneneutral.accfirst} & \num{e6.decider.jev.doneneutral.accsecond} & \num{rv.cas.decider.jev.doneneutral.acc} & {\scriptsize\cellnote{[\num{rv.cas.decider.jev.doneneutral.acclo}, \num{rv.cas.decider.jev.doneneutral.acchi}]}} & \num{rv.cas.decider.jev.doneneutral.gain}\mk{} & {\scriptsize\cellnote{[\num{rv.cas.decider.jev.doneneutral.gainlo}, \num{rv.cas.decider.jev.doneneutral.gainhi}]}} & \num{rv.cas.decider.jev.doneneutral.esc} & \num{rv.cas.decider.jev.doneneutral.costfrac} \\*
\msw{thisthat}this-that-model-1.0 & \num{e6.thisthat.jev.doneneutral.accfirst} & \num{e6.thisthat.jev.doneneutral.accsecond} & \num{rv.cas.thisthat.jev.doneneutral.acc} & {\scriptsize\cellnote{[\num{rv.cas.thisthat.jev.doneneutral.acclo}, \num{rv.cas.thisthat.jev.doneneutral.acchi}]}} & \num{rv.cas.thisthat.jev.doneneutral.gain}\mk{} & {\scriptsize\cellnote{[\num{rv.cas.thisthat.jev.doneneutral.gainlo}, \num{rv.cas.thisthat.jev.doneneutral.gainhi}]}} & \num{rv.cas.thisthat.jev.doneneutral.esc} & \num{rv.cas.thisthat.jev.doneneutral.costfrac} \\*
\msw{nimble}Nimble-9B & \num{e6.nimble.jev.doneneutral.accfirst} & \num{e6.nimble.jev.doneneutral.accsecond} & \num{rv.cas.nimble.jev.doneneutral.acc} & {\scriptsize\cellnote{[\num{rv.cas.nimble.jev.doneneutral.acclo}, \num{rv.cas.nimble.jev.doneneutral.acchi}]}} & \num{rv.cas.nimble.jev.doneneutral.gain}\mk{} & {\scriptsize\cellnote{[\num{rv.cas.nimble.jev.doneneutral.gainlo}, \num{rv.cas.nimble.jev.doneneutral.gainhi}]}} & \num{rv.cas.nimble.jev.doneneutral.esc} & \num{rv.cas.nimble.jev.doneneutral.costfrac} \\*
\bottomrule
\end{tabular}
\par\vspace{6pt}
\begin{tabular}{P{0.2600}N{0.0850}N{0.0850}N{0.0850}P{0.1300}N{0.0750}P{0.1100}N{0.0850}N{0.0850}}
\toprule
\headrow \hdtop{D2 CLINC-150} & \hdgroup{two}{Single stage} & \hdgroup{four}{Cascade} & \hdgroup{two}{Escalation} \\
\bandrules{2-3,4-7,8-9}
\headrow \hdsub{First stage} & \hdsub{First} & \hdsub{Second} & \hdsub{Accuracy} & \hdsub{} & \hdsub{Gain\mk{}} & \hdsub{} & \hdsub{Esc.} & \hdsub{Cost} \\
\midrule
\tabhead
\groupspan{nine}{Escalation to Gemma-4-31B, verbal}
\msw{jev}Jev 1.13.0 & \num{e6.jev.compgem.dtwo.accfirst} & \num{e6.jev.compgem.dtwo.accsecond} & \num{rv.cas.jev.compgem.dtwo.acc} & {\scriptsize\cellnote{[\num{rv.cas.jev.compgem.dtwo.acclo}, \num{rv.cas.jev.compgem.dtwo.acchi}]}} & \num{rv.cas.jev.compgem.dtwo.gain}\mk{} & {\scriptsize\cellnote{[\num{rv.cas.jev.compgem.dtwo.gainlo}, \num{rv.cas.jev.compgem.dtwo.gainhi}]}} & \num{rv.cas.jev.compgem.dtwo.esc} & \num{rv.cas.jev.compgem.dtwo.costfrac} \\
\msw{layaen}Laya (English) & -- & -- & -- &  & --\mk{} &  & -- & -- \\
\msw{layaml}Laya (multilingual) & \num{e6.layaml.compgem.dtwo.accfirst} & \num{e6.layaml.compgem.dtwo.accsecond} & \num{rv.cas.layaml.compgem.dtwo.acc} & {\scriptsize\cellnote{[\num{rv.cas.layaml.compgem.dtwo.acclo}, \num{rv.cas.layaml.compgem.dtwo.acchi}]}} & \num{rv.cas.layaml.compgem.dtwo.gain}\mk{} & {\scriptsize\cellnote{[\num{rv.cas.layaml.compgem.dtwo.gainlo}, \num{rv.cas.layaml.compgem.dtwo.gainhi}]}} & \num{rv.cas.layaml.compgem.dtwo.esc} & \num{rv.cas.layaml.compgem.dtwo.costfrac} \\
\msw{kevsmall}Kev-0.8B & \num{e6.kevsmall.compgem.dtwo.accfirst} & \num{e6.kevsmall.compgem.dtwo.accsecond} & \num{rv.cas.kevsmall.compgem.dtwo.acc} & {\scriptsize\cellnote{[\num{rv.cas.kevsmall.compgem.dtwo.acclo}, \num{rv.cas.kevsmall.compgem.dtwo.acchi}]}} & \num{rv.cas.kevsmall.compgem.dtwo.gain}\mk{\symfull} & {\scriptsize\cellnote{[\num{rv.cas.kevsmall.compgem.dtwo.gainlo}, \num{rv.cas.kevsmall.compgem.dtwo.gainhi}]}} & \num{rv.cas.kevsmall.compgem.dtwo.esc} & \num{rv.cas.kevsmall.compgem.dtwo.costfrac} \\
\msw{kevlarge}Kev-9B & \num{e6.kevlarge.compgem.dtwo.accfirst} & \num{e6.kevlarge.compgem.dtwo.accsecond} & \num{rv.cas.kevlarge.compgem.dtwo.acc} & {\scriptsize\cellnote{[\num{rv.cas.kevlarge.compgem.dtwo.acclo}, \num{rv.cas.kevlarge.compgem.dtwo.acchi}]}} & \num{rv.cas.kevlarge.compgem.dtwo.gain}\mk{} & {\scriptsize\cellnote{[\num{rv.cas.kevlarge.compgem.dtwo.gainlo}, \num{rv.cas.kevlarge.compgem.dtwo.gainhi}]}} & \num{rv.cas.kevlarge.compgem.dtwo.esc} & \num{rv.cas.kevlarge.compgem.dtwo.costfrac} \\
\msw{decider}decider-2b & \num{e6.decider.compgem.dtwo.accfirst} & \num{e6.decider.compgem.dtwo.accsecond} & \num{rv.cas.decider.compgem.dtwo.acc} & {\scriptsize\cellnote{[\num{rv.cas.decider.compgem.dtwo.acclo}, \num{rv.cas.decider.compgem.dtwo.acchi}]}} & \num{rv.cas.decider.compgem.dtwo.gain}\mk{} & {\scriptsize\cellnote{[\num{rv.cas.decider.compgem.dtwo.gainlo}, \num{rv.cas.decider.compgem.dtwo.gainhi}]}} & \num{rv.cas.decider.compgem.dtwo.esc} & \num{rv.cas.decider.compgem.dtwo.costfrac} \\
\msw{thisthat}this-that-model-1.0 & \num{e6.thisthat.compgem.dtwo.accfirst} & \num{e6.thisthat.compgem.dtwo.accsecond} & \num{rv.cas.thisthat.compgem.dtwo.acc} & {\scriptsize\cellnote{[\num{rv.cas.thisthat.compgem.dtwo.acclo}, \num{rv.cas.thisthat.compgem.dtwo.acchi}]}} & \num{rv.cas.thisthat.compgem.dtwo.gain}\mk{} & {\scriptsize\cellnote{[\num{rv.cas.thisthat.compgem.dtwo.gainlo}, \num{rv.cas.thisthat.compgem.dtwo.gainhi}]}} & \num{rv.cas.thisthat.compgem.dtwo.esc} & \num{rv.cas.thisthat.compgem.dtwo.costfrac} \\
\msw{nimble}Nimble-9B & \num{e6.nimble.compgem.dtwo.accfirst} & \num{e6.nimble.compgem.dtwo.accsecond} & \num{rv.cas.nimble.compgem.dtwo.acc} & {\scriptsize\cellnote{[\num{rv.cas.nimble.compgem.dtwo.acclo}, \num{rv.cas.nimble.compgem.dtwo.acchi}]}} & \num{rv.cas.nimble.compgem.dtwo.gain}\mk{} & {\scriptsize\cellnote{[\num{rv.cas.nimble.compgem.dtwo.gainlo}, \num{rv.cas.nimble.compgem.dtwo.gainhi}]}} & \num{rv.cas.nimble.compgem.dtwo.esc} & \num{rv.cas.nimble.compgem.dtwo.costfrac} \\
\groupspan{nine}{Escalation to Jev 1.13.0}
\msw{layaen}Laya (English) & -- & -- & -- &  & --\mk{} &  & -- & -- \\
\msw{layaml}Laya (multilingual) & \num{e6.layaml.jev.dtwo.accfirst} & \num{e6.layaml.jev.dtwo.accsecond} & \num{rv.cas.layaml.jev.dtwo.acc} & {\scriptsize\cellnote{[\num{rv.cas.layaml.jev.dtwo.acclo}, \num{rv.cas.layaml.jev.dtwo.acchi}]}} & \num{rv.cas.layaml.jev.dtwo.gain}\mk{} & {\scriptsize\cellnote{[\num{rv.cas.layaml.jev.dtwo.gainlo}, \num{rv.cas.layaml.jev.dtwo.gainhi}]}} & \num{rv.cas.layaml.jev.dtwo.esc} & \num{rv.cas.layaml.jev.dtwo.costfrac} \\
\msw{kevsmall}Kev-0.8B & \num{e6.kevsmall.jev.dtwo.accfirst} & \num{e6.kevsmall.jev.dtwo.accsecond} & \num{rv.cas.kevsmall.jev.dtwo.acc} & {\scriptsize\cellnote{[\num{rv.cas.kevsmall.jev.dtwo.acclo}, \num{rv.cas.kevsmall.jev.dtwo.acchi}]}} & \num{rv.cas.kevsmall.jev.dtwo.gain}\mk{\symfull} & {\scriptsize\cellnote{[\num{rv.cas.kevsmall.jev.dtwo.gainlo}, \num{rv.cas.kevsmall.jev.dtwo.gainhi}]}} & \num{rv.cas.kevsmall.jev.dtwo.esc} & \num{rv.cas.kevsmall.jev.dtwo.costfrac} \\
\msw{kevlarge}Kev-9B & \num{e6.kevlarge.jev.dtwo.accfirst} & \num{e6.kevlarge.jev.dtwo.accsecond} & \num{rv.cas.kevlarge.jev.dtwo.acc} & {\scriptsize\cellnote{[\num{rv.cas.kevlarge.jev.dtwo.acclo}, \num{rv.cas.kevlarge.jev.dtwo.acchi}]}} & \num{rv.cas.kevlarge.jev.dtwo.gain}\mk{} & {\scriptsize\cellnote{[\num{rv.cas.kevlarge.jev.dtwo.gainlo}, \num{rv.cas.kevlarge.jev.dtwo.gainhi}]}} & \num{rv.cas.kevlarge.jev.dtwo.esc} & \num{rv.cas.kevlarge.jev.dtwo.costfrac} \\
\msw{decider}decider-2b & \num{e6.decider.jev.dtwo.accfirst} & \num{e6.decider.jev.dtwo.accsecond} & \num{rv.cas.decider.jev.dtwo.acc} & {\scriptsize\cellnote{[\num{rv.cas.decider.jev.dtwo.acclo}, \num{rv.cas.decider.jev.dtwo.acchi}]}} & \num{rv.cas.decider.jev.dtwo.gain}\mk{} & {\scriptsize\cellnote{[\num{rv.cas.decider.jev.dtwo.gainlo}, \num{rv.cas.decider.jev.dtwo.gainhi}]}} & \num{rv.cas.decider.jev.dtwo.esc} & \num{rv.cas.decider.jev.dtwo.costfrac} \\*
\msw{thisthat}this-that-model-1.0 & \num{e6.thisthat.jev.dtwo.accfirst} & \num{e6.thisthat.jev.dtwo.accsecond} & \num{rv.cas.thisthat.jev.dtwo.acc} & {\scriptsize\cellnote{[\num{rv.cas.thisthat.jev.dtwo.acclo}, \num{rv.cas.thisthat.jev.dtwo.acchi}]}} & \num{rv.cas.thisthat.jev.dtwo.gain}\mk{} & {\scriptsize\cellnote{[\num{rv.cas.thisthat.jev.dtwo.gainlo}, \num{rv.cas.thisthat.jev.dtwo.gainhi}]}} & \num{rv.cas.thisthat.jev.dtwo.esc} & \num{rv.cas.thisthat.jev.dtwo.costfrac} \\*
\msw{nimble}Nimble-9B & \num{e6.nimble.jev.dtwo.accfirst} & \num{e6.nimble.jev.dtwo.accsecond} & \num{rv.cas.nimble.jev.dtwo.acc} & {\scriptsize\cellnote{[\num{rv.cas.nimble.jev.dtwo.acclo}, \num{rv.cas.nimble.jev.dtwo.acchi}]}} & \num{rv.cas.nimble.jev.dtwo.gain}\mk{} & {\scriptsize\cellnote{[\num{rv.cas.nimble.jev.dtwo.gainlo}, \num{rv.cas.nimble.jev.dtwo.gainhi}]}} & \num{rv.cas.nimble.jev.dtwo.esc} & \num{rv.cas.nimble.jev.dtwo.costfrac} \\*
\bottomrule
\end{tabular}
\tablenote{First and Second are the accuracies of each stage alone, Accuracy is the cascade accuracy with thresholds chosen on data disjoint from the scored items, and Gain is its difference from the better single stage in points. Esc.\ is the escalated share and Cost the cost fraction from each decision's own cost, where the escalation unit on D1 is the decision, a request's cost is shared among its decisions and a comparator is charged per question. Brackets hold cluster bootstrap intervals at \num{p.boot.level} percent whose \num{p.boot.refitreps} replicates refit the thresholds, unadjusted for the number of cascades. A dash marks a pair without results, because Laya (English) refused the full intent list, and \symfull\ marks a gain whose interval excludes zero.}

%% file: sections/engineering.tex
\section{Implications for engineering decision gates}
\label{sec:engineering}

A decision gate is the point in a software system where a program asks a model for one of a few declared options and acts on the answer without a person in the loop. The workflows of D1 are gates of this kind, and the intent benchmark stands for the routing gate of a virtual assistant, which must map a request to one of many handlers and send unknown requests elsewhere. This section works through one gate on the intent benchmark with outcomes stated at the level of the application, and then states design rules that each name the condition under which they hold.

\begin{table}
\centering
\caption{Worked case of an intent-routing gate for a virtual assistant with \num{p.kall} handlers on the CLINC-150 test items. Every threshold is chosen on held-out data, and the intervals come from the cluster bootstrap.}
\label{tab:gate}
\input{tables/t13_gate}
\end{table}

Table~\ref{tab:gate} compares eight policies for one routing gate. Letting the hosted model route every request sends every out-of-scope request to some handler, because the policy has no reject option. A held-out threshold that targets five percent in-scope risk lets the hosted model route \num{rv.gate.jev.cov} of the in-scope requests at a realized risk of \num{rv.gate.jev.risk}, but it still accepts \num{rv.gate.jev.far} of the out-of-scope requests. The same rule applied to Kev-9B routes fewer requests, \num{rv.gate.kevlarge.cov}, and accepts only \num{rv.gate.kevlarge.far} of the unknown ones, so the choice depends on whether a misrouted unknown request or a request left for review costs more. The likelihood readout of Qwen3.6-27B routes \num{rv.gate.comptwoll.cov} of the in-scope requests and accepts \num{rv.gate.comptwoll.far} of the unknown ones, and Gemma-4-31B routes \num{rv.gate.compgem.cov} and accepts \num{rv.gate.compgem.far}, both at a higher cost per request.

The cascade from this-that-model-1.0 to the hosted model routes every request, out-of-scope ones included, at the hosted model's in-scope accuracy for \num{rv.cas.thisthat.jev.dtwo.costfrac} of its cost with the GPU fully occupied. The cascade from the same first stage to Gemma-4-31B reaches the higher in-scope accuracy of \num{rv.cas.thisthat.compgem.dtwo.acc} at \num{rv.cas.thisthat.compgem.dtwo.usd} US dollars per thousand requests, about the hosted price. The latency of each decision is composed from the measured latencies of the stages it passes through. Where the gate's own labeled traffic exists, a DeBERTa-v3-large fine-tuned on it reaches a higher in-scope accuracy than any policy that answers every request (Table~\ref{tab:baselines}).

The first condition is the availability of labels. When the gate has a stable label set and labeled traffic whose labels the input determines, a classifier fine-tuned on that traffic had the highest observed accuracy on every benchmark with such labels except power. A fine-tuned DeBERTa-v3-large answers in about \num{bl.debertalargeclinc.d2k150.latencymsp50l4} milliseconds per request on an L4 GPU. When no labels exist or the option set changes faster than a classifier can be retrained, a decision model or a generative model answers from the rubric alone. Among these, the hosted decision model was matched on workflows and exceeded on CLINC-150 by Gemma-4-31B, so its advantage lies in cost, latency and the probability interface rather than in accuracy.

A generative model should be read through the likelihood of each option key or have its reply format verified, because stated probabilities depend on a format that the parser must match, which the parser could not read in \num{vr.compmis.emotion.failshare} of the emotion replies of Mistral-Small-24B. The choice should still be checked on a labeled sample of the gate's own traffic, because the ranking of open checkpoints changed between benchmarks and exposure to a benchmark's training data lifted a model on one emotion corpus but not on another.

The second condition is the error rate the gate may accept. The threshold has to be set for one model on one task on held-out labeled data and its risk verified on further held-out data, because the same model accepts \num{rv.ho.jev.dtwo.five.cov} of the intent decisions but only \num{rv.ho.jev.doneneutral.five.cov} of the workflow decisions at the same target. On typed-decisions its realized risk of \num{rv.ho.jev.doneneutral.five.risk} already exceeds the target. A gate that needs a guaranteed risk can choose its threshold by Learn-then-Test \citep{angelopoulos2021learn,bates2021distributionfree}, which kept the hosted model at \num{rc.jev.dtwo.cov} of the intent decisions with a guarantee at \num{rc.confidence} percent confidence. The guarantee therefore costs coverage, and a model whose probabilities are saturated near one, such as the key likelihoods of Gemma-4-31B, receives no guaranteed coverage until it is recalibrated.

The third condition is a change of task or option count. A stored temperature belongs to the conditions it was fitted under, as the Kev temperatures show when they raise the calibration error of Kev-9B from \num{e1.kevlarge.dtwo.eceraw} to \num{e1.kevlarge.dtwo.ece} with \num{p.kall} options. One temperature fitted per task on held-out labeled decisions is cheap, and Proposition~\ref{prop:temperature} applies it without a second forward pass for a checkpoint whose temperature is known, so recalibration should precede any threshold choice.

The fourth condition is the design of the option schema. Where option names are chosen freely, neutral identifiers with all meaning in the rubric reduce the naming effect for the hosted model to near its retest floor. Where names are dictated by an existing schema, the four larger decoder-based checkpoints moved by at most \num{e2.decoderheads.kswap.maxflips} flips per hundred under a polarity swap. The encoder-based checkpoints and Kev-0.8B also moved under digits and random strings, and every generative comparator tested moved by at least \num{e2.compgem.toxicity.kswap.flips} on every binary set. Their untuned backbones flipped far more answers on the binary social-science questions, so this robustness comes from training beyond the base model and has to be verified for each checkpoint rather than assumed for a backbone family. The adapter that renders options into model text belongs to the gate as well, since describing the options without their keys changed the intent accuracy of this-that-model-1.0 by \num{rp.dtwo.descvsoriginal.diff} points. The option count and description length have to fit the model's input budget, and where they do not, the domain-then-intent split of Proposition~\ref{prop:hierarchical} restores part of the loss.

The fifth condition is the presence of requests outside the gate's scope. A threshold tuned for in-scope risk does not control out-of-scope acceptance even when the top probability ranks unknown requests well. An explicit out-of-scope option among the declared options lowered the false acceptance of the hosted model from \num{rv.gate.jev.far} to \num{oo.jev.gfar} at the same in-scope target and that of Gemma-4-31B from \num{rv.gate.compgem.far} to \num{oo.compgem.gfar}. Four of the open decision checkpoints, Qwen3-14B and Mistral-Small-24B seldom chose it. The gate therefore needs such an option, validated for the model at hand, or a second threshold chosen on out-of-scope examples, and its false acceptance has to be reported beside the in-scope risk.

The sixth condition is the price and accuracy gap between two stages. Proposition~\ref{prop:cascade} shows that the escalated share and the accuracy difference on the escalated decisions set the gain, and the held-out cascades show the two ways in which escalation pays. A cheap first stage that knows the gate's labels can escalate a minority of its decisions to a costlier stage that is more accurate on them. The intent-trained this-that-model-1.0 did so toward the hosted model at under half the hosted cost, and toward Gemma-4-31B at that model's accuracy and about the hosted price. A first stage without that training gained accuracy only where both stages were weak, as Kev-0.8B escalating to the verbal readout of Qwen3.6-27B did, and Nimble-9B escalating to Qwen3-14B paid more than twice the cost of its second stage. The second stage should therefore be the most accurate model the budget allows on the doubted decisions, and the first stage the cheapest model that is confident where it is right.

%% file: tables/t13_gate.tex
\tablestylewide
\begin{tabular}{P{0.2100}N{0.1000}N{0.0950}N{0.0750}N{0.0750}N{0.0900}N{0.1050}N{0.0850}N{0.0750}N{0.0900}}
\toprule
\headrow \hdtop{} & \hdgroup{five}{In-scope requests} & \hdtop{Out of scope} & \hdgroup{three}{Cost and latency} \\
\bandrules{2-6,7-7,8-10}
\headrow \hdsub{Policy} & \hdsub{Accuracy} & \hdsub{Answered} & \hdsub{Risk} & \hdsub{Bound} & \hdsub{Escalated} & \hdsub{Accepted} & \hdsub{USD} & \hdsub{Median} & \hdsub{\num{p.pninetyfive}th pct.} \\
\midrule
\tabhead
\groupspan{ten}{Every request answered}
(a) \msw{jev}Jev 1.13.0, no gate & \num{e1.jev.dtwo.acc}\par{\scriptsize\cellnote{[\num{e1.jev.dtwo.acclo}, \num{e1.jev.dtwo.acchi}]}} & all & -- & -- & none & all & \num{e5.jev.d2k150.usd} & \num{e5.jev.d2k150.pfifty} & \num{e5.jev.d2k150.pninetyfive} \\
\groupspan{ten}{Gate at the in-scope risk target}
(b) \msw{jev}Jev 1.13.0 & -- & \num{rv.gate.jev.cov}\par{\scriptsize\cellnote{[\num{rv.gate.jev.covlo}, \num{rv.gate.jev.covhi}]}} & \num{rv.gate.jev.risk} & \num{rv.gate.jev.riskub} & none & \num{rv.gate.jev.far}\par{\scriptsize\cellnote{[\num{rv.gate.jev.farlo}, \num{rv.gate.jev.farhi}]}} & \num{e5.jev.d2k150.usd} & \num{e5.jev.d2k150.pfifty} & \num{e5.jev.d2k150.pninetyfive} \\
(c) \msw{kevlarge}Kev-9B & -- & \num{rv.gate.kevlarge.cov}\par{\scriptsize\cellnote{[\num{rv.gate.kevlarge.covlo}, \num{rv.gate.kevlarge.covhi}]}} & \num{rv.gate.kevlarge.risk} & \num{rv.gate.kevlarge.riskub} & none & \num{rv.gate.kevlarge.far}\par{\scriptsize\cellnote{[\num{rv.gate.kevlarge.farlo}, \num{rv.gate.kevlarge.farhi}]}} & \num{e5.kevlarge.d2k150.usd} & \num{e5.kevlarge.d2k150.pfifty} & \num{e5.kevlarge.d2k150.pninetyfive} \\
(d) \msw{decider}decider-2b & -- & \num{rv.gate.decider.cov}\par{\scriptsize\cellnote{[\num{rv.gate.decider.covlo}, \num{rv.gate.decider.covhi}]}} & \num{rv.gate.decider.risk} & \num{rv.gate.decider.riskub} & none & \num{rv.gate.decider.far}\par{\scriptsize\cellnote{[\num{rv.gate.decider.farlo}, \num{rv.gate.decider.farhi}]}} & \num{e5.decider.d2k150.usd} & \num{e5.decider.d2k150.pfifty} & \num{e5.decider.d2k150.pninetyfive} \\
(e) \msw{comptwoll}Qwen3.6-27B, likelihood & -- & \num{rv.gate.comptwoll.cov}\par{\scriptsize\cellnote{[\num{rv.gate.comptwoll.covlo}, \num{rv.gate.comptwoll.covhi}]}} & \num{rv.gate.comptwoll.risk} & \num{rv.gate.comptwoll.riskub} & none & \num{rv.gate.comptwoll.far}\par{\scriptsize\cellnote{[\num{rv.gate.comptwoll.farlo}, \num{rv.gate.comptwoll.farhi}]}} & \num{e5.comptwoll.d2k150.usd} & \num{e5.comptwoll.d2k150.pfifty} & \num{e5.comptwoll.d2k150.pninetyfive} \\
(f) \msw{compgem}Gemma-4-31B, verbal & -- & \num{rv.gate.compgem.cov}\par{\scriptsize\cellnote{[\num{rv.gate.compgem.covlo}, \num{rv.gate.compgem.covhi}]}} & \num{rv.gate.compgem.risk} & \num{rv.gate.compgem.riskub} & none & \num{rv.gate.compgem.far}\par{\scriptsize\cellnote{[\num{rv.gate.compgem.farlo}, \num{rv.gate.compgem.farhi}]}} & \num{e5.compgem.d2k150.usd} & \num{e5.compgem.d2k150.pfifty} & \num{e5.compgem.d2k150.pninetyfive} \\*
\groupspan{ten}{Cascade}
(g) \msw{thisthat}this-that-model-1.0 escalating to Jev 1.13.0 & \num{rv.cas.thisthat.jev.dtwo.acc}\par{\scriptsize\cellnote{[\num{rv.cas.thisthat.jev.dtwo.acclo}, \num{rv.cas.thisthat.jev.dtwo.acchi}]}} & all & -- & -- & \num{rv.cas.thisthat.jev.dtwo.esc}\par{\scriptsize\cellnote{[\num{rv.cas.thisthat.jev.dtwo.esclo}, \num{rv.cas.thisthat.jev.dtwo.eschi}]}} & all & \num{rv.cas.thisthat.jev.dtwo.usd} & \num{rv.cas.thisthat.jev.dtwo.pfifty} & \num{rv.cas.thisthat.jev.dtwo.pninetyfive} \\*
(h) \msw{thisthat}this-that-model-1.0 escalating to Gemma-4-31B & \num{rv.cas.thisthat.compgem.dtwo.acc}\par{\scriptsize\cellnote{[\num{rv.cas.thisthat.compgem.dtwo.acclo}, \num{rv.cas.thisthat.compgem.dtwo.acchi}]}} & all & -- & -- & \num{rv.cas.thisthat.compgem.dtwo.esc}\par{\scriptsize\cellnote{[\num{rv.cas.thisthat.compgem.dtwo.esclo}, \num{rv.cas.thisthat.compgem.dtwo.eschi}]}} & all & \num{rv.cas.thisthat.compgem.dtwo.usd} & \num{rv.cas.thisthat.compgem.dtwo.pfifty} & \num{rv.cas.thisthat.compgem.dtwo.pninetyfive} \\*
\bottomrule
\end{tabular}
\tablenote{Answered is the share of in-scope requests a policy answers without review, and Risk the error rate among them with its one-sided \num{p.boot.level} percent upper bound. Escalated is the share sent to the second stage, and Accepted the share of out-of-scope requests routed to a handler, which is all of them when no request can be rejected. USD is the cost per thousand requests and the latencies are in ms per request, amortized over a batch for the comparators. The cascades have no reject option and so route every out-of-scope request, and their latencies compose recorded in-scope latencies offline. Beneath a value is its cluster bootstrap interval at \num{p.boot.level} percent.}

%% file: sections/discussion.tex
\section{Discussion}
\label{sec:discussion}

Placing every model under one harness changes several conclusions that single-model studies reached on their own items. The developers of this-that-model-1.0 report an accuracy of \num{lit.cheng2026thisthatmodel.thisThatAccuracy68q} for their model against \num{lit.cheng2026thisthatmodel.thisThatAccuracy68q.cmp} for the hosted model on a third party's questions \citep{cheng2026thisthatmodel}. On the identical items of this study the same checkpoint trails the hosted model by \num{pd.thisthat.doneneutral.absdiff} points on typed-decisions and \num{pd.thisthat.dtwo.absdiff} on CLINC-150, and given the option descriptions alone it is level with the hosted model on CLINC-150. The Laya card reports a zero-shot typed-decisions accuracy of \num{lit.greyLaya.layaEnglishZeroShotAccuracy} for its English checkpoint \citep{laya2026card}, and the harness reproduces it at \num{e1.layaen.donenative.acc}, so developer and harness numbers agree when the items and question texts match. The Kev-9B card reports an out-of-domain calibration error of \num{lit.greyKev.kev9bOodEce} with its stored temperature \citep{kev9b2026card}, a developer-reported value, yet on CLINC-150 that temperature yields \num{e1.kevlarge.dtwo.ece}. The same-item cardinality sweep ties the increase to the option count, since the error of Kev-9B rises with the number of options on identical utterances.

The closest single-model evaluation, posted after the literature sweep, scores the hosted model with one frozen template per dataset \citep{deusser2026evaluating}. Its accuracy of \num{lit.deusser2026evaluating.jevBankingAcc} on Banking77 and in-scope accuracy of \num{lit.deusser2026evaluating.jevClincInScope} on CLINC-150 lie close to the \num{nt.jev.banking.acc} and \num{e1.jev.dtwo.acc} measured here, and its pooled calibration error of \num{lit.deusser2026evaluating.jevPooledEce} matches the low errors found here on typed-decisions and CLINC-150. Its explicit out-of-scope option recovers \num{lit.deusser2026evaluating.jevOosRecall} of the out-of-scope requests against \num{oo.jev.oosrecall} here, a difference that the wording of the option and the share of out-of-scope requests in each test set may explain. Its finding that none of the leads of Qwen3.8-27B over the hosted model lies outside the bootstrap intervals agrees with the result here that no Qwen readout is significantly more accurate than the hosted model. It tests the smaller Gemma-4-E4B, whereas Gemma-4-31B here leads the hosted model on CLINC-150, so the ranking of a generative model against the hosted model depends on the checkpoint as well as on the readout.

An independent leaderboard offers a second check on the harness itself. The Decision Index scores Banking77 by macro-F1 on the full test split, with each model reading its own native prompt \citep{web_decision_index}. It reports \num{lit.webDecisionIndex.jevBankingF1} for the hosted model and \num{lit.webDecisionIndex.kevlargeBankingF1} for Kev-9B, within a point of the \num{nt.jev.banking.fone} and \num{nt.kevlarge.banking.fone} measured here on a sample of the same split. For Kev-0.8B and decider-2b it reports \num{lit.webDecisionIndex.kevsmallBankingF1} and \num{lit.webDecisionIndex.deciderBankingF1}, against \num{nt.kevsmall.banking.fone} and \num{nt.decider.banking.fone} here. Gaps of that size fit the sensitivity to the adapter that the rendering run of this-that-model-1.0 shows, but the leaderboard releases no answers per item that would trace them.

The option-name results extend the single study that measured a polarity swap and relocate its contrast from hosted against open to one checkpoint against another. \citet{sun2026typesafe} report that exchanging the rubrics behind no and yes changes \num{lit.sun2026typesafe.hostedJevFlipRate} percent of the hosted model's answers and \num{lit.sun2026typesafe.markerNoYesFlipRate} percent for an open marker head on the encoder family of Laya. The hosted model moves further here, by \num{e2.jev.done.kswap.flips} to \num{e2.jev.toxicity.kswap.flips} flips per hundred, and Laya by \num{e2.layaen.done.kswap.flips}, close to their marker head. The heads of Kev-9B, Nimble-9B, decider-2b and this-that-model-1.0, which read the answer from a slot, a pointer or a single token after the rubric, move by at most \num{e2.kevlarge.done.kswap.flips} on typed-decisions, and Kev-0.8B by \num{e2.kevsmall.done.kswap.flips}. Their finding that neutral identifiers remove the effect holds for these decoder-based models and approximately for the hosted model, whose flips fall to near its retest floor, while Laya also moves under digits and random strings.

The calibration results agree in direction with the largest prior evaluation and differ in magnitude. \citet{ibrahim2026evaluating} report a median calibration error of \num{lit.ibrahim2026evaluating.jevMedianECE} for the hosted model over their social-science tasks and \num{lit.ibrahim2026evaluating.jevMedianECETempScaled} after a transferred temperature. On the four shared tasks the pooled error as shipped is \num{e1.jev.dthree.ece} and falls to \num{e1.jev.dthree.ecescaled} with a temperature fitted within each task. Their grid places Nimble-9B at a median macro-F1 of \num{lit.ibrahim2026evaluating.BespokeNimble9BMedianF1} and decider-2b at \num{lit.ibrahim2026evaluating.decider2bMedianF1} over a wider task set, against \num{e1.nimble.dthree.macrofone} and \num{e1.decider.dthree.macrofone} here, the latter raised by the emotion task whose label source decider-2b was trained on. The Laya card reports that the checkpoint ships over-confident and recovers after temperature scaling \citep{laya2026card}, which the reliability diagrams of Figure~\ref{fig:reliability} reproduce.

The evaluations of judges and of call screening both located the value of a decision model in how its confidence separates easy decisions from hard ones. \citet{li2026jevasajudge} report an accuracy of \num{lit.li2026jevasajudge.jevAccuracyHighConfidenceQGe09} percent for confident judgments against \num{lit.li2026jevasajudge.jevAccuracyLowConfidenceQLt06} percent for unconfident ones, and \citet{ren2026openjev} attribute the gain of their open readout to calibration rather than accuracy. The risk-coverage results show the same separation on the workflow and intent benchmarks and its near absence on the social-science tasks, so the value of a confidence gate has to be measured per task before a threshold is deployed.

The cascade results place prior reports of cheap escalation in a narrower frame, because their savings depend on the relative cost of the second stage. \citet{ibrahim2026evaluating} report that escalating low-confidence items from the hosted model to a frontier model keeps most of its accuracy at a cost fraction of \num{lit.ibrahim2026evaluating.jevGeminiCascadeCostFracAt0.8}, and that the same policy toward an open model costs \num{lit.ibrahim2026evaluating.jevGemmaCascadeCostFracAt0.8} times that model alone at a strict threshold. \citet{li2026jevasajudge} report a frozen cascade that reaches \num{lit.li2026jevasajudge.jevFrozenCascadeAccuracyGpt6Tau09} percent accuracy at \num{lit.li2026jevasajudge.jevFrozenCascadeFeePctGpt6Tau09} percent of the frontier fee.

Here every open comparator costs more per typed-decisions decision than the hosted model, so escalation from the hosted model saves nothing against the hosted model alone, and toward Gemma-4-31B it buys that model's higher intent accuracy at \num{rv.cas.jev.compgem.dtwo.costfrac} of its cost. Escalation from a small intent-trained open model to the hosted model on the intent benchmark matches the hosted accuracy at under half the hosted cost, with a gain whose interval includes zero. Toward Gemma-4-31B the same first stage reaches the accuracy of that model at about the hosted price. The saving therefore follows the price gap between the two stages and the difference in what they answer correctly, and it has to be measured for each pair before a policy is deployed.

The comparison with generative models agrees with the largest prior evaluation and narrows it. \citet{ibrahim2026evaluating} found that the hosted model trails the best of many language models on most social-science tasks but costs less. Under one harness the same holds on intents, where Gemma-4-31B leads the hosted model by \num{rv.pd.compgem.dtwo.diff} points at \num{e5.compgem.d2k150.usd} against \num{e5.jev.d2k150.usd} US dollars per thousand decisions. On workflows and on the social-science tasks no comparator is significantly more accurate than the hosted model, and Qwen3-14B and Mistral-Small-24B stay significantly below it on workflows. Part of the difference between the generative models lies in how they are read, since key likelihoods avoid the unreadable replies that halved the verbal accuracy of Mistral-Small-24B on emotion. The calibration of a likelihood readout cannot be assumed either, because the key likelihoods of Gemma-4-31B saturate near one and leave no threshold with a guaranteed risk. Reports that a language model outperforms a decision model, or the reverse, therefore have to name the checkpoint, the readout and the task.

%% file: sections/conclusion.tex
\section{Conclusion}
\label{sec:conclusion}

This paper compared eight decision-model checkpoints from six families, four generative language models from three developers read in eight ways, and conventional classifiers under matched conditions on three dataset families and two further benchmarks, with every decision model and comparator receiving the same semantic requests. On accuracy (RQ1), the ranking depended on the availability of task labels and on the particular checkpoint. A DeBERTa-v3-large encoder fine-tuned on each task's labels had the highest observed accuracy on every task with such labels except power, and smaller trained classifiers remained competitive on intents. Without task labels no decision model was significantly more accurate than the hosted model on the workflow and intent benchmarks. Gemma-4-31B matched it on workflows and exceeded it on CLINC-150, while Qwen3-14B and Mistral-Small-24B stayed significantly below it. Exposure explained the largest lead of an open checkpoint, on the hashtag-labeled emotion task whose training split decider-2b had seen, and that lead vanished on GoEmotions. Kev-9B gained no significant lead on Banking77 although it lists that benchmark among its training sets. The stated probabilities of generative models depended on the reply format, which the parser could not read for up to \num{vr.compmis.emotion.failshare} of the decisions of one model, whereas key likelihoods avoided that problem. On calibration (RQ2), calibration as shipped depended on the pairing of model and task, and a stored temperature fitted on few options raised the error of both Kev checkpoints with \num{p.kall} options. One held-out temperature per task lowered the error of nearly every model without adding resolution. A Learn-then-Test threshold guaranteed the risk at the price of coverage, keeping the hosted model at \num{rc.jev.dtwo.cov} of the intent decisions against \num{rv.ho.jev.dtwo.five.cov} under a held-out threshold, and saturated key likelihoods received no guaranteed coverage. A threshold tuned for in-scope risk left the hosted model accepting \num{rv.gate.jev.far} of out-of-scope intent requests, and an explicit out-of-scope option lowered that share to \num{oo.jev.gfar} at nearly the same in-scope coverage.

On robustness (RQ3), the encoder-based checkpoints collapsed once the option text exceeded their input budget, while the other decision models degraded gradually and consistently across distractor permutations. Swapping yes and no against fixed rubrics flipped most binary answers of the hosted model and of Laya and between \num{e2.compgemll.done.kswap.flips} and \num{e2.comptwoll.done.kswap.flips} typed-decisions answers per hundred for the generative comparators. Four decoder-based checkpoints moved by at most \num{e2.decoderheads.kswap.maxflips} flips per hundred, and their untuned backbones flipped far more on the binary social-science tasks, so this robustness comes from their training. Describing options without their keys moved this-that-model-1.0 by \num{rp.dtwo.descvsoriginal.diff} points on the intent benchmark, so the adapter belongs to the gate as well. On escalation (RQ4), a cascade gained accuracy beyond its interval only where both stages were weak. An intent-trained first stage matched the hosted model's accuracy at \num{rv.cas.thisthat.jev.dtwo.costfrac} of its cost, and escalating to Gemma-4-31B instead it matched that model's higher accuracy at \num{rv.cas.thisthat.compgem.dtwo.usd} US dollars per thousand decisions, about the hosted price. For engineers who build automated decision gates, these findings become rules that choose the model class by the availability of labels, read generative models through key likelihoods, and set, verify or guarantee thresholds per model and task. The same rules recalibrate per gate, validate option names, rendering and input budget together with the model, check out-of-scope requests separately, and escalate only where the second stage is more accurate on the doubted decisions.

The workflow labels are the consensus of a teacher model and the intent and social-science labels come from the original collections, so accuracy measures agreement with those references, although the rater-agreed GoEmotions labels gave the same ranking of the model classes. The public benchmarks stand in for operating gates rather than logs of a deployed system. The training data of the hosted model and of Laya are not released, so their exposure to the benchmark items cannot be checked, and the hosted model can change without notice, which the pinned version documents but cannot prevent. Most of the reviewed literature on these models consists of preprints and developer reports rather than peer-reviewed articles.

Future work can test these rules on logs from operating gates with labels adjudicated by people, and can rerun the released harness on each new release of the models it covers.

%% file: sections/appendix_proofs.tex
\section{Proofs of the propositions}
\label{sec:proofs}

This appendix proves the three propositions of Section~\ref{sec:formulation}. Each proof uses only the definitions stated there. The analysis code uses each result exactly as stated, the temperature recovery to form the raw and scaled arms, the hierarchical identity to score the two-stage intent condition and the cascade identity to read the held-out cascades of Table~\ref{tab:cascade}.

\begin{pf}[Proposition~\ref{prop:temperature}]
For logits \(\mathbf{z}\) and a temperature \(T\), write \(Z_T = \sum_{j=1}^{K} \exp(z_j/T)\) for the normalizing sum, so that \(p^{(T)}_k = \exp(z_k/T)/Z_T\). Raising each component to the power \(T/T'\) gives
\begin{equation*}
  \bigl(p^{(T)}_k\bigr)^{T/T'} = \frac{\exp(z_k/T')}{Z_T^{\,T/T'}} .
\end{equation*}
The denominator \(Z_T^{\,T/T'}\) does not depend on \(k\), so dividing by the sum of these values over \(k\) cancels it and leaves \(\exp(z_k/T')/\sum_j \exp(z_j/T') = p^{(T')}_k\). The logits themselves are never needed. When a model serves a known stored temperature \(T_s\), setting \(T=T_s\) and \(T'=1\) recovers the raw arm, and a further temperature \(T''\) fitted on held-out data is applied in the same way to form the scaled arm. \qed
\end{pf}

\begin{pf}[Proposition~\ref{prop:hierarchical}]
Let \(d(y)\) denote the domain of the gold intent \(y\), let \(\hat d\) be the predicted domain, and let \(\hat\imath_{d}\) be the intent the model selects when it is asked only among the intents of domain \(d\). The hierarchical policy returns \(\hat\imath_{\hat d}\), which lies in domain \(\hat d\). If \(\hat d \neq d(y)\), then \(\hat\imath_{\hat d}\) belongs to a domain that does not contain \(y\), because every intent belongs to exactly one domain, so the answer is wrong. If \(\hat d = d(y)\), the policy returns \(\hat\imath_{d(y)}\), which is correct exactly when the within-gold-domain answer is correct. Hence the policy is correct on an item if and only if \(\mathbf{1}[\hat d = d(y)]\,\mathbf{1}[\hat\imath_{d(y)} = y] = 1\), and averaging this product over items gives its accuracy. Both indicators are observed for every item, because the two-stage condition asks the domain question and the within-gold-domain intent question independently. The identity treats the answer to the offline within-gold-domain request as the answer the policy would receive, which requires that request to be identical to the one the policy would send after a correct domain answer. It then holds exactly for a model that returns the same distribution for the same request, as the determinism check of Section~\ref{sec:protocol} tests for the open checkpoints, and it is an offline estimate for the hosted model, whose repeated answers vary. \qed
\end{pf}

\begin{pf}[Proposition~\ref{prop:cascade}]
Let \(a_{1,i}\) and \(a_{2,i}\) be the correctness indicators of the two stages on decision \(i\) out of \(N\), and let \(E=\{i : c_i<\tau\}\) be the escalated set, so that \(e(\tau)=|E|/N\). The cascade takes the second stage's answer on \(E\) and the first stage's answer elsewhere, so its accuracy is
\begin{equation*}
  A_{\mathrm{c}}(\tau) = \frac{1}{N}\Bigl(\sum_{i\notin E} a_{1,i} + \sum_{i\in E} a_{2,i}\Bigr)
  = A_1 + \frac{1}{N}\sum_{i\in E}\bigl(a_{2,i}-a_{1,i}\bigr)
  = A_1 + e(\tau)\,\Delta_{\mathrm{esc}}(\tau),
\end{equation*}
where the middle expression holds for every threshold and the last one requires \(E\) to be nonempty, with \(\Delta_{\mathrm{esc}}(\tau) = |E|^{-1}\sum_{i\in E}(a_{2,i}-a_{1,i})\) the accuracy difference on the escalated decisions. When \(E\) is empty the sum vanishes and the cascade equals the first stage, so the cascade exceeds \(A_1\) only if \(E\) is nonempty and \(\Delta_{\mathrm{esc}}(\tau)>0\). Writing the same sum around the second stage gives
\begin{equation*}
  A_{\mathrm{c}}(\tau) - A_2 = \frac{1}{N}\sum_{i\notin E}\bigl(a_{1,i}-a_{2,i}\bigr) = \bigl(1-e(\tau)\bigr)\,\Delta_{\mathrm{keep}}(\tau),
\end{equation*}
where the last expression requires at least one kept decision and \(\Delta_{\mathrm{keep}}(\tau)\) is the accuracy of the first stage minus that of the second on the kept decisions, so the cascade exceeds \(A_2\) only if \(\Delta_{\mathrm{keep}}(\tau)>0\). For the cost, every decision passes through the first stage and only the escalated decisions reach the second. The mean cost per decision is \(\bar k_1 + e(\tau)\,\bar k_2^{E}\), where \(\bar k_1\) is the mean first-stage cost over all decisions and \(\bar k_2^{E}\) the mean second-stage cost over the escalated ones. Dividing by the mean cost of the second stage alone gives the cost fraction of Eq.~\eqref{eq:costfraction}, which reduces to \((k_1 + e(\tau)\,k_2)/k_2\) when every decision costs the same. \qed
\end{pf}

%% file: sections/appendix.tex
\section{Question texts and frozen conditions}
\label{sec:appendix-question-texts}

This appendix lists the exact question text and every frozen condition a model was asked to answer, so a reader can reproduce a request byte for byte from the released input files rather than from a paraphrase in the body of the paper. Every condition below is written once to a JSON Lines file before any model runs. Each file's SHA-256 digest is recorded in a manifest alongside the seed, \num{p.seed}, used to build it, and that digest is what lets a reader confirm the file they replay is the one every model in this study actually answered.

D1 asks \num{p.questionsperstate} questions per state across its four workflows. Each question is phrased as the dataset's own authors wrote it, for example whether a customer-service reply should be escalated, what urgency an incident warrants on an ordered scale, or whether an invoice should be paid, held or rejected. Three D1 conditions are used. \texttt{d1\_native} keeps the dataset's own Choice option names, such as \texttt{billing} or \texttt{technical}, and exists only as the harness-validation anchor of Section~\ref{sec:protocol}. \texttt{d1\_neutral} replaces every Choice option name with a neutral identifier, $o_1$ through $o_K$, while leaving every rubric sentence describing that option unchanged, and it is the condition scored in every accuracy and calibration table in this paper. \texttt{d1\_calib} repeats the neutral rendering on \num{p.calibperworkflow} training-split states per workflow and supplies the disjoint slice the D1 temperature is fitted on.

D2 asks one question, phrased identically across every cardinality condition, of which intent a request to a virtual assistant expresses, or, in the hierarchical condition, of which domain it belongs to. The option descriptions are the intent names with underscores replaced by spaces, so no formatting artifact distinguishes a neutral identifier from a natural-language label. Four flat conditions, \texttt{d2\_k5}, \texttt{d2\_k20}, \texttt{d2\_k50} and \texttt{d2\_k150}, ask the same \num{n.d2k5.requests} in-scope utterances with a growing, nested option set that always contains the gold intent, and \texttt{d2\_k150} additionally carries the \num{p.oos} out-of-scope utterances under the full option set. Two further conditions, \texttt{d2\_hier\_dom} and \texttt{d2\_hier\_int}, ask the domain and then the intent within the gold domain. The condition \texttt{d2\_k150\_oos} repeats \texttt{d2\_k150} with one further option named out of scope, the reference label of the out-of-scope utterances.

D3 asks one fixed question per task, quoted here as the replication package and the original collection render it. Conversations gone awry asks whether the previous conversation will eventually derail into a personal attack. Wikipedia corpus power asks whether the named participant in the conversation is in a position of power. Emotion asks which of six basic emotions a mental health professional would assign to the passage, which is the wording of the replication package, although the reference label is the dataset's distant hashtag label. Wikipedia politeness asks which of three politeness levels, judged against formal workplace norms, best describes the conversation.

Each D3 task is asked once under neutral option keys. The two binary tasks are additionally asked under four option-naming conditions, keyed \texttt{k01}, \texttt{kny}, \texttt{kswap} and \texttt{krand} for the digit pair, the rubric-aligned pair, the rubric-reversed pair and a pair of random five-letter strings. For the binary D3 tasks, the leading word that names the rubric's own key, such as a description that opens with the word True, is stripped in the four naming conditions. The option name is then the only cue to polarity. D1's noul questions are recast under the same four naming conditions with their rubric text unchanged, because their rubrics do not name a key, which gives the \texttt{e2\_d1\_*} and \texttt{e2\_d3\_*} condition families used by the option-naming experiment of Section~\ref{sec:measures}.

Qwen3-14B receives each question as a separate chat turn built from the same state, question and option texts, with the thinking mode of Qwen3 switched off. The prompt lists every option as its key and description and asks for one JSON object that holds the key of the best option and a list of key and probability pairs. The list covers every option when a question has at most \num{p.comp.fullk} options and the \num{p.comp.topn} options the model judges most likely otherwise. Structured decoding constrains the reply to that schema under greedy decoding, with a context limit of \num{p.comp.maxlen} tokens and at most \num{p.comp.maxtokens} generated tokens.

The parser clips negative values at zero, assigns zero to the options the reply omits and renormalizes the reported values to sum to one, falling back to a uniform distribution when every reported key misses the option set. A reply that cannot be parsed is scored as a uniform distribution, the same rule that applies when every reported key misses the option set, and a prompt that exceeds the context limit is recorded as a failed request. The prompt tells the model that its values will be renormalized, and the chosen key it reports is not used, since the option with the highest renormalized value is taken as its answer.

Of the \num{rv.fail.compopen.answeredquestions} answered questions of Qwen3-14B, \num{rv.fail.compopen.renormalized} returned values that did not sum to one, \num{rv.fail.compopen.percentscale} of them on a percent scale, and every reply held parseable JSON, although a few named a key together with its description (Table~\ref{tab:readouts}). The resulting numbers are verbalized probabilities that the model states in text, not the probabilities it assigns to its own output tokens, so every table keeps the comparators in a separate group. The calibration of a verbal readout is read as that of stated confidence. The exact prompt and schema are part of the released adapter code.

Qwen3.6-27B receives the same chat turn with its thinking mode off, and its verbal readout follows the protocol above with a longer context limit, which the released adapter records. Its likelihood readout appends the opening characters of the requested JSON reply up to the answer field and scores each option key as the continuation, summing the log-probabilities of the key's tokens and of the token that closes the field. The scores are normalized over the options at a temperature of one, and every needed token lay within the log-probabilities the server returned. Both readouts choose the same option for every derailment conversation, nearly always the same one of the two, so their derailment accuracies coincide.

The thinking mode switches the reasoning of Qwen3.6-27B on and samples it at a temperature of \num{m.think.temp} with a nucleus mass of \num{m.think.topp} and a fixed seed. The same parser reads the JSON object that follows the reasoning and scores a reply without one as a uniform distribution. Gemma-4-31B and Mistral-Small-24B receive the same chat turn through their own chat templates, and both readouts follow the protocol of Qwen3.6-27B unchanged. The backbone controls use the same key scoring after a plain prompt that lists the state, the question and each option as its key and description, followed by an instruction to reply with one key.

\section{Supporting results}
\label{sec:appendix-per-task}

This appendix reports the estimates behind several statements in Section~\ref{sec:results} with their uncertainty, together with the run manifest. Table~\ref{tab:calibtasks} gives the calibration error on every task rather than pooled over the social-science tasks. The scaled error lies below the shipped error for nearly every model and task, but on the social-science tasks it stays between \num{rv.ts.layaml.politeness.ecescaled} and \num{rv.ts.thisthat.toxicity.ecescaled} for the decision models. The intervals refit the temperatures inside each bootstrap replicate. Because the bootstrap distribution of the calibration error is biased upward when the error is small, a few point estimates lie below their own interval. Several fits on those tasks reach the upper end of the temperature grid, which flattens the probabilities of a model whose confidence carries little information.

\longtab{Expected calibration error on every task as shipped and after temperature scaling, with an interval for the scaled error whose bootstrap replicates refit the cross-fitted temperatures.}{tab:calibtasks}{tables/t09_calib_tasks}

Table~\ref{tab:selective} contrasts the coverage at \num{p.riskfive} percent risk chosen on the scored decisions with the coverage a threshold chosen on disjoint data achieves on them. The hosted model keeps its coverage on CLINC-150 and gains some on typed-decisions, at a realized risk above the target there, while the typed-decisions coverage of Kev-9B falls from \num{rv.sel.kevlarge.doneneutral.covfive} in sample to \num{rv.ho.kevlarge.doneneutral.five.cov} once the threshold is fixed in advance. Every one-sided upper bound on the realized risk lies above the target except that of the stated probabilities of Gemma-4-31B on typed-decisions, which is why the main text treats these coverages as estimates. Table~\ref{tab:paired} gives the paired accuracy difference between every model and the hosted model on each dataset with its state-clustered interval, and bold marks the differences that remain significant after Holm correction within each dataset.

\longtab{Selective prediction at \num{p.riskfive} percent risk on D1 and D2, with the threshold chosen on the scored decisions and with the threshold chosen on data disjoint from them.}{tab:selective}{tables/t10_selective}

Table~\ref{tab:riskcontrol} replaces the held-out threshold with one chosen by Learn-then-Test, whose coverage carries a finite-sample guarantee on the risk of the accepted decisions. The threshold is the lowest of \num{rc.gridsize} candidates fixed before any data are seen whose accepted calibration decisions pass the binomial test after the Bonferroni correction, and it is applied unchanged to the scored decisions. Coverage falls wherever the calibration data leave few accepted decisions at high confidence, so most models receive none on typed-decisions while the hosted model keeps \num{rc.jev.dtwo.cov} of the intent decisions. A model whose top probabilities sit near one, as the key likelihoods of Gemma-4-31B do, receives no coverage on either benchmark.

\longtab{Selective prediction with a guaranteed risk on D1 and D2, with the threshold chosen by Learn-then-Test on the calibration data.}{tab:riskcontrol}{tables/t21_risk_control}

\longtab{Paired accuracy differences between each model and Jev 1.13.0 on identical decisions, in percentage points.}{tab:paired}{tables/t11_paired}

Table~\ref{tab:readouts} gives, for every verbal readout, the share of decisions whose reply the fixed parser could not use and the accuracy with the fixed parser and with a lenient re-parse. A reply is unusable when it holds no parseable JSON after any reasoning or names an option by its key joined to its description, and the fixed parser scores it as a uniform distribution. The lenient re-parse accepts such keys and leaves every other reply unchanged, so the difference between the two accuracies is the cost of the format alone. The unusable share differs most between checkpoints on emotion, from \num{vr.compgem.emotion.failshare} for Gemma-4-31B to \num{vr.compmis.emotion.failshare} for Mistral-Small-24B.

\longtab{Reliability of the verbal readouts, with the share of unusable replies and the accuracy under the fixed and a lenient parser.}{tab:readouts}{tables/t20_readouts}

Table~\ref{tab:cascadeqwen} lists the held-out cascades whose second stage is one of the two Qwen comparators, in the format of Table~\ref{tab:cascade}. Every decision model serves as first stage, the thresholds are chosen out of fold as for the main cascades, and the costs are computed per item from the recorded tokens and GPU time. Most of these cascades gain no accuracy beyond the interval of the better single stage. Four gains have intervals that exclude zero, the two positive gains that Section~\ref{sec:results-cascade} discusses and two small losses when a Kev checkpoint escalates to the likelihood readout of Qwen3.6-27B on CLINC-150.

\longtab{Confidence-gated cascades on D1 and D2 that escalate every decision model to Qwen3-14B or to either readout of Qwen3.6-27B, with thresholds chosen on data disjoint from the scored items.}{tab:cascadeqwen}{tables/t22_cascade_qwen}

Table~\ref{tab:backbones} compares each decoder-based decision model with the untuned backbone it adapts, on the same decisions and under the same paired bootstrap.

\longtab{Decoder-based decision models against their untuned backbones on identical decisions.}{tab:backbones}{tables/t17_backbones}

Table~\ref{tab:oosoption} repeats the intent-routing gate with an explicit out-of-scope option among the declared options.

\longtab{The intent benchmark with an explicit out-of-scope option, answered directly and through the gate at the in-scope risk target.}{tab:oosoption}{tables/t18_oos_option}

Table~\ref{tab:baselines} adds the calibration, coverage, out-of-scope separation and latency of the conventional classifiers on the workflow and intent benchmarks, beside the hosted model.

\longtab{Conventional classifier baselines on the same test items beside the hosted decision model. The supervised classifiers are trained on their benchmark's training split and the entailment classifier without task training.}{tab:baselines}{tables/t14_baselines}

\begin{figure}[pos=htbp]
\centering
\includegraphics{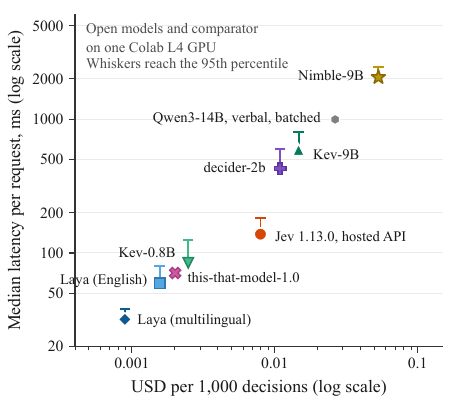}
\caption{Cost per thousand typed-decisions decisions against median latency per request, with whiskers to the \num{p.pninetyfive}th percentile, for the hosted model through its API and the other models on one Colab L4 GPU.}
\label{fig:cost}
\end{figure}

Table~\ref{tab:names} gives every option-name condition on each binary set together with the measured test-retest floors. Figure~\ref{fig:cost} places the cost per thousand typed-decisions decisions against the median latency of each decision model and Qwen3-14B, whose values Table~\ref{tab:cost} reports with those of the other comparators.

\longtab{Option-name conditions on the binary questions, with flips per hundred decisions against the aligned identifiers, the area under the curve under those identifiers and the test-retest floor.}{tab:names}{tables/t06_names}

Table~\ref{tab:scores} gives the multi-class Brier score and the negative log-likelihood of every model on each benchmark, the two proper scoring rules that complement the calibration error.

\longtab{Multi-class Brier score and negative log-likelihood of every model on each benchmark, from the probabilities as shipped.}{tab:scores}{tables/t16_scores}

\begin{figure}[pos=htbp]
\centering
\includegraphics[width=0.84\textwidth]{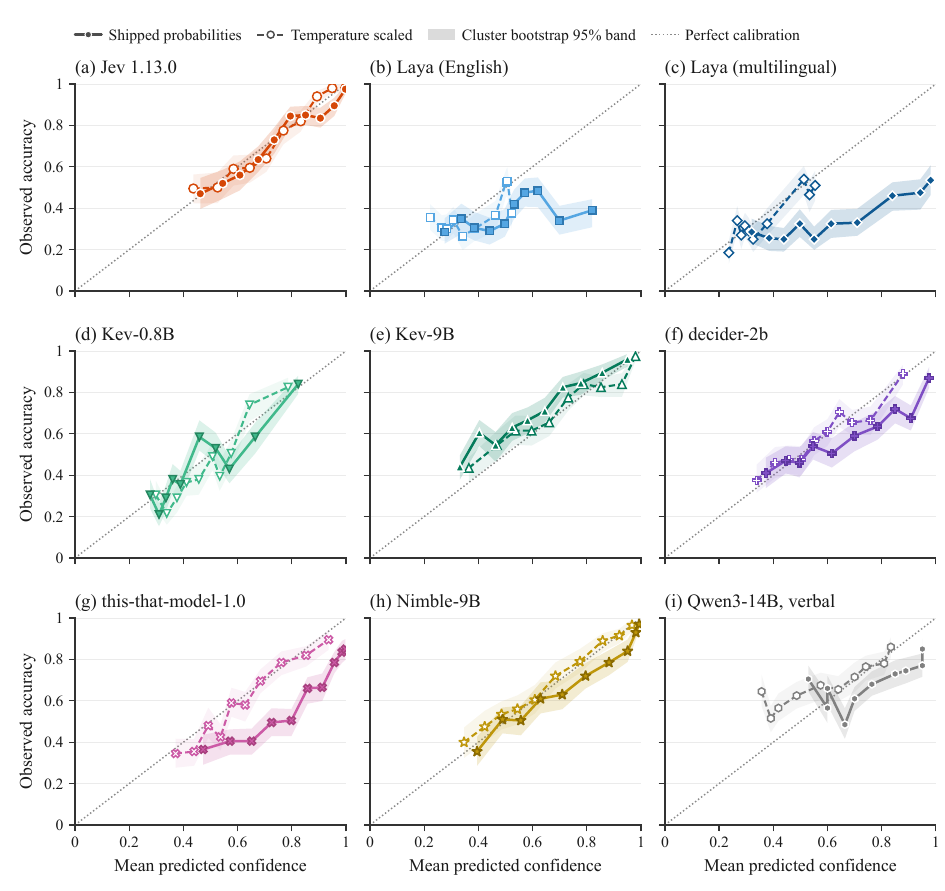}
\caption{Reliability diagrams on D1 with \num{p.bins} equal-mass bins, shipped and temperature scaled, with cluster bootstrap bands.}
\label{fig:reliability}
\end{figure}

Table~\ref{tab:manifest} lists the identifier, resolved revision, served temperature, run date and runtime of every model, which the released manifest records in full together with the digest of every frozen input. The hosted service reports neither a weight revision nor a temperature, so its entry records the pinned model name that every response returned. Figure~\ref{fig:reliability} then shows the reliability diagrams on typed-decisions from which the shipped and scaled calibration values of that benchmark are computed.

\longtab{Run manifest with the identifier, resolved revision, served temperature, run date and runtime of every model.}{tab:manifest}{tables/t12_manifest}

The calibration errors of Table~\ref{tab:calibration} summarize each model by one number. Each panel of Figure~\ref{fig:reliability} relates confidence to accuracy for one model. Each panel bins the decisions of one model into \num{p.bins} equal-mass bins by top-label probability and plots the observed accuracy of each bin against its mean confidence. A curve on the diagonal is therefore calibrated, a curve below it is over-confident and a curve above it is under-confident. The shaded band is a cluster bootstrap interval that resamples whole states.

The hosted model and Kev-0.8B lie close to the diagonal as shipped, and Kev-9B lies slightly above it, which marks mild under-confidence. The decider-2b, this-that-model-1.0 and Nimble-9B checkpoints lie below the diagonal as shipped, and the scaled temperature moves their curves onto it or slightly above it. Qwen3-14B is also over-confident as shipped and remains noisy after scaling. The two Laya checkpoints differ from every other model, because their observed accuracy stays between roughly a quarter and a half across the whole confidence range. Scaling pulls their confidences toward the same narrow range rather than separating correct from incorrect decisions, which is the loss of resolution discussed in Section~\ref{sec:results-accuracy}. Their shipped curves show the over-confidence that the Laya card reports.

%% file: sections/declarations.tex
\section*{Declaration of competing interest}

The authors declare that they have no known competing financial interests or personal relationships that could have appeared to influence the work reported in this paper. The authors have no affiliation with the developers of any model evaluated in this study.

\section*{Funding}

This research did not receive any specific grant from funding agencies in the public, commercial, or not-for-profit sectors.

\section*{CRediT authorship contribution statement}

\textbf{Amir Rafe:} Conceptualization, methodology, software, data curation, formal analysis, investigation, visualization, writing of the original draft, review and editing. \textbf{Subasish Das:} Conceptualization, methodology, supervision, and review and editing.

\section*{Code and data availability}

The harness, the frozen input files with their SHA-256 digests, the raw answers of every model, and the analysis and plotting scripts are available at \url{https://github.com/pozapas/system-one-models-survey}. D1, D2, the four D3 tasks, GoEmotions and Banking77 are drawn from their own public releases under the licenses stated in Table~\ref{tab:datasets}. The repository redistributes D1, D2 and Banking77 with attribution in the frozen, neutral-identifier form that the harness constructs. It releases the four D3 tasks and GoEmotions as item identifiers together with scripts that rebuild the frozen inputs from the pinned public releases, because the D3 licenses do not grant redistribution of the text and the GoEmotions texts are comments written by Reddit users. For the same reason, reasoning traces of the thinking mode that quote the text of these items are withheld from the released answers, whose probabilities are kept.